\documentclass[lettersize,journal]{IEEEtran}

\usepackage[colorlinks,urlcolor=black,linkcolor=blue,citecolor=blue]{hyperref}

\def\BibTeX{{\rm B\kern-.05em{\sc i\kern-.025em b}\kern-.08em
    T\kern-.1667em\lower.7ex\hbox{E}\kern-.125emX}}

\usepackage{rotating}
\usepackage{amssymb}
\usepackage{algorithm}
\usepackage{algpseudocode}
\usepackage{color}
\usepackage{epsf}
\usepackage{psfrag}
\usepackage{epsfig}
\usepackage{graphicx}
\usepackage{amsfonts}
\usepackage{textcomp}
\usepackage{comment}
\usepackage{footnote}
\usepackage{tablefootnote}
\usepackage[flushleft]{threeparttable}
\usepackage{paralist}
\usepackage{mdwlist}
\usepackage{amssymb}
\usepackage{amsmath}
\usepackage{url}
\usepackage{verbatim}
\usepackage{alltt}
\usepackage{balance}
\usepackage{multirow}
\usepackage{epstopdf}
\usepackage{lipsum}
\usepackage{float}
\usepackage{booktabs}
\usepackage{slashbox}
\usepackage[table]{xcolor}
\usepackage{array}
\usepackage{boldline}
\usepackage{caption,setspace}
\usepackage{mathtools}
\usepackage[table]{xcolor}
\usepackage{csquotes}

\usepackage{stackengine}[2013-10-15]

\usepackage[T1]{fontenc}
\usepackage{frcursive}
\usepackage{calligra}
\usepackage{wedn}
\usepackage{bm}
\usepackage{aurical}
\usepackage{upgreek}

\definecolor{darkgreen}{RGB}{0,204,0}

\usepackage{cite}
\usepackage{graphicx}
\usepackage{adjustbox}

\usepackage{esint}
\usepackage{tipa}
\usepackage{soul}

\DeclareMathOperator{\E}{\mathbb{E}}

\newcolumntype{?}{!{\vrule width 1pt}}
\newcolumntype{+}{!{\vrule width 1.25pt}}

\def\hlineb#1{%
\noalign{\ifnum0=`}\fi\hrule \@height #1 %
\futurelet\reserved@a\@xhline}

\newcommand{\argmax}{\arg\!\max}

\usepackage{cancel}

\ifCLASSINFOpdf
\else
\fi
\begin{document}

\title{Demystifying Visual Features of Movie Posters for Multi-Label Genre Identification}

\author{Utsav Kumar Nareti,~\IEEEmembership{Student Member,~IEEE}, 
    Chandranath Adak,~\IEEEmembership{Senior Member,~IEEE},
    Soumi Chattopadhyay,~\IEEEmembership{Senior Member,~IEEE}
\thanks{
%
Utsav Kumar Nareti and Chandranath Adak are with the Dept. of CSE, Indian Institute of Technology Patna, Bihar 801106, India. 
(email: {\{utsav\_2221cs28, chandranath\}@iitp.ac.in}).\\ 
Soumi Chattopadhyay is with the Dept. of CSE, Indian Institute of Technology Indore, Madhya Pradesh 453552, India. 
(email: soumi@iiti.ac.in).\\
Soumi Chattopadhyay acknowledges the partial support from YFRSG-PRIUS Grant, IIT Indore (IITI/YFRSG-PRIUS/2023-24/Phase-I/01).\\
This work has been accepted to the IEEE Transactions on Computational Social Systems for publication. Copyright may be transferred without notice, after which this version may no longer be accessible.
}
}

\markboth{U. K. Nareti \MakeLowercase{\textit{et al.}}: Demystifying Visual Features of Movie Posters for Multi-Label Genre Identification}
{U. K. Nareti \MakeLowercase{\textit{et al.}}: Demystifying Visual Features of Movie Posters for Multi-Label Genre Identification}


\maketitle

\begin{abstract}
In the film industry, movie posters have been an essential part of advertising and marketing for many decades, and continue to play a vital role even today in the form of digital posters through online, social media and OTT (over-the-top) platforms. Typically, movie posters can effectively promote and communicate the essence of a film, such as its genre, visual style/tone, vibe and storyline cue/theme, which are essential to attract potential viewers. 
Identifying the genres of a movie often has significant practical applications in recommending the film to target audiences. 
Previous studies on genre identification have primarily focused on sources such as plot synopses, subtitles, metadata, movie scenes, and trailer videos; however, posters precede the availability of these sources, and provide pre-release implicit information to generate mass interest. 
In this paper, we work for automated multi-label movie genre identification only from poster images, without any aid of additional textual/meta-data/video information about movies, which is one of the earliest attempts of its kind. Here, we present a deep transformer network with a probabilistic module to identify the movie genres exclusively from the poster. 
For experiments, we procured 13882 number of posters of 13 genres from the Internet Movie Database (IMDb), where our model performances were encouraging and even outperformed some major contemporary architectures.
\end{abstract}


\section{Introduction}\label{sec:intro}

\noindent
\IEEEPARstart{I}{n} 
the contemporary landscape of the film industry, 
digital platforms have revolutionized how content is consumed, elevating the role of movie posters. 
These visual canvases, once primarily relegated to theater exhibits, newspaper ads, and DVD covers, have emerged as powerful tools for attracting audiences in the era of online streaming \cite{30_9057706}. 
A movie poster is no longer just a piece of promotional artwork; it has become a gateway to a cinematic experience, a glimpse into the world of a film, and a crucial factor in a viewer's decision-making process.
Beyond their aesthetic appeal, posters convey subtle cues about the film's genre, style, and thematic content. A well-crafted poster can encapsulate the essence of a movie, enticing viewers with tantalizing glimpses of its narrative and emotional landscape. As viewers increasingly turn to online platforms to discover and enjoy films, movie posters have assumed a pivotal role in the digital realm, guiding users to their desired content \cite{4_winoto2010role}.

In the digital age, with the sheer volume of content available, accurate genre categorization is essential for audiences to navigate the extensive catalogs of online streaming platforms and find movies that match their preferences \cite{netflix_frey2021}. 
The reliance on genre categorization underscores the crucial role of automated genre identification in improving film discoverability and enhancing the overall user experience. 
In the literature, while automated movie genre identification mostly used video trailers \cite{25_montalvo2022trailers12k,26_yadav2020unified} and textual plot synopses\cite{10_hoang2018predicting,13_wehrmann2018self},  
very few works engaged movie posters \cite{28_10.1145/3132515.3132516}. 
However, movie posters play a crucial role in genre identification, and subsequently attracting potential target audiences, since they precede the release of the film itself, even before trailers and synopses become available. 
Moreover, compared to the video/ textual modality, posters serve as prevalent thumbnails on OTT (over-the-top) platforms, and are extensively shared across social media and various advertising/ promotional channels.

\begin{figure}
\centering
\scriptsize
\begin{adjustbox}{width=0.49\textwidth}
\begin{tabular}{c|c|c|c}
\includegraphics[width=0.2\linewidth, height=0.3\linewidth]{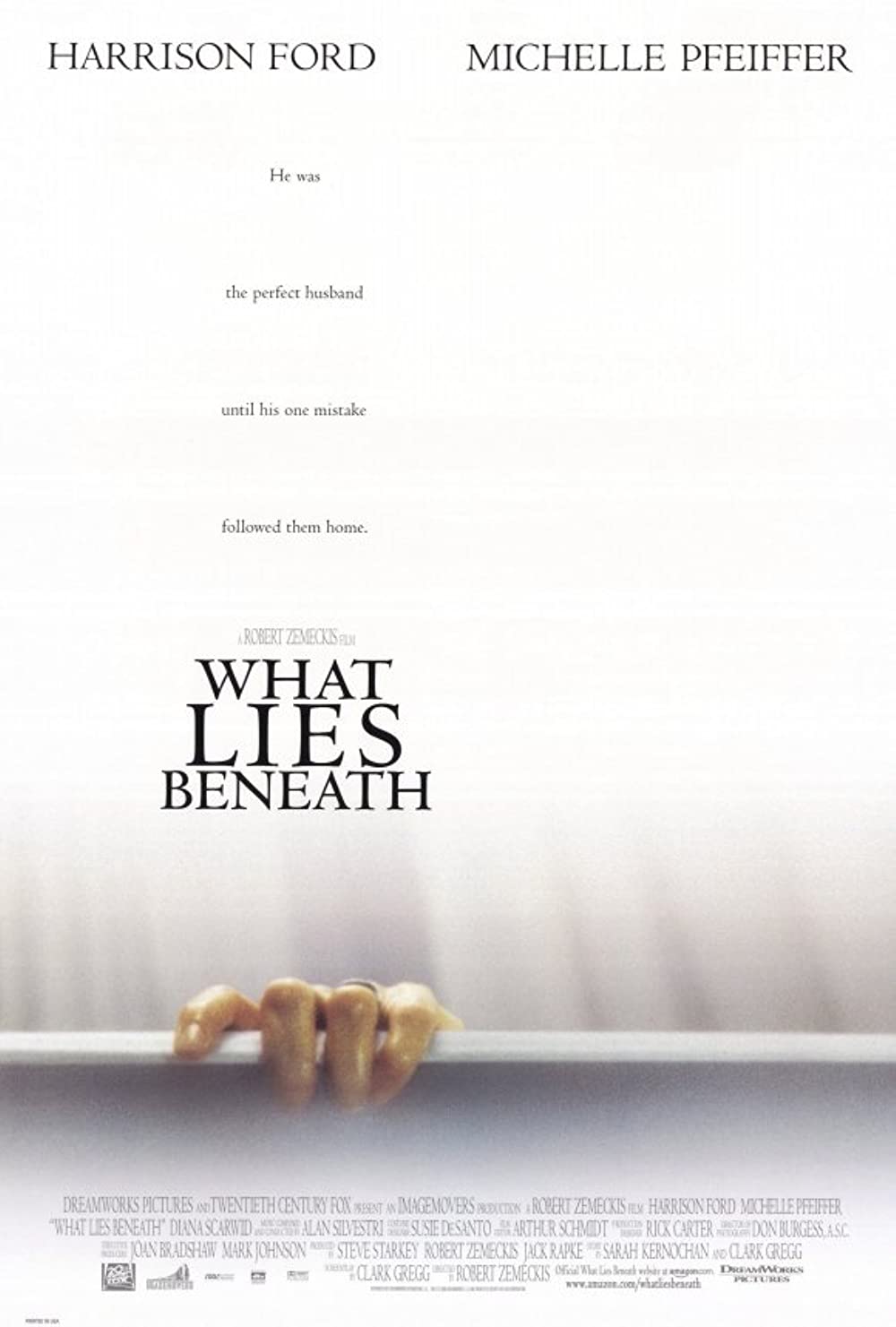} & 
\includegraphics[width=0.2\linewidth, height=0.3\linewidth]{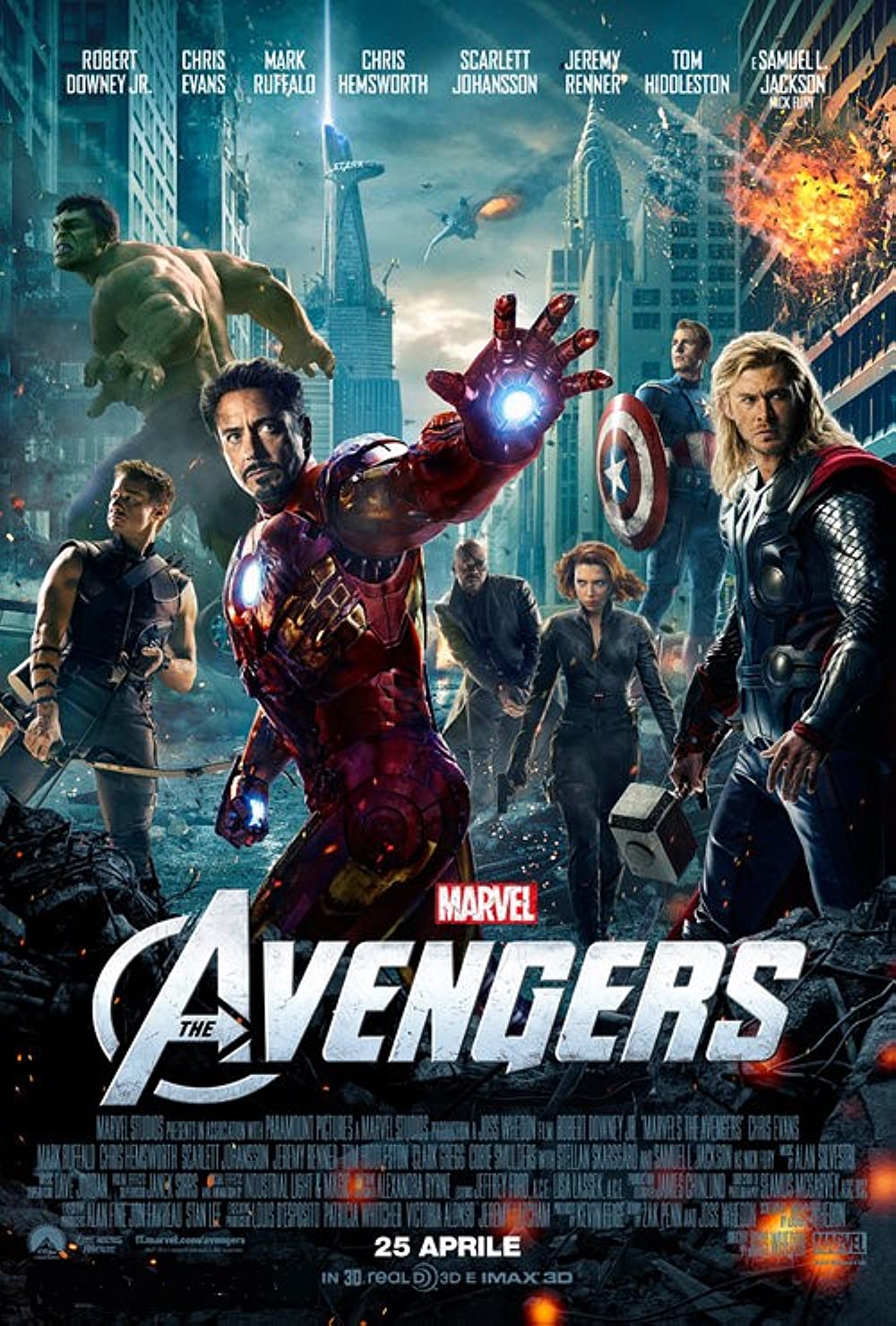} & 
\includegraphics[width=0.2\linewidth, height=0.3\linewidth]{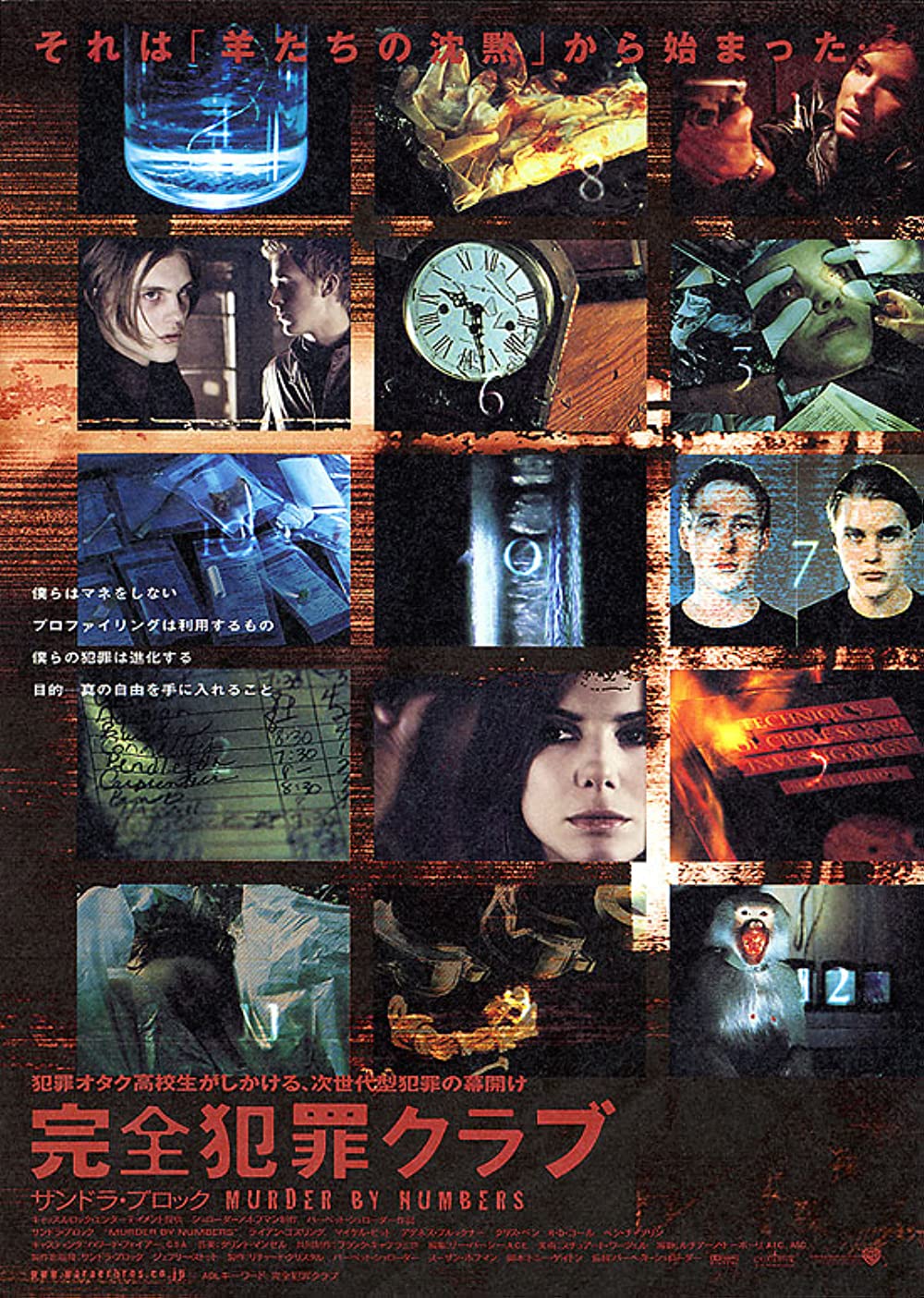} & 
\includegraphics[width=0.2\linewidth, height=0.3\linewidth]{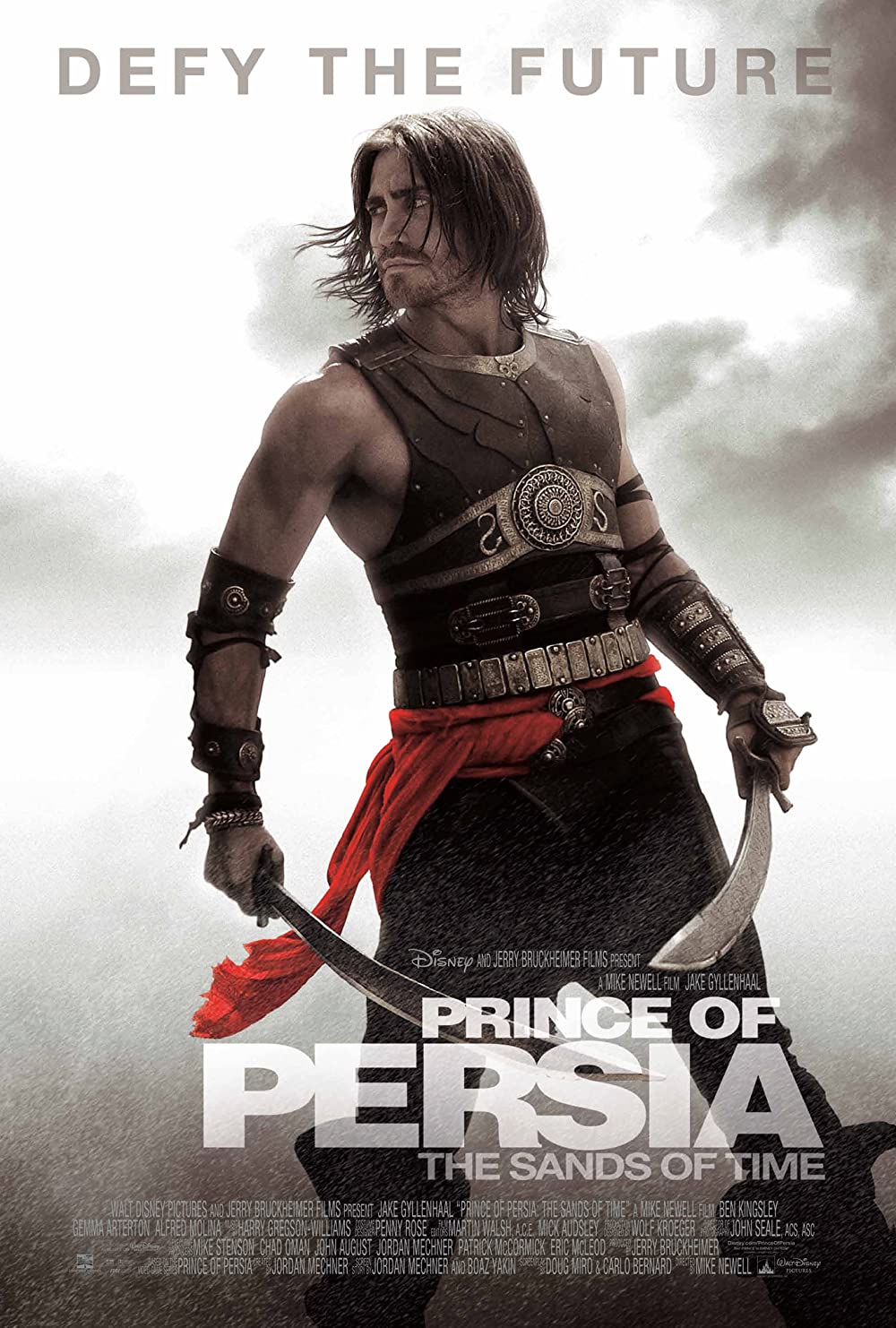} \\  
\textcolor{blue}{{(a)}} Drama, & \textcolor{blue}{{(b)}} Action, & \textcolor{blue}{{(c)}} Crime, & \textcolor{blue}{{(d)}} Action,   \\
Horror, Mystery & Sci-Fi, \---  & Mystery, Thriller & Adventure, Fantasy \\
\end{tabular}
\end{adjustbox}
\caption{Example of movie posters with genres}
\label{fig:intro_challenge}
\end{figure}

In Fig. \ref{fig:intro_challenge}, we present some movie poster samples with corresponding genres. 
Analyzing only poster images brings several challenges, since a single poster may have 
limited information (Fig. \ref{fig:intro_challenge}: (a)), 
intricate backgrounds (Fig. \ref{fig:intro_challenge}: (b)), 
incorporated multiple small images in a collage (Fig. \ref{fig:intro_challenge}: (c)), 
or included solely the cast member photos (Fig. \ref{fig:intro_challenge}: (d)). 
We here analyze only the poster image without any aid of other modalities to identify its genre, considerably more challenging than other computer vision tasks, 
since genres are intangible implicit features and rely on individual human perception, i.e., 
one person may classify a poster as one genre while another may classify it differently \cite{30_9057706}. 
Additionally, a poster can belong to multiple genres, introducing the challenge of multi-label classification \cite{Herrera2016} and potentially exacerbating data imbalance concerns \cite{Johnson2019SurveyOD}. 
In a poster, a genre may be suppressed by other genres, e.g., in Fig. \ref{fig:intro_challenge}: (d), \emph{action} and \emph{adventure} genres are more explicit than \emph{fantasy}. 
Moreover, the information present in a poster brings additional challenges in identifying the genre, which we briefly mention in 
Supplementary Appendix B. 
In this study, we obtained posters from 
IMDb, 
where each poster can be multi-labeled \cite{Herrera2016} with up to three genres.

\textcolor{black}{The primary motivation for addressing genre identification from posters lies in the pivotal role that a poster plays on OTT platforms. 
The poster is often the first piece of content a user sees, which prompts them to click on the item to explore its textual content or watch the trailer, ultimately deciding whether they are interested in watching the movie. 
Therefore, it is crucial to design the poster in a way that effectively conveys the content or genre of the film. 
If the poster fails to do so, potential viewers interested in a particular genre might skip the movie, leading content creators to redesign the poster to better reflect its content. 
To address this need, an automated system is necessary to determine whether a poster accurately identifies the genres. 
However, identifying genres solely from a poster, without considering other modalities, is a relatively underexplored area in the literature \cite{28_10.1145/3132515.3132516,25_montalvo2022trailers12k,26_yadav2020unified,10_hoang2018predicting,13_wehrmann2018self}.
Previous research in this domain has not achieved satisfactory accuracy, often due to ineffective handling of the complexities of multi-label genre identification or by considering a limited number of genres. 
Therefore, the goal of this paper is to effectively identify multiple genres from a poster image alone, without any aid from textual, video, or audio modalities.  
To achieve this, we propose a novel approach incorporating several key techniques and contributions, as outlined below:}

\textcolor{black}{\emph{{(i)} Novel model for understanding poster-genre relationships:}  
We propose a residual dense transformer model by leveraging densely connected transformer encoders that use deep feature embeddings as input instead of raw image patches \cite{dosovitskiy2020image}. 
This approach enhances the ability of the model to grasp the global context, and 
decipher complex visual relationships within the poster image and its multiple genres.}

\textcolor{black}{\emph{{(ii)} Advanced multi-label classification technique:} To address the multi-label classification challenges inherent in posters associated with multiple genres, we introduce an ensemble technique combined with an asymmetric loss function. This method is particularly effective when positive labels are less prevalent than negative ones \cite{ridnik2021asymmetric}, improving the accuracy and reliability of genre predictions.}

\textcolor{black}{\emph{{(iii)} Probabilistic module for variable genre identification:} Our model includes a novel probabilistic module designed to handle the variability in the number of genres associated with each poster. This module aims to eliminate extraneous genres and accurately determine the correct number of genres, enhancing the precision of genre identification.}

\textcolor{black}{\emph{{(iv)} Extensive experiments and ablation studies:} We conducted comprehensive experiments on poster images obtained from IMDb to evaluate the effectiveness of our model. By comparing our approach with contemporary architectures and performing detailed ablation studies, we provide valuable insights into how visual elements of posters correlate with movie genres. These findings have significant implications for improving film recommendation systems and advancing the digital evolution of the film industry.}

The rest of the paper is organized as follows. 
Section \ref{1_1sec:related} reviews related literature, 
Section \ref{3sec:method} discusses the proposed methodology, 
Section \ref{4sec:result} analyzes the experimental results, and 
Section \ref{5sec:conclusion} concludes the paper.

\section{Related Work}
\label{1_1sec:related}
\noindent 
Our primary focus in this paper is identifying multi-label movie genres exclusively from posters, a task that is relatively limited in the literature \cite{30_9057706}. However, some past works used trailer \cite{23_Simes2016MovieGC}, clips \cite{21_4107177}, and facial frames \cite{26_yadav2020unified} as visual inputs. 
Additionally, numerous studies focused on textual inputs, such as movie plot summary \cite{10_hoang2018predicting}/ synopsis \cite{11_kar-etal-2018-folksonomication} and screenplay \cite{17_gorinski-lapata-2018-whats}. 
Furthermore, past endeavors engaged multimodal approaches, combining visual, textual, and audio inputs \cite{38_bribiesca2021multimodal,35_cascante2019moviescope}. 
We now briefly summarize significant past studies, along with Table \ref{tab:related_work}.

\begin{table}[!b]
 \centering
 \caption{\small Summary of related works for movie genre identification}
   \begin{adjustbox}{width=0.49\textwidth} 
 \begin{tabular}{c | c | c | c | c | c | c}
   \cline{2-7} 
& {\small Method} &  Input & Architecture/  & \#Genre/ & Dataset & {\small Multi-}\\
& & & Technique & \#Tags & & {\small label?} \\  


\hline
\multirow{11}{*}{\rotatebox{90}{Visual}} &  \cite{20_zhou2010movie}  & Trailer & {\footnotesize GIST, CENTRIST, kNN} & 4 & ($\wp$)  & $\upchi$ \\ \cline{2-7}
&   \cite{23_Simes2016MovieGC}  & Trailer & CNN & 4 & LMTD & $\upchi$ \\ \cline{2-7}
& \cite{24_wehrmann2017movie}   & Trailer & CNN & 9 & LMTD & \checkmark \\ \cline{2-7}
&  \cite{25_montalvo2022trailers12k}  & Trailer &  Transformer & 10 & Trailers12k  & 
    \checkmark\\ \cline{2-7}
&  \cite{21_4107177}   & Clip & SVM &  4 & ($\wp$) & $\upchi$\\ \cline{2-7} 
& \cite{26_yadav2020unified}  & Facial frame & Inception-LSTM & 6 & EmoGDB ($\wp$)  &  
    \checkmark \\ \cline{2-7}
&  \cite{27_pobar2017multi}  & Poster & NB & 
   18 & TMDb ($\wp$) & \checkmark \\ \cline{2-7}
&  \cite{28_10.1145/3132515.3132516}  & Poster & CNN, YOLO & 23 & IMDb ($\wp$)  &  
  \checkmark\\ \cline{2-7}
&  \cite{29_8990490}  & Poster & CNN & 4 & ($\wp$) & $\upchi$ \\ \cline{2-7}
&  \cite{30_9057706}  & Poster & CNN, Gram layer & 12 & IMDb ($\wp$) &  \checkmark\\ 
  \cline{2-7}
\rowcolor[HTML]{F3F3F3}\cellcolor[HTML]{FFFFFF}
& \textbf{Ours} & Poster &  ERDT,  PrERDT & 13 & IMDb & \checkmark \\ 

\hline \hlineB{2}
\multirow{6}{*}{\rotatebox{90}{Textual}}  & \cite{9_ertugrul2018movie} &  Plot summary & BLSTM & 4 & ($\wp$) & $\upchi$\\ \cline{2-7}
  &  \cite{10_hoang2018predicting}  & Plot summary &  GRU & 20 & IMDb ($\wp$) & \checkmark\\ \cline{2-7}
  &  \cite{11_kar-etal-2018-folksonomication}  & Synopsis & CNN-FE & 71 & MPST  & \checkmark\\ \cline{2-7}
  &  \cite{12_battu-etal-2018-predicting}  & Synopsis & CNN, LSTM & 9 & MLMRD  & $\upchi$\\ \cline{2-7}
  
  &  \cite{13_wehrmann2018self}  & Synopsis & Self-Attention & 9 & LMTD & \checkmark \\
  \cline{2-7}

  &  \cite{17_gorinski-lapata-2018-whats}  & Screenplay & MLE, LSTM & 31 & Jinni ($\wp$) & \checkmark\\ 


\hline \hlineB{2} 
\multirow{9}{*}{\rotatebox{90}{Multimodal}}
&  \multirow{2}{*}{\cite{38_bribiesca2021multimodal}}  & Synopsis, Metadata,  & \multirow{2}{*}{MulT-GMU} & \multirow{2}{*}{13} & Moviescope  & \multirow{2}{*}{\checkmark} \\ 
  & & Poster, Trailer, Audio & && ($\wp$) & \\ \cline{2-7}
&  \multirow{2}{*}{\cite{35_cascante2019moviescope}}  & Synopsis, Metadata,  & fastText, fastVideo, & \multirow{2}{*}{13} & Moviescope & 
   \multirow{2}{*}{\checkmark} \\   
   ~  & ~ & Poster, Trailer, Audio & CRNN, VGG-16 & & ($\wp$) & \\ \cline{2-7}
&  \multirow{2}{*}{\cite{36_mangolin2022multimodal}}  & Synopsis, Subtitle, & {\footnotesize Textural feature, LSTM} & \multirow{2}{*}{18} & \multirow{2}{*}{TMDb ($\wp$)} & \multirow{2}{*}{\checkmark} \\
  & & Poster, Trailer, Audio & {\footnotesize kNN, SVM, MLP, DT} & & & \\   \cline{2-7}  
&  \cite{37_arevalo2017gated}  & {\footnotesize Synopsis, Metadata, Poster} & {\footnotesize Word2Vec, CNN, GMU} & 23 & MM-IMDb & 
  \checkmark \\   \cline{2-7}  
&  \cite{31_1048494}  & Trailer, Audio & {\footnotesize Rule-based classifier} & 4 & ($\wp$) & $\upchi$\\ \cline{2-7}
&  \cite{32_brezeale2006using}  & Subtitle, Video & DCT, BoW, SVM & 18 & ($\wp$) & $\upchi$\\   
\hline
\multicolumn{7}{r}{($\wp$): Publicly unavailable}


  %
  
 \end{tabular}
 \end{adjustbox}
 
 \label{tab:related_work}
\end{table}

\textbf{\em Visual Input:} 
Visual data related to a movie, e.g., poster, teaser, trailer, or clip, can convey cues about the film genre.
Many studies in the literature emphasized trailers, while only a few works have explored the use of movie posters for genre identification.
From a movie trailer, Zhou et al. \cite{20_zhou2010movie} chose keyframes and extracted GIST, CENTRIST, and W-CENTRIST features, followed by a nearest neighbor (kNN)-based classifier to identify the movie genre. 
Sim{\~o}es et al. \cite{23_Simes2016MovieGC} employed a CNN model to detect 4 different genres within a selection of trailers procured from LMTD (Labeled Movie Trailer Dataset).
Wehrmann et al.\cite{24_wehrmann2017movie} also used a CNN leveraging trailer frames across time to detect 9 genres from some trailers of LMTD. 
In \cite{25_montalvo2022trailers12k}, genres were identified from trailer clips using 
dual image and video transformer architecture. 
%
In \cite{21_4107177}, spatio-temporal features were extracted from video clips, followed by using a 
hierarchical SVM. 
Initially, the videos were categorized into broader categories, e.g., movie, news, sports, commercial, and music videos, after which the specific genre was identified.
Yadav et al. \cite{26_yadav2020unified} predicted emotions of facial frames of trailers followed by genre identification using an Inception-LSTM-based architecture.
Pobar et al. \cite{27_pobar2017multi} used 
Naïve Bayes (NB) with GIST descriptor and classeme to predict the genres from movie posters. 
In \cite{28_10.1145/3132515.3132516}, YOLO was used to detect objects on posters, and a CNN model was engaged for corresponding genre identification. 
Turkish movie genres were identified in \cite{29_8990490} from posters using a basic CNN architecture. 
Wi et al. \cite{30_9057706} employed a Gram layer to extract style features and merged with a CNN to classify genres from posters only.


\textbf{\em Textual Input:}
Textual data in movies, including plots/ synopses, subtitles, and user-generated reviews on social media, offer valuable insights into genre identification.    
Ertugrul et al. \cite{9_ertugrul2018movie} employed BLSTM 
to classify movie genres based on sentences extracted from plot summaries. 
In \cite{10_hoang2018predicting}, GRU 
was for a similar input/output. 
%
Kar et al. \cite{11_kar-etal-2018-folksonomication} engaged plot synopses and 
used CNN encoded with the flow of emotions (CNN-FE)  
to predict movie tags, 
i.e., genres and associated plot-related attributes (e.g., violence, suspenseful, melodrama).
Battu et al. \cite{12_battu-etal-2018-predicting} identified genres from multilingual movie synopses and attempted to predict movie ratings using CNN, LSTM, and GRU-based models.
In \cite{13_wehrmann2018self}, CNN with a self-attention mechanism was used for genre classification from textual synopses. 
%
%
Gorinski et al. \cite{17_gorinski-lapata-2018-whats} used a multi-label encoder (MLE) and LSTM-based decoder to predict various movie attributes, including genre, mood, plot, and style from screenplays.


\textbf{\em Multimodal Input:}
In the past, often, two or more modalities (e.g., text,
image, video, audio) were combined and used for the genre
identification.
Arevalo et al. \cite{37_arevalo2017gated} proposed GMU (Gated Multimodal Unit) to fuse features extracted from text (synopsis, metadata) and image (poster) using Word2Vecc and CNN, respectively. 
Bribiesca et al. \cite{38_bribiesca2021multimodal} engaged text (synopsis, metadata), image (poster), video (trailer), audio, and fed to MulT-GMU, which is a transformer architecture with GMU.
Bonilla et al. \cite{35_cascante2019moviescope} proposed a multimodal fusion using fastText, fastVideo, VGG-16, CRNN to fuse text (plot, metadata), video (trailer), image (poster), audio, respectively. 
In \cite{36_mangolin2022multimodal}, various textural features were extracted from the text (synopsis, subtitle), video (trailer), image (poster), audio, and fed to various classifiers, e.g., LSTM, kNN, SVM, MLP, DT (Decision Tree) followed by a fusion step to classify genres. 
Rasheed \cite{31_1048494} et al. computed average shot length, visual disturbance, audio energy from trailer video/audio and used a rule-based classifier to classify into four genres.
In \cite{32_brezeale2006using}, DCT (Discrete Cosine Transform) and BoW (Bag-of-Word) were used to extract features from trailer videos and subtitles, respectively, followed by SVM for detecting movie genres.

\textbf{\em Positioning of our work:} 
In the existing literature, there is a notable scarcity of research focused on genre identification exclusively from poster 
images.
In \cite{27_pobar2017multi}, a Naïve Bayes classifier was employed, while other studies \cite{30_9057706,28_10.1145/3132515.3132516,29_8990490} 
predominantly utilized CNN. 
However, these models underperformed due to ineffectively tackling the intricacies of multi-label genres and challenges associated with posters. 
Additionally, some studies considered only a few genres \cite{29_8990490}.
Our study is one of the earliest attempts to perform genre identification solely from poster images using a transformer-based architecture adept in handling multi-label genres and eliminating extraneous genre labels using a probabilistic module. Moreover, we have conducted a thorough genre-wise analysis and studied model mispredictions that have not been addressed in prior works.



\section{Proposed Methodology}
\label{3sec:method}


\subsection{Problem Formulation}
\label{subsec:prob_form}
\begin{figure}[!t]
    \centering
    \includegraphics[width=0.8\linewidth]{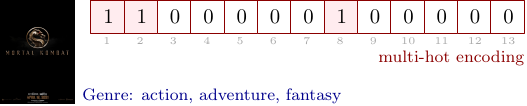}
    \caption{Multi-hot encoding of a movie poster genre}
    \label{fig:multi_hot}
\end{figure}

\noindent 
We are given:
\begin{enumerate}[(i)]
    \item A set of $\delta$ genres ${\cal{G}} = \{{\cal{G}}_1, {\cal{G}}_2, \ldots, {\cal{G}}_\delta\}$
    \item A set of $n$ movie poster images ${\cal{I}} = \{{\cal{I}}_1, {\cal{I}}_2, \ldots, {\cal{I}}_n\}$
    \item Each movie poster image ${\cal{I}}_i \in {\cal{I}}$ is associated with a set of $\kappa_i$ number of genres ${\cal{G}}^{<i>} = \{{\cal{G}}^{<i>}_1, {\cal{G}}^{<i>}_2, \ldots, {\cal{G}}^{<i>}_{\kappa_i}\} \subseteq {\cal{G}}$, $1 \le \kappa_i \le \delta$
\end{enumerate}
\noindent
In this paper, we represent the ground-truth genre in terms of a multi-hot encoding vector (an example is shown in Fig. \ref{fig:multi_hot}). The encoding of ${\cal{I}}_i$ is represented by $\Lambda^{<i>}$ of length $\delta$, as defined below: 
\begin{equation}
\Lambda^{<i>}_j=
\begin{cases}
    1~,  & ~~ \text{if ${\cal{I}}_i$ is associated with ${\cal{G}}_j$} \\
    0~,  & ~~ \text{otherwise}
\end{cases}
\end{equation}
\noindent
Given an unknown movie poster ${\cal{I}}_u$, the objective here is to identify the genres associated with ${\cal{I}}_u$. 
Since a sample can have more than one positive class, we formulate this problem as a \emph{multi-label classification} task \cite{Herrera2016}, and predict multiple genre labels for each movie poster image.

\subsection{Solution Architecture}
\label{subsec:solution_arch}

\noindent
We employ a transformer network for our task due to its capacity for reducing inductive bias and its effectiveness in capturing global dependencies and contextual understanding compared to CNNs.
However, it is important to note that in our approach, we do not directly utilize the ViT (Vision Transformer) paradigm, which involves feeding raw image patches directly into the transformer encoder \cite{dosovitskiy2020image}; instead, we feed deep feature embeddings and perform dense connection among the transformer encoders. 
We now begin with presenting our architecture, Residual Dense Transformer.


\subsubsection{\textbf{Residual Dense Transformer (RDT)}}

\begin{figure}[!t]
\centering
\includegraphics[width=0.98\linewidth]{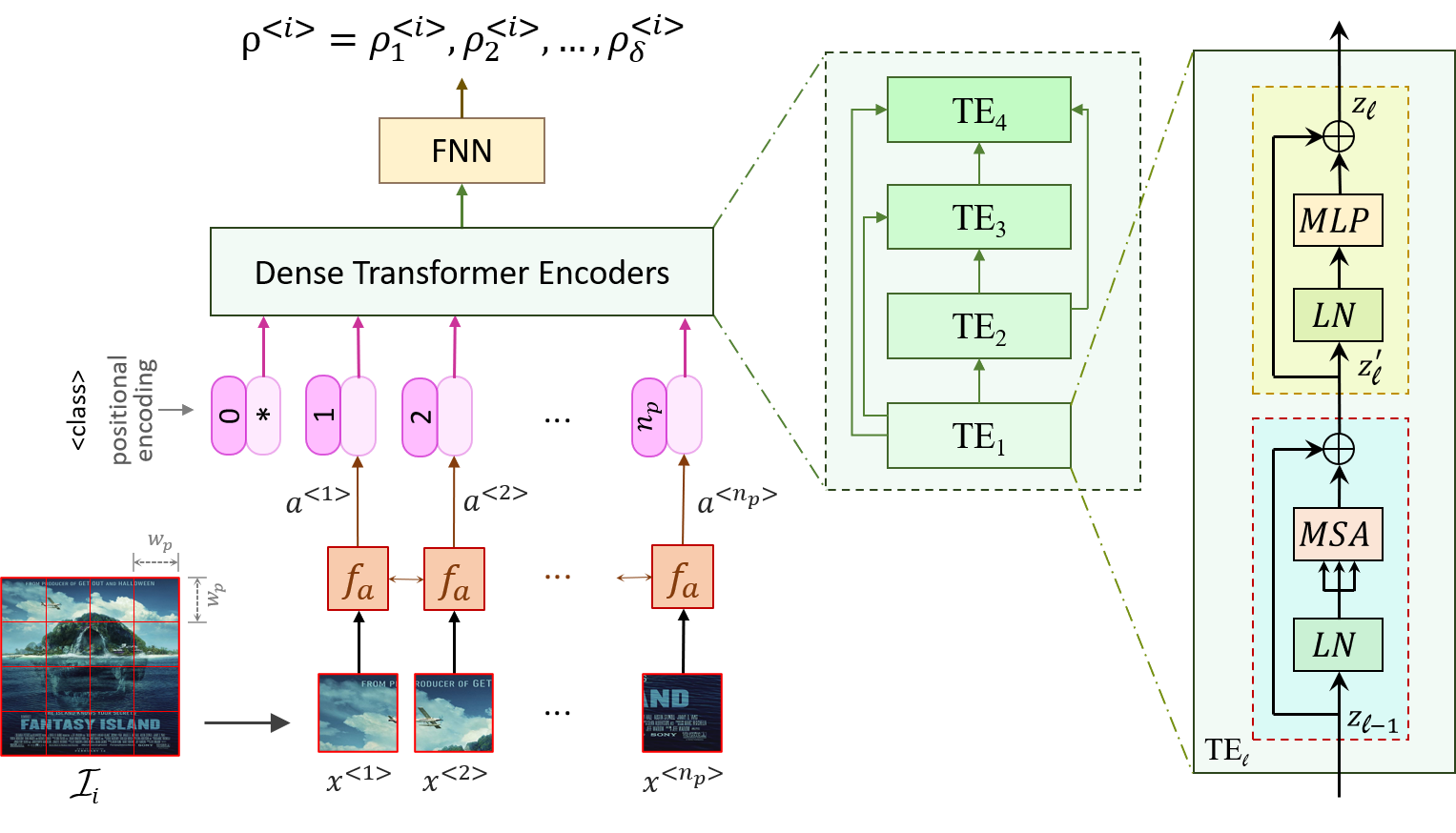}
\caption{Workflow of Residual Dense Transformer (RDT)}
\label{fig:RDT_workflow}
\end{figure}

RDT comprises three main modules: 
deep feature embedding, 
densely connected transformer encoders comprising multi-head self-attention and multi-layer perceptron, 
and a feed-forward neural network \cite{vit_survey_pami}. The pictorial representation of the workflow of RDT is shown in Fig. \ref{fig:RDT_workflow}, and the modules are discussed below.

\textbf{\emph{(i) Deep feature embedding:}}
The transformer network takes input into a sequence of token embedding \cite{vit_survey_pami}. 
Here, an input image is first resized into 
${\mathcal{I}}_i \in \mathbb{R}^{w_z \times w_z \times c_p}$ 
that is converted into a sequence of patches 
$x^{<p>} \in \mathbb{R}^{w_p \times w_p \times c_p}$, 
for $p
=1,2,\ldots,n_p$. 
From each patch $x^{<p>}$, we extract deep features $a^{<p>}$ using a convolutional architecture $f_a$. For our task, up to the average\_pool layer of ResNet50V2 \cite{resnetv2} as $f_a$ works better among some contemporary architectures \cite{DL_Survey_Alom}. The employed $f_a$'s share weights among patches. 
Empirically, we set $w_z = 1024$, $w_p = 256$, $n_p = (w_z/w_p)^2 = 16$, and $c_p=3$ that denotes the RGB channel count of $\mathcal{I}$. 

Further, each $a^{<p>}$ is flattened and mapped into a $D$-dimensional vector, i.e., embedding $z_0$ through transformer layers by the below linear projection.
\begin{equation}
\small
    z_0 = \left[ a_{class}~;~ a^{<1>}\E~;~ a^{<2>}\E~;~ \ldots~;~ a^{<n_p>}\E \right] +\E_{pos}
\end{equation}
where, $\E \in \mathbb{R}^{w_p \times w_p \times c_p \times D}$ is the patch embedding projection,  
$\E_{pos} \in \mathbb{R}^{(n_p + 1) \times D}$ is the positional encoding that holds the patches' position information \cite{attention_need}, and  
$a_{class} = z_0^0$ is a learnable embedding \cite{dosovitskiy2020image}.

\textbf{\emph{(ii) Dense transformer encoders:}} 
After mapping the image patches to the deep feature embedding space with positional encoding, we employ densely connected transformer encoders sequentially \cite{attention_need}. Here, the $\ell^{th}$ transformer encoder (TE$_\ell$) inputs concatenated feature encodings ($\mathcal{X}$) of all preceding encoders: 
\begin{equation}
 \mathcal{X}(\text{TE}_\ell) = \left[ \mathcal{X}(\text{TE}_1); \mathcal{X}(\text{TE}_2); \ldots; \mathcal{X}(\text{TE}_{\ell-1}) \right] 
\end{equation}
The building blocks of a TE are shown in Fig. \ref{fig:RDT_workflow}, which includes alternating layers of $MSA$ (Multi-head Self-Attention) and $MLP$ (Multi-Layer Perceptron) blocks \cite{dosovitskiy2020image,d2l}.

\paragraph{Multi-head Self-Attention ($MSA$)}
The core of the TE is its $MSA$ mechanism consisting of $H$ parallel attention layers, i.e., attention heads, where each head utilizes $SA$ (Scaled dot-product Attention) \cite{attention_need}. 
The $SA$ takes input comprising $D_K$ dimensional queries and keys, and $D_V$ dimensional values \cite{attention_need}, and is computed as follows: 
\begin{equation}
    SA(Q,K,V) = softmax \left( {QK^T}\diagup{\sqrt{D_K}} \right)V
\end{equation}
where, a set of queries, keys, and values are packed to form $Q$, $K$, $V$ matrices, respectively. 
$MSA$ empowers the capability to focus on information across diverse representations at various positions. Here, concurrent self-attention computations for each head collectively output as below: 
\begin{equation}
\centering
\begin{split}
    MSA(Q,K,V) = \left[head_1; head_2; \ldots; head_H\right]W_O;\\
    head_h = SA(QW_h^Q, KW_h^K, VW_h^V)
\end{split}
\end{equation}
where, $W_h^Q \in \mathbb{R}^{D\times D_K}$, 
$W_h^K \in \mathbb{R}^{D\times D_K}$, 
$W_h^V \in \mathbb{R}^{D\times D_V}$, 
{\small $W_O \in \mathbb{R}^{HD_V\times D}$} are parameter matrices; {\small $D_K = D_V = \lfloor D/H \rfloor$}.

\paragraph{Multi-Layer Perceptron ($MLP$)} 
The $MLP$ block consists of two fully connected layers with $2D$ and $D$ nodes, respectively, and employs the GELU (Gaussian Error Linear Unit) non-linear activation function, similar to \cite{dosovitskiy2020image}. 
Before and after the $MSA$ / $MLP$ blocks, $LN$ (Layer Normalization) \cite{ln} and residual connections \cite{resnetv2} are engaged, respectively (Fig. \ref{fig:RDT_workflow}), which can be represented as below: 
\begin{equation}
\begin{split}
    z_\ell = MLP(LN(z'_\ell)) + z'_\ell; \\
    z'_\ell = MSA(LN(z_{\ell-1})) + z_{\ell-1}; ~\ell=1,2,\ldots,L
\end{split}
\end{equation}
where, $L$ is the total count of engaged TEs. 
After multiple TEs, the <class> token is imbued with contextual information. 
The learnable embedding state at the outcome of the TE$_L$, i.e., $z_L^0$, serves as the image representation $y'$ \cite{dosovitskiy2020image}; 
$ y' = LN(z_L^0) $.

\textbf{\emph{(iii) Feed-forward neural network (FNN):}} 
The final stage of our model comprises an FNN consisting of one hidden layer with $D/2$ nodes having ReLU activation function \cite{d2l}, followed by an output layer. 
The output layer contains $\delta$ number of nodes with sigmoid as output function \cite{d2l}. 
To mitigate the challenge in multi-label classification, where positive labels are lesser than negative ones, we leverage the asymmetric loss function (ASL) to train our model \cite{ridnik2021asymmetric}. 
We use Adam optimizer here due to its adaptive learning rates and efficient memory usage \cite{adam2014}.  

Finally, for a poster image ${\cal{I}}_i$, RDT generates a confidence score vector 
${\uprho^{<i>}} = (\rho_1^{<i>}, \rho_2^{<i>}, \ldots, \rho_{\delta}^{<i>})$. 
The top-3 genres based on the confidence scores are selected as the associated genres of ${\cal{I}}_i$.


\subsubsection{\textbf{Ensmebled Residual Dense Transformer (ERDT)}}
\label{subsubsec:proposed_ensemble}
\noindent
Given that data imbalances are a common issue in multi-label classification problems \cite{Herrera2016}, here, we propose an ensemble strategy to mitigate this challenge. In our proposed ensemble method, we consider three fundamental models:
(a) R: the residual network with sigmoid as the output function and ASL as the loss function, (b) RT: the residual transformer network, a simplified version of the RDT that does not include dense connections, (c) RDT: the proposed model. 

For a poster ${\cal{I}}_i$, we first obtain three confidence score vectors 
${\uprho^{<i, 1>}} = (\rho_1^{<i, 1>}, \rho_2^{<i, 1>}, \ldots, \rho_{\delta}^{<i, 1>})$, 
${\uprho^{<i, 2>}} = (\rho_1^{<i, 2>}, \rho_2^{<i, 2>}, \ldots, \rho_{\delta}^{<i, 2>})$, and 
${\uprho^{<i, 3>}} = (\rho_1^{<i, 3>}, \rho_2^{<i, 3>}, \ldots, \rho_{\delta}^{<i, 3>})$ 
from R, RT, and RDT models, respectively, 
which are then combined using \emph{weighted mean} ensemble scheme \cite{ensemble}, as shown in Eq. \ref{eq:weight}, to produce the confidence score vector $\uprho^{<i>} = (\rho_1^{<i>}, \rho_2^{<i>}, \ldots, \rho_{\delta}^{<i>})$ for the ERDT model.
\begin{equation}\label{eq:weight}
\centering
    \rho_j^{<i>} = {\sum_{k = 1}^3 \alpha_k~ \rho_j^{<i, k>}} ; ~~\forall j \in \{1, 2, \ldots, \delta\}
\end{equation}
where, $0 \le \alpha_k \le 1$; 
${\sum_{k = 1}^3 \alpha_k}=1$; $\alpha_k$'s represent the weights that were tuned by a grid-search technique \cite{hpTuning}.


\subsubsection{\textbf{Probabilistic Module}}
\label{subsubsec:proposed_probabilistic}
As discussed earlier, a movie poster can encompass multiple genres. 
Given an input poster ${\cal{I}}_i$, the multi-label classifier generates a confidence score vector $\uprho^{<i>} = (\rho_1^{<i>}, \rho_2^{<i>}, \ldots, \rho_{\delta}^{<i>})$.
The top three genres with the highest confidence score are then predicted as the associated genres with ${\cal{I}}_i$. 
However, the poster can be associated with fewer than three genres. To address this issue, here, we propose a probabilistic module. The objective of this module is to determine whether the poster is associated with more than one genre and, if so, select the 2$^{nd}$ and 3$^{rd}$ genres accordingly.

The crux of this module is to compute the association between genres, which are captured by the following equations: 
\begin{equation}\label{eq:second_genre_association}
    P(g_k | g_j) = 
        {|{\cal{Z}}_j \cap {\cal{Z}}_k|}~/~{|{\cal{Z}}_j|}
\end{equation}
\begin{equation}\label{eq:third_genre_association}
P(g_l | g_j,g_k) =
    {|{\cal{Z}}_j \cap {\cal{Z}}_k \cap {\cal{Z}}_l|}~/~{|{\cal{Z}}_j \cap {\cal{Z}}_k|}
\end{equation}
%
%
where, ${\cal{Z}}_j$ is the set of posters that are associated with genre ${\cal{G}}_j$. Eqn. \ref{eq:second_genre_association} expresses the likelihood of a poster being associated with ${\cal{G}}_k$, considering that the poster is already associated with ${\cal{G}}_j$. 
Eqn. \ref{eq:third_genre_association} denotes the probability of a poster being associated with ${\cal{G}}_l$, given that the poster is already associated with both ${\cal{G}}_j$ and ${\cal{G}}_k$. We calculate the conditional probabilities in advance for all possible combinations of genres.

Once the multi-label classifier generates the confidence score vector $\uprho^{<i>}$ for the input poster ${\cal{I}}_i$, the genre with the highest confidence score is chosen to be the 1$^{st}$ genre of ${\cal{I}}_i$. 
We refer to this 1$^{st}$ genre as the dominant genre.

The 2$^{nd}$ genre of ${\cal{I}}_i$ is determined based on the dominant genre. Let us assume that ${\cal{G}}_j$ is the dominant genre for ${\cal{I}}_i$ (i.e., ${\cal{G}}^{<i>}_1 = {\cal{G}}_j$). The 2$^{nd}$ genre for ${\cal{I}}_i$ is selected by Eqn. \ref{eq:2_argmax}. 
\begin{equation} \label{eq:2_argmax}
 \begin{split}
     {\cal{G}}^{<i>}_2 = \underset{k \ne j,~ 1 \le k \le \delta}\argmax \left(\rho_k^{<i>} \times {\widetilde{P}}(g_k|g_j) \right);  \\
     \text{if }  \underset{k \ne j,~ 1 \le k \le \delta}\max \left(\rho_k^{<i>} \times {\widetilde{P}}(g_k|g_j) \right) > \tau
 \end{split}
\end{equation}
where, $\widetilde{P}(g_k|g_j)$ is the normalized probability value computed over $\left( P(g_1|g_j), P(g_2|g_j), \ldots, P(g_{\delta}|g_j) \right)$, and $\tau$ is a tunable threshold, determined empirically. 
Here, for each genre other than ${\cal{G}}_j$, an association score is computed. The association score for genre ${\cal{G}}_k$ is determined by multiplying its confidence score $\rho_k^{<i>}$ with its normalized conditional probability value ${\widetilde{P}}(g_k|g_j)$, provided that ${\cal{I}}_i$ is linked to ${\cal{G}}_j$. 
If the maximum association score across all genres except ${\cal{G}}_j$ exceeds $\tau$, the corresponding genre is assigned in ${\cal{G}}^{<i>}_2$. 

If the 2$^{nd}$ genre of ${\cal{I}}_i$ (assuming ${\cal{G}}^{<i>}_2 = {\cal{G}}_k$) is chosen, we then proceed to determine whether to select the 3$^{rd}$ genre. 
The selection of the 3$^{rd}$ genre of ${\cal{I}}_i$ is captured by Eqn. \ref{eq:3_argmax}. 
\begin{equation} \label{eq:3_argmax}
\small
 \begin{split}
     {\cal{G}}^{<i>}_3 = \underset{l \ne j, ~l \ne k,~ 1 \le l \le \delta}\argmax \left(\rho_l^{<i>} \times {\widetilde{P}}(g_l|g_j) \times {\widetilde{P}}(g_l|g_j, g_k) \right);  \\
     \text{if }  \underset{l \ne j, ~l \ne k,~ 1 \le l \le \delta}\max \left(\rho_l^{<i>} \times {\widetilde{P}}(g_l|g_j) \times {\widetilde{P}}(g_l|g_j, g_k) \right) > \tau'
 \end{split}
\end{equation}
where, $\widetilde{P}(g_l|g_j,g_k)$ represents normalized probability calculated from $\left( P(g_1|g_j,g_k), P(g_2|g_j,g_k), \ldots, P(g_{\delta}|g_j,g_k) \right)$, and $\tau'$ is a tunable threshold, set empirically.
It is important to emphasize that selecting the 2$^{nd}$ and 3$^{rd}$ genres is greatly influenced by the correct prediction of the dominant genre. 
Therefore, an incorrect prediction for the dominant genre may lead to the inaccurate selection of the 2$^{nd}$ and 3$^{rd}$ genres. 

In our experimental analysis, we show the performance of this probabilistic module with respect to the correct prediction of the dominant genre, 
which is denoted by \emph{hit ratio} (${\cal{H}}it$), i.e.,  
${\cal{H}}it = {|DB_t^c|}~\diagup~{|DB_t|}$; 
where, $DB_t^c$ is the set of test samples for which the dominant genre is correctly identified, and $|DB_t|$ is the total number of employed test samples.

Finally, we integrate the probabilistic module with ERDT to obtain our final proposed model, PrERDT.

\section{Experiments and Discussion}
\label{4sec:result}

\subsection{Employed Dataset} \label{subsec:DB}

\begin{table}[!b]
\centering   
\footnotesize
\caption{Poster and movie counts across genre labels}\label{tab1}
\begin{tabular}{c|c|c|c}
\hline
Class Id & Genre label & Poster count & Movie count\\
\hline 
1 & Action &  4985 & 1426\\
2 & Adventure & 3702 & 1024\\
3 & Animation &  1196 & 325\\
4 & Biography &  1076 & 348\\
5 & Comedy & 4380 & 1517\\
6 & Crime &  3052 & 1003\\
7 & Drama & 6609 & 2217\\
8 & Fantasy &  1379 & 423\\
9 & Horror & 2646 & 860\\
10 & Mystery &  2285 & 750\\
11 & Romance &  2406 & 913\\
12 & Sci-Fi &  1542 & 458\\
13 & Thriller &  3455 & 1092\\
\hline
\end{tabular}
\label{tab:class_no}
\end{table}

\begin{figure}[!t]
\centering
\scriptsize
(a) \includegraphics[width=0.45\linewidth]{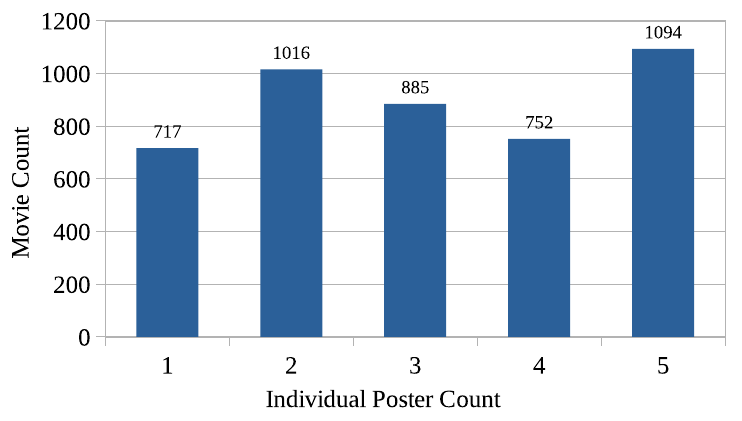} (b)\includegraphics[width=0.45\linewidth]{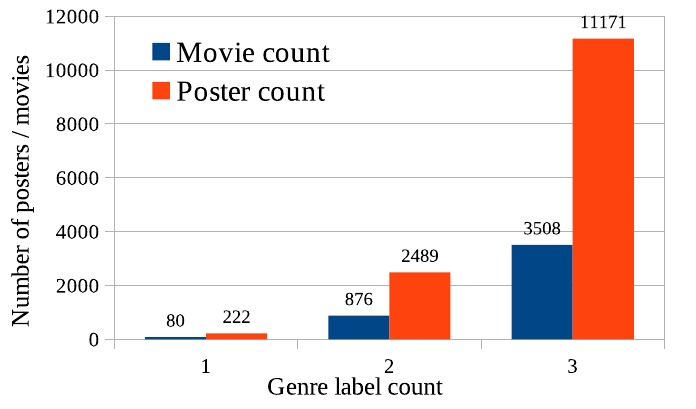}
\caption{\small (a) Distribution of movie count w.r.t. individual poster count, 
(b) Distribution of movie \& poster counts w.r.t. genre-label count}
\label{fig:movie_poster_count}
\end{figure}

\noindent
The primary objective of this study is to analyze the poster images to identify their multi-labeled movie genres.
Consequently, obtaining a dataset featuring posters with multiple genres proved challenging, as there were scarce off-the-shelf options available. 
Therefore, we procured authentic movie poster images with corresponding genres from IMDb (\href{https://developer.imdb.com/non-commercial-datasets}{\emph{https://developer.imdb.com/non-commercial-datasets}}). 
A movie may be of multiple genres; however, on IMDb, a maximum of 3 genres are labeled for an individual movie. 
Currently, our dataset considers 13 distinct genres (i.e., $\delta =$ 13), as mentioned in Table \ref{tab:class_no}. 
The ground-truth genre of a movie poster is available in terms of multi-hot encoding (refer to Section \ref{subsec:prob_form}). 
We gathered posters of 4464 individual movies, each with 1 to 5 posters; the distribution of movie count with respect to the individual poster count is presented in Fig. \ref{fig:movie_poster_count}: (a). 
Overall, our dataset comprises 13882 distinct posters, each having 1 to 3 genre labels.
The movie and poster counts with respect to genre label count are shown in Fig. \ref{fig:movie_poster_count}: (b). Here, we can see that about $\frac{3}{4}^{th}$ posters/movies of our dataset have 3 genre labels. 
For individual genre label/ class id, the corresponding poster and movie counts are shown in Table \ref{tab:class_no}. 
As a matter of fact, in this table, some poster/movie counts overlap across genres due to having multi-label genres.
Here, genre \emph{drama} is included in the highest number of posters, i.e., 6609; whereas \emph{biography} has 1076 posters, which is the lowest in our dataset.
Table \ref{tab:class_no} comprehends the data imbalance issue \cite{Johnson2019SurveyOD}. 
In Fig. B.1 (refer to Supplementary Appendix B), we present some poster images from our employed dataset along with the genre class numbers. 
Fig. A.1 of Supplementary Appendix A illustrates a co-occurrence matrix for movie genre labels associated with the posters \cite{30_9057706}.

From IMDb, while selecting the movies/poster, we followed the following strategies: 
\begin{itemize}[---]
     \item sorted out movies based on release year since 2000.
     \item picked movies having more than 10000 user votes and more than 60 minutes of runtime.
     \item crawled and filtered movies based on the 13 genres employed here (refer to Table \ref{tab:class_no}).
\end{itemize}

The dataset was split up into training ($DB_{tr}$), validation ($DB_v$), and testing ($DB_t$) disjoint sets with an approx ratio of $8:1:1$ considering the presence of all 13 genres in each set equivalently.
$DB_{tr}$, $DB_v$, and $DB_t$ contain 10942, 1470, and 1470 posters, respectively. 


\subsection{Experimental Details}
\label{subsec:Exp_details}

\noindent
We executed the experimentation using the TensorFlow-2.5 framework having Python 3.9.13 on an Ubuntu 20.04.2 LTS-based machine with specifications including AMD EPYC 7552 Processor running at 2.20 GHz with 48 CPU cores and 256 GB RAM, NVIDIA A100-PCIE GPU with 40 GB of memory.
In this paper, all the presented results were obtained from $DB_t$.

The hyper-parameters of our model were tuned and set during the model training with a focus on optimizing performance over $DB_v$. 
For the transformer networks, we fixed the following hyper-parameters empirically: 
transformer\_layers ($L$) = 4, 
embedding\_dimension ($D$) = 256, 
num\_heads ($H$) = 6. 
In ASL, we set focusing parameters $\gamma^{+} = 0$, $\gamma^{-}=1$, and probability margin $m=0.2$. 
For Adam optimizer, we chose 
initial\_learning\_rate = $10^{-3}$; 
exponential decay rates for 1$^{st}$ and 2$^{nd}$ moment estimates, i.e., $\beta1=0.9$, $\beta2=0.999$; 
zero-denominator removal parameter ($\upvarepsilon$) = $10^{-8}$. 
For the early stopping strategy, we set the patience parameter to 10 epochs, and we maintained a fixed mini-batch size of 32. 
We empirically chose $\tau = 0.3$ and  $\tau' = 0.03$ for our probabilistic module (refer to Eqn.s \ref{eq:2_argmax}, \ref{eq:3_argmax}).

We evaluated the model performance based on the \emph{macro}-level analysis, considering standard metrics in multi-label classification, such as 
precision ($\cal{P}$) \%, 
recall ($\cal{R}$) \%, 
specificity (${\cal{S}}p$) \%, 
balanced accuracy (${\cal{BA}}$) \%, 
F-measure (${\cal{FM}}$) \%, and 
Hamming loss (${\cal{HL}}$) \cite{eval_method2}.

\subsection{Comparison with State-of-the-Art (SOTA) Models}

\begin{table}[]
\centering
\caption{\small{{Comparison with baseline and SOTA models}}}
\begin{adjustbox}{width=0.48\textwidth} 
\begin{tabular}{c|l|c|c|c|c|c|c}
\cline{2-8}
{}  & Method & $\cal{P}$~\textuparrow & $\cal{R}$~\textuparrow & ${\cal{S}}p$~\textuparrow & ${\cal{BA}}$~\textuparrow & ${\cal{FM}}$~\textuparrow   & ${\cal{HL}}$~\textdownarrow \\ \hline
{\multirow{6}{*}{Baseline}} 
& ResNet50V2 \cite{resnetv2}                   & \textbf{52.04} & 50.44& 85.36  & 67.90 & 48.97 & 0.18524   \\ 
& DenseNet121 \cite{8099726}                   & 26.10 & 27.18& 78.62  & 52.90 & 22.63 & 0.30120 \\  
& EfficientNetB2 \cite{tan2019efficientnet}    & 51.37 & \textbf{\underline{52.53}}   & \textbf{\underline{85.95}}  & \textbf{\underline{69.24}} & \textbf{\underline{51.40}} & \textbf{\underline{0.18503}}   \\  
& ViT \cite{dosovitskiy2020image}              & 29.11 & 27.40& 78.61  & 53.01 & 22.58 & 0.25243   \\  
& InceptionV3 \cite{inceptionv3}               & 21.03 & 28.80& 79.12  & 53.96 & 22.43 & 0.25599   \\  
& MobileNetV2 \cite{mobilenetv2}               & 37.34 & 33.22& 80.57  & 56.90 & 30.37 & 0.23799   \\ \hline
{\multirow{4}{*}{SOTA}} 
& Chu et al. \cite{28_10.1145/3132515.3132516}  & 19.73 & 27.32& -  & - & 20.89 & -   \\  
& Gozuacik et al. \cite{29_8990490}             & 36.76 & 35.12& \textbf{80.91}  & \textbf{58.02} & 33.49 & \textbf{0.24290} \\ 
& Pobar et al. \cite{27_pobar2017multi}         & 28.76 & 47.76& 68.18  & 57.97 & 34.72 & 0.34688   \\ 
&  Wi et al. \cite{30_9057706}                  & \textbf{\underline{52.89}} & \textbf{51.18}   & \textbf{-}  & - & \textbf{49.61} & -   \\ \hline
{\multirow{3}{*}{Proposed}} & RDT   & 55.01 &	57.25 &	87.08 &	72.16 &	55.69 &	0.17069   \\  
& ERDT                      & 55.95 & \textbf{57.88} & 87.35 & \textbf{72.61} & \textbf{56.40} & 0.16546   \\ 
& PrERDT                    & \textbf{57.77} & 54.65 & \textbf{88.71} & 71.68 & 55.26 & \textbf{0.16216}   \\ \hline \hline
\rowcolor[HTML]{F3F3F3} 
& {\cellcolor[HTML]{F3F3F3}RDT}  & 2.12 &	4.72 &	1.13 &	2.93 &	4.29 &	7.75\% \\ 
\rowcolor[HTML]{F3F3F3} 
{\small{Improvement}}$^\ast$ & {\cellcolor[HTML]{F3F3F3}ERDT}  & 3.06 & 5.35 & 1.40 & 3.38  & 5.00 & 10.58\% \\ 
\rowcolor[HTML]{F3F3F3} 
& {\cellcolor[HTML]{F3F3F3}PrERDT} & 4.88 & 2.12 & 2.76 & 2.44  & 3.86 & 12.36\% \\ \hline  
\multicolumn{8}{r}{$^\ast$Improvement of our models over the \textbf{\underline{second-best}} baseline/ SOTA} 
\end{tabular}
\end{adjustbox}
\label{tab:comparison}
\end{table}

\noindent
Table \ref{tab:comparison} presents the performance of the three models: RDT, ERDT and PrERDT, proposed in this paper. Here, we compare the performance of our proposed models with some major contemporary deep architectures, called baseline, and some related SOTA models. 
It may be noted the actual baseline models are designed for multi-class classification problems \cite{DL_Survey_Alom}, and are typically not well-suited for handling multi-label classification challenges. 
Therefore, we improvised the baseline models by incorporating sigmoid as the output function and ASL as the loss function (refer to Section \ref{subsec:solution_arch}) to enable them to address multi-label classification. 
As evident from Table \ref{tab:comparison}, all three models, RDT, ERDT and PrERDT, outperformed the baseline and SOTA models in terms of all the performance evaluation metrics employed in this paper.

\subsection{Ensemble Study}
\label{subsec:ensemble}

\begin{table}
\centering
\caption{\small Performance by various models for ensemble analysis}
\begin{adjustbox}{width=0.40\textwidth} 
\begin{tabular}{l|c|c|c|c|c|c}
\hline
\multirow{1}{*}{Model}& $\cal{P}$~\textuparrow & $\cal{R}$~\textuparrow & ${\cal{S}}p$~\textuparrow  & ${\cal{BA}}$~\textuparrow & ${\cal{FM}}$~\textuparrow & ${\cal{HL}}$~\textdownarrow  \\ 
\hline
R & {52.04}  & {50.44}  & {85.36}  & {67.90}  & {48.97}  & 0.18524  \\ 
RT  & {52.54}  & {55.53}  & {86.67}  & {71.10}  & {53.57}  & 0.17833  \\ 
RDT & {55.01}  & {57.25}  & {87.08}  & {72.16}  & {55.69}  & 0.17069  \\ 
R + RT & {54.76}  & {56.28}  & {87.04}  & {71.66}  & {54.69}  & 0.16902   \\ 
R + RDT  & {55.35}  & {56.46}  & {86.97}  & {71.72}  & {55.20}  & 0.16954  \\ 
RT + RDT & {54.96}  & {57.44}  & {87.28}  & {72.36}  & {55.70}  & 0.16755  \\ 
ERDT  & {\textbf{55.95}}  & {\textbf{57.88}}  & {\textbf{87.35}}  & \textbf{72.61}  & {\textbf{56.40}}  & \textbf{0.16546}  \\ \hline
\multicolumn{7}{l}{R: ResNet50V2, RT: Residual Transformer network,}\\
\multicolumn{7}{l}{RDT: Residual Dense Transformer, R + RT: Ensembling R and RT,}\\
\multicolumn{7}{l}{ERDT (R + RT + RDT): Ensmebled Residual Dense Transformer}
\end{tabular}
\end{adjustbox}
\label{tab:ensemble}
\end{table}

\noindent
Table \ref{tab:ensemble} presents the performance of various models (excluding the probabilistic module) participating in the ensemble. From this table, we have the following observations:

\emph{(i) Overall performance of ERDT:} ERDT outperformed all other models listed in the first column of Table \ref{tab:ensemble}. 

\emph{(ii) Comparison between RDT and R+RT:} 
It is worth noting that the overall performance of the RDT was better than the R+RT model in terms of ${\cal{BA}}$ and ${\cal{FM}}$.

\emph{(iii) Comparison between RDT and R+RDT:}
As evident from Table \ref{tab:ensemble}, the performance of R+RDT is degraded compared to that of RDT in terms of ${\cal{BA}}$ and ${\cal{FM}}$. 
From Table \ref{tab:genre_big}: (b), the reason for this can be explained. Table \ref{tab:genre_big}: (b) shows the rank of each model based on the balanced accuracy for all 13 genres separately. According to these ranks, RDT performed better than R in all 13 genres. 
The negative influence of R caused the degradation of the performance of R+RDT model in 8 out of 13 cases, which was reflected further in the overall performance of R+RDT model across all genres.

\emph{(iv) Significance of participating models in the ensemble:}
As per the ranking shown in Table \ref{tab:genre_big}: (b), RT performed better than RDT for two genres (i.e., \emph{mystery}, and \emph{sci-fi}). Hence, we first choose to combine RT and RDT to improve the overall performance of RDT. Our experimental results show that in 6 of 13 genres, RT+RDT, indeed, performed better than RT and RDT individually. RT+RDT also outperformed RT and RDT independently in terms of overall performance across all genres.
Furthermore, as per our observation from Table \ref{tab:genre_big}: (b), for a few genres (e.g., \emph{adventure}, \emph{animation}, \emph{thriller}), the performances of R, RT, and RDT are comparable. Hence, we select R, RT, and RDT as the fundamental models for our ensemble. It may be noted from Table \ref{tab:genre_big}: (b) that in 6 out of 13 genres, ERDT performed better than RT+RDT. Moreover, in 5 out of the 8 cases for which R+RDT performed worse than RDT, the ERDT model performed better than RDT due to the influence of RT. In terms of the overall performance, ERDT also turned out to be the best, comparing all the fundamental models used for our ensemble and their other possible combinations. This justifies our choice of fundamental models for the ensemble. 
In Fig. C.1 of Supplementary Appendix C, we show ERDT's qualitative performances using heat-map encodings.

\subsection{Genre-wise Analysis}

\begin{figure}
    \centering
    \includegraphics[width=0.8\linewidth]{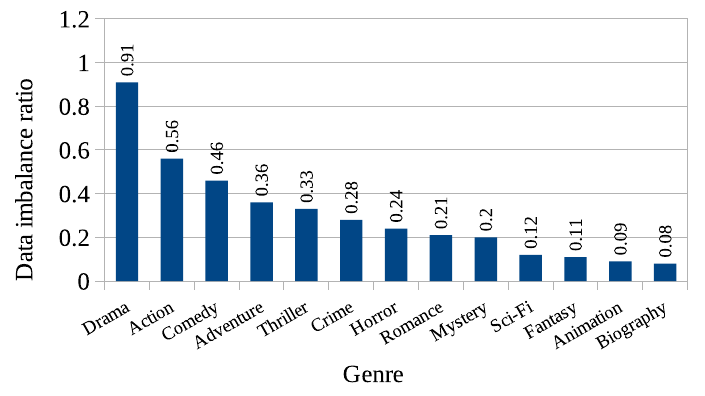}
    \caption{Genre-wise data imbalance ratio}
    \label{fig:dataImbalance}
\end{figure}


\noindent
According to Table \ref{tab:ensemble}, we have evaluated the models based on their ${\cal{BA}}$ (and ${\cal{FM}}$), resulting in the following ranking: ERDT $\succ$ RT+RDT $\succ$ RDT $\succ$ R+RDT $\succ$ R+RT $\succ$ RT $\succ$ R;  
where, the notation Model-1 $\succ$ Model-2 indicates that the ${\cal{BA}}$ of Model-1 surpasses that of Model-2.
Here, we perform a detailed analysis of the models' performance across different genres, focusing on ${\cal{BA}}$. Table \ref{tab:genre_big}: (a) provides a breakdown of the genre-wise performance analysis for all the models. From Tables \ref{tab:genre_big}: (a), (b), we yield the following findings:

{\emph{(i) Comparison among fundamental models:}} 
As evident from Table \ref{tab:genre_big}: (b), RT demonstrated superior performance when compared to R in 12 out of 13 genres, highlighting its improvement over the latter. Similarly, RDT, being an enhancement over RT, outperformed RT in 11 out of 13 genres.

{\emph{(ii) Comparison between the proposed fundamental model with other ensemble models:}} 
RDT outperformed R+RT in 7 out of 13 genres. We observed performance improvements for the R+RDT and RT+RDT models over the RDT model in 5 and 8 of the 13 genres, respectively (refer to Table \ref{tab:genre_big}: (b)). 
The ERDT model performed better than the RT+RDT model in 6 of 13 genres.
Interestingly, despite the RT+RDT model outperforming ERDT in more genres when considering the count, the quantitative measure of improvement for ERDT was significantly higher than the degradation observed for ERDT in all the genres where RT+RDT surpassed the ERDT model.

\begin{figure}
    \centering
    \scriptsize 
    (a) \includegraphics[width=0.45\linewidth]{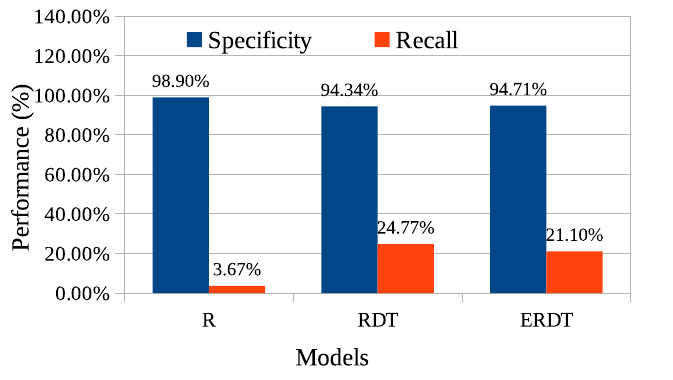}
    (b) \includegraphics[width=0.45\linewidth]{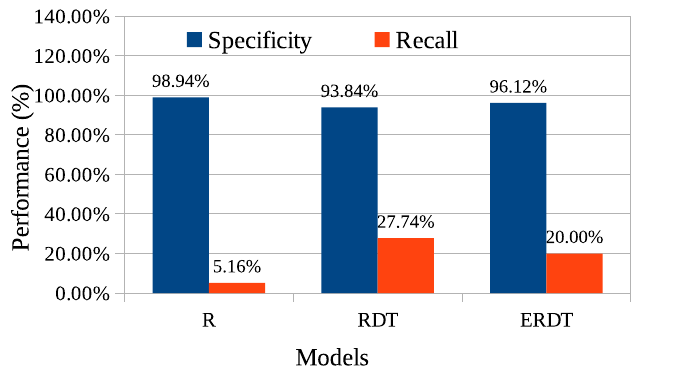}
    \caption{Comparative performance of R, RDT, and ERDT for imbalanced genres: (a) {Biography}, (b) {Fantasy}}
    \label{fig:imbalance_fb}
\end{figure}

\emph{(iii) Performance of R, RDT, and ERDT on imbalanced genres:} 
Fig. \ref{fig:dataImbalance} represents the ratio between positive and negative samples for each genre, denoted as \emph{imbalance ratio}. This figure clearly illustrates that certain genres, such as \emph{biography}, \emph{animation}, and \emph{fantasy}, suffer from significant data imbalance issues. Here, our analysis focuses on evaluating the performance of R, RDT, and ERDT specifically for these imbalanced genres. Fig. \ref{fig:imbalance_fb} visually depicts the comparative performance of these models in terms of specificity and recall for the \emph{biography} and \emph{fantasy} genres, both of which exhibit high levels of data imbalance.
As observed in Fig.s \ref{fig:imbalance_fb}: (a), (b), the specificity of R is notably high, while its recall is considerably low in these imbalanced genres. 
This indicates that R struggles to address the challenge posed by imbalanced data effectively since R is unable to identify the posters belonging to these genres.
In contrast, the RDT model performed better by identifying more posters from the imbalanced genres compared to the R model, resulting in improved recall. 
However, this gain in recall came at the cost of reduced specificity. 
The trade-off between recall and specificity is observed for the ERDT model.
In other words, ERDT exhibits higher specificity compared to RDT but lower than R. Additionally, ERDT demonstrates higher recall compared to R but lower than RDT.
As a result, the ${\cal{BA}}$ of the ERDT model surpasses that of both the R and RDT models.

\subsection{Ablation Study}

\begin{table}[!b]
\centering
\caption{\small Ablation Study}
 \begin{adjustbox}{width=0.45\textwidth} 
\begin{tabular}{l|c|c|c |c|c|c|c}
\hline 
Model & $\cal{P}$~\textuparrow & $\cal{R}$~\textuparrow & ${\cal{S}}p$~\textuparrow &  ${\cal{BA}}$~\textuparrow & ${\cal{FM}}$~\textuparrow & ${\cal{HL}}$~\textdownarrow & {\st{$\ell$}}~\textdownarrow \\
\hline 
R        & 52.04 & 50.44 & 85.36   &  67.90 & 48.97 & 0.18524 & {\textbf{1.616}$\times 10^{-2}$}\\
T        & 29.11 & 27.40 & 78.61   &  53.01 & 22.58 & 0.25243 & {1.697$\times 10^{-2}$}\\
RT       & 52.54 & 55.53 & 86.67    & 71.10 & 53.57 & 0.17833 & {1.748$\times 10^{-2}$}\\
DT       & 33.78 & 29.81 & 79.14    & 54.48 & 25.42 & 0.26509 & {1.730$\times 10^{-2}$}\\
RDT      & 55.01 & 57.25 & 87.08    & 72.16 & 55.69 & 0.17069 & {1.801$\times 10^{-2}$}\\
R + RDT  & 55.35 & 56.46 & 86.97    & 71.72 & 55.20 & 0.16954 & {1.821$\times 10^{-2}$}\\
RT + RDT & 54.96 & 57.44 & 87.28    & 72.36 & 55.70 & 0.16755 & {1.846$\times 10^{-2}$}\\
ERDT     & 55.95 & \textbf{57.88}  & 87.35    & \textbf{72.61}    & \textbf{56.40}    & 0.16546 & {1.864$\times 10^{-2}$}\\
PrERDT   & \textbf{57.77}     & 54.65 & \textbf{88.71}           & 71.68& 55.26 & \textbf{0.16216} & {1.871$\times 10^{-2}$}\\
\hline 
\multicolumn{8}{r}{{\st{$\ell$}: Average inference time in second}}
\end{tabular}
\end{adjustbox}
\label{tab:ablation}
\end{table}


\noindent
Table \ref{tab:ablation} shows the ablation study of our proposed architecture. The observation from this table is listed below:

{\emph{(i) Ablation study for RDT:}} As discussed earlier, RDT is the composition of a residual network and dense transformer. Therefore, here, we first compare the performance of RDT with other component models, such as the Residual network (R), Transformer network (T), Dense Transformer (DT), and Residual Transformer network (RT). As evident from Table \ref{tab:ablation},  RDT outperformed R, T, RT, and DT. This comprehends the impact of our proposed fundamental model RDT. 

{\emph{(ii) Incremental performance improvement due to improvisation of the models:}} It is worth noting from Table \ref{tab:ablation} that as RT is an improvisation over R and T individually, the performance of RT is better than each of R and T. Similarly, as DT is an improvisation over T, the performance of DT is better than T. Finally, RDT is an improvisation over RT and DT models; therefore, the performance of RDT is better than them.

{\emph{(iii) Ablation study for ERDT:}} Table \ref{tab:ablation} shows that our ensemble model ERDT performed better than any other combination of the fundamental models (i.e., R, RT, and RDT) used in the proposed ensemble. The analysis for PrERDT is presented in the next subsection. 

\emph{(iv) Computational performance analysis:}
The last column of Table \ref{tab:ablation} presents the inference time of PrERDT and its various ablated modules. This table indicates that our proposed RDT framework is $1.04\times$ faster than PrERDT. It is also noteworthy that the proposed ensemble architecture (i.e., PrERDT) significantly enhances prediction performance while only marginally increasing inference time by approximately $7 \times 10^{-4}$ second compared to RDT. 

\subsection{Analysis of Probabilistic Module}
\label{subsec:res_Pr}

\begin{figure}[!b]
    \centering
    \includegraphics[width=0.8\linewidth]{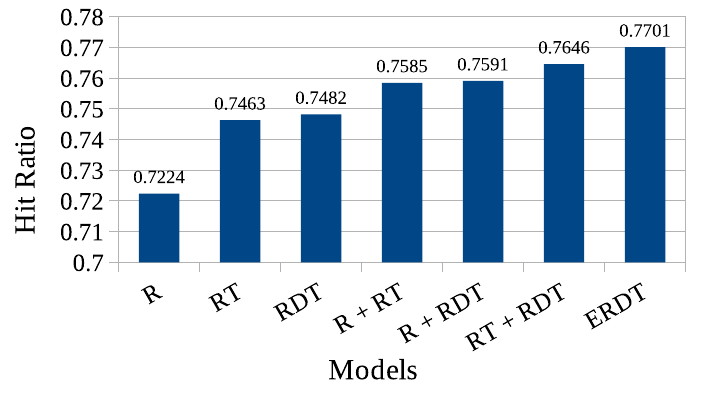}
    \caption{Performance analysis on hit ratio $({\cal{H}}it)$}
    \label{fig:hitRatio}
\end{figure}

\noindent
%
The last row of Table \ref{tab:ablation} shows the performance of PrERDT, where we used the probabilistic module on ERDT. 
In PrERDT, the recall decreased more than the improvement in the precision and specificity (refer to the last two rows of Table \ref{tab:ablation}). 
Consequently, the ${\cal{BA}}$ and the ${\cal{FM}}$ were decreased for PrERDT. 
However, the motivation for introducing the probabilistic module is to enhance precision while making only minimal concessions in recall, ultimately leading to improved ${\cal{BA}}$ and ${\cal{FM}}$.
The reason behind obtaining the counterintuitive outcome can be elucidated by referring to Fig. \ref{fig:hitRatio}. 
The performance of our probabilistic module highly relies on accurately identifying the first genre through ERDT. 
However, according to Fig. \ref{fig:hitRatio}, the hit ratio (refer to Section \ref{subsubsec:proposed_probabilistic}) of the ERDT model for identifying the first genre is 0.7701. 
Consequently, the PrERDT model sometimes discarded the correctly predicted second and third genres due to its dependency on the erroneously predicted first genre, thus causing a decline in recall.

To validate the correctness of our hypothesis, we conducted an additional experiment on different subsets of test data where the hit ratio is notably high. 
Table \ref{tab:prob_analysis} shows the performance of our probabilistic module for five different subsets of test data characterized by a high hit ratio.  
As observed from Table \ref{tab:prob_analysis}, the precision for the PrERDT model exhibited improvement when compared to the ERDT without compromising the recall value. Consequently, this enhancement translated into improved ${\cal{BA}}$ and ${\cal{FM}}$ for the PrERDT model. 
These findings underscore the effectiveness of our probabilistic module.
The analysis of performance stagnation in PrERDT, attributed to incomprehensible posters and indistinguishable genres, as well as its limitations, is discussed in 
Supplementary Appendix D.

\begin{table}[]
\caption{\small Significance of the probabilistic module}
\begin{adjustbox}{width=0.47\textwidth} 
\begin{tabular}{c|c|c|c|c|c|c|c}
\hline 
\multicolumn{1}{l|}{Model} & ${\cal{H}}it$ & $\cal{P}$ & $\cal{R}$ & ${\cal{S}}p$ &  ${\cal{BA}}$ & ${\cal{FM}}$ & ${\cal{HL}}$ \\ \hline
 
{ERDT} & \multirow{2}{*}{0.8889} & {79.48}  & {91.78}   & {95.60} & {93.69} & {84.27} & 0.04102   \\ 
{PrERDT}   & & {80.17}  & {91.78}   & {95.70} & {93.74} & {84.64} & 0.04017   \\ \hline
 
{ERDT} & \multirow{2}{*}{0.9000} & {87.25} & {96.44}  & {95.84} & {96.14} & {91.08} & 0.03760 \\ 
{PrERDT}   && {87.74} & {96.44}  & {95.94} & {96.19} & {91.38} & 0.03675   \\ \hline

{ERDT} & \multirow{2}{*}{0.9111} & {87.81} & {87.28}  & {95.86} & {91.57} & {85.72} & 0.03760 \\  
{PrERDT}   &  & {87.94} & {87.28}  & {96.05} & {91.66} & {85.79} & 0.03675   \\ \hline
 
{ERDT} & \multirow{2}{*}{0.9133} &  {84.70} & {92.51}  & {95.73} & {94.12} & {88.10} & 0.03880  \\  
{PrERDT}   & & {84.94} & {92.51}  & {95.77} & {94.14} & {88.23} & 0.03846   \\ \hline

 
{ERDT} & \multirow{2}{*}{0.9444} & {83.49} & {92.82}  & {94.46} & {93.64} & {86.21} & 0.04444   \\ 
{PrERDT}   && {84.78} & {92.82}  & {94.54} & {93.68} & {87.24} & 0.04358   \\ \hline
\end{tabular}
\end{adjustbox}
\label{tab:prob_analysis}
\end{table}

\subsection{Performance Analysis based on Genre Label Count}
\label{subsec:genre_label_count}
\noindent

\begin{table}[]
\centering
\caption{\small Performance study (${\cal{BA}}$ \%)  on genre label count}
\begin{adjustbox}{width=0.47\textwidth}
\begin{tabular}{l|c|c|c|c|c|c}
\cline{2-7} 
& \multicolumn{2}{c|}{} & \multicolumn{2}{c|}{} & \multicolumn{2}{c}{}  \\[\dimexpr-\normalbaselineskip+1.5pt]
        & \multicolumn{2}{c|}{$DB_t^{<1>}$}       & \multicolumn{2}{c|}{$DB_t^{<2>}$}       & \multicolumn{2}{c}{$DB_t^{<3>}$}       \\ \hline 
\cellcolor[HTML]{FFFFFF} Model        & {\cellcolor[HTML]{EFEFEF}w/o Pr} & \cellcolor[HTML]{EFEFEF}Pr & {\cellcolor[HTML]{EFEFEF}w/o Pr} & \cellcolor[HTML]{EFEFEF}Pr & {\cellcolor[HTML]{EFEFEF}w/o Pr} & \cellcolor[HTML]{EFEFEF}Pr \\ \hline
{R}        & {50.68} & 50.70 & {63.88} & 62.70 & {67.87} & 66.79 \\ \hline
{RT}       & {48.67} & 49.09 & {68.08} & 68.29 & {70.93} & 70.13 \\ \hline
{RDT}      & {50.73} & 51.31 & {69.57} & 66.61 & {72.07} & 70.86 \\ \hline
{R + RT}   & {49.45} & 49.08 & {67.90} & 66.95 & {71.60} & 70.46 \\ \hline
{R + RDT}  & {50.70} & \textbf{51.62} & {68.24} & 65.05 & {71.28} & 70.14 \\ \hline
{RT + RDT} & {49.87} & 49.73 & {69.20} & 68.61 & {72.20} & 71.37 \\ \hline
{ERDT}     & \textbf{51.58} & 49.36 & \textbf{70.26} & \textbf{69.69} & \textbf{72.34} & \textbf{71.41} \\ \hline
\multicolumn{7}{r}{Pr : with probabilistic, and w/o Pr : without probabilistic module}
\end{tabular}
\end{adjustbox}
\label{tab:genre_123}
\end{table}

\noindent
As mentioned earlier in Section \ref{subsec:DB} and shown in Fig. \ref{fig:movie_poster_count}: (b), each poster in our dataset is associated with either 1, 2, or 3 genre labels.  
In this experiment, we partitioned the test data ($DB_t$) into three disjoint subsets $DB_t^{<1>}$, $DB_t^{<2>}$, $DB_t^{<3>}$ with posters associated with 1, 2, and 3 genres, respectively, 
and present the results in Table \ref{tab:genre_123}. 
From Table \ref{tab:genre_123}, it can be comprehended that posters having 3 genres yielded the best performance. Here also, in most of the cases, ERDT demonstrated the best performance. 



\begin{table*}[]
\centering
\caption{\small Genre-wise performance analysis and ranking of the models}
\begin{adjustbox}{angle=90, height=\textheight}
\begin{tabular}{c|l|c|c|c|c|c|c|c|c|c|c|c|c|c|c|c|c|c|c|c|c|c|c|c|c|c|c} 
\multicolumn{28}{c}{}\\
\multicolumn{28}{c}{\Large{(a) Genre-wise performance study}}\\
\multicolumn{28}{c}{}\\
\cline{3-28}
\multicolumn{1}{c}{} & \multicolumn{1}{c}{} & \multicolumn{26}{c}{Genre}\\ \cline{3-28}
\multicolumn{2}{r}{{}} & \multicolumn{2}{c|}{{Action}} & \multicolumn{2}{c|}{{Adventure}} & \multicolumn{2}{c|}{{Animation}} & \multicolumn{2}{c|}{{Biography}} & \multicolumn{2}{c|}{{Comedy}} & \multicolumn{2}{c|}{{Crime}} & \multicolumn{2}{c|}{{Drama}} & \multicolumn{2}{c|}{{Fantasy}} & \multicolumn{2}{c|}{{Horror}} & \multicolumn{2}{c|}{{Mystery}} & \multicolumn{2}{c|}{{Romance}} & \multicolumn{2}{c|}{{Sci-Fi}} & \multicolumn{2}{c}{{Thriller}} \\ \cline{2-28} 
\rowcolor[HTML]{F3F3F3}\cellcolor[HTML]{FFFFFF} & \cellcolor[HTML]{FFFFFF}Model & {w/o Pr} & {Pr} & {w/o Pr} & {Pr} & {w/o Pr} & {Pr} & {w/o Pr} & {Pr} & {w/o Pr} & {Pr} & {w/o Pr} & {Pr} & {w/o Pr} & {Pr} & {w/o Pr} & {Pr} & {w/o Pr} & {Pr} & {w/o Pr} & {Pr} & {w/o Pr} & {Pr} & {w/o Pr} & {Pr} & {w/o Pr} & {Pr} \\ \hline 

\multirow{7}{*}{{$\cal{P}$}} & {R} & 65.86 & 68.86 & 66.26 & 68.72 & 86.76 & \textbf{90.44} & 21.05 & 18.18 & 59.69 & 60.59 & 46.28 & 47.12 & 56.83 & 57.51 & 36.36 & 40.00 & 50.18 & 59.46 & \textbf{40.58} & \textbf{45.28} & 48.25 & 48.89 & 55.79 & 58.75 & 42.58 & 45.50 \\ 
& {RT} & 75.22 & 76.09 & 67.47 & 69.67 & 87.12 & 89.81 & 18.27 & 22.35 & 66.17 & 66.81 & 46.11 & 47.45 & \textbf{63.17} & 63.30 & 29.33 & 39.54 & 58.94 & 61.11 & 35.97 & 35.75 & 45.10 & 46.28 & 44.50 & 48.00 & 45.71 & 47.16 \\ 
& {RDT} & 76.03 & 77.49 & \textbf{76.01} & \textbf{78.10} & 85.29 & 86.79 & 25.96 & 29.11 & 67.99 & 68.40 & 46.33 & 47.71 & 61.62 & 62.44 & 34.68 & 33.75 & 54.60 & 62.34 & 33.50 & 35.80 & 49.59 & 50.00 & 54.55 & 59.62 & 48.95 & 48.23 \\ 
& {R + RT} & 74.75 & 76.10 & 69.55 & 70.91 & \textbf{89.24} & 90.39 & 19.44 & 16.67 & 66.67 & 67.17 & 48.80 & 48.22 & 62.01 & 62.66 & \textbf{38.30} & 46.43 & \textbf{59.32} & 64.21 & 38.22 & 41.21 & \textbf{50.43} & \textbf{50.65} & 48.33 & 52.94 & 46.84 & 48.62 \\ 
& {R + RDT} & 75.04 & 76.74 & 75.60 & 77.18 & 86.67 & 87.58 & \textbf{27.85} & \textbf{29.51} & 66.31 & 66.23 & 45.61 & 46.44 & 60.32 & 61.06 & 37.35 & 36.00 & 54.52 & 61.23 & 35.64 & 38.04 & 49.58 & 50.00 & \textbf{56.25} & \textbf{60.19} & 48.86 & 50.00 \\
& {RT + RDT} & \textbf{76.40} & \textbf{78.14} & 71.68 & 72.82 & 86.39 & 87.81 & 22.68 & 23.38 & 67.22 & \textbf{67.25} & \textbf{49.27} & 49.69 & 62.83 & 63.00 & 35.14 & 45.65 & 58.97 & \textbf{65.42} & 36.32 & 37.98 & 48.78 & 49.38 & 49.15 & 55.40 & 49.59 & 50.29 \\
& ERDT & 75.97 & 77.84 & 73.72 & 73.42 & 87.95 & 87.20 & 24.21 & 24.32 & \textbf{68.03} & 66.74 & 48.38 & \textbf{50.32} & 61.80 & \textbf{63.36} & 37.81 & \textbf{47.27} & 58.39 & 65.30 & 36.57 & 36.45 & 48.96 & 49.58 & 53.29 & 57.14 & \textbf{52.24} & \textbf{52.10} \\ \hline
 
\rowcolor[HTML]{EFEFEF} 
 & {R} & \textbf{72.83} & 67.77 & 66.67 & 61.93 & 79.88 & 75.00 & 3.67 & 1.84 & 75.49 & 74.27 & 44.77 & 42.81 & \textbf{87.69} & 86.00 & 5.16 & 2.58 & 52.99 & 32.84 & 26.17 & 22.43 & 57.90 & 57.90 & 28.65 & 25.41 & \textbf{53.80} & 45.38 \\
\rowcolor[HTML]{EFEFEF} 
 & {RT} & 68.09 & 66.35 & 75.10 & \textbf{73.25} & 86.59 & 85.98 & 17.43 & 17.43 & 75.97 & 73.79 & 52.29 & 48.69 & 77.85 & 77.23 & 14.19 & 10.97 & 54.10 & 45.15 & \textbf{42.52} & \textbf{39.25} & 60.53 & 58.95 & \textbf{52.43} & \textbf{45.41} & 44.84 & 42.94 \\
\rowcolor[HTML]{EFEFEF} 
 & {RDT} & 69.67 & 65.25 & 69.75 & 67.49 & 88.42 & 84.15 & \textbf{24.77} & \textbf{21.10} & 74.76 & 73.54 & 53.60 & 50.98 & 85.23 & 84.92 & \textbf{27.74} & \textbf{17.42} & \textbf{66.42} & \textbf{53.73} & 31.31 & 29.44 & \textbf{63.16} & \textbf{63.16} & 38.92 & 33.51 & 50.54 & 44.29 \\
\rowcolor[HTML]{EFEFEF} 
 & {R + RT} & 72.04 & 68.40 & \textbf{76.13} & 72.22 & 85.98 & 85.98 & 12.84 & 7.34 & \textbf{78.64} & \textbf{75.49} & 53.27 & 48.69 & 83.39 & 82.62 & 11.61 & 8.39 & 58.21 & 45.52 & 40.19 & 38.32 & 62.11 & 61.58 & 47.03 & 38.92 & 50.27 & \textbf{47.83} \\
\rowcolor[HTML]{EFEFEF} 
 & {R + RDT} & 71.72 & 67.77 & 71.40 & 67.49 & 87.20 & 85.98 & 20.18 & 16.51 & 75.49 & 73.79 & 50.98 & 49.02 & 87.23 & \textbf{86.62} & 20.00 & 11.61 & 65.30 & 51.87 & 31.31 & 28.97 & 62.11 & 62.11 & 38.92 & 33.51 & 52.17 & 46.74 \\
\rowcolor[HTML]{EFEFEF} 
 & {RT + RDT} & 71.09 & \textbf{68.88} & 75.51 & 72.22 & \textbf{89.02} & \textbf{87.81} & 20.18 & 16.51 & 78.16 & 75.24 & \textbf{55.23} & \textbf{51.96} & 83.23 & 82.00 & 16.77 & 13.55 & 60.08 & 52.24 & 37.85 & 36.92 & \textbf{63.16} & \textbf{63.16} & 47.03 & 41.62 & 49.46 & 46.47 \\

\rowcolor[HTML]{EFEFEF} 
\multirow{-7}{*}{{$\cal{R}$}} & ERDT & 70.93 & 68.25 & 73.87 & 71.61 & \textbf{89.02} & 87.20 & 21.10 & 16.51 & 76.94 & 74.52 & 53.60 & \textbf{51.96} & 85.39 & 83.54 & 20.00 & 16.77 & 64.93 & 53.36 & 36.92 & 34.58 & 62.11 & 61.58 & 43.78 & 43.24 & \textbf{53.80} & 47.28 \\ \hline 

\multirow{7}{*}{${\cal{S}}p$} & {R} & 71.45 & 76.82 & 83.23 & 86.08 & 98.47 & \textbf{99.00} & \textbf{98.90} & \textbf{99.34} & 80.15 & 81.19 & \textbf{86.34} & \textbf{87.37} & 47.20 & 49.63 & \textbf{98.94} & \textbf{99.54} & 88.27 & \textbf{95.01} & \textbf{93.47} & \textbf{95.38} & 90.78 & 91.02 & \textbf{96.73} & \textbf{97.43} & 75.77 & 81.85 \\ 
& {RT} & 83.03 & 84.23 & 82.11 & 84.25 & 98.39 & 98.77 & 93.75 & 95.15 & 84.88 & 85.73 & 83.93 & 85.82 & \textbf{64.02} & \textbf{64.51} & 95.97 & 98.02 & \textbf{91.60} & 93.59 & 87.10 & 87.98 & 89.06 & 89.84 & 90.58 & 92.92 & 82.21 & 83.94 \\
& {RDT} & \textbf{83.39} & \textbf{85.66} & \textbf{89.13} & \textbf{90.65} & 98.09 & 98.39 & 94.34 & 95.89 & \textbf{86.29} & \textbf{86.77} & 83.68 & 85.31 & 57.93 & 59.51 & 93.84 & 95.97 & 87.69 & 92.76 & 89.41 & 91.00 & 90.47 & 90.63 & 95.33 & 96.73 & 82.40 & 84.12 \\
& {R + RT} & 81.60 & 83.75 & 83.54 & 85.37 & \textbf{98.70} & 98.85 & 95.74 & 97.06 & 84.69 & 85.63 & 85.31 & 86.25 & 59.51 & 60.98 & 97.79 & 98.86 & 91.10 & 94.34 & 88.93 & 90.68 & \textbf{90.94} & \textbf{91.09} & 92.76 & 95.02 & 80.94 & 83.12 \\
& {R + RDT}& 81.96 & 84.47 & 88.52 & 90.04 & 98.32 & 98.47 & 95.81 & 96.84 & 85.07 & 85.35 & 83.93 & 85.14 & 54.51 & 56.22 & 96.12 & 97.57 & 87.85 & 92.68 & 90.45 & 91.96 & 90.63 & 90.78 & 95.64 & 96.81 & 81.76 & 84.39 \\
& {RT + RDT}  & \textbf{83.39} & 85.42 & 85.26 & 86.69 & 98.24 & 98.47 & 94.49 & 95.66 & 85.16 & 85.73 & 85.05 & 86.17 & 60.98 & 61.83 & 96.35 & 98.10 & 90.68 & 93.84 & 88.69 & 89.73 & 90.16 & 90.39 & 93.00 & 95.18 & 83.21 & 84.66 \\
 & {R + RT + RDT} & 83.03 & 85.30 & 86.99 & 87.20 & 98.47 & 98.39 & 94.71 & 95.89 & 85.92 & 85.54 & 84.97 & 86.51 & 58.17 & 61.71 & 96.12 & 97.79 & 89.68 & 93.68 & 89.09 & 89.73 & 90.39 & 90.70 & 94.47 & 95.33 & \textbf{83.58} & \textbf{85.48} \\ \hline

\rowcolor[HTML]{EFEFEF}
 & {R} & 72.14 & 72.30 & 74.95 & 74.01 & 89.17 & 87.00 & 51.28 & 50.59 & 77.82 & 77.73 & 65.56 & 65.09 & 67.44 & 67.82 & 52.05 & 51.06 & 70.63 & 63.92 & 59.82 & 58.91 & 74.34 & 74.46 & 62.69 & 61.42 & 64.79 & 63.62 \\
\rowcolor[HTML]{EFEFEF} 
 & {RT} & 75.56 & 75.29 & 78.61 & 78.75 & 92.49 & 92.38 & 55.59 & 56.29 & 80.42 & 79.76 & 68.11 & 67.26 & 70.94 & 70.87 & 55.08 & 54.50 & 72.85 & 69.37 & \textbf{64.81} & 63.62 & 74.79 & 74.40 & \textbf{71.51} & 69.16 & 63.53 & 63.44 \\
\rowcolor[HTML]{EFEFEF}
 & {RDT} & 76.53 & 75.45 & 79.44 & 79.07 & 93.25 & 91.27 & \textbf{59.56} & \textbf{58.49} & 80.53 & 80.16 & 68.64 & 68.14 & 71.58 & 72.22 & \textbf{60.79} & 56.69 & 77.05 & 73.25 & 60.36 & 60.22 & \textbf{76.81} & \textbf{76.89} & 67.12 & 65.12 & 66.47 & 64.21 \\
\rowcolor[HTML]{EFEFEF}
 & {R + RT} & 76.82 & 76.08 & 79.83 & 78.79 & 92.34 & 92.41 & 54.29 & 52.20 & \textbf{81.66} & \textbf{80.56} & 69.29 & 67.47 & 71.45 & 71.80 & 54.70 & 53.62 & 74.65 & 69.93 & 64.56 & \textbf{64.50} & 76.52 & 76.34 & 69.89 & 66.97 & 65.61 & 65.47 \\
 \rowcolor[HTML]{EFEFEF} 
 & {R + RDT} & 76.84 & 76.12 & 80.01 & 78.82 & 92.76 & 92.22 & 58.00 & 56.68 & 80.28 & 79.57 & 67.50 & 67.08 & 70.87 & 71.42 & 58.02 & 54.59 & 76.58 & 72.27 & 60.84 & 60.47 & 76.37 & 76.44 & 67.28 & 65.16 & 66.97 & 65.57 \\
\rowcolor[HTML]{EFEFEF} 
 & {RT + RDT} & \textbf{77.24} & \textbf{77.15} & 80.39 & \textbf{79.45} & 93.63 & \textbf{93.14} & 57.34 & 56.09 & \textbf{81.66} & 80.49 & \textbf{70.14} & 69.06 & \textbf{72.10} & 71.91 & 56.56 & 55.82 & 75.38 & 73.04 & 63.27 & 63.32 & 76.66 & 76.77 & 70.01 & 68.40 & 66.33 & 65.57 \\

\rowcolor[HTML]{EFEFEF}
\multirow{-7}{*}{{${\cal{BA}}$}} & ERDT & 76.98 & 76.78 & \textbf{80.43} & 79.40 & \textbf{93.75} & 92.79 & 57.91 & 56.20 & 81.43 & 80.03 & 69.28 & \textbf{69.24} & 71.78 & \textbf{72.62} & 58.06 & \textbf{57.28} & \textbf{77.30} & \textbf{73.52} & 63.00 & 62.15 & 76.25 & 76.14 & 69.13 & \textbf{69.29} & \textbf{68.69} & \textbf{66.38} \\ \hline 
\multirow{7}{*}{${\cal{FM}}$} & {R} & 69.17 & 68.31 & 66.46 & 65.15 & 83.18 & 82.00 & 6.25 & 3.33 & 66.67 & 66.74 & 45.52 & 44.86 & 68.97 & 68.93 & 9.04 & 4.85 & 51.54 & 42.31 & 31.82 & 30.00 & 52.63 & 53.01 & 37.86 & 35.47 & 47.54 & 45.44 \\
 
& {RT} & 71.48 & 70.89 & 71.08 & 71.41 & 86.85 & 87.85 & 17.84 & 19.59 & 70.73 & 70.13 & 49.01 & 48.07 & 69.75 & 69.58 & 19.13 & 17.17 & 56.42 & 51.93 & 38.97 & 37.42 & 51.69 & 51.85 & \textbf{48.14} & 46.67 & 45.27 & 44.95 \\
 & {RDT} & 72.71 & 70.84 & 72.75 & 72.41 & 86.83 & 85.45 & \textbf{25.35} & \textbf{24.47} & 71.21 & 70.88 & 49.70 & 49.29 & 71.53 & 71.97 & \textbf{30.82} & 22.98 & 59.93 & 57.72 & 32.37 & 32.31 & 55.56 & \textbf{55.81} & 45.43 & 42.91 & 49.73 & 46.18 \\
& {R + RT} & 73.37 & 72.05 & 72.69 & 71.56 & 87.58 & \textbf{88.13} & 15.47 & 10.19 & 72.16 & \textbf{71.09} & 50.94 & 48.46 & 71.13 & 71.27 & 17.82 & 14.21 & 58.76 & 53.28 & \textbf{39.18} & \textbf{39.71} & \textbf{55.66} & 55.58 & 47.67 & 44.86 & 48.49 & 48.22 \\ 
 & {R + RDT} & 73.34 & 71.98 & 73.44 & 72.01 & 86.93 & 86.77 & 23.40 & 21.18 & 70.60 & 69.81 & 48.15 & 47.70 & 71.32 & 71.63 & 26.05 & 17.56 & 59.42 & 56.16 & 33.33 & 32.89 & 55.14 & 55.40 & 46.01 & 43.06 & 50.46 & 48.32 \\
 & {RT + RDT} & \textbf{73.65} & \textbf{73.22} & 73.55 & \textbf{72.52} & 87.69 & 87.81 & 21.36 & 19.36 & \textbf{72.28} & 71.02 & \textbf{52.08} & 50.80 & 71.61 & 71.26 & 22.71 & 20.90 & 59.52 & 58.09 & 37.07 & 37.44 & 55.05 & 55.43 & 48.07 & 47.53 & 49.52 & 48.31 \\
 & ERDT & 73.37 & 72.73 & \textbf{73.79} & 72.50 & \textbf{88.49} & 87.20 & 22.55 & 19.67 & 72.21 & 70.41 & 50.85 & \textbf{51.13} & \textbf{71.71} & \textbf{72.06} & 26.16 & \textbf{24.76} & \textbf{61.48} & \textbf{58.73} & 36.74 & 35.49 & 54.76 & 54.93 & 48.07 & \textbf{49.23} & \textbf{53.01} & \textbf{49.57} \\ \hline
\multicolumn{28}{r}{Pr : with probabilistic module, w/o Pr : without probabilistic module}

\end{tabular}
\end{adjustbox}
\begin{adjustbox}{angle=90, height=\textheight}
\begin{tabular}{l|c|c||c|c|c|c|c|c|c|c|c|c|c|c|c|c|c|c|c|c|c|c|c|c|c|c|c|c}
\multicolumn{29}{c}{}\\
\multicolumn{29}{c}{{(b) Genre-wise ranking ($\Re$) of models (w/o Pr) based on balanced accuracy (${\cal{BA}}$) }}\\ 
\multicolumn{29}{c}{}\\
\cline{4-29}
\multicolumn{1}{c}{} & \multicolumn{1}{c}{} & \multicolumn{1}{c}{} & \multicolumn{26}{c}{Genre}\\ \cline{2-29}
\multicolumn{1}{c}{} & 
\multicolumn{2}{c||}{{Overall}} &
\multicolumn{2}{c|}{{Action}} & \multicolumn{2}{c|}{{Adventure}} & \multicolumn{2}{c|}{{Animation}} & \multicolumn{2}{c|}{{Biography}} & \multicolumn{2}{c|}{{Comedy}} & \multicolumn{2}{c|}{{Crime}} & \multicolumn{2}{c|}{{Drama}} & \multicolumn{2}{c|}{{Fantasy}} & \multicolumn{2}{c|}{{Horror}} & \multicolumn{2}{c|}{{Mystery}} & \multicolumn{2}{c|}{{Romance}} & \multicolumn{2}{c|}{{Sci-Fi}} & \multicolumn{2}{c}{{Thriller}} \\
\hline 

\rowcolor[HTML]{F3F3F3}\cellcolor[HTML]{FFFFFF}Model
 & ${\cal{BA}}$ & $\Re$ & ${\cal{BA}}$ & $\Re$ & ${\cal{BA}}$ & $\Re$ & ${\cal{BA}}$ & $\Re$ & ${\cal{BA}}$ & $\Re$ & ${\cal{BA}}$ & $\Re$ & ${\cal{BA}}$ & $\Re$ & ${\cal{BA}}$ & $\Re$ & ${\cal{BA}}$ & $\Re$ & ${\cal{BA}}$ & $\Re$ & ${\cal{BA}}$ & $\Re$ & ${\cal{BA}}$ & $\Re$ & ${\cal{BA}}$ & $\Re$ & ${\cal{BA}}$ & $\Re$ \\ \hline
 
R & 67.90 & 7 & 72.14 & 7 & 74.95 & 7 & 89.17 & 7 & 51.28 & 7 & 77.82 & 7 & 65.56 & 7 & 67.44 & 7 & 52.05 & 7 & 70.63 & 7 & 59.82 & 7 & 74.34 & 7 & 62.69 & 7 & 64.79 & 6 \\ 
RT & 71.10 & 6 & 75.56 & 6 & 78.61 & 6 & 92.49 & 5 & 55.59 & 5 & 80.42 & 5 & 68.11 & 5 & 70.94 & 5 & 55.08 & 5 & 72.85 & 6 & 64.81 & 1 & 74.79 & 6 & 71.51 & 1 & 63.53 & 7 \\ 
RDT & 72.16 & 3 & 76.53 & 5 & 79.44 & 5 & 93.25 & 3 & 59.56 & 1 & 80.53 & 4 & 68.64 & 4 & 71.58 & 3 & 60.79 & 1 & 77.05 & 2 & 60.36 & 6 & 76.81 & 1 & 67.12 & 6 & 66.47 & 3 \\
R + RT & 71.66 & 5 & 76.82 & 4 & 79.83 & 4 & 92.34 & 6 & 54.29 & 6 & 81.66 & 1 & 69.29 & 2 & 71.45 & 4 & 54.70 & 6 & 74.65 & 5 & 64.56 & 2 & 76.52 & 3 & 69.89 & 3 & 65.61 & 5 \\ 
R + RDT & 71.72 & 4 & 76.84 & 3 & 80.01 & 3 & 92.76 & 4 & 58.00 & 2 & 80.28 & 6 & 67.50 & 6 & 70.87 & 6 & 58.02 & 3 & 76.58 & 3 & 60.84 & 5 & 76.37 & 4 & 67.28 & 5 & 66.97 & 2 \\
RT + RDT & 72.36 & 2 & 77.24 & 1 & 80.39 & 2 & 93.63 & 2 & 57.34 & 4 & 81.66 & 2 & 70.14 & 1 & 72.10 & 1 & 56.56 & 4 & 75.38 & 4 & 63.27 & 3 & 76.66 & 2 & 70.01 & 2 & 66.33 & 4 \\ 
ERDT & 72.61 & 1 & 76.98 & 2 & 80.43 & 1 & 93.75 & 1 & 57.91 & 3 & 81.43 & 3 & 69.28 & 3 & 71.78 & 2 & 58.06 & 2 & 77.30 & 1 & 63.00 & 4 & 76.25 & 5 & 69.13 & 4 & 68.69 & 1 \\ \hline
\end{tabular}

\end{adjustbox}
\label{tab:genre_big}
\end{table*}

\section{Conclusion}
\label{5sec:conclusion}
\noindent
In this paper, we worked on multi-label genre identification solely from movie poster images. 
We did not take any aid from any other visual/ textual/ audio modalities. 
We initially proposed a Residual Dense Transformer (RDT) with asymmetric loss to handle imbalanced data; then improvised the model using an ensembled variation of RDT, i.e., ERDT to tackle multi-label genre identification. 
We also added a probabilistic module to our models (e.g., PrERDT) to eliminate unnecessary genres. 
For experiments, we procured 13882 number of poster images from IMDb. 
Our models exhibited encouraging performances and bit some major SOTA architectures.
In the future, we will focus on enhancing the performance for some specific genres, e.g., \emph{biography}, \emph{fantasy}, \emph{mystery}, where our current models have shown subpar results. 
Currently, PrERDT lags behind ERDT due to a lower hit ratio. 
We will also endeavor to improve the performance of ERDT in multi-label classification, so that the hit ratio improves, and eventually boosts the efficacy of PrERDT.



\balance 

\bibliographystyle{IEEEtran}  
\bibliography{ref} 

\begin{thebibliography}{10}
\providecommand{\url}[1]{#1}
\csname url@samestyle\endcsname
\providecommand{\newblock}{\relax}
\providecommand{\bibinfo}[2]{#2}
\providecommand{\BIBentrySTDinterwordspacing}{\spaceskip=0pt\relax}
\providecommand{\BIBentryALTinterwordstretchfactor}{4}
\providecommand{\BIBentryALTinterwordspacing}{\spaceskip=\fontdimen2\font plus
\BIBentryALTinterwordstretchfactor\fontdimen3\font minus \fontdimen4\font\relax}
\providecommand{\BIBforeignlanguage}[2]{{%
\expandafter\ifx\csname l@#1\endcsname\relax
\typeout{** WARNING: IEEEtran.bst: No hyphenation pattern has been}%
\typeout{** loaded for the language `#1'. Using the pattern for}%
\typeout{** the default language instead.}%
\else
\language=\csname l@#1\endcsname
\fi
#2}}
\providecommand{\BIBdecl}{\relax}
\BIBdecl

\bibitem{30_9057706}
J.~A. Wi, S.~Jang, and Y.~Kim, ``Poster-based multiple movie genre classification using inter-channel features,'' \emph{IEEE Access}, vol.~8, pp. 66\,615--66\,624, 2020.

\bibitem{4_winoto2010role}
P.~Winoto \emph{et~al.}, ``The role of user mood in movie recommendations,'' \emph{Expert Systems with Applications}, vol.~37, no.~8, pp. 6086--6092, 2010.

\bibitem{netflix_frey2021}
M.~Frey, \emph{Netflix recommends: algorithms, film choice, and the history of taste}.\hskip 1em plus 0.5em minus 0.4em\relax Univ of California Press, 2021.

\bibitem{25_montalvo2022trailers12k}
R.~M.~Lezama \emph{et~al.}, ``Trailers12k: Evaluating transfer learning for movie trailer genre classification,'' \emph{arXiv:2210.07983}, 2022.

\bibitem{26_yadav2020unified}
A.~Yadav and D.~K. Vishwakarma, ``A unified framework of deep networks for genre classification using movie trailer,'' \emph{Applied Soft Computing}, vol.~96, p. 106624, 2020.

\bibitem{10_hoang2018predicting}
Q.~Hoang, ``Predicting movie genres based on plot summaries,'' \emph{arXiv:1801.04813}, 2018.

\bibitem{13_wehrmann2018self}
J.~Wehrmann \emph{et~al.}, ``Self-attention for synopsis-based multi-label movie genre classification,'' in \emph{Int. FLAIRS Conf.}, 2018, pp. 236--241.

\bibitem{28_10.1145/3132515.3132516}
W.-T. Chu and H.-J. Guo, ``Movie genre classification based on poster images with deep neural networks,'' in \emph{Multimodal Understanding of Social, Affective and Subjective Attributes}.\hskip 1em plus 0.5em minus 0.4em\relax {ACM}, 2017, p. 39–45.

\bibitem{Herrera2016}
F.~Herrera \emph{et~al.}, \emph{Multilabel Classification: Problem Analysis, Metrics and Techniques}.\hskip 1em plus 0.5em minus 0.4em\relax Springer, 2016.

\bibitem{Johnson2019SurveyOD}
J.~M. Johnson and T.~M. Khoshgoftaar, ``Survey on deep learning with class imbalance,'' \emph{Journal of Big Data}, vol.~6, pp. 1--54, 2019.

\bibitem{dosovitskiy2020image}
A.~Dosovitskiy \emph{et~al.}, ``An image is worth 16x16 words: Transformers for image recognition at scale,'' in \emph{ICLR}, 2021.

\bibitem{ridnik2021asymmetric}
E.~B.-Baruch \emph{et~al.}, ``{Asymmetric loss for multi-label classification},'' in \emph{ICCV}, 2021, pp. 82--91.

\bibitem{23_Simes2016MovieGC}
G.~S. Sim{\~o}es \emph{et~al.}, ``Movie genre classification with convolutional neural networks,'' \emph{IJCNN}, pp. 259--266, 2016.

\bibitem{21_4107177}
X.~Yuan \emph{et~al.}, ``{Automatic Video Genre Categorization using Hierarchical SVM},'' in \emph{{ICIP}}, 2006, pp. 2905--2908.

\bibitem{11_kar-etal-2018-folksonomication}
S.~Kar \emph{et~al.}, ``{F}olksonomication: Predicting tags for movies from plot synopses using emotion flow encoded neural network,'' in \emph{{{ICCL}}}, 2018, pp. 2879--2891.

\bibitem{17_gorinski-lapata-2018-whats}
P.~J. Gorinski and M.~Lapata, ``What{'}s this movie about? a joint neural network architecture for movie content analysis,'' in \emph{NAACL: Human Language Technologies, vol. 1}.\hskip 1em plus 0.5em minus 0.4em\relax ACL, 2018, pp. 1770--1781.

\bibitem{38_bribiesca2021multimodal}
I.~R. Bribiesca \emph{et~al.}, ``Multimodal weighted fusion of transformers for movie genre classification,'' in \emph{Workshop on MAI}, 2021, pp. 1--5.

\bibitem{35_cascante2019moviescope}
P.~C.~Bonilla \emph{et~al.}, ``Moviescope: Large-scale analysis of movies using multiple modalities,'' \emph{arXiv:1908.03180}, 2019.

\bibitem{20_zhou2010movie}
H.~Zhou \emph{et~al.}, ``Movie genre classification via scene categorization,'' in \emph{ACM Multimedia}, 2010, pp. 747--750.

\bibitem{24_wehrmann2017movie}
J.~Wehrmann and R.~C. Barros, ``Movie genre classification: A multi-label approach based on convolutions through time,'' \emph{Applied Soft Computing}, vol.~61, pp. 973--982, 2017.

\bibitem{27_pobar2017multi}
M.~Pobar \emph{et~al.}, ``Multi-label poster classification into genres using different problem transformation methods,'' in \emph{CAIP}, 2017, pp. 367--378.

\bibitem{29_8990490}
N.~Gozuacik \emph{et~al.}, ``Turkish movie genre classification from poster images using convolutional neural networks,'' in \emph{ELECO}, 2019, pp. 930--934.

\bibitem{9_ertugrul2018movie}
A.~M. Ertugrul \emph{et~al.}, ``{Movie genre classification from plot summaries using bidirectional LSTM},'' in \emph{ICSC}, 2018, pp. 248--251.

\bibitem{12_battu-etal-2018-predicting}
V.~Battu \emph{et~al.}, ``Predicting the genre and rating of a movie based on its synopsis,'' in \emph{PACLIC}, 1{--}3 2018.

\bibitem{36_mangolin2022multimodal}
R.~B. Mangolin \emph{et~al.}, ``A multimodal approach for multi-label movie genre classification,'' \emph{Multimedia Tools and Applications}, vol.~81, no.~14, pp. 19\,071--19\,096, 2022.

\bibitem{37_arevalo2017gated}
J.~Arevalo \emph{et~al.}, ``Gated multimodal units for information fusion,'' \emph{ICLR Workshop}, 2017.

\bibitem{31_1048494}
Z.~Rasheed and M.~Shah, ``Movie genre classification by exploiting audio-visual features of previews,'' in \emph{ICPR}, 2002, pp. 2: 1086--1089.

\bibitem{32_brezeale2006using}
D.~Brezeale and D.~J. Cook, ``Using closed captions and visual features to classify movies by genre,'' in \emph{MDM/KDD}, 2006, pp. 1--5.

\bibitem{vit_survey_pami}
K.~Han \emph{et~al.}, ``A survey on vision transformer,'' \emph{IEEE TPAMI}, vol.~45, no.~1, pp. 87--110, 2023.

\bibitem{resnetv2}
K.~He, X.~Zhang, S.~Ren, and J.~Sun, ``Identity mappings in deep residual networks,'' in \emph{ECCV}, 2016, pp. 630--645.

\bibitem{DL_Survey_Alom}
M.~Z. Alom \emph{et~al.}, ``{The history began from AlexNet: A comprehensive survey on deep learning approaches},'' \emph{arXiv:1803.01164}, 2018.

\bibitem{attention_need}
A.~Vaswani \emph{et~al.}, ``Attention is all you need,'' \emph{NIPS}, vol.~30, 2017.

\bibitem{d2l}
A.~Zhang, Z.~C. Lipton, M.~Li, and A.~J. Smola, ``Dive into deep learning,'' 2023.

\bibitem{ln}
J.~L. Ba \emph{et~al.}, ``Layer normalization,'' \emph{arXiv:1607.06450}, 2016.

\bibitem{adam2014}
D.~P. Kingma and J.~Ba, ``{Adam: A Method for Stochastic Optimization},'' \emph{ICLR (Poster)}, 2015.

\bibitem{ensemble}
L.~Rokach, ``{Ensemble-based Classifiers},'' \emph{Artificial intelligence review}, vol.~33, no.~1, pp. 1--39, 2010.

\bibitem{hpTuning}
L.~Zahedi \emph{et~al.}, ``Search algorithms for automated hyper-parameter tuning,'' in \emph{{ICDATA}}, 2021, pp. 1--10.

\bibitem{eval_method2}
M.-L. Zhang and Z.-H. Zhou, ``A review on multi-label learning algorithms,'' \emph{IEEE TKDE}, vol.~26, no.~8, pp. 1819--1837, 2014.

\bibitem{8099726}
G.~Huang \emph{et~al.}, ``Densely connected convolutional networks,'' in \emph{CVPR}, 2017, pp. 2261--2269.

\bibitem{tan2019efficientnet}
M.~Tan and Q.~Le, ``Efficientnet: Rethinking model scaling for convolutional neural networks,'' in \emph{ICML}, 2019, pp. 6105--6114.

\bibitem{inceptionv3}
C.~Szegedy \emph{et~al.}, ``Rethinking the inception architecture for computer vision,'' in \emph{CVPR}, 2016, pp. 2818--2826.

\bibitem{mobilenetv2}
M.~Sandler \emph{et~al.}, ``{MobileNetV2: Inverted Residuals and Linear Bottlenecks},'' in \emph{CVPR}, 2018, pp. 4510--4520.

\end{thebibliography}

\vspace{-0.5in}
\begin{IEEEbiography}[{\includegraphics[width=0.8in,height=0.8in,clip,keepaspectratio]{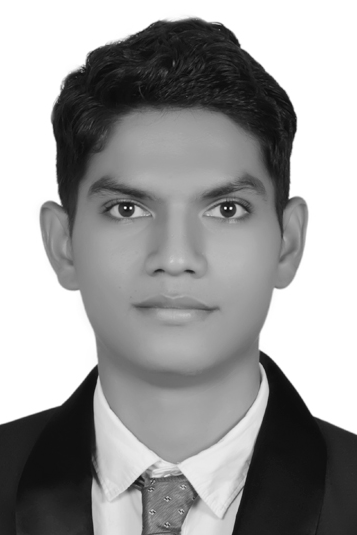}}]
{Utsav Kumar Nareti} (Student Member, IEEE) is pursuing his Ph.D. in CSE from IIT Patna, India. 
He received his B.Tech and M.Tech dual degree in IT, from IIIT Allahabad, India in 2021.
His research interests include computer vision and deep learning.
\end{IEEEbiography}
\vspace{-0.5in}
\begin{IEEEbiography}[{\includegraphics[width=0.8in,height=0.8in,clip,keepaspectratio]{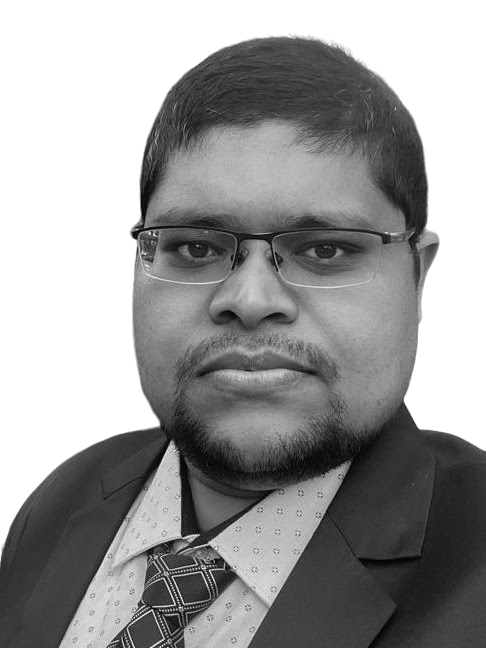}}]
{Chandranath Adak} (Senior Member, IEEE) received his Ph.D. in
Analytics from the UTS, Australia in 2019.
He is currently an assistant professor at IIT Patna, India. His research interests include computer vision, deep learning, and data analytics.
\end{IEEEbiography}
\vspace{-0.5in}
\begin{IEEEbiography}[{\includegraphics[width=0.8in,height=0.8in,clip,keepaspectratio]{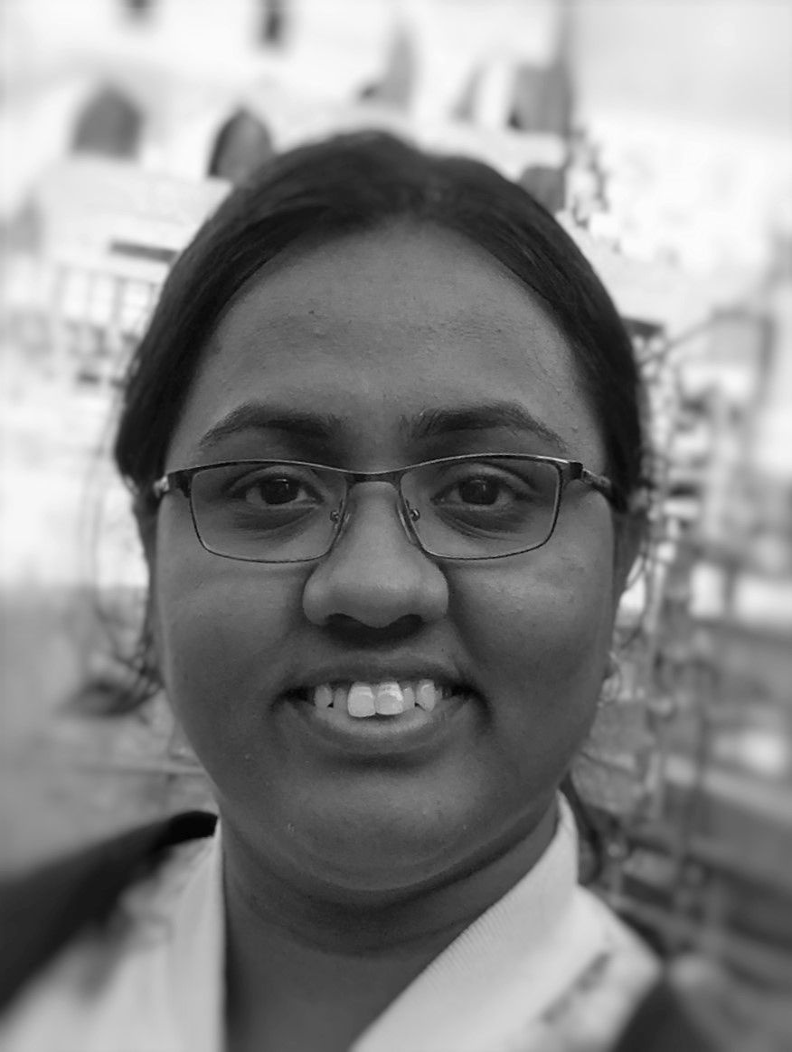}}]
{Soumi Chattopadhyay} (Senior Member, IEEE) received her Ph.D. from the Indian Statistical Institute in 2019. Currently, she is an assistant professor at IIT Indore, India. Her research interests include services computing, artificial intelligence, machine/ deep learning.
\end{IEEEbiography}

\pagebreak 

\appendices

\counterwithin{figure}{section}
\counterwithin{table}{section}

\section{Genre label co-occurrence matrix}
\label{app:co_occurence}

\begin{figure}[!h]
    \centering
    \includegraphics[width=0.9\linewidth]{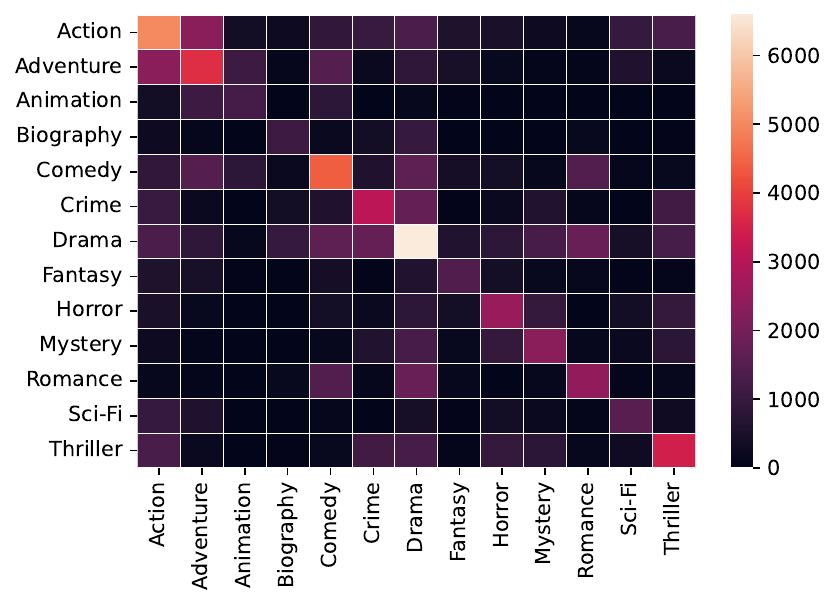}
    \caption{Co-occurrence matrix for movie poster genres}
    \label{fig:co_ocurrance_matrix}
\end{figure}

\noindent
Fig. \ref{fig:co_ocurrance_matrix} provides a visual representation of the relationships and occurrences among various movie poster genres as noted in Section 
IV-A of the main manuscript.

\section{Dataset Challenges}
\label{app:data_challenge}

\begin{figure*}
\centering
\footnotesize
\begin{tabular}{c|c|c|c || c|c|c|c}
\hline 
&&& &&&&  \\[\dimexpr-\normalbaselineskip+1.5pt]
\includegraphics[width=0.1\linewidth, height=0.145\linewidth]{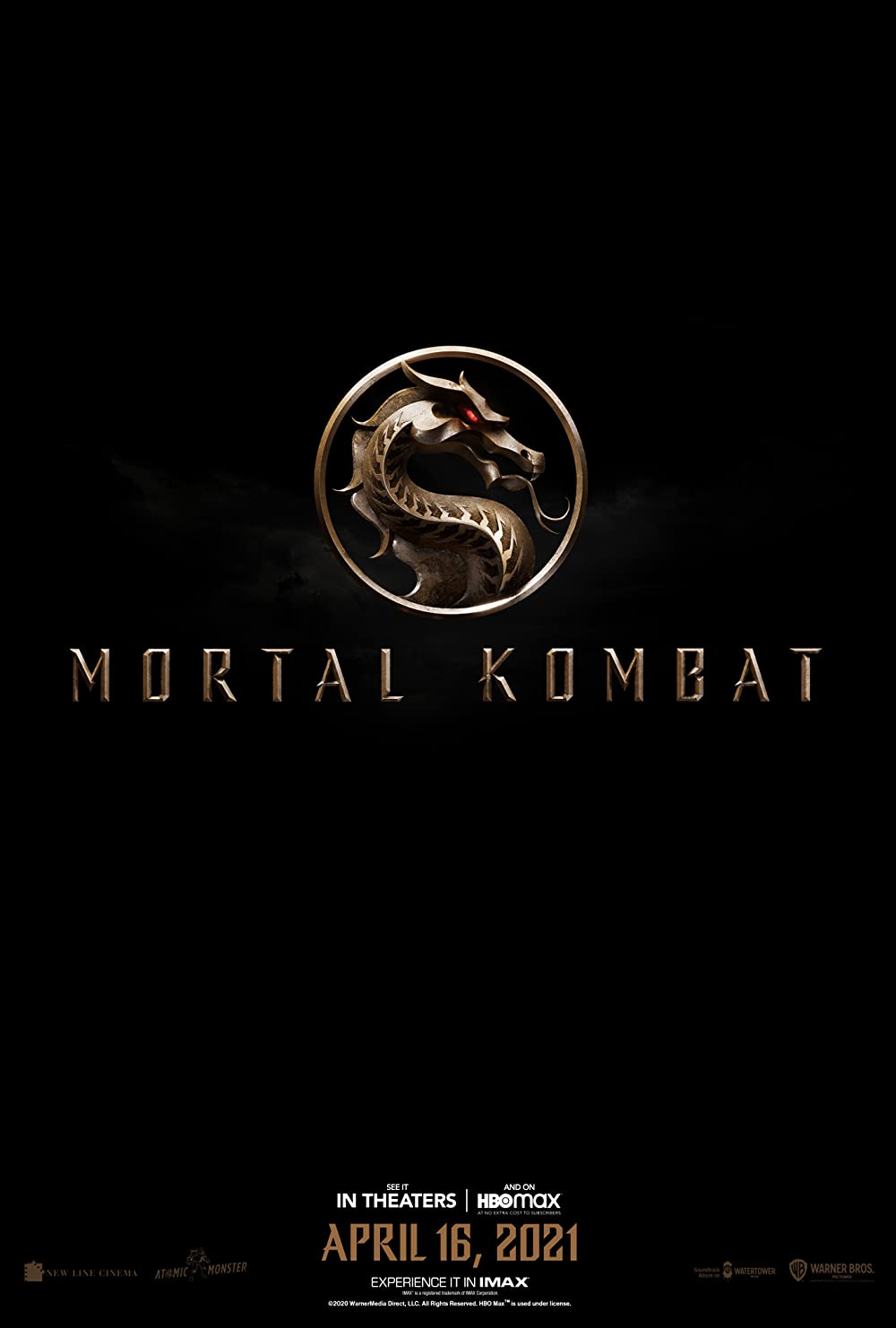} & 
\includegraphics[width=0.1\linewidth, height=0.145\linewidth]{figs/challenge/2.jpg} & 
\includegraphics[width=0.1\linewidth, height=0.145\linewidth]{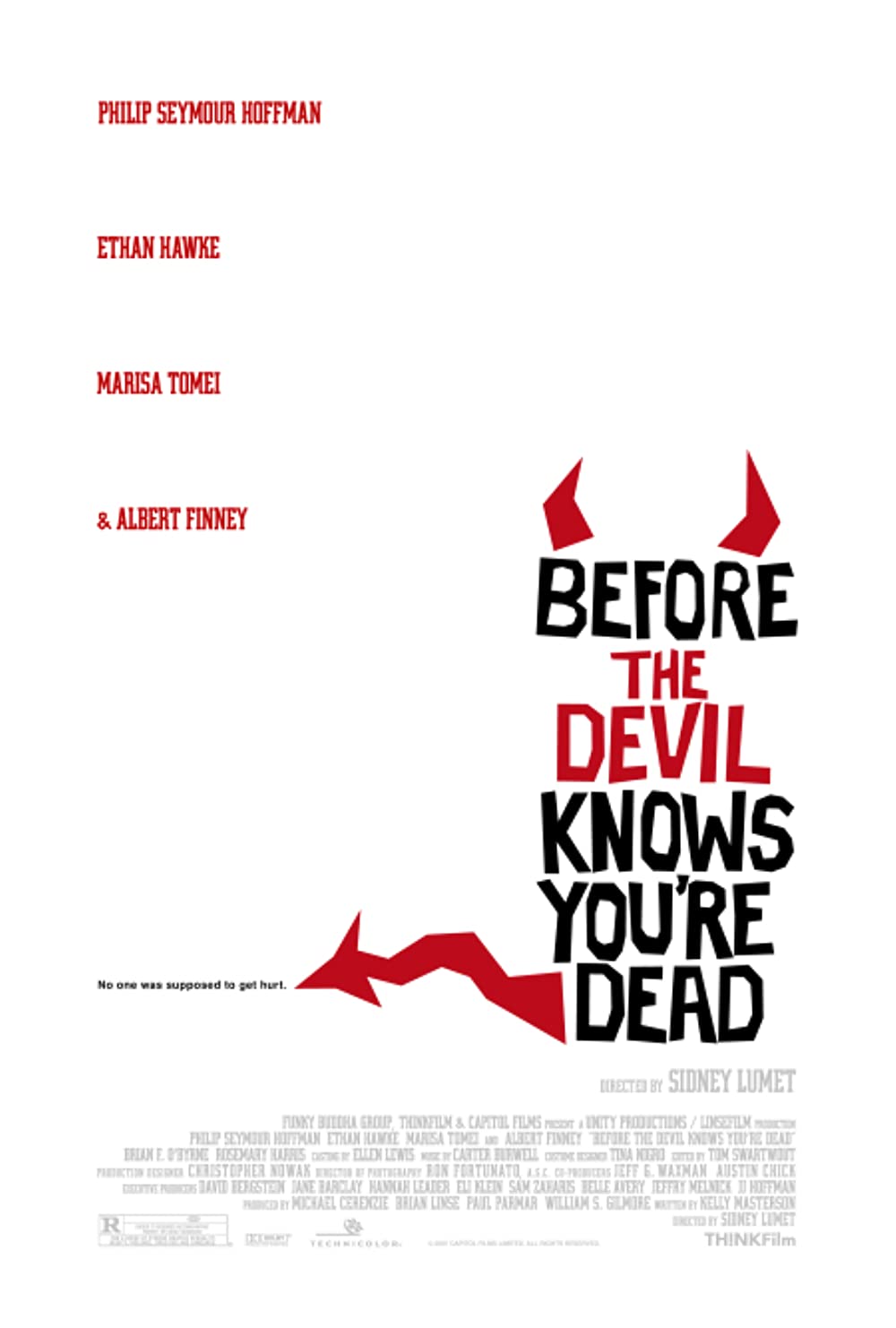} &
\includegraphics[width=0.1\linewidth, height=0.145\linewidth]{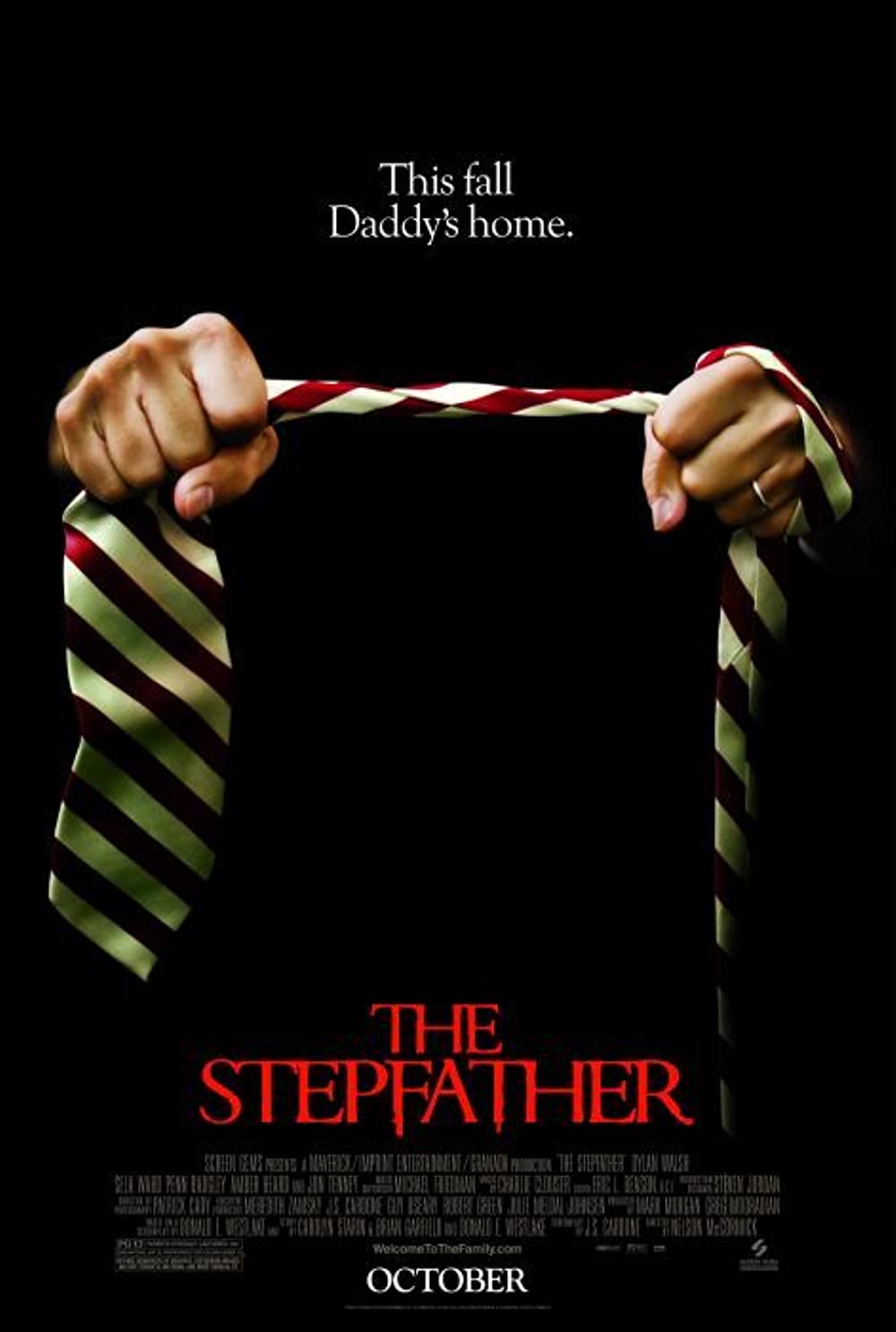} & 
\includegraphics[width=0.1\linewidth, height=0.145\linewidth]{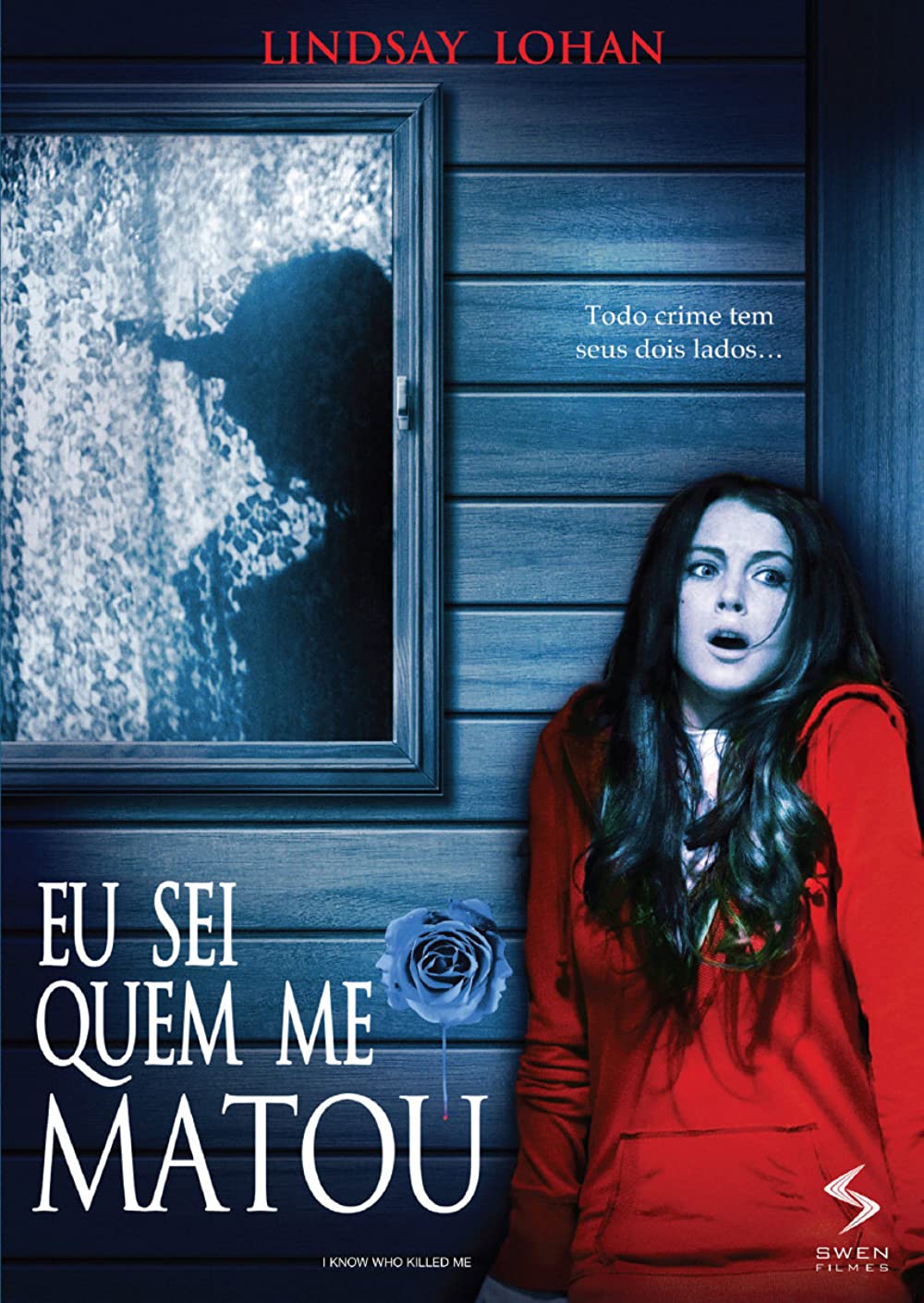} & 
\includegraphics[width=0.1\linewidth, height=0.145\linewidth]{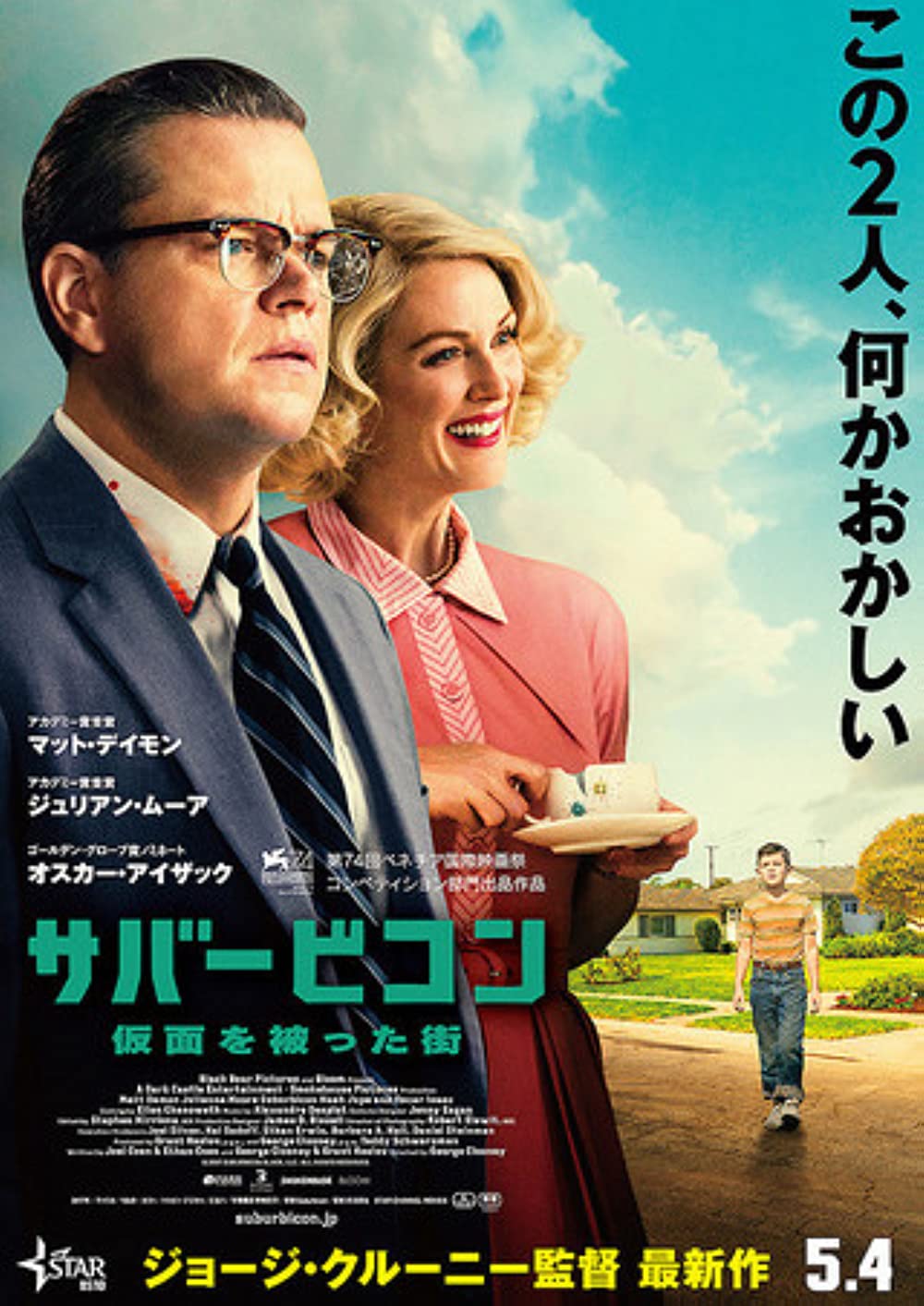} & 
\includegraphics[width=0.1\linewidth, height=0.145\linewidth]{figs/challenge/7.jpg} & 
\includegraphics[width=0.1\linewidth, height=0.145\linewidth]{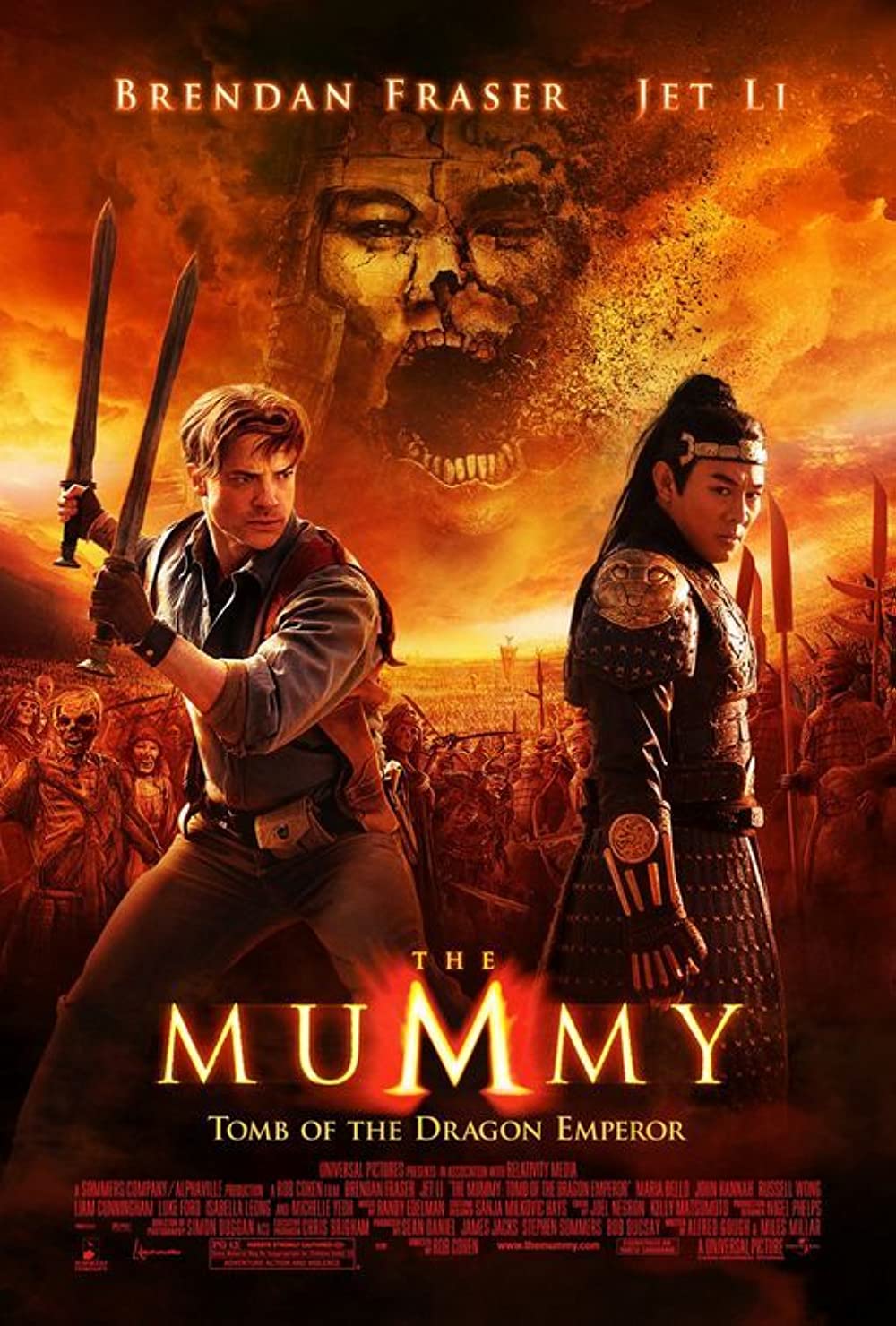} \\  
\textcolor{blue}{\emph{i.}} (1, 2, 8) & \textcolor{blue}{\emph{ii.}} (7, 9, 10) & \textcolor{blue}{\emph{iii.}} (6, 7, 13) & \textcolor{blue}{\emph{iv.}} (10, 13, -) & 
\textcolor{blue}{\emph{v.}} (9, 10, 13) & \textcolor{blue}{\emph{vi.}} (5, 6, 7) & \textcolor{blue}{\emph{vii.}} (1, 12, -) & \textcolor{blue}{\emph{viii.}} (1, 2, 8)  \\ 
\multicolumn{4}{c||}{less info} & \multicolumn{2}{c|}{moderate/ adequate info} & \multicolumn{2}{c}{complex background}\\ 
\hline 
&&& &&&&  \\[\dimexpr-\normalbaselineskip+1.5pt]
\includegraphics[width=0.1\linewidth, height=0.145\linewidth]{figs/challenge/9.jpg} & 
\includegraphics[width=0.1\linewidth, height=0.145\linewidth]{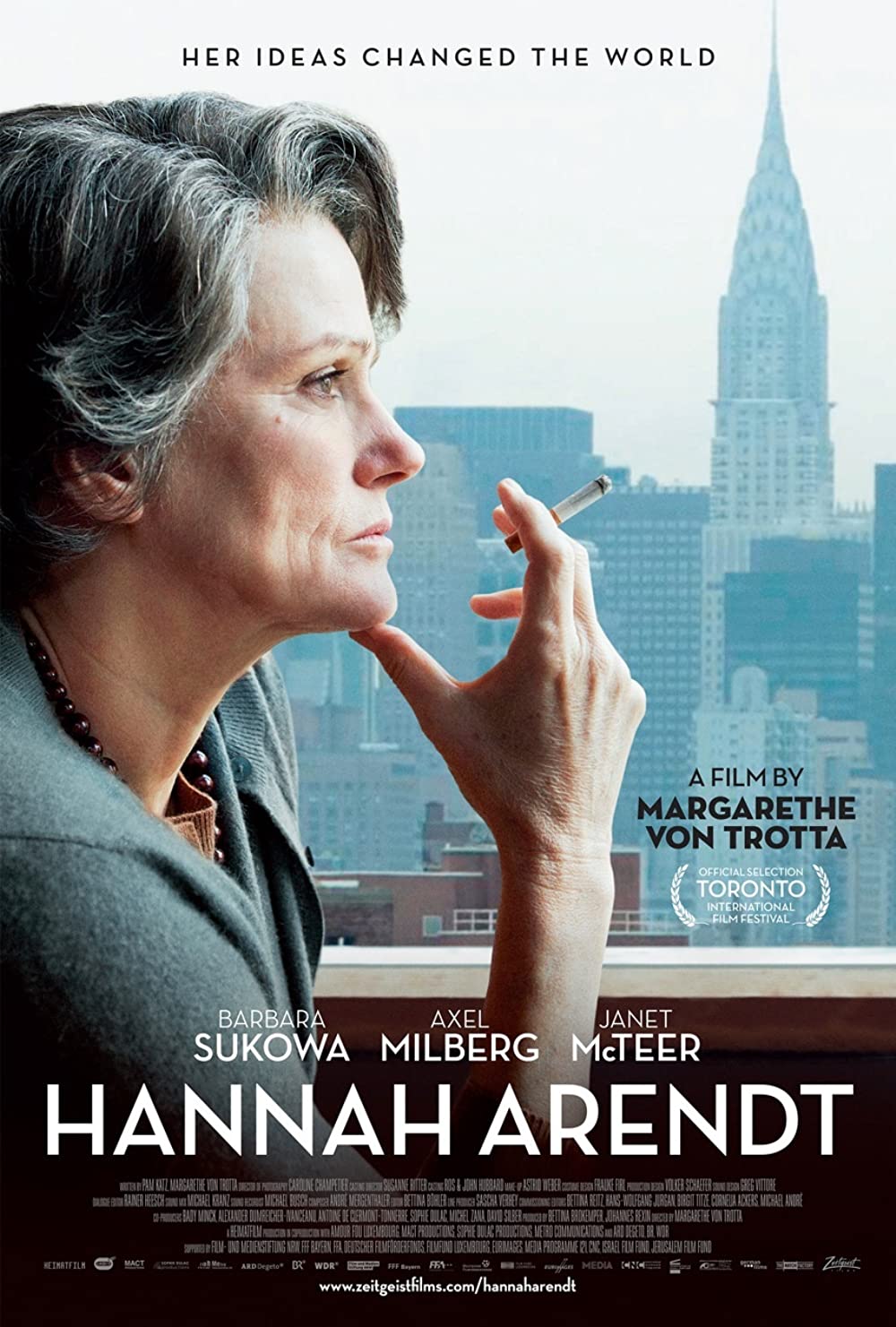} & 
\includegraphics[width=0.1\linewidth, height=0.145\linewidth]{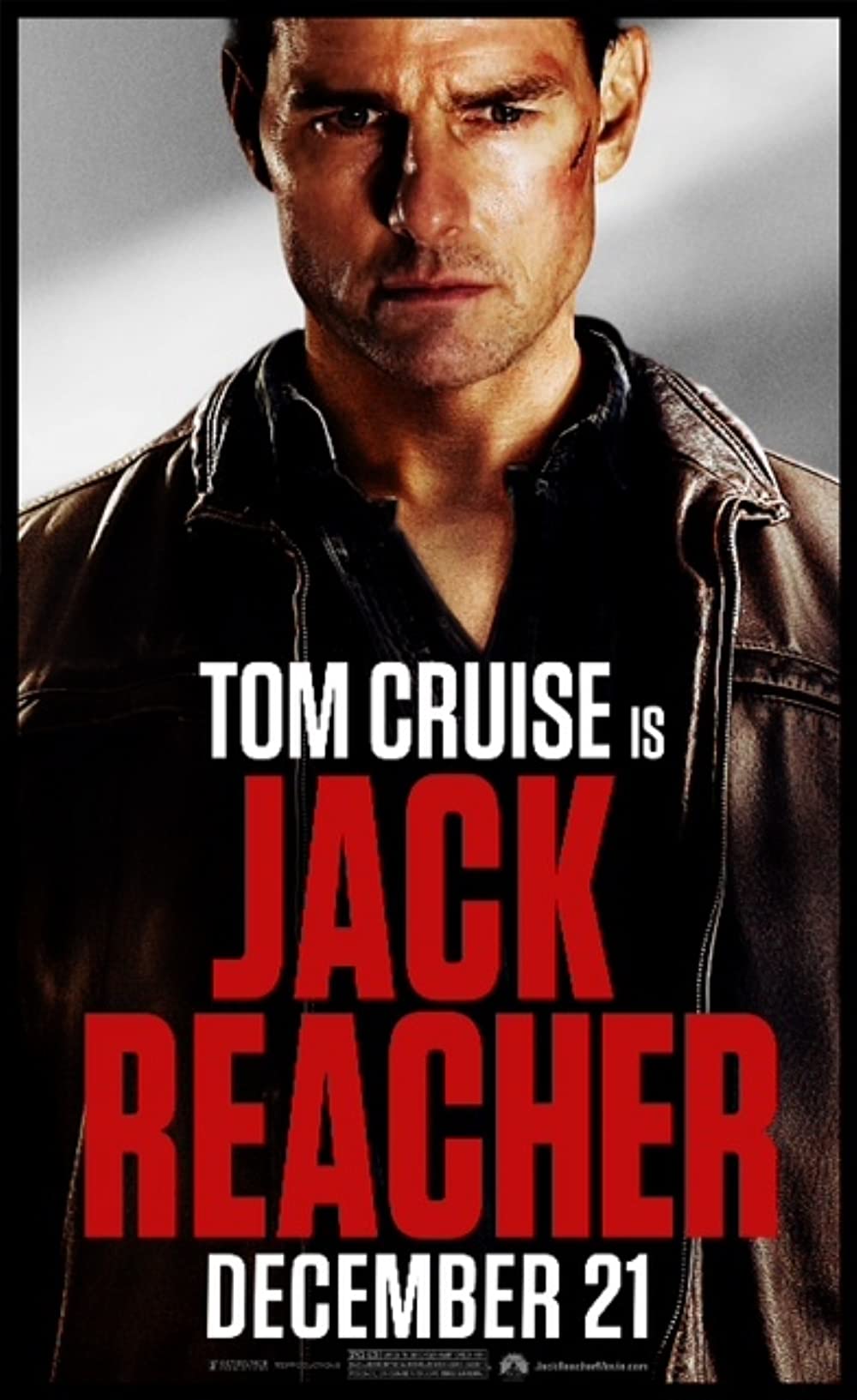} & 
\includegraphics[width=0.1\linewidth, height=0.145\linewidth]{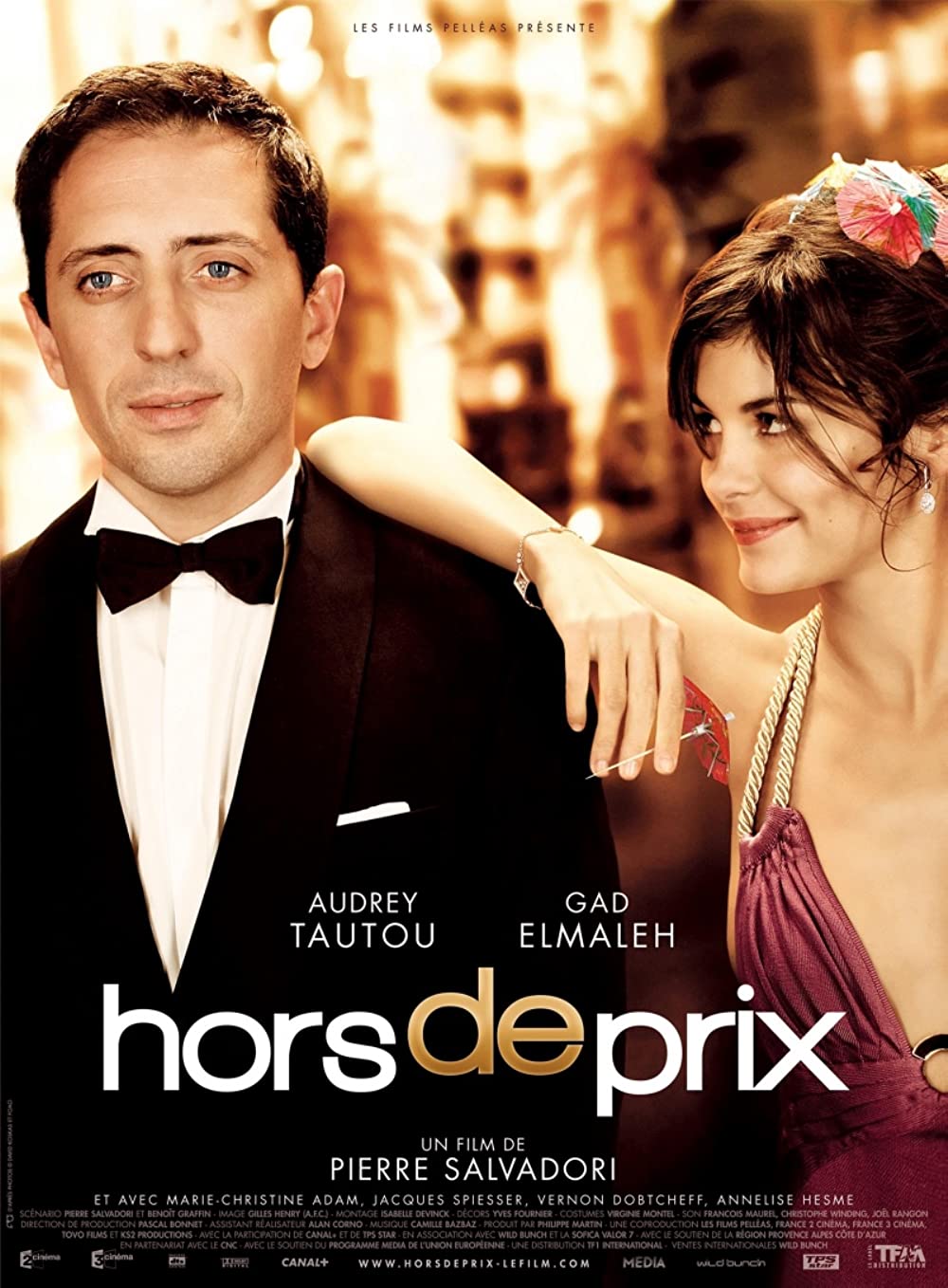} &
\includegraphics[width=0.1\linewidth, height=0.145\linewidth]{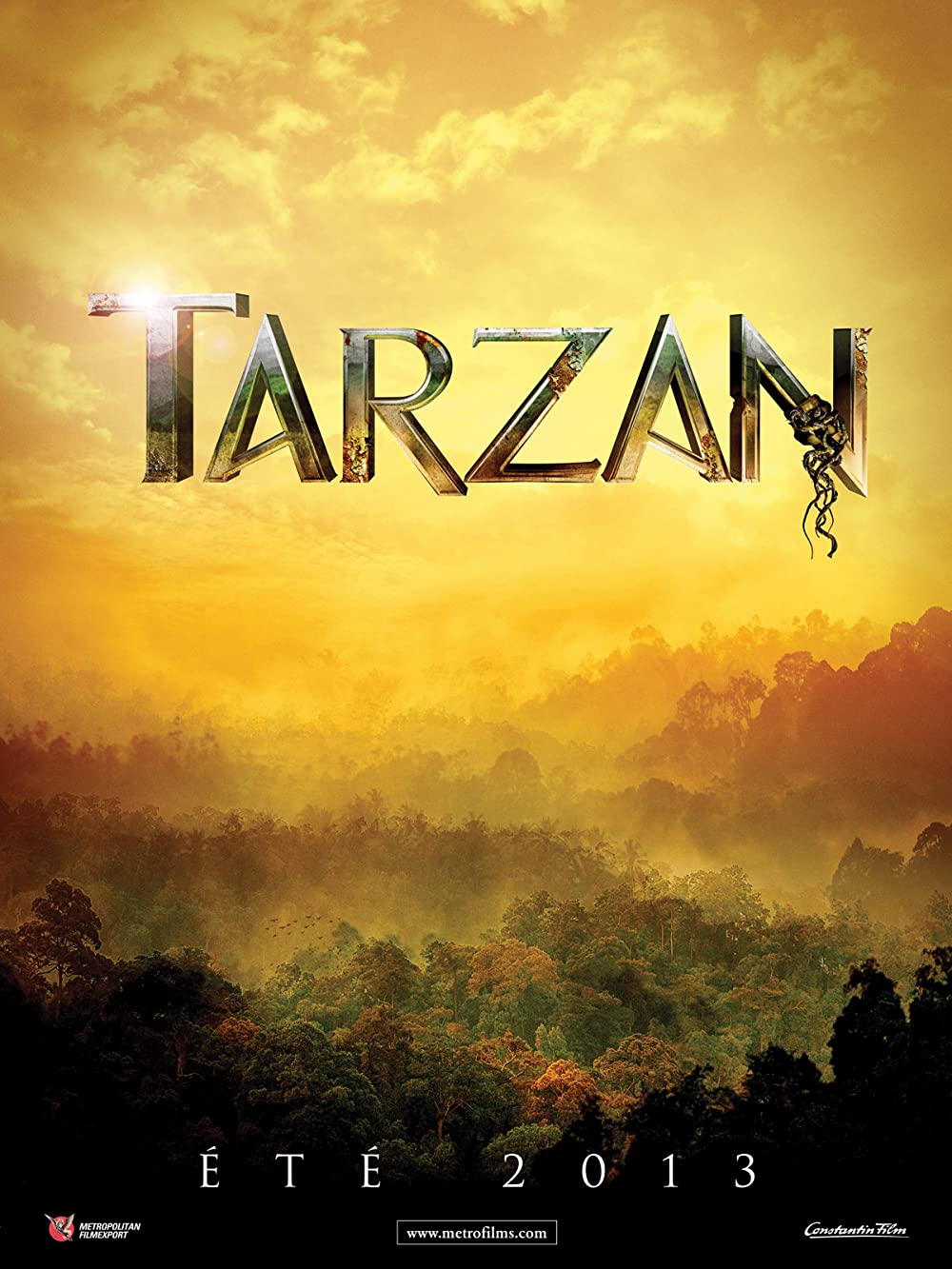} & 
\includegraphics[width=0.1\linewidth, height=0.145\linewidth]{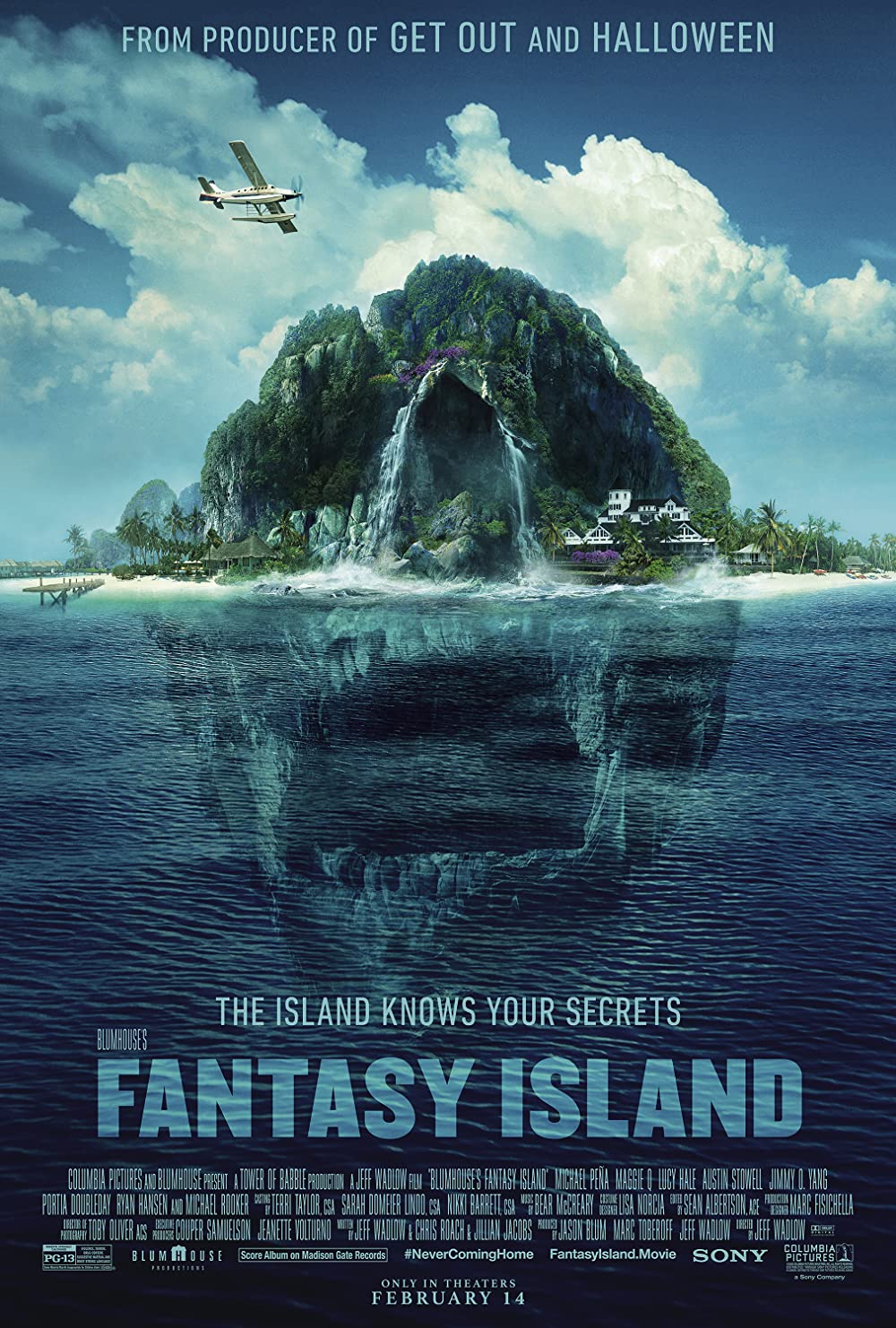} & 
\includegraphics[width=0.1\linewidth, height=0.145\linewidth]{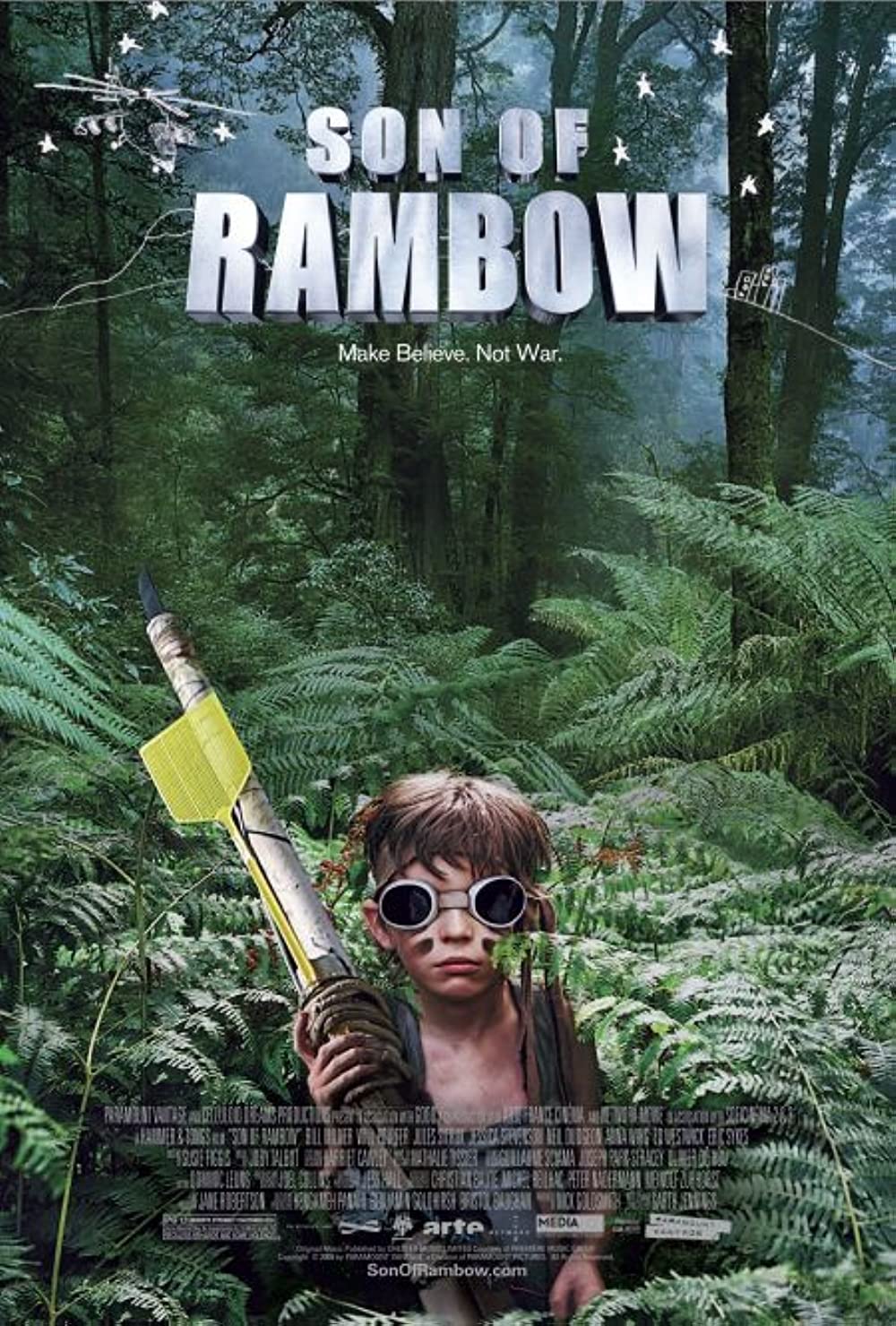} & 
\includegraphics[width=0.1\linewidth, height=0.145\linewidth]{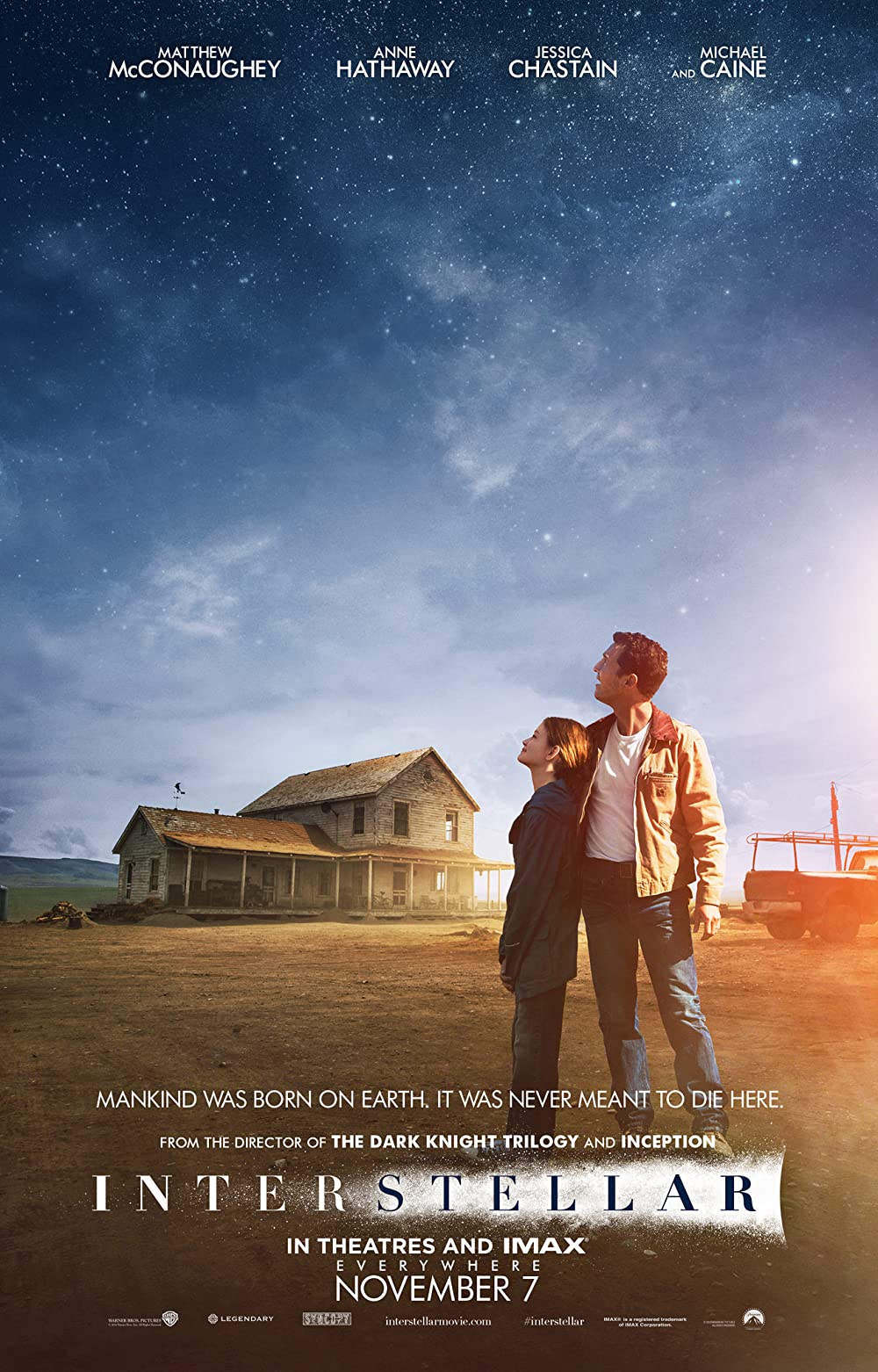} \\ 
\textcolor{blue}{\emph{ix.}} (1, 2, 8) & \textcolor{blue}{\emph{x.}}  (4, 7, -) & \textcolor{blue}{\emph{xi.}}  (1, 10, 13) & \textcolor{blue}{\emph{xii.}}  (11, 5, -) & 
\textcolor{blue}{\emph{xiii.}}  (3, 1, 2) & \textcolor{blue}{\emph{xiv.}}  (8, 9, 10) & \textcolor{blue}{\emph{xv.}}  (1, 2, 5) & \textcolor{blue}{\emph{xvi.}}  (2, 7, 12)  \\ 
\multicolumn{4}{c||}{cast image} & \multicolumn{2}{c|}{scene} & \multicolumn{2}{c}{cast + scene}\\ 
\hline 
&&& &&&&  \\[\dimexpr-\normalbaselineskip+1.5pt]
\includegraphics[width=0.1\linewidth, height=0.145\linewidth]{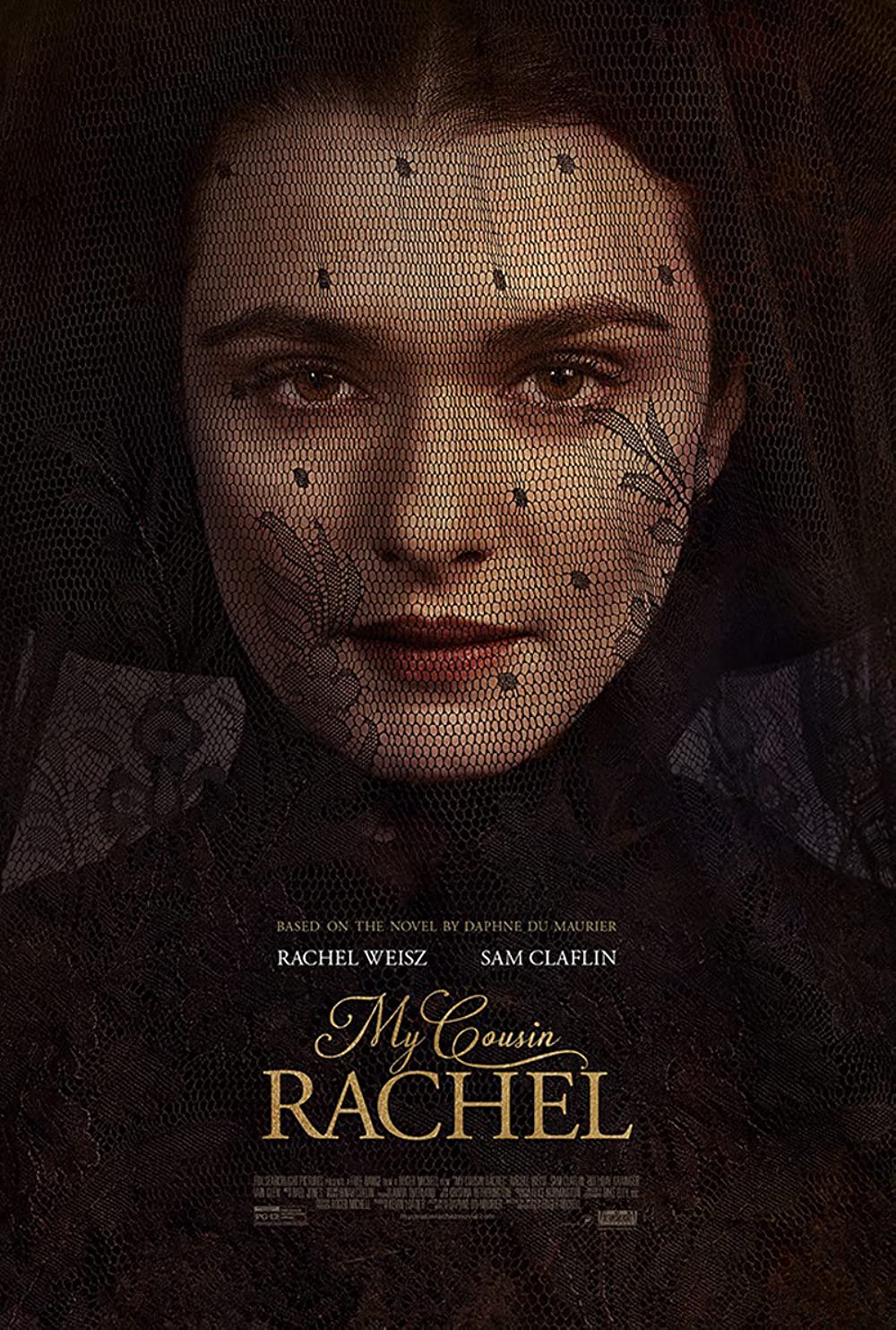} & 
\includegraphics[width=0.1\linewidth, height=0.145\linewidth]{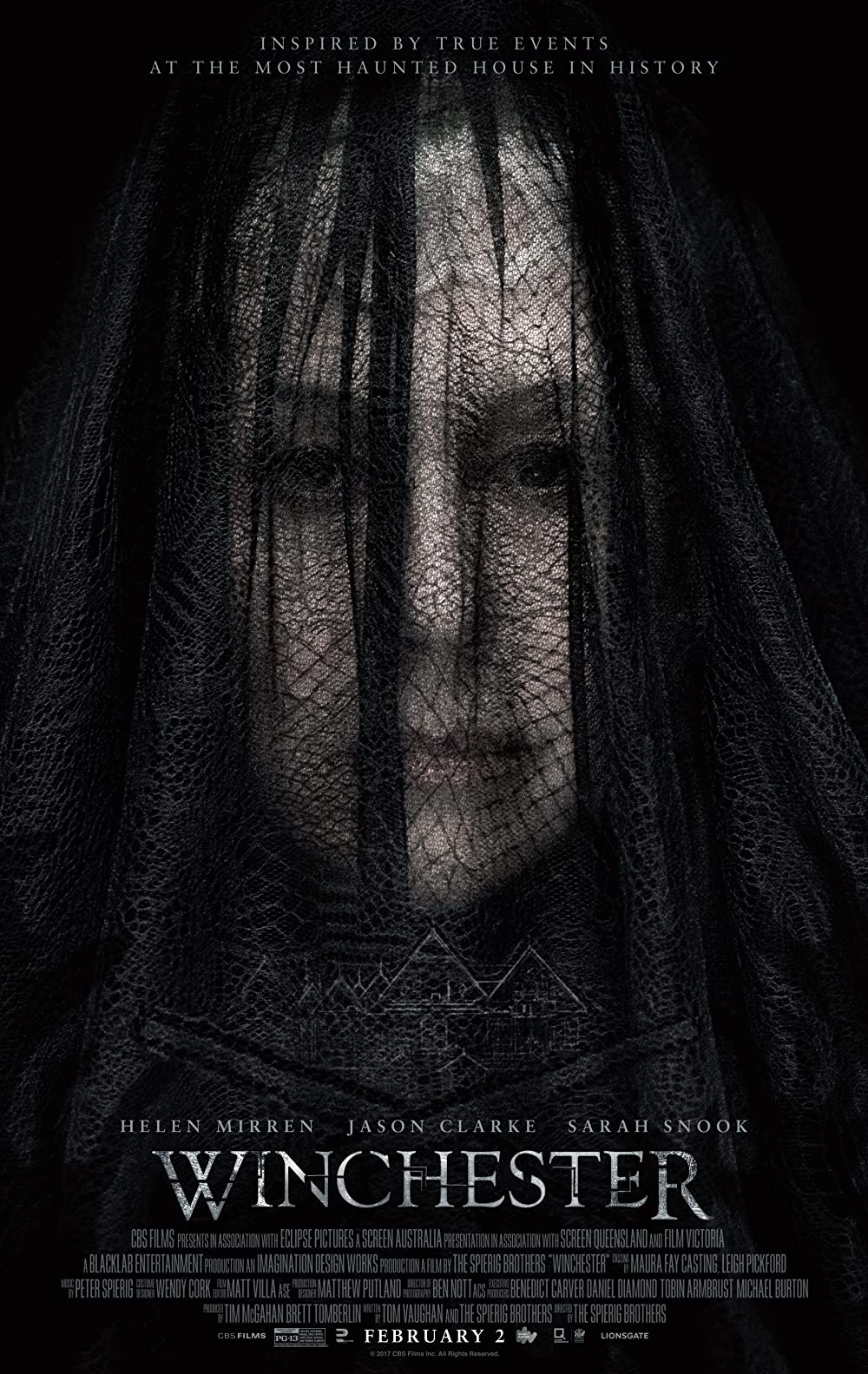} & 
\includegraphics[width=0.1\linewidth, height=0.145\linewidth]{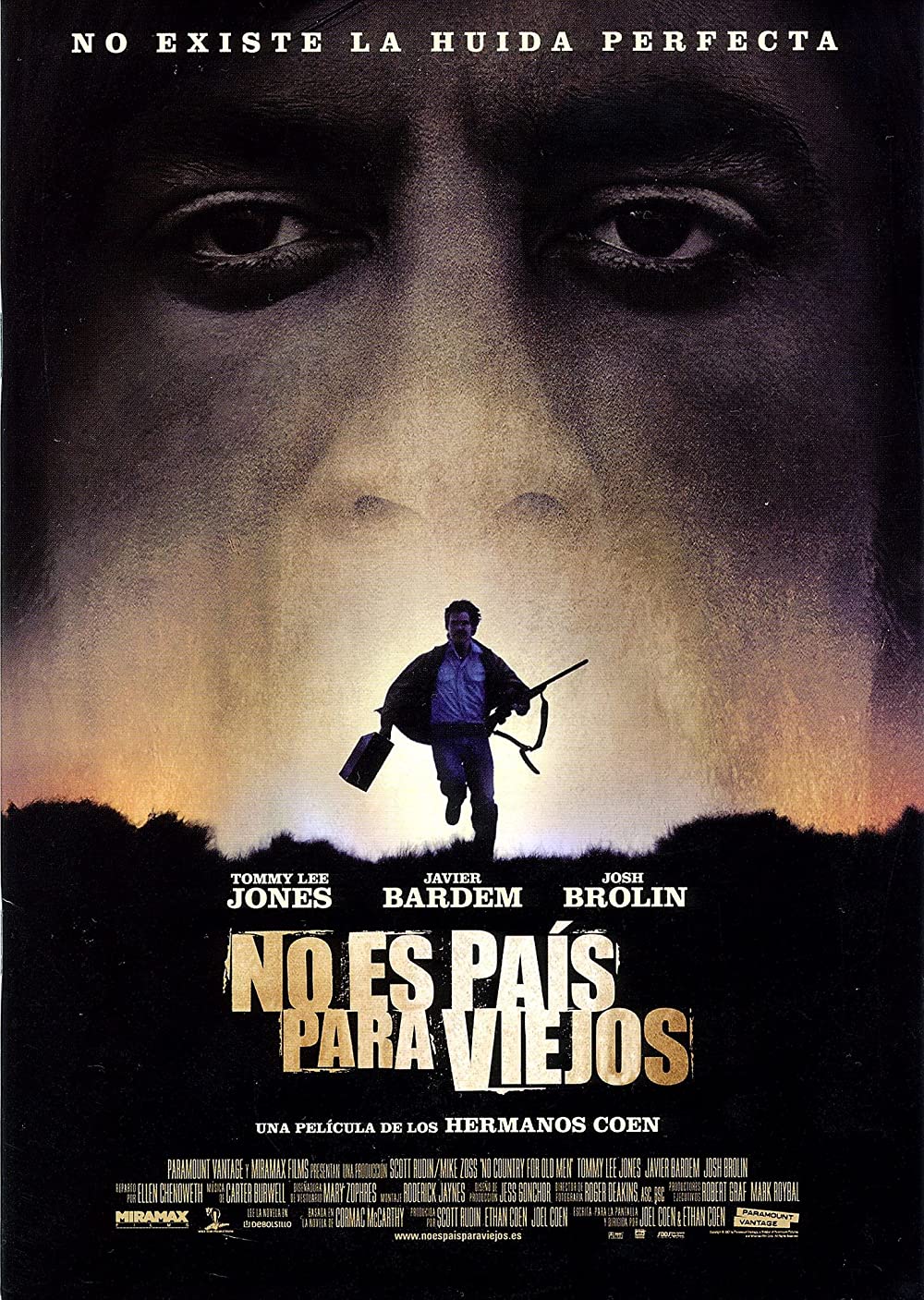} & 
\includegraphics[width=0.1\linewidth, height=0.145\linewidth]{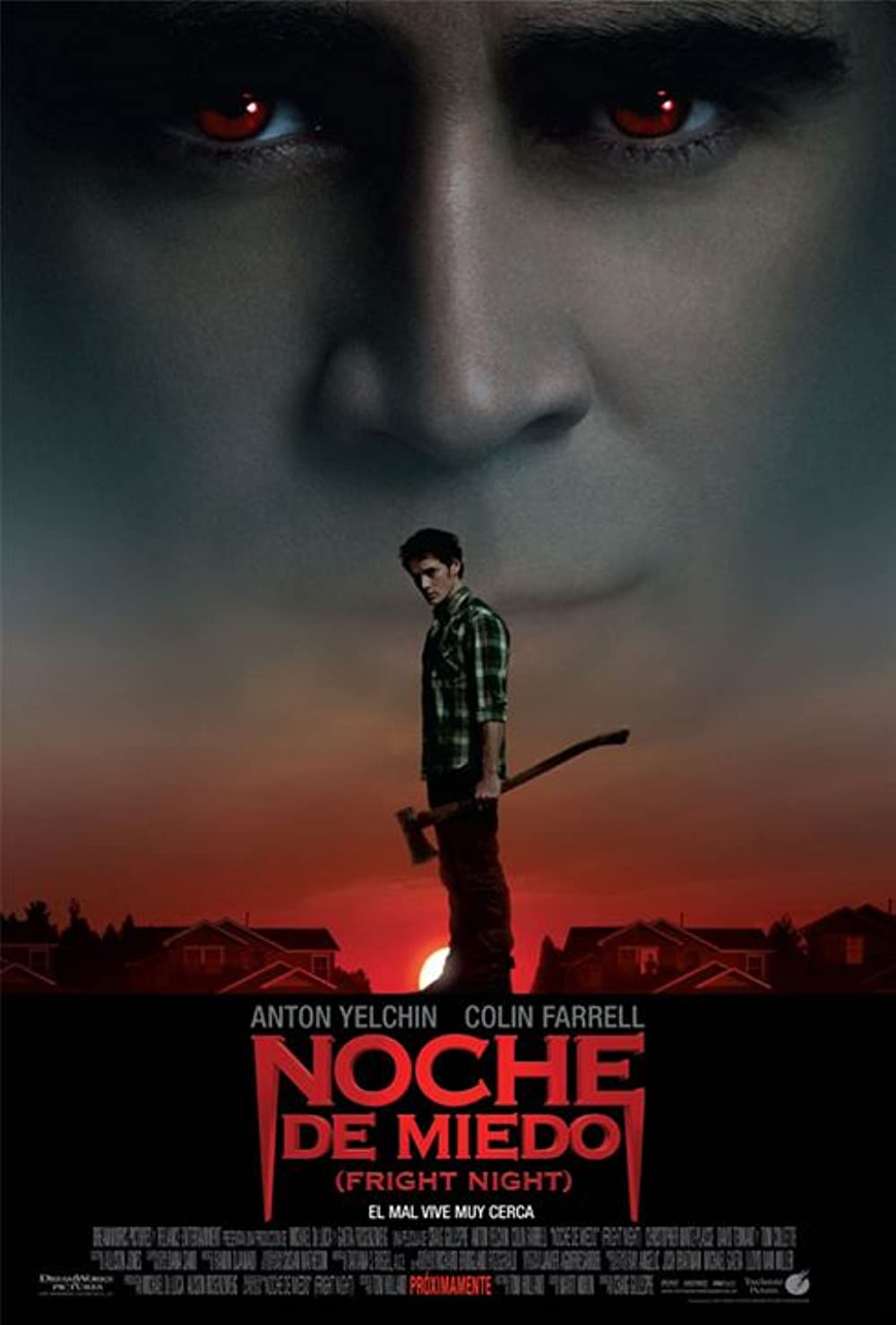} &
\includegraphics[width=0.1\linewidth, height=0.145\linewidth]{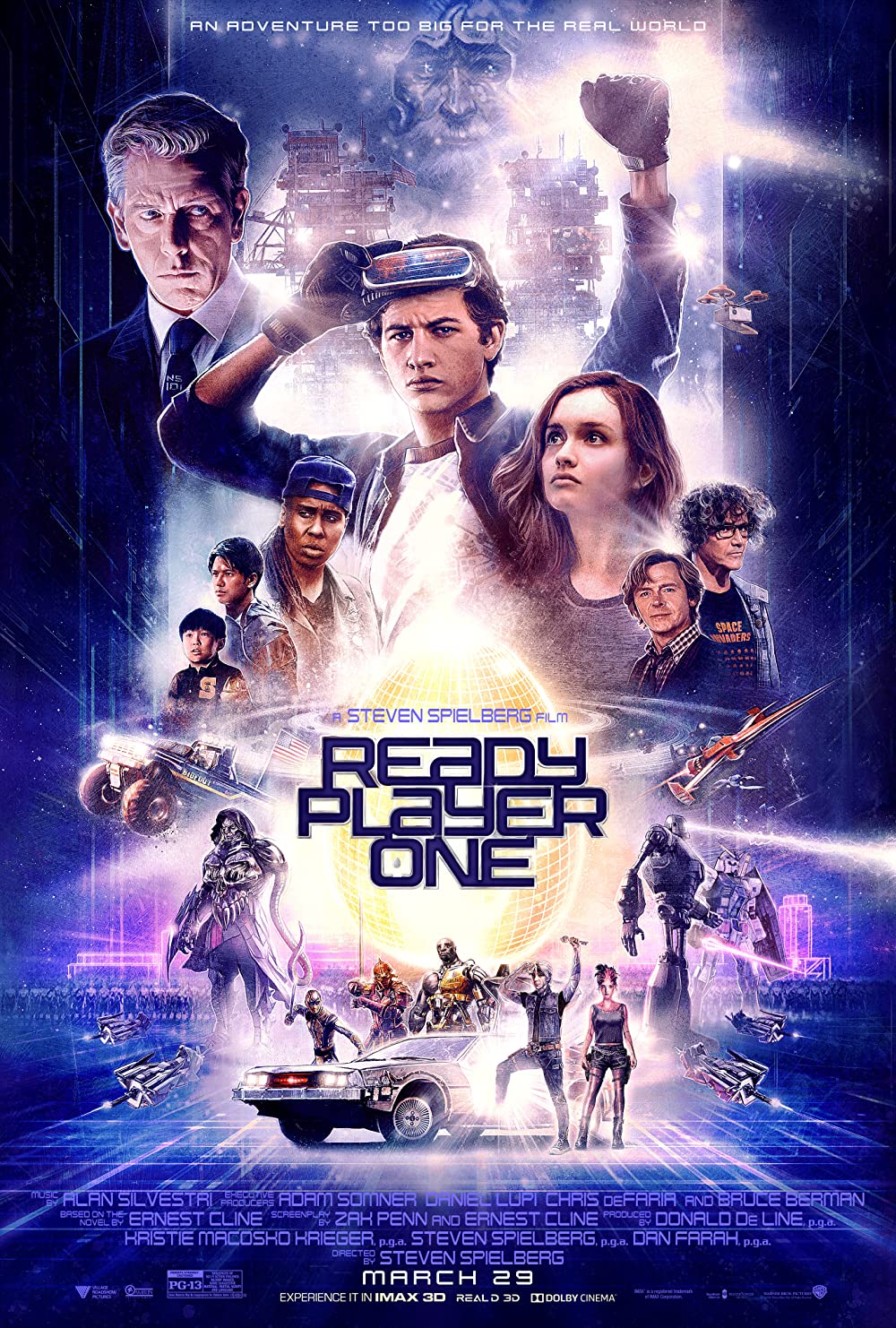} & 
\includegraphics[width=0.1\linewidth, height=0.145\linewidth]{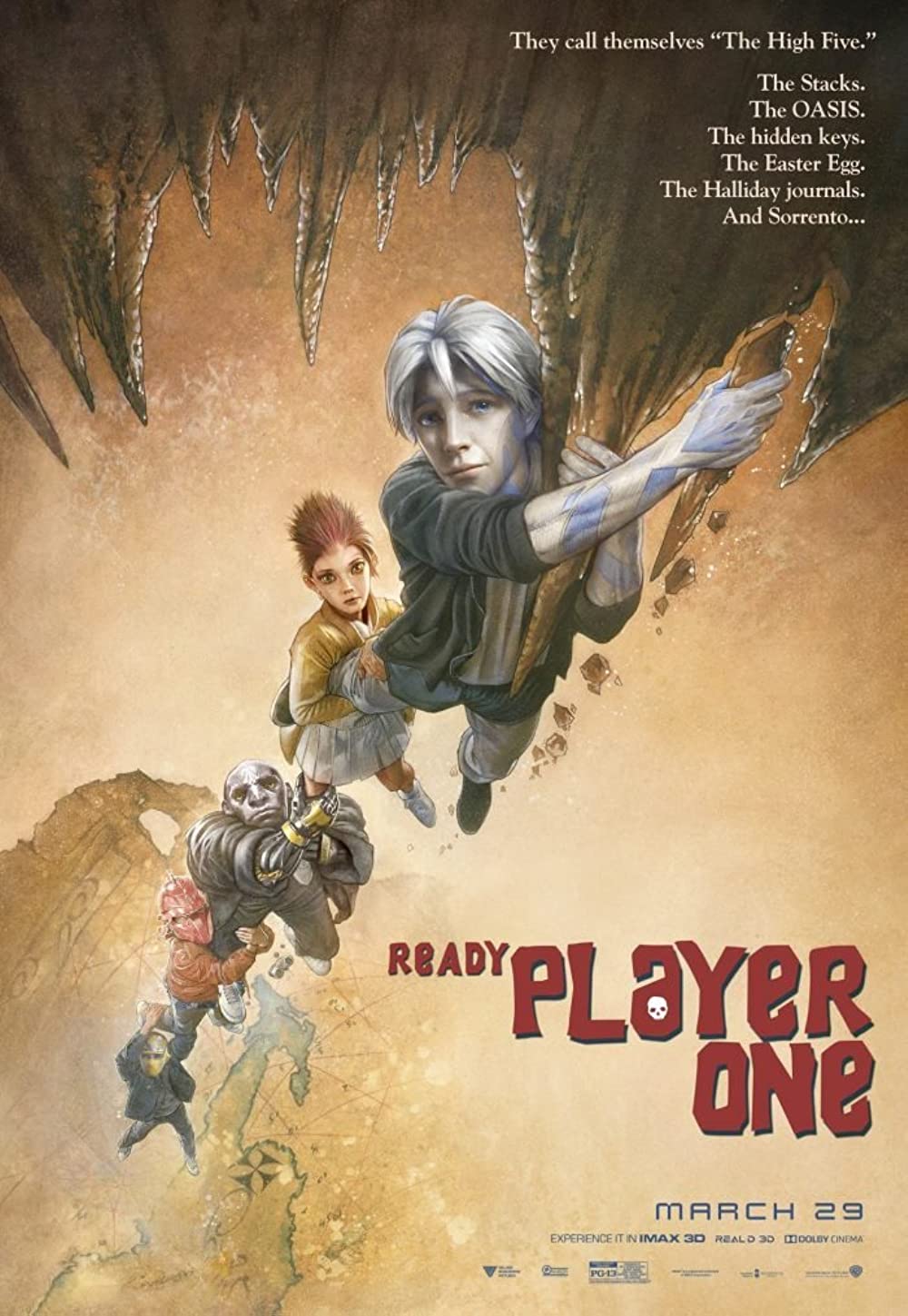} & 
\includegraphics[width=0.1\linewidth, height=0.145\linewidth]{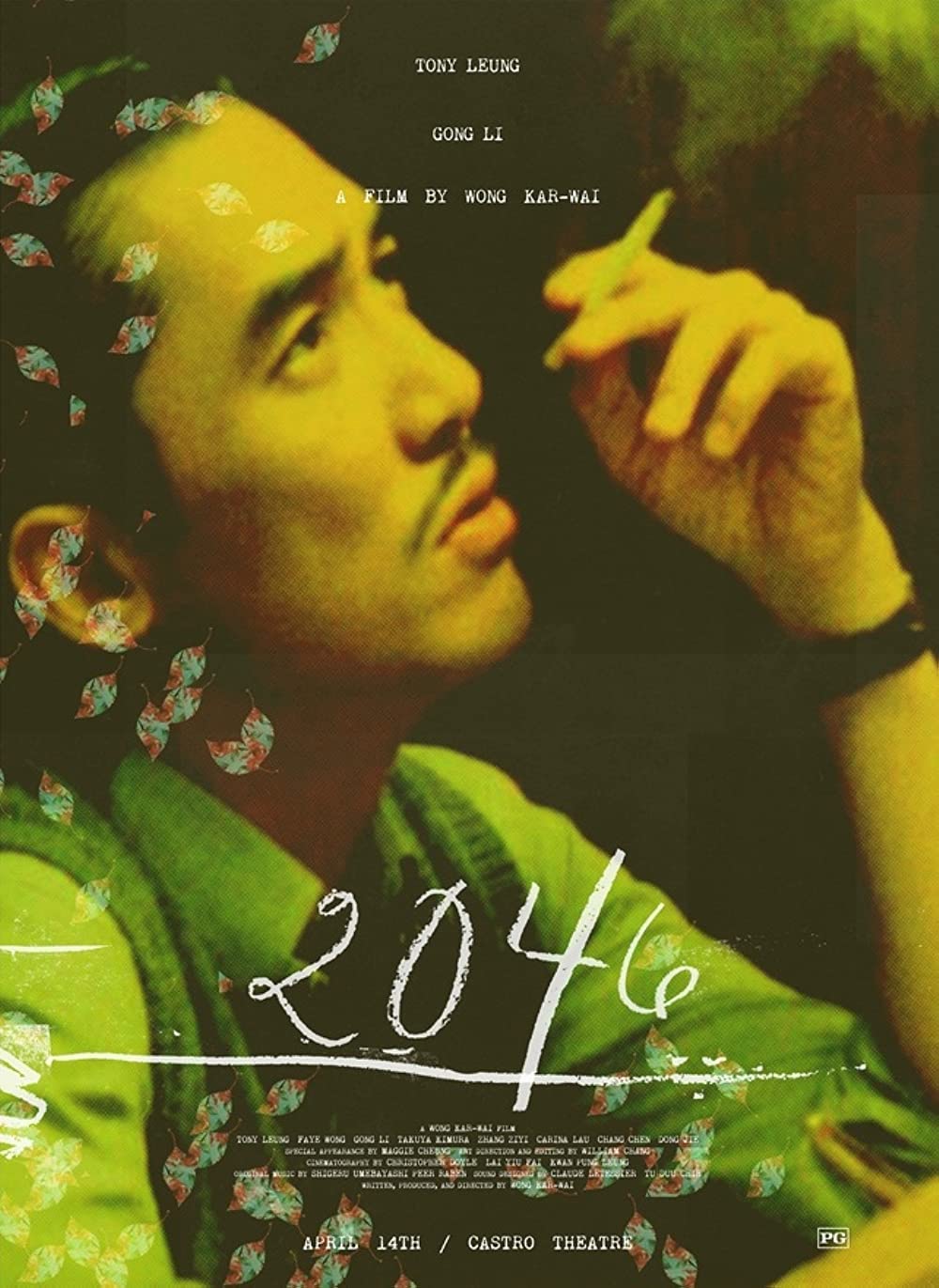} & 
\includegraphics[width=0.1\linewidth, height=0.145\linewidth]{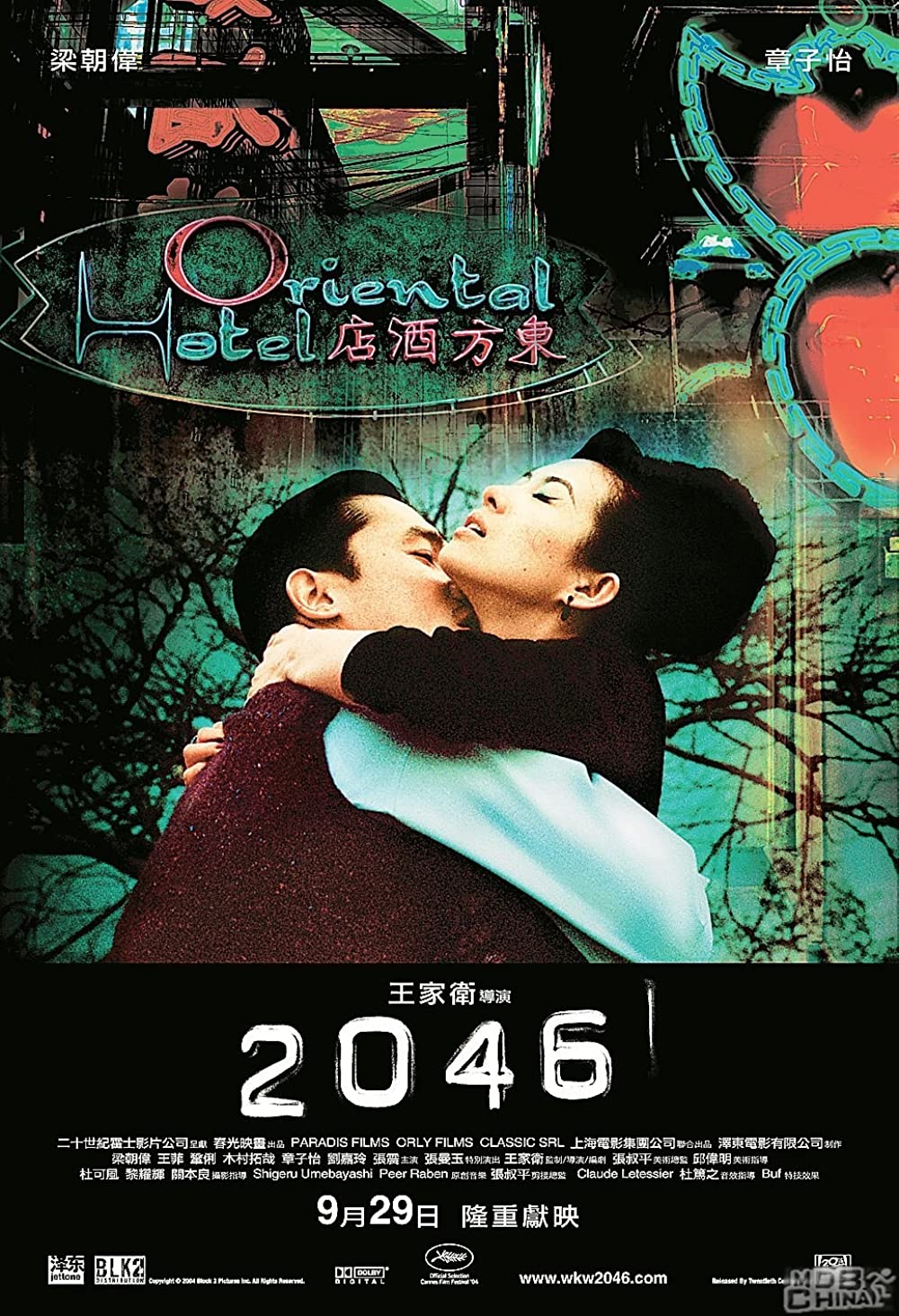} \\ 
\textcolor{blue}{\emph{xvii.}} (10, 11, 7) & \textcolor{blue}{\emph{xviii.}}  (9, 8, 7) & \textcolor{blue}{\emph{xix.}}  (6, 7, 13) & \textcolor{blue}{\emph{xx.}}  (5, 9, -) & 
\textcolor{blue}{\emph{xxi.}}  (1, 2, 12) & \textcolor{blue}{\emph{xxii.}}  (1, 2, 12) & \textcolor{blue}{\emph{xxiii.}}  (7, 11, 12) & \textcolor{blue}{\emph{xxiv.}}  (7, 11, 12)  \\ 
\multicolumn{2}{c|}{inter-variation} & \multicolumn{2}{c||}{inter-variation} & \multicolumn{2}{c|}{intra-variation} & \multicolumn{2}{c}{intra-variation}\\
\hline 
&&& &&&&  \\[\dimexpr-\normalbaselineskip+1.5pt]
\includegraphics[width=0.1\linewidth, height=0.145\linewidth]{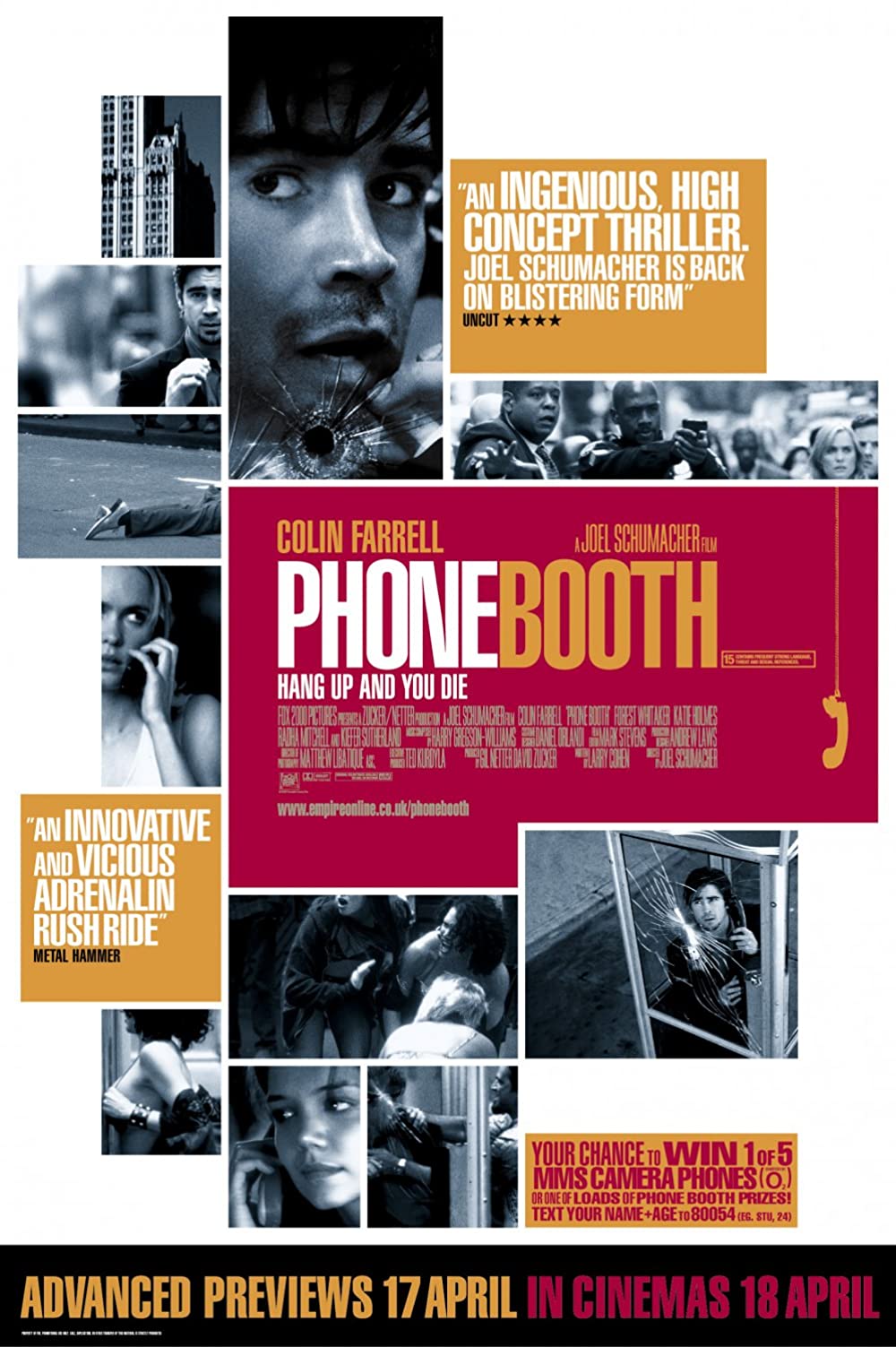} & 
\includegraphics[width=0.1\linewidth, height=0.145\linewidth]{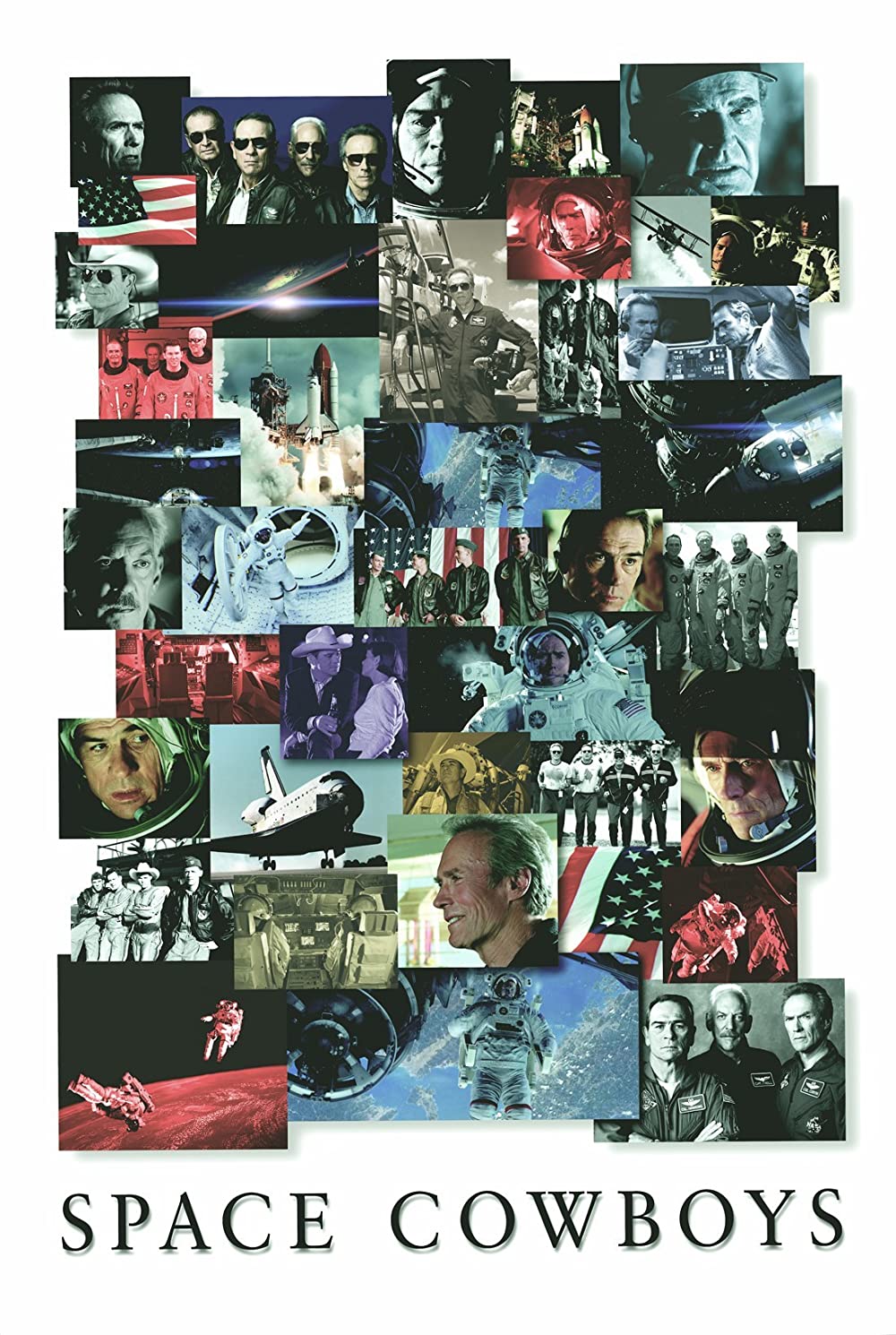} & 
\includegraphics[width=0.1\linewidth, height=0.145\linewidth]{figs/challenge/27.jpg} & 
\includegraphics[width=0.1\linewidth, height=0.145\linewidth]{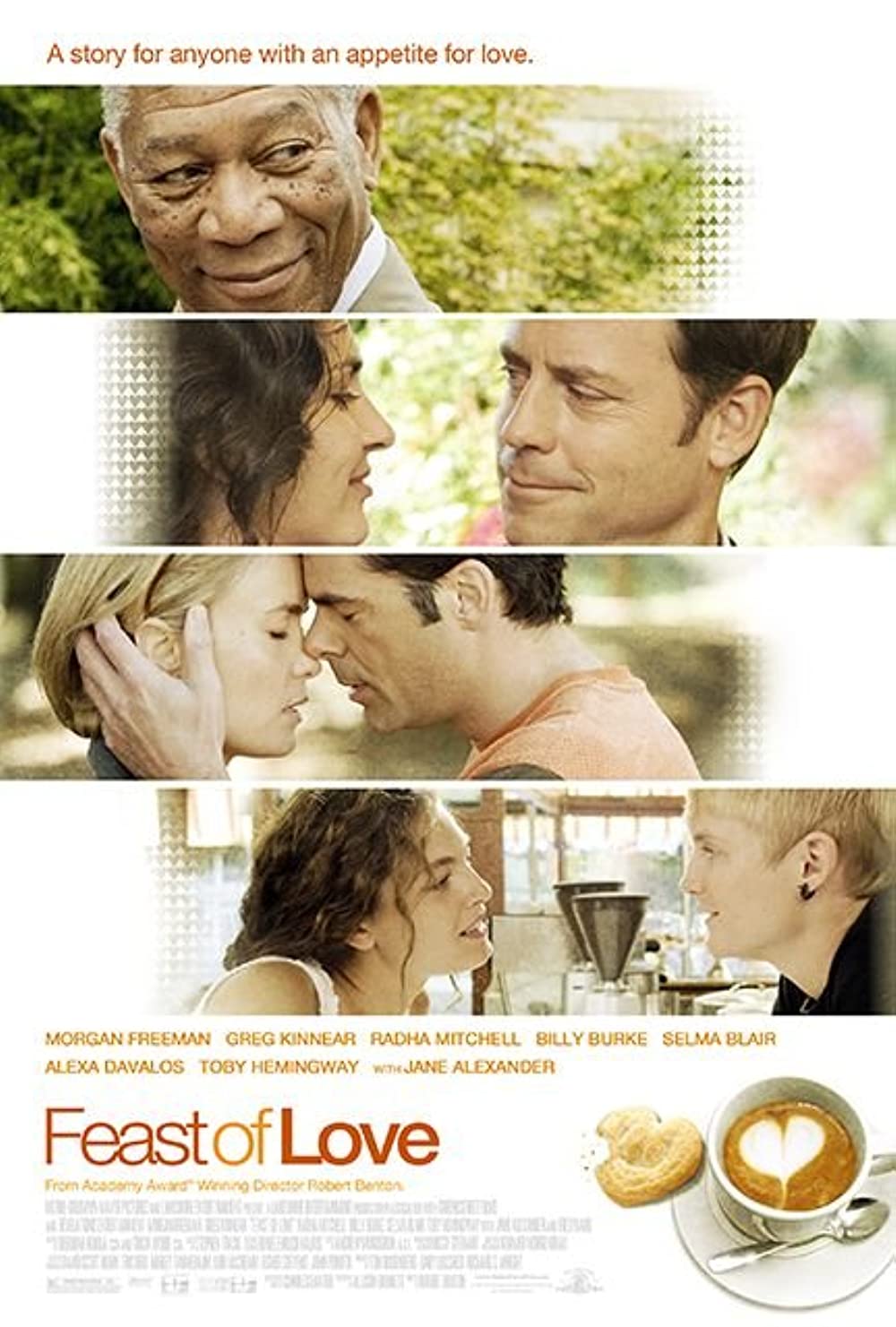} &
\includegraphics[width=0.1\linewidth, height=0.145\linewidth]{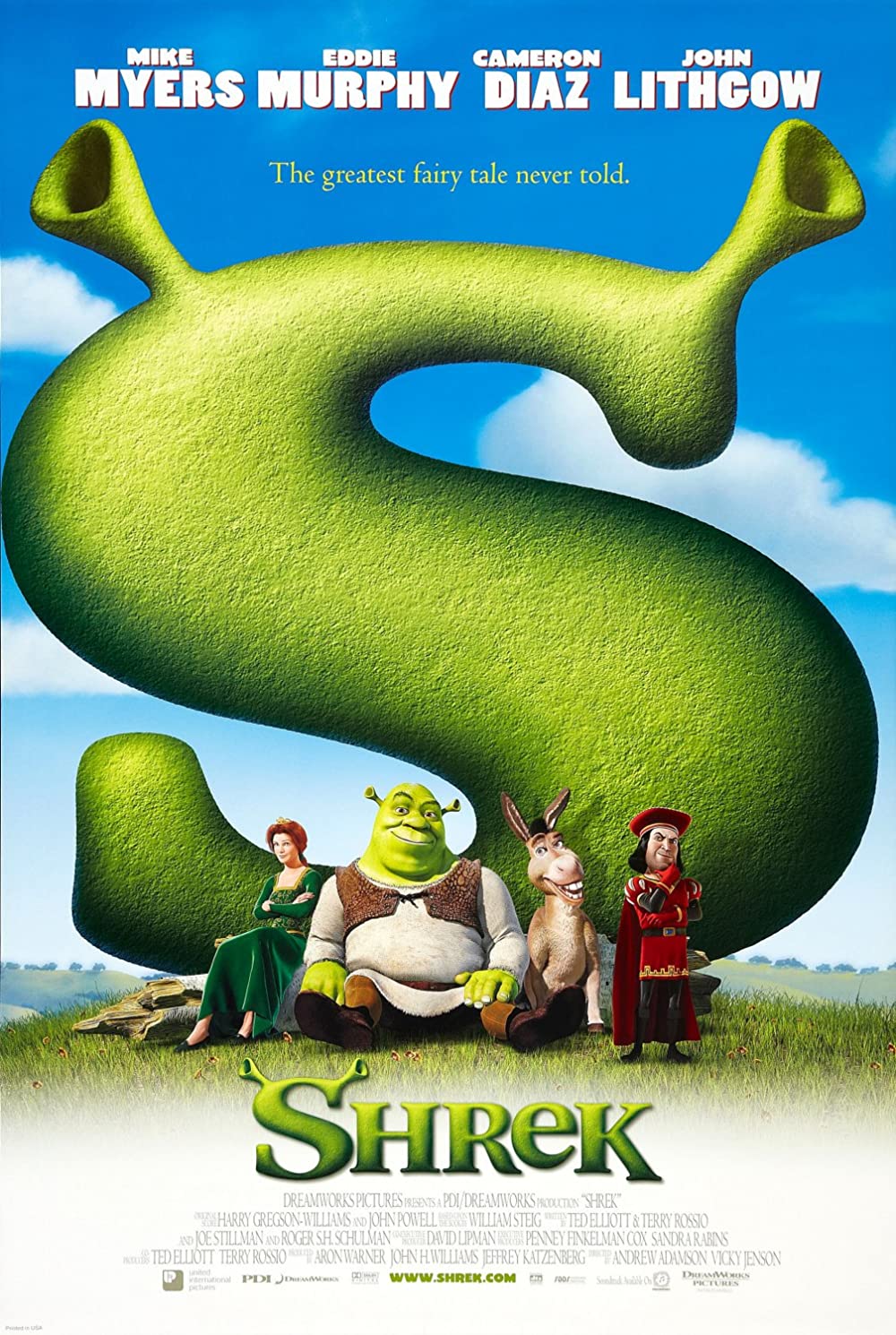} & 
\includegraphics[width=0.1\linewidth, height=0.145\linewidth]{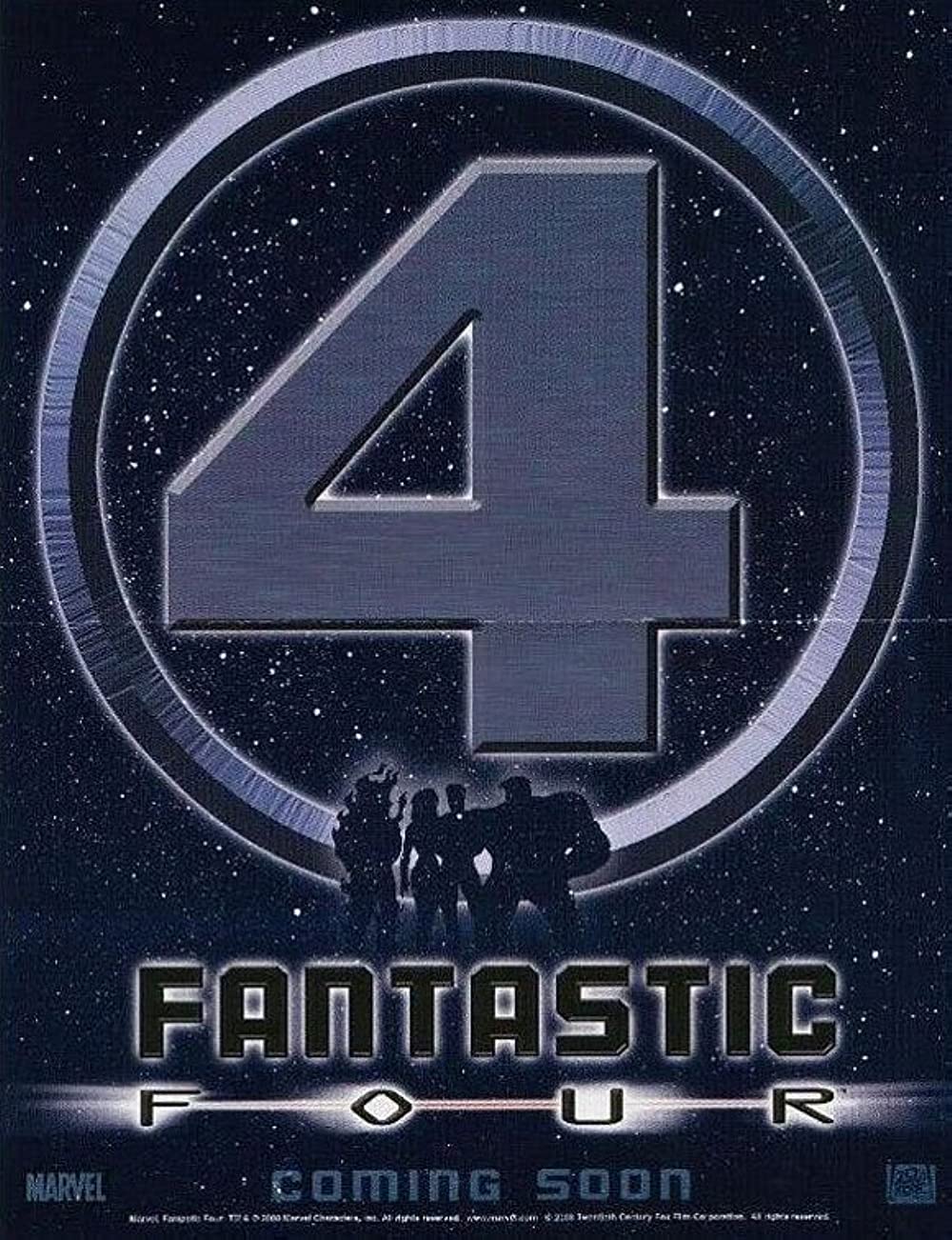} & 
\includegraphics[width=0.1\linewidth, height=0.145\linewidth]{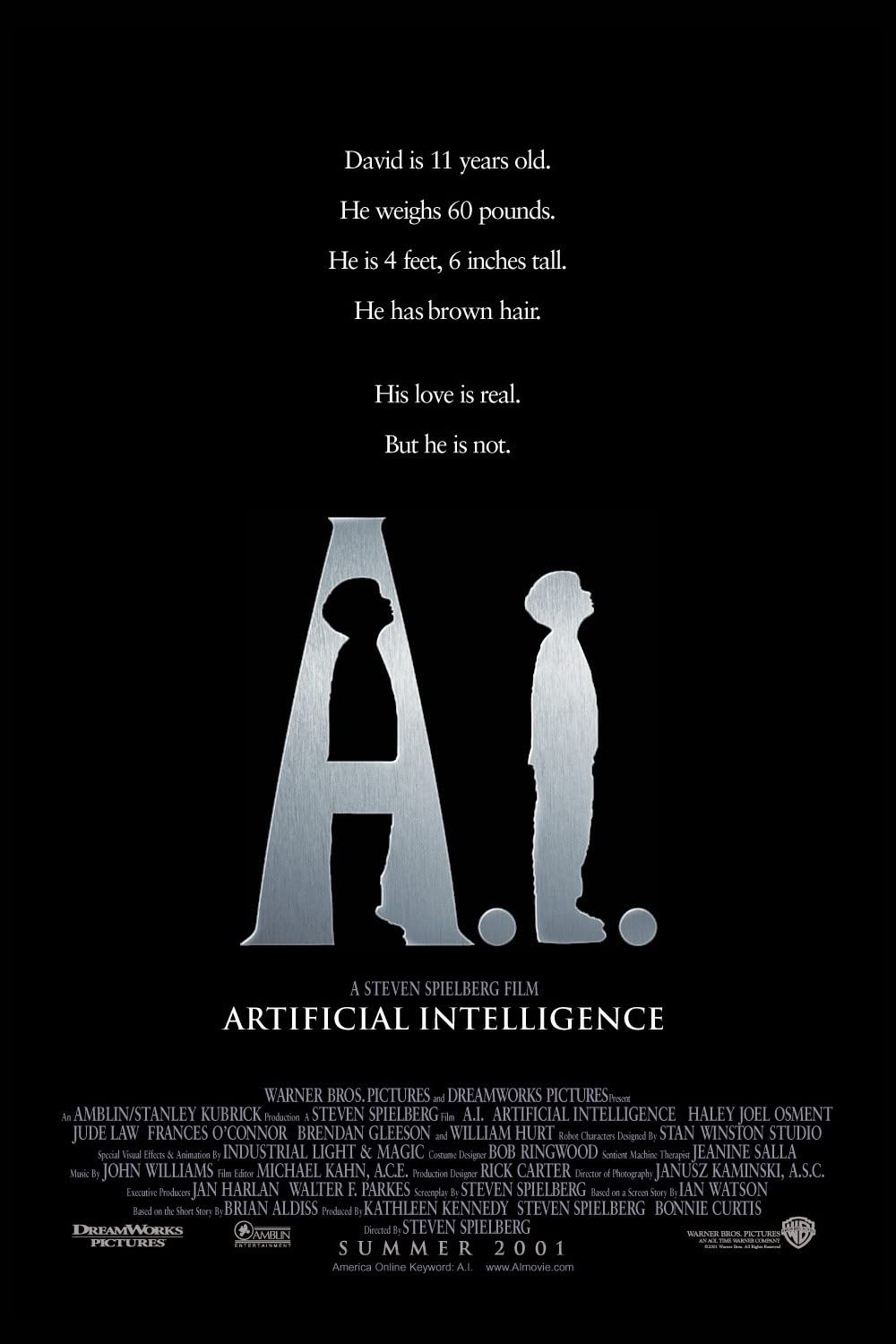} & 
\includegraphics[width=0.1\linewidth, height=0.145\linewidth]{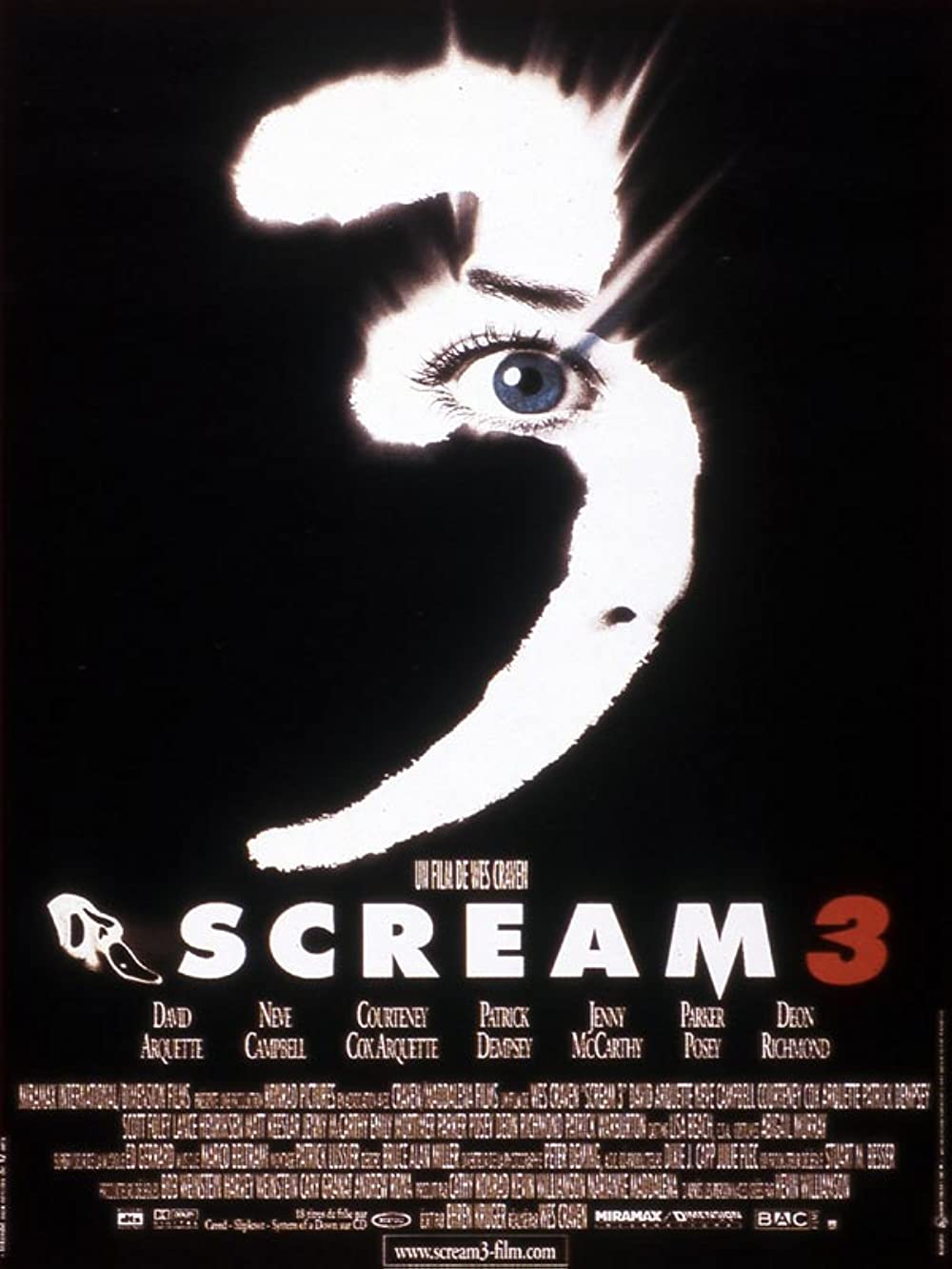} \\ 
\textcolor{blue}{\emph{xxv.}} (6, 13, -) & \textcolor{blue}{\emph{xxvi.}}  (1, 2, 13) & \textcolor{blue}{\emph{xxvii.}}  (6, 10, 13) & \textcolor{blue}{\emph{xxviii.}}  (7, 11, 12) & 
\textcolor{blue}{\emph{xxix.}}  (3, 2, 5) & \textcolor{blue}{\emph{xxx.}}  (1, 2, 8) & \textcolor{blue}{\emph{xxxi.}} (7, 12, -) & \textcolor{blue}{\emph{xxxii.}}  (9, 10, -) \\ 
\multicolumn{4}{c||}{collage} & \multicolumn{2}{c|}{text as image} & \multicolumn{2}{c}{image as text}\\
\hline
\end{tabular}
\caption{Examples of some challenging poster images with mentioned issues;  
underneath each image, within the first bracket, the genre class ids 
(refer to Table II of main manuscript) 
are also shown}
\label{fig:challenge}
\end{figure*}

\noindent
As mentioned in Section 
I of the main manuscript, 
the information contained within the posters introduces additional complexities when it comes to identifying the movie genre, and we briefly outline these challenges as follows.

\emph{(a) Background}:
The background of movie posters serves a significant purpose in establishing a sense of atmosphere or setting, piques curiosity, and helps individuals to make informed decisions about whether the movie aligns with their interests and preferences.
For example, \emph{horror} movie posters (Fig. \ref{fig:challenge}: \emph{v, xviii}) often utilize specific background elements with dark shadows and eerie lighting to entice the viewers with a chilling, suspenseful, and frightening mood. 
Based on the information available on the poster background, we can further categorize it as below:

    \begin{itemize}[---]
        \item \emph{Less information:} 
        Sometimes, a poster background may contain little to no information, which brings challenges to automated genre identification (Fig.s  \ref{fig:challenge}: \emph{i-iv}).
        
        \item \emph{Moderate \& adequate information:} 
        Often, the background has sufficient visual characteristics to convey its genre 
        (Fig.s \ref{fig:challenge}: \emph{v-vi}).
        
        \item \emph{Complex background:}
        In some cases, the background of a poster becomes complex due to having enormous and/or composite visual effects/elements (Fig.s  \ref{fig:challenge}: \emph{vii-viiii}).
    \end{itemize}

\emph{(b) Foreground}:
The foreground in a poster plays a crucial role in capturing the viewer's attention and creating visual interest. By strategically placing dynamic foreground elements, such as the main characters or key plot elements, the poster can effectively convey the theme or atmosphere of the movie. In the case of a \emph{romantic}, \emph{comedy}, featuring the two leads in a playful stance in the foreground may help in establishing the genre and the central focus of the film, which is the relationship between those characters (Fig. \ref{fig:challenge}: \emph{xii}). 

    \begin{itemize}[---]
        \item \emph{Cast image:} Often, the lead casts' portrait, full/half body images cover the entire poster (Fig.s \ref{fig:challenge}: \emph{ix-xii}), which makes our task challenging due to relying only upon visual elements without taking any aid from face recognition and object detection modules. 

        \item \emph{Scene image:} The scene images present in the entire poster sometimes brings challenges due to complex visual elements, visual clutter and lack of cohesive composition (Fig.s  \ref{fig:challenge}: \emph{xiii-xiv}).
        
        \item \emph{Cast in Scene:} The hybridization of cast and scene images can also be observed, where the cast image may be fused with scene images (Fig.s  \ref{fig:challenge}: \emph{xv-xvi}).         
    \end{itemize}

\emph{(c) Inter variance}: 
Often, different movie posters may have visual similarities while belonging to diverse genres. 
This can be done to challenge audience expectations, create intrigue, or highlight genre mashups. Such instances make the genre identification task quite difficult. 
For example, movie genre of Fig.s  \ref{fig:challenge}: \emph{xviii} poster is \emph{horror}, but 
Fig. \ref{fig:challenge}: \emph{xvii} does not, although visually quite similar; similarly, Fig.s  \ref{fig:challenge}: \emph{xix} and \emph{xx} show non-identical genres, while sharing similar visual elements.

\emph{(d) Intra variance}:
Generally, a movie has multiple posters of various designs, where different designs may emphasize multiple aspects of the same movie to effectively market it to a diverse range of viewers, which brings additional challenges in identifying the genre. 
For example, 
Fig.s  \ref{fig:challenge}: \emph{xxi-xxii} are the posters from the same movie; however, they show variation in visual elements.
Similarly, Fig.s \ref{fig:challenge}: \emph{xxiii-xxiv} show intra-variation.

\emph{(e) Collage}: 
Sometimes, a movie poster combines various instances of the abovementioned background and foreground information, and creates a collage made from tiny images. Such collage posters make genre identification challenging due to amalgamating a pool of information (Fig.s  \ref{fig:challenge}: \emph{xxv-xxviii}).

\emph{(f) Text $\leftrightarrow$ Image}:
In certain movie posters, some texts or titles are sometimes displayed as images rather than traditional typography 
(Fig.s  \ref{fig:challenge}: \emph{xxix-xxxii}). 
This artistic approach is often used to convey a specific theme or style associated with the movie. This brings additional challenges to our task, since we are not taking any aid from the text recognition module.


\section{Qualitative result: Heat map encoding}
\label{app:heatmap}

\begin{figure*}
\centering
\begin{adjustbox}{width=0.8\textwidth}
\begin{tabular}{c|c|c|c|c|c|c|c|c|c|c|c|c|c | | c|c|c|c|c|c|c|c|c|c|c|c|c|}
\cline{2-27} 
& \multicolumn{13}{c||}{Ground-truth heat-map of multi-hot encoding} 
& \multicolumn{13}{c}{ERDT predicted confidence score heat-map encoding} 
\\ \hline
{Class ID} & {1} & {2} & {3} & {4} & {5} & {6} & {7} & {8} & {9} & {10} & {11} & {12} & {13} & {1} & {2} & {3} & {4} & {5} & {6} & {7} & {8} & {9} & {10} & {11} & {12} & {13} \\ \hline

\includegraphics[width=0.015\linewidth, height=0.018\linewidth]{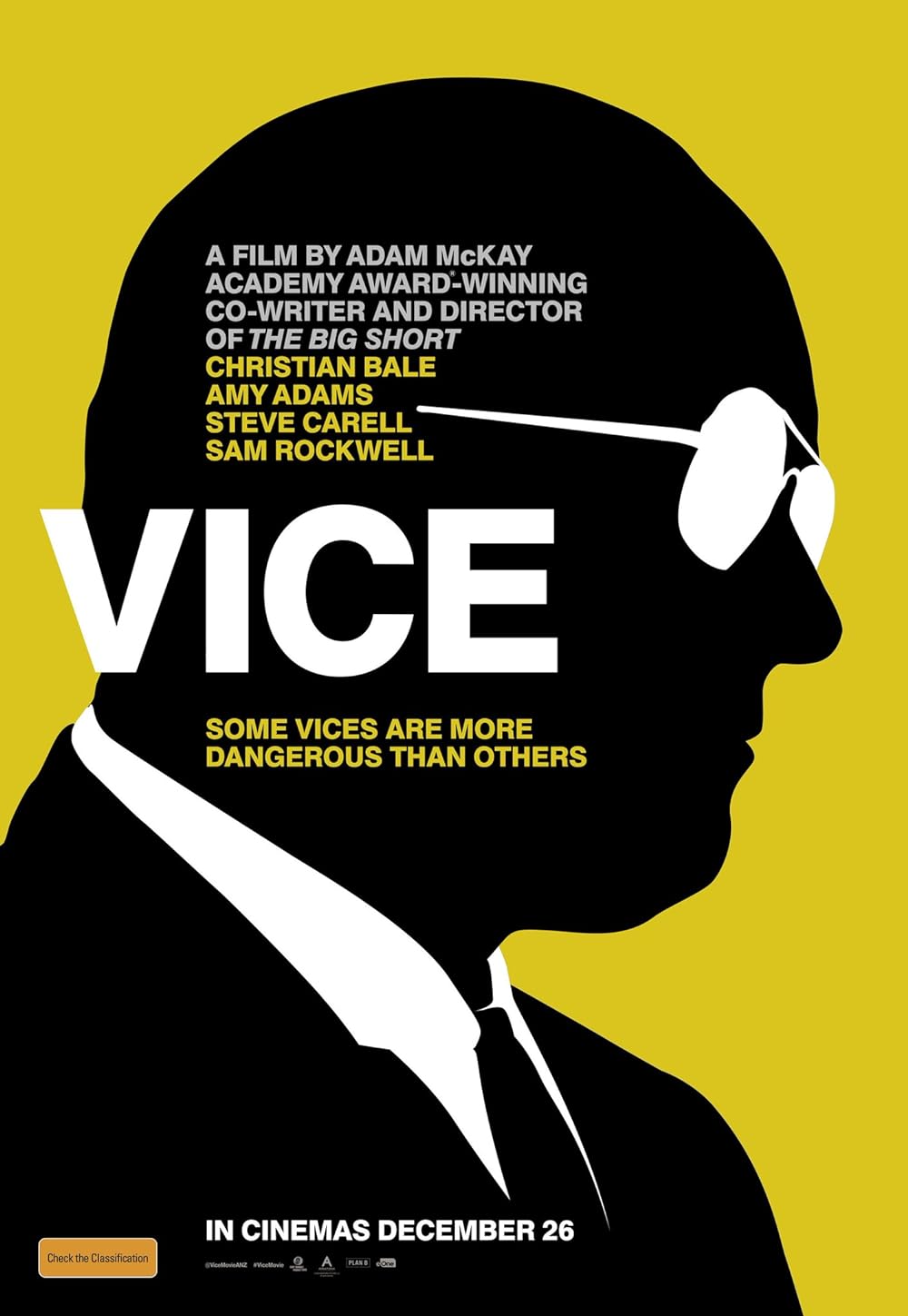} & \cellcolor[gray]{0} & \cellcolor[gray]{0} & \cellcolor[gray]{0} & \cellcolor[gray]{1} & \cellcolor[gray]{1} & \cellcolor[gray]{0} & \cellcolor[gray]{1} & \cellcolor[gray]{0} & \cellcolor[gray]{0} & \cellcolor[gray]{0} & \cellcolor[gray]{0} & \cellcolor[gray]{0} & \cellcolor[gray]{0} & 
\cellcolor[gray]{0.1318} & \cellcolor[gray]{0.07256} & \cellcolor[gray]{0.31924} & \cellcolor[gray]{0.83949} & \cellcolor[gray]{0.67122} & \cellcolor[gray]{0.38304} & \cellcolor[gray]{0.84476} & \cellcolor[gray]{0.01191} & \cellcolor[gray]{0.07019} & \cellcolor[gray]{0.18933} & \cellcolor[gray]{0.27507} & \cellcolor[gray]{0.04365} & \cellcolor[gray]{0.1869} \\ \hline

\includegraphics[width=0.015\linewidth, height=0.018\linewidth]{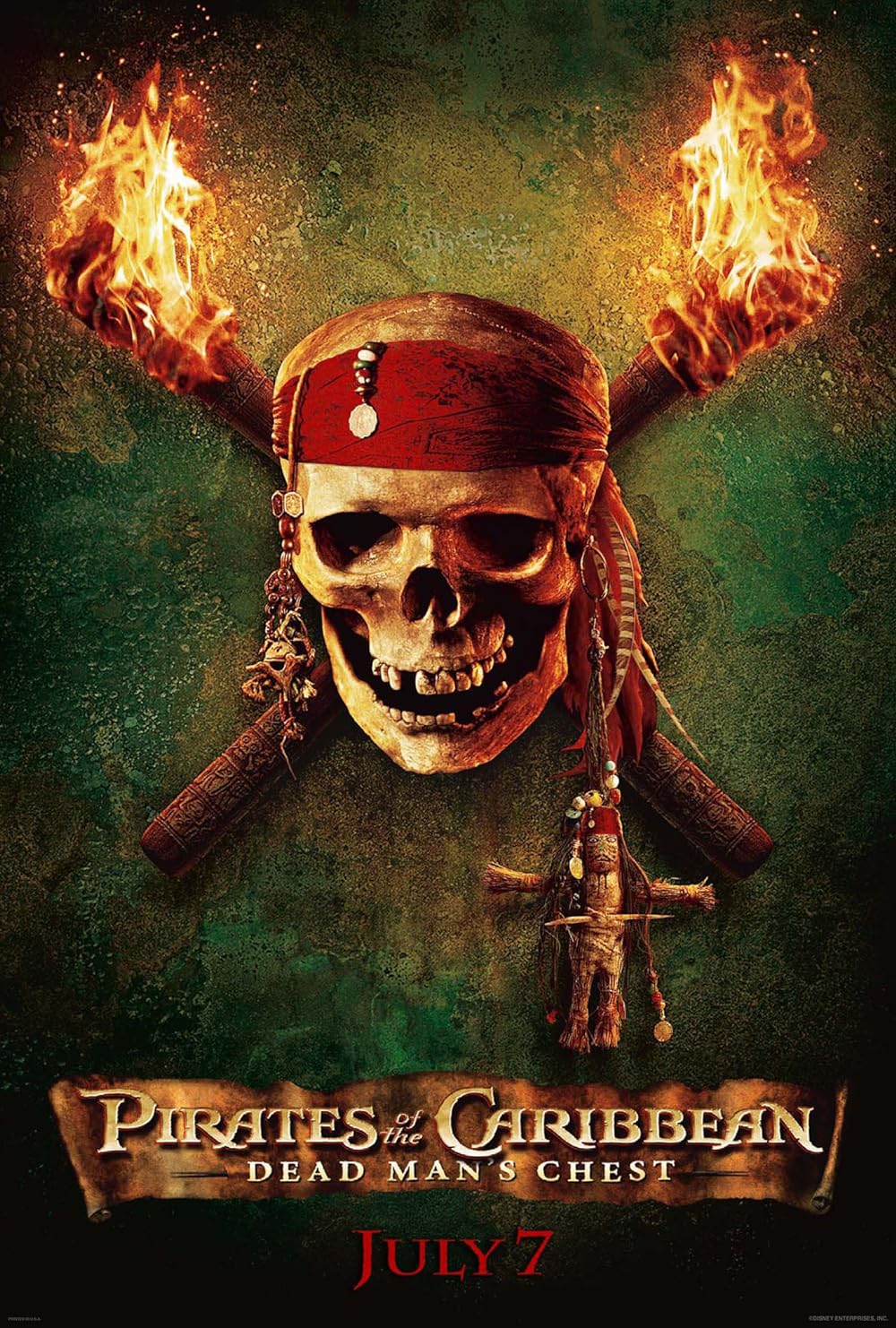} & \cellcolor[gray]{1} & \cellcolor[gray]{1} & \cellcolor[gray]{0} & \cellcolor[gray]{0} & \cellcolor[gray]{0} & \cellcolor[gray]{0} & \cellcolor[gray]{0} & \cellcolor[gray]{1} & \cellcolor[gray]{0} & \cellcolor[gray]{0}& \cellcolor[gray]{0}& \cellcolor[gray]{0}& \cellcolor[gray]{0}
& \cellcolor[gray]{0.86652} & \cellcolor[gray]{0.93002} & \cellcolor[gray]{0.00792} & \cellcolor[gray]{0.00138} & \cellcolor[gray]{0.31529} & \cellcolor[gray]{0.08141} & \cellcolor[gray]{0.08774} & \cellcolor[gray]{0.81049} & \cellcolor[gray]{0.55357} & \cellcolor[gray]{0.08914} & \cellcolor[gray]{0.00226} & \cellcolor[gray]{0.03485} & \cellcolor[gray]{0.12434} \\ \hline

\includegraphics[width=0.015\linewidth, height=0.018\linewidth]{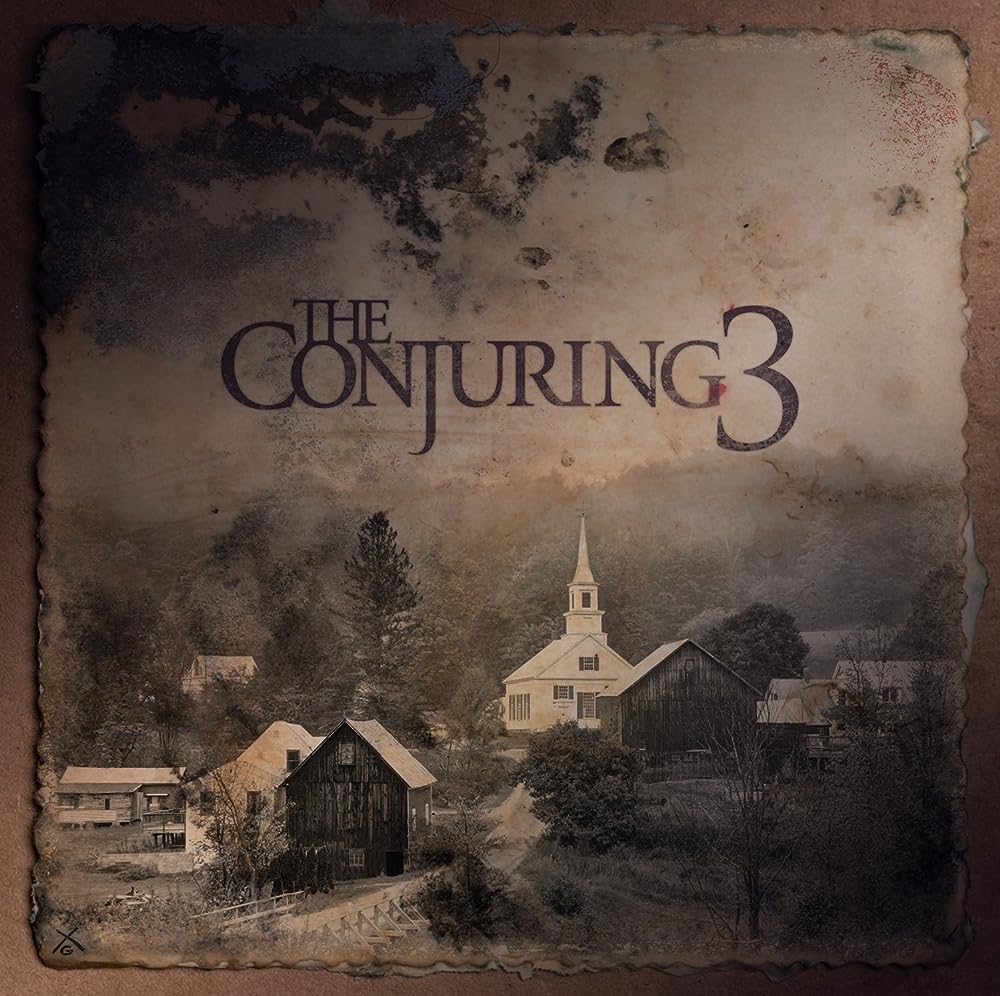} & \cellcolor[gray]{0} & \cellcolor[gray]{0} & \cellcolor[gray]{0} & \cellcolor[gray]{0} & \cellcolor[gray]{0} & \cellcolor[gray]{0} & \cellcolor[gray]{0} & \cellcolor[gray]{0} & \cellcolor[gray]{1} & \cellcolor[gray]{1}& \cellcolor[gray]{0}& \cellcolor[gray]{0}& \cellcolor[gray]{1}
& \cellcolor[gray]{0.01888} & \cellcolor[gray]{0.0528} & \cellcolor[gray]{0.00028} & \cellcolor[gray]{0.04366} & \cellcolor[gray]{0.07821} & \cellcolor[gray]{0.14444} & \cellcolor[gray]{0.60987} & \cellcolor[gray]{0.09555} & \cellcolor[gray]{0.98682} & \cellcolor[gray]{0.8901 } & \cellcolor[gray]{0.00355} & \cellcolor[gray]{0.01362} & \cellcolor[gray]{0.68263} \\ \hline

\includegraphics[width=0.015\linewidth, height=0.018\linewidth]{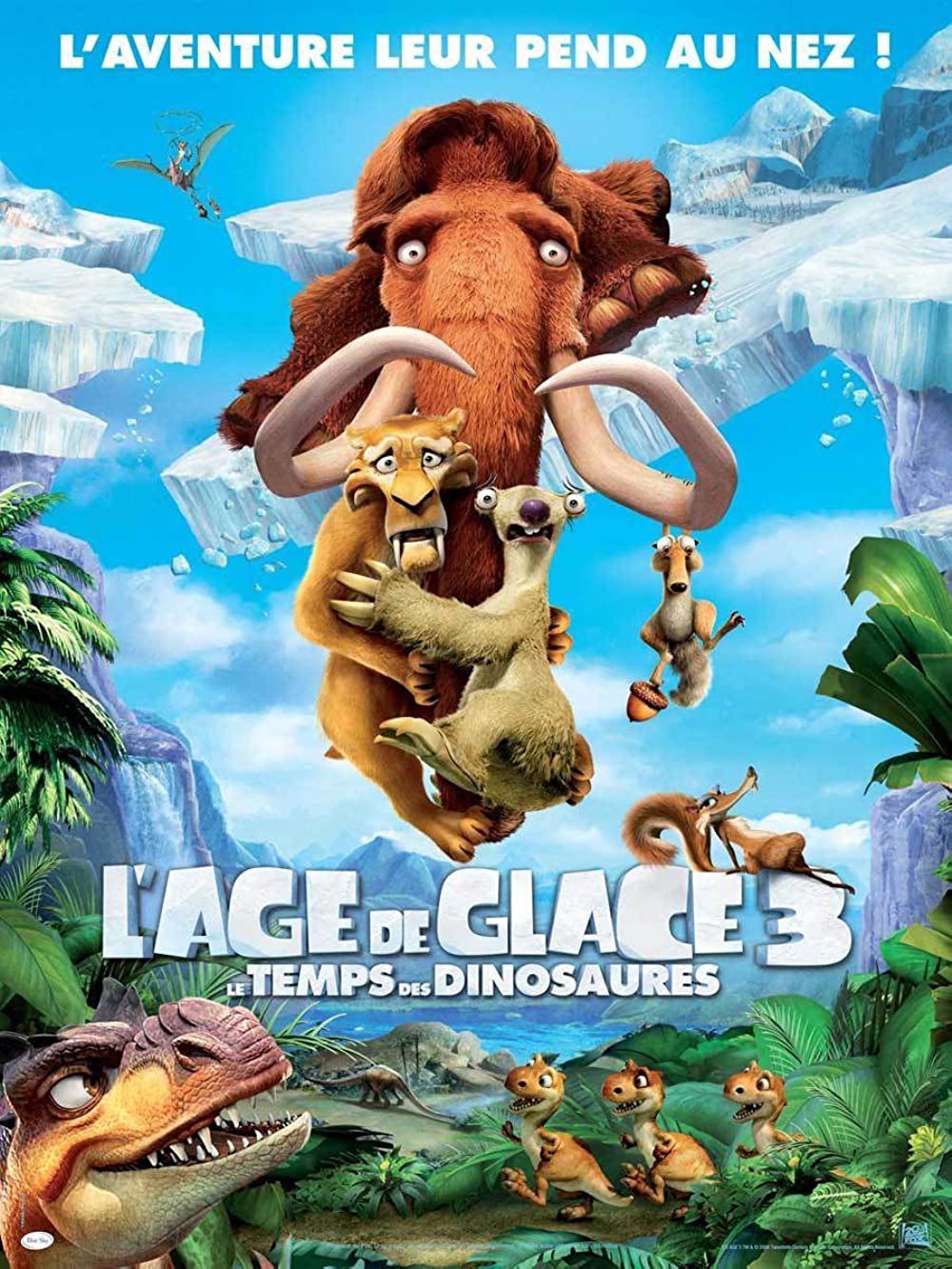} & \cellcolor[gray]{0} & \cellcolor[gray]{1} & \cellcolor[gray]{1} & \cellcolor[gray]{0} & \cellcolor[gray]{1} & \cellcolor[gray]{0} & \cellcolor[gray]{0} & \cellcolor[gray]{0} & \cellcolor[gray]{0} & \cellcolor[gray]{0}& \cellcolor[gray]{0}& \cellcolor[gray]{0}& \cellcolor[gray]{0}
& \cellcolor[gray]{0.08275} & \cellcolor[gray]{0.99552} & \cellcolor[gray]{0.99896} & \cellcolor[gray]{0.00241} & \cellcolor[gray]{0.99594} & \cellcolor[gray]{0.0168} & \cellcolor[gray]{0.05968} & \cellcolor[gray]{0.01442} & \cellcolor[gray]{0.00154} & \cellcolor[gray]{0.0016} & \cellcolor[gray]{0.01673} & \cellcolor[gray]{0.0029} & \cellcolor[gray]{0.00232} \\ \hline

\includegraphics[width=0.015\linewidth, height=0.018\linewidth]{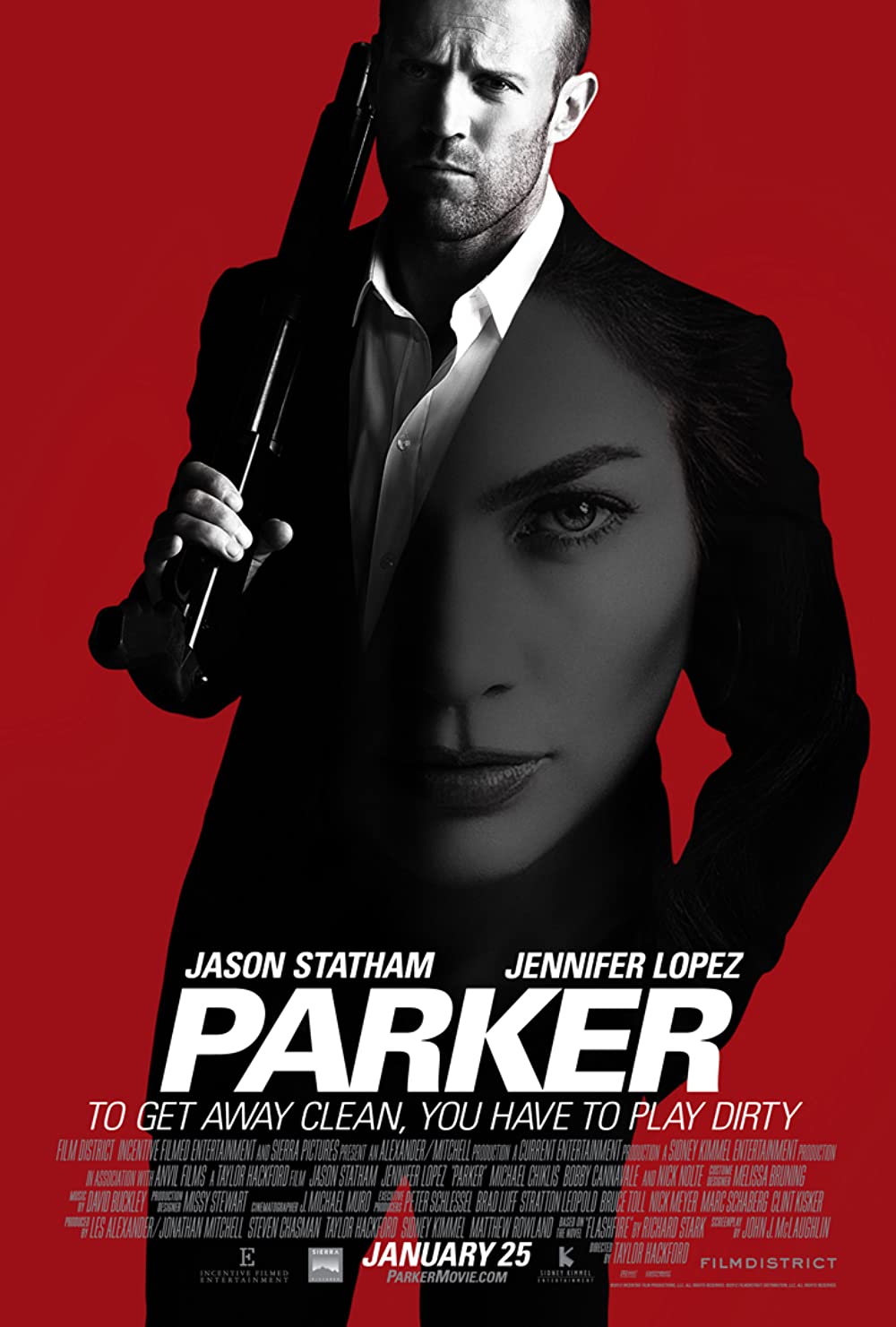} & \cellcolor[gray]{1} & \cellcolor[gray]{0} & \cellcolor[gray]{0} & \cellcolor[gray]{0} & \cellcolor[gray]{0} & \cellcolor[gray]{1} & \cellcolor[gray]{0} & \cellcolor[gray]{0} & \cellcolor[gray]{0} & \cellcolor[gray]{0}& \cellcolor[gray]{0}& \cellcolor[gray]{0}& \cellcolor[gray]{1}
& \cellcolor[gray]{0.78933} & \cellcolor[gray]{0.07239} & \cellcolor[gray]{0.00038} & \cellcolor[gray]{0.20418} & \cellcolor[gray]{0.09774} & \cellcolor[gray]{0.9001} & \cellcolor[gray]{0.36881} & \cellcolor[gray]{0.01163} & \cellcolor[gray]{0.02129} & \cellcolor[gray]{0.19708} & \cellcolor[gray]{0.04643} & \cellcolor[gray]{0.00966} & \cellcolor[gray]{0.80034} \\ \hline

\includegraphics[width=0.015\linewidth, height=0.018\linewidth]{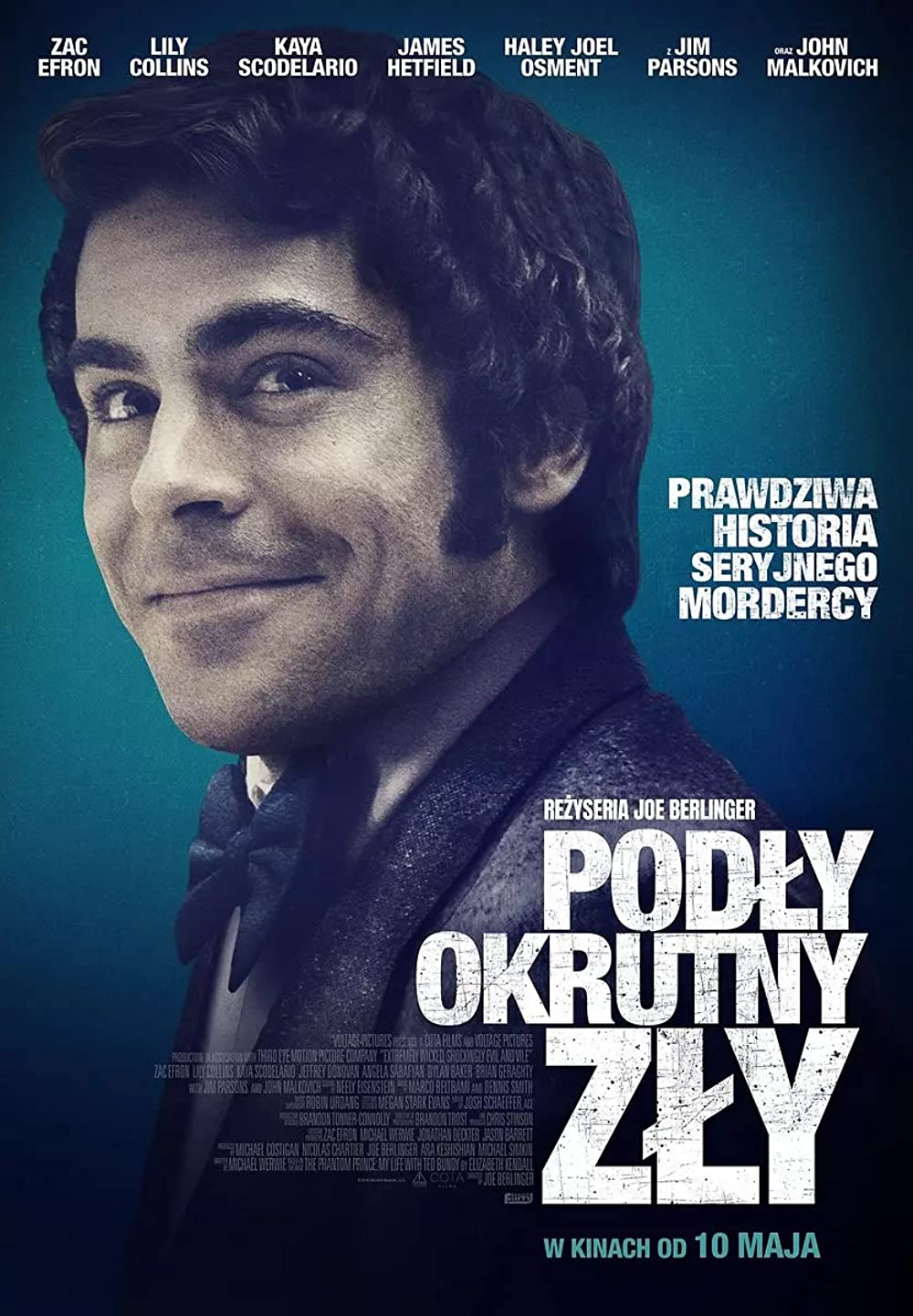} & \cellcolor[gray]{0} & \cellcolor[gray]{0} & \cellcolor[gray]{0} & \cellcolor[gray]{1} & \cellcolor[gray]{0} & \cellcolor[gray]{1} & \cellcolor[gray]{1} & \cellcolor[gray]{0} & \cellcolor[gray]{0} & \cellcolor[gray]{0}& \cellcolor[gray]{0}& \cellcolor[gray]{0}& \cellcolor[gray]{0}
& \cellcolor[gray]{0.2192} & \cellcolor[gray]{0.01735} & \cellcolor[gray]{0.00022} & \cellcolor[gray]{0.70319} & \cellcolor[gray]{0.02977} & \cellcolor[gray]{0.89074} & \cellcolor[gray]{0.82937} & \cellcolor[gray]{0.04146} & \cellcolor[gray]{0.10944} & \cellcolor[gray]{0.12263} & \cellcolor[gray]{0.07811} & \cellcolor[gray]{0.03972} & \cellcolor[gray]{0.6675} \\ \hline

\includegraphics[width=0.015\linewidth, height=0.018\linewidth]{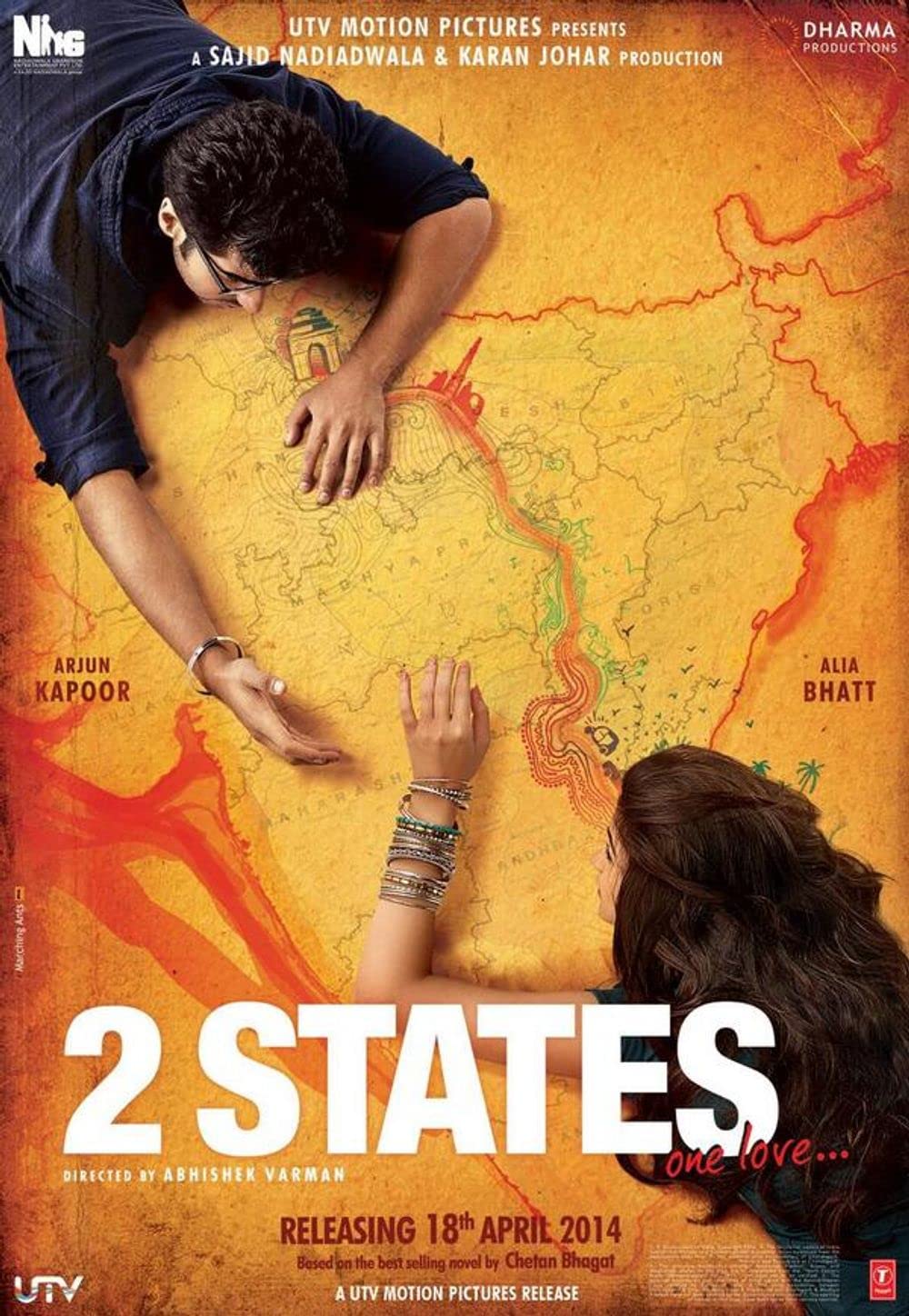} & \cellcolor[gray]{0} & \cellcolor[gray]{0} & \cellcolor[gray]{0} & \cellcolor[gray]{0} & \cellcolor[gray]{1} & \cellcolor[gray]{0} & \cellcolor[gray]{1} & \cellcolor[gray]{0} & \cellcolor[gray]{0} & \cellcolor[gray]{0}& \cellcolor[gray]{1}& \cellcolor[gray]{0}& \cellcolor[gray]{0}
& \cellcolor[gray]{0.10357} & \cellcolor[gray]{0.03438} & \cellcolor[gray]{0.00214} & \cellcolor[gray]{0.01869} & \cellcolor[gray]{0.83892} & \cellcolor[gray]{0.06817} & \cellcolor[gray]{0.78019} & \cellcolor[gray]{0.05941} & \cellcolor[gray]{0.05603} & \cellcolor[gray]{0.04074} & \cellcolor[gray]{0.96291} & \cellcolor[gray]{0.00325} & \cellcolor[gray]{0.10278} \\ \hline

\includegraphics[width=0.015\linewidth, height=0.018\linewidth]{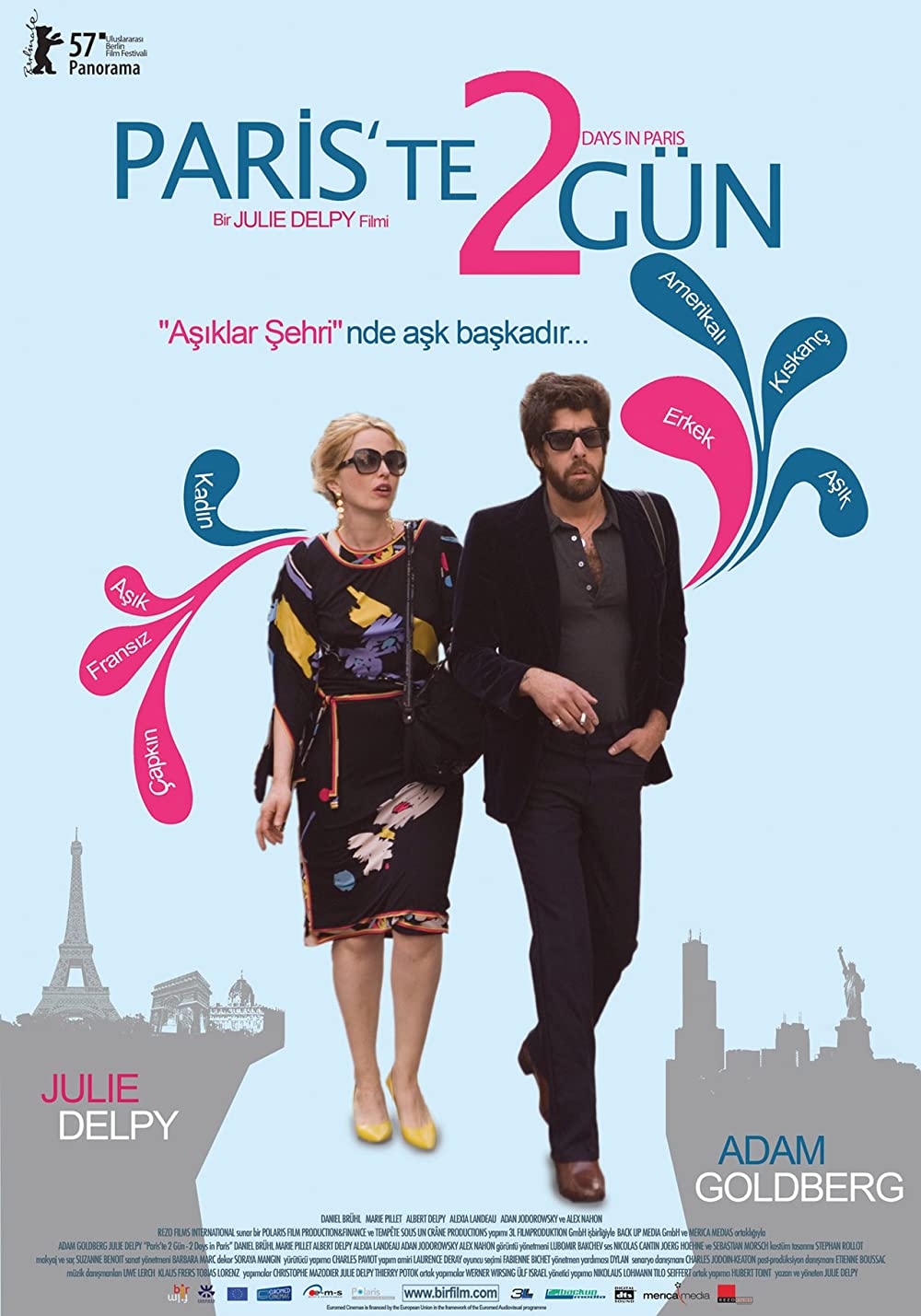} & \cellcolor[gray]{0} & \cellcolor[gray]{0} & \cellcolor[gray]{0} & \cellcolor[gray]{0} 
& \cellcolor[gray]{1} & \cellcolor[gray]{0} & \cellcolor[gray]{1} & 
\cellcolor[gray]{0} & \cellcolor[gray]{0} & \cellcolor[gray]{0}& 
\cellcolor[gray]{1} & \cellcolor[gray]{0}& \cellcolor[gray]{0} & 
\cellcolor[gray]{0.01919} & \cellcolor[gray]{0.024} & \cellcolor[gray]{0.00578} & \cellcolor[gray]{0.02222} & \cellcolor[gray]{0.98731} & \cellcolor[gray]{0.2746} & \cellcolor[gray]{0.7772} & \cellcolor[gray]{0.03267} & \cellcolor[gray]{0.00816} & \cellcolor[gray]{0.01432} & \cellcolor[gray]{0.90762} & \cellcolor[gray]{0.00193} & \cellcolor[gray]{0.02003} \\ \hline

\includegraphics[width=0.015\linewidth, height=0.018\linewidth]{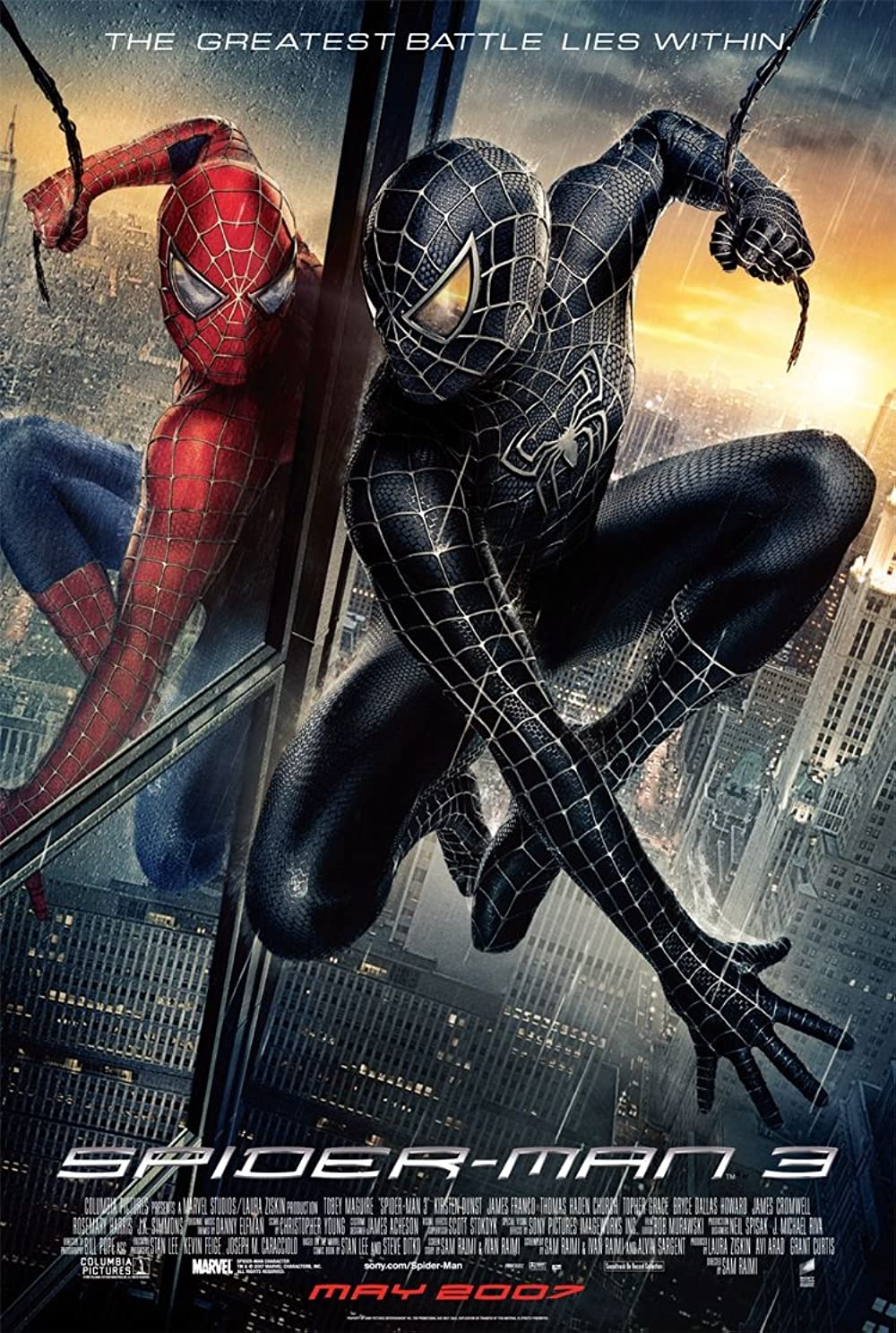} & \cellcolor[gray]{1} & \cellcolor[gray]{1} & \cellcolor[gray]{0} & \cellcolor[gray]{0} & \cellcolor[gray]{0} & \cellcolor[gray]{0} & \cellcolor[gray]{0} & \cellcolor[gray]{0} & \cellcolor[gray]{0} & \cellcolor[gray]{0}& \cellcolor[gray]{0}
& \cellcolor[gray]{1}& \cellcolor[gray]{0}
& \cellcolor[gray]{0.99585} & \cellcolor[gray]{0.96187} & \cellcolor[gray]{0.00958} & \cellcolor[gray]{0.00898} & \cellcolor[gray]{0.02394} & \cellcolor[gray]{0.25031} & \cellcolor[gray]{0.04266} & \cellcolor[gray]{0.23091} & \cellcolor[gray]{0.00762} & \cellcolor[gray]{0.01966} & \cellcolor[gray]{0.00159} & \cellcolor[gray]{0.96609} & \cellcolor[gray]{0.1607} \\ \hline

\includegraphics[width=0.015\linewidth, height=0.018\linewidth]{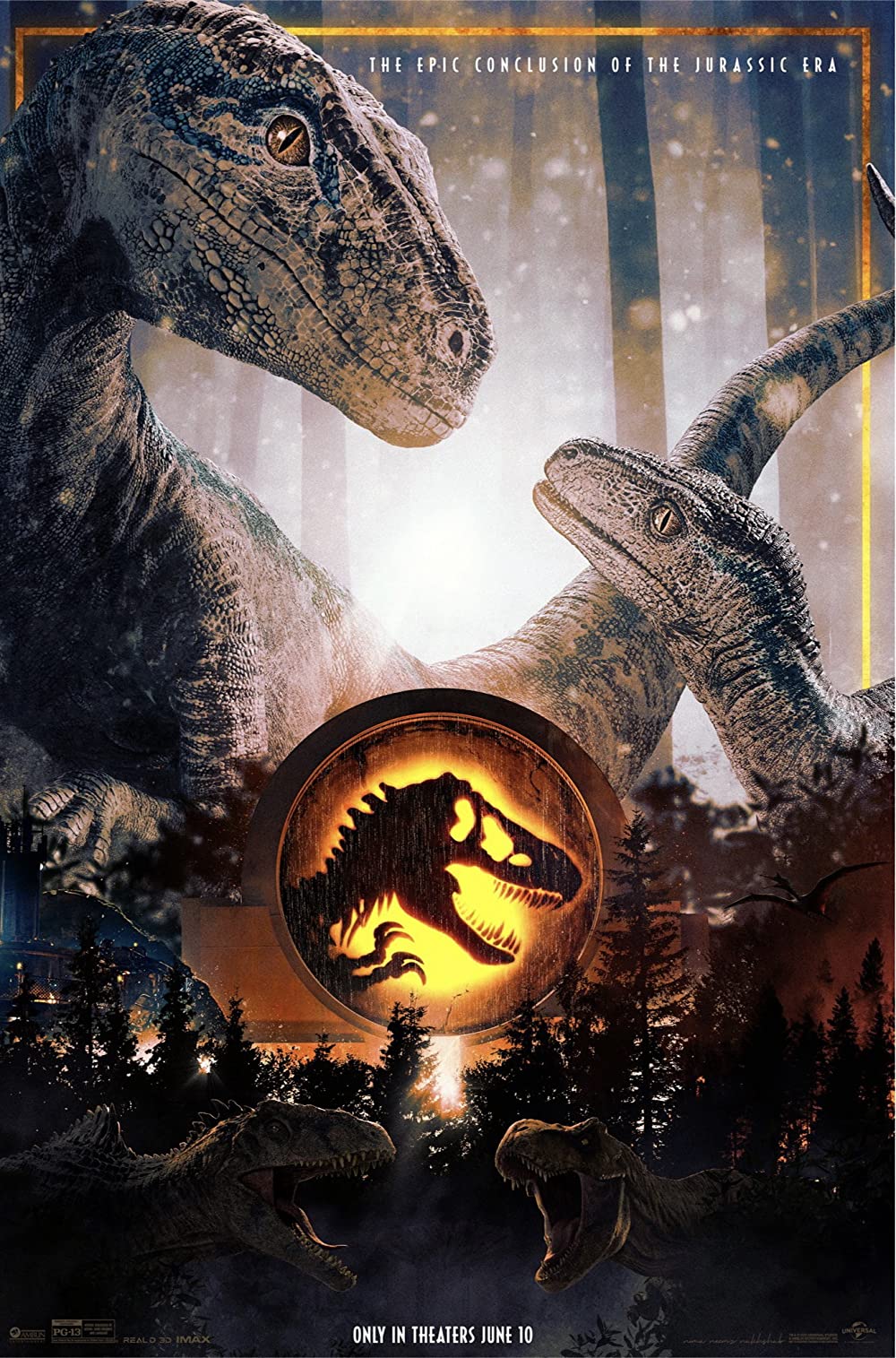} & \cellcolor[gray]{1} & \cellcolor[gray]{1} & \cellcolor[gray]{0} & \cellcolor[gray]{0} & \cellcolor[gray]{0} & \cellcolor[gray]{0} & \cellcolor[gray]{0} & \cellcolor[gray]{0} & \cellcolor[gray]{0} & \cellcolor[gray]{0}& \cellcolor[gray]{0}& \cellcolor[gray]{1}& \cellcolor[gray]{0}
& \cellcolor[gray]{0.96284} & \cellcolor[gray]{0.95474} & \cellcolor[gray]{0.00578} & \cellcolor[gray]{0.01344} & \cellcolor[gray]{0.04755} & \cellcolor[gray]{0.06989} & \cellcolor[gray]{0.29438} & \cellcolor[gray]{0.37829} & \cellcolor[gray]{0.13212} & \cellcolor[gray]{0.04073} & \cellcolor[gray]{0.00289} & \cellcolor[gray]{0.70634} & \cellcolor[gray]{0.14478} \\ \hline

\includegraphics[width=0.015\linewidth, height=0.018\linewidth]{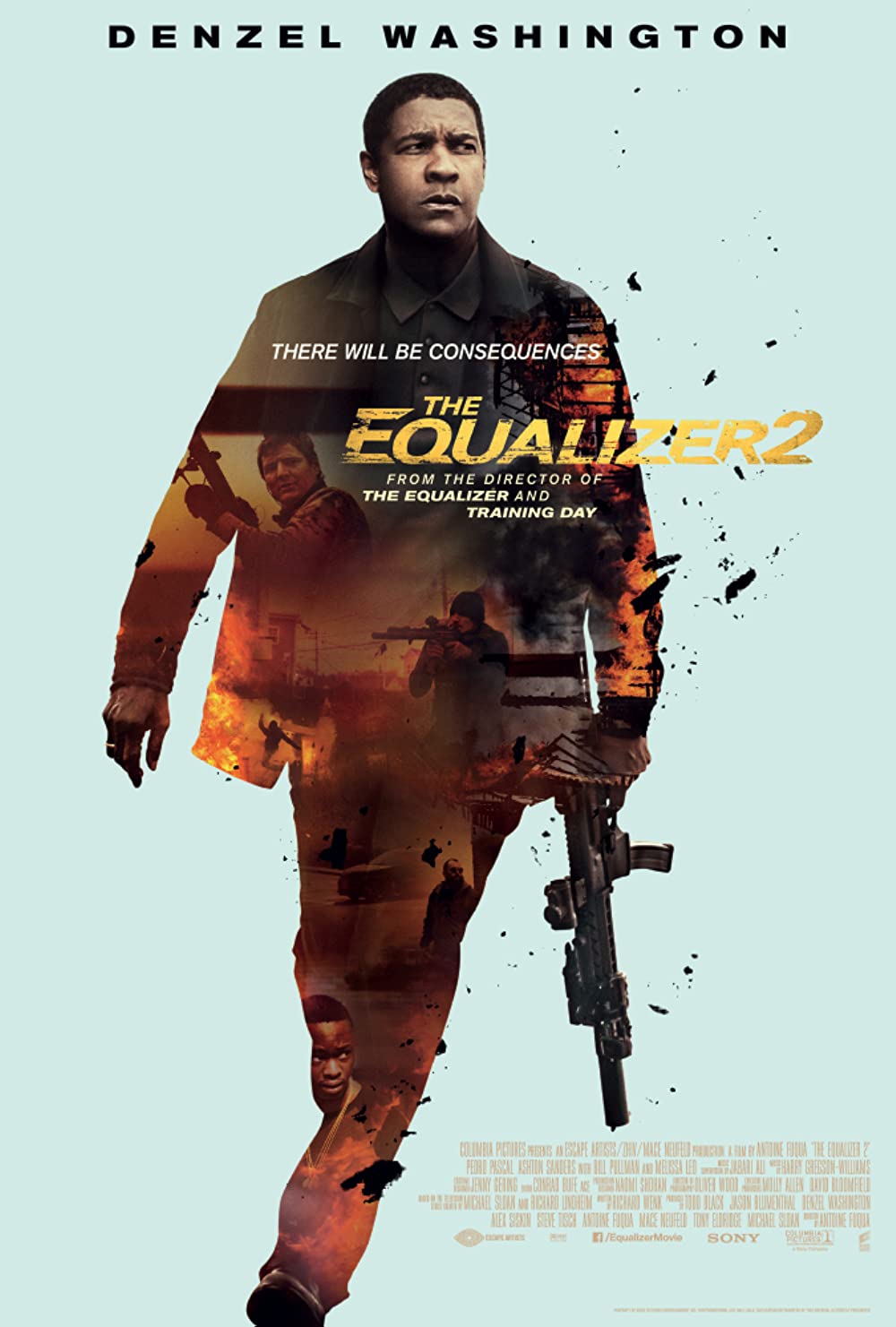} & \cellcolor[gray]{1} & \cellcolor[gray]{0} & \cellcolor[gray]{0} & \cellcolor[gray]{0} & \cellcolor[gray]{0} & \cellcolor[gray]{1} & \cellcolor[gray]{0} & \cellcolor[gray]{0} & \cellcolor[gray]{0} & \cellcolor[gray]{0}& \cellcolor[gray]{0}& \cellcolor[gray]{0}& \cellcolor[gray]{1}
& \cellcolor[gray]{0.96325} & \cellcolor[gray]{0.17143} & \cellcolor[gray]{0.00083} & \cellcolor[gray]{0.013 } & \cellcolor[gray]{0.14366} & \cellcolor[gray]{0.76794} & \cellcolor[gray]{0.2642} & \cellcolor[gray]{0.06296} & \cellcolor[gray]{0.06267} & \cellcolor[gray]{0.04101} & \cellcolor[gray]{0.01043} & \cellcolor[gray]{0.0263 } & \cellcolor[gray]{0.81012} \\ \hline

\includegraphics[width=0.015\linewidth, height=0.018\linewidth]{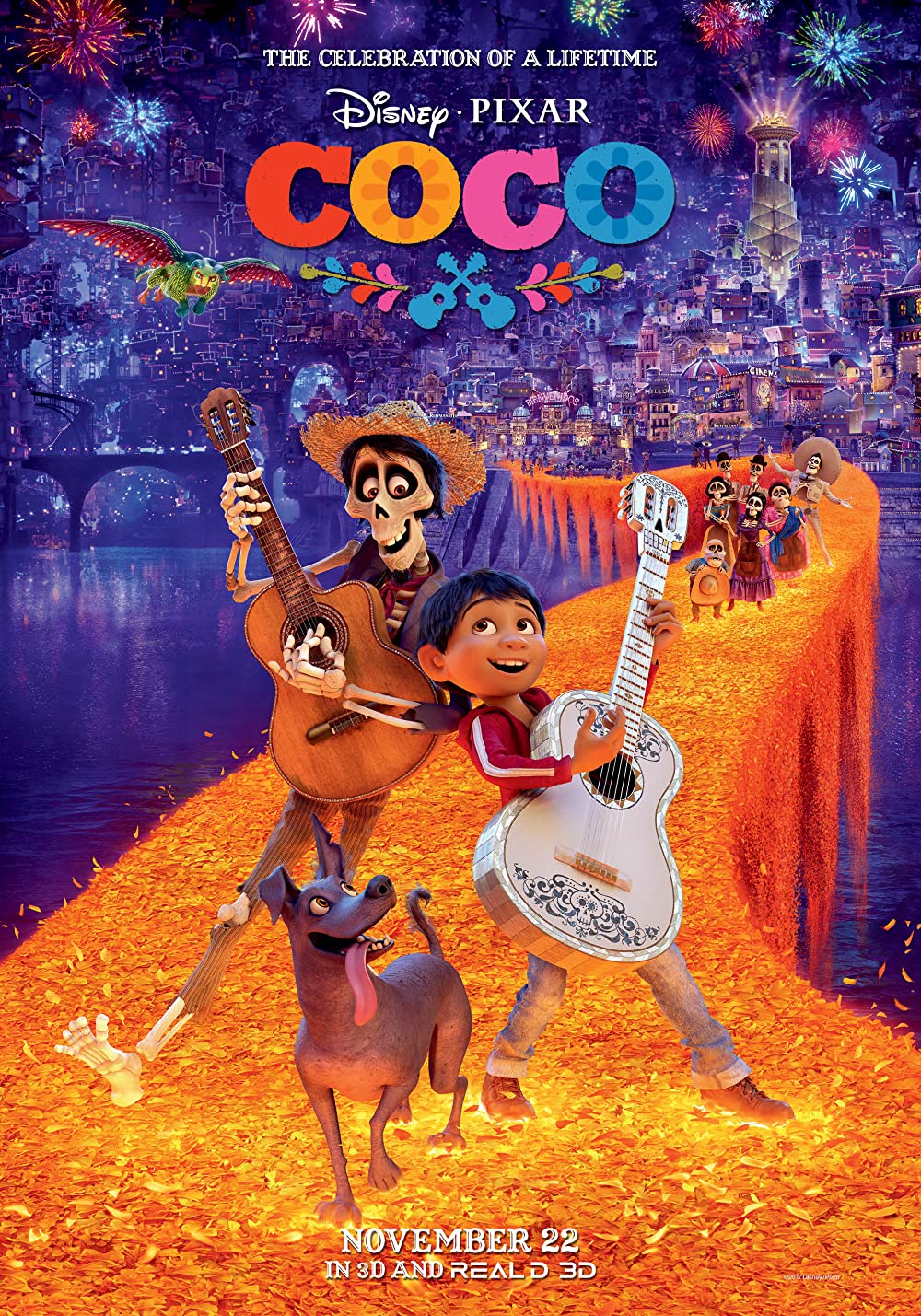} & \cellcolor[gray]{0} & \cellcolor[gray]{1} & \cellcolor[gray]{1} & \cellcolor[gray]{0} & \cellcolor[gray]{1} & \cellcolor[gray]{0} & \cellcolor[gray]{0} & \cellcolor[gray]{0} & \cellcolor[gray]{0} & \cellcolor[gray]{0}& \cellcolor[gray]{0}& \cellcolor[gray]{0}& \cellcolor[gray]{0}
& \cellcolor[gray]{0.39225} & \cellcolor[gray]{0.92189} & \cellcolor[gray]{0.97194} & \cellcolor[gray]{0.00044} & \cellcolor[gray]{0.97305} & \cellcolor[gray]{0.07064} & \cellcolor[gray]{0.10204} & \cellcolor[gray]{0.13403} & \cellcolor[gray]{0.01699} & \cellcolor[gray]{0.00256} & \cellcolor[gray]{0.03189} & \cellcolor[gray]{0.00489} & \cellcolor[gray]{0.01564} \\ \hline 

\includegraphics[width=0.015\linewidth, height=0.018\linewidth]{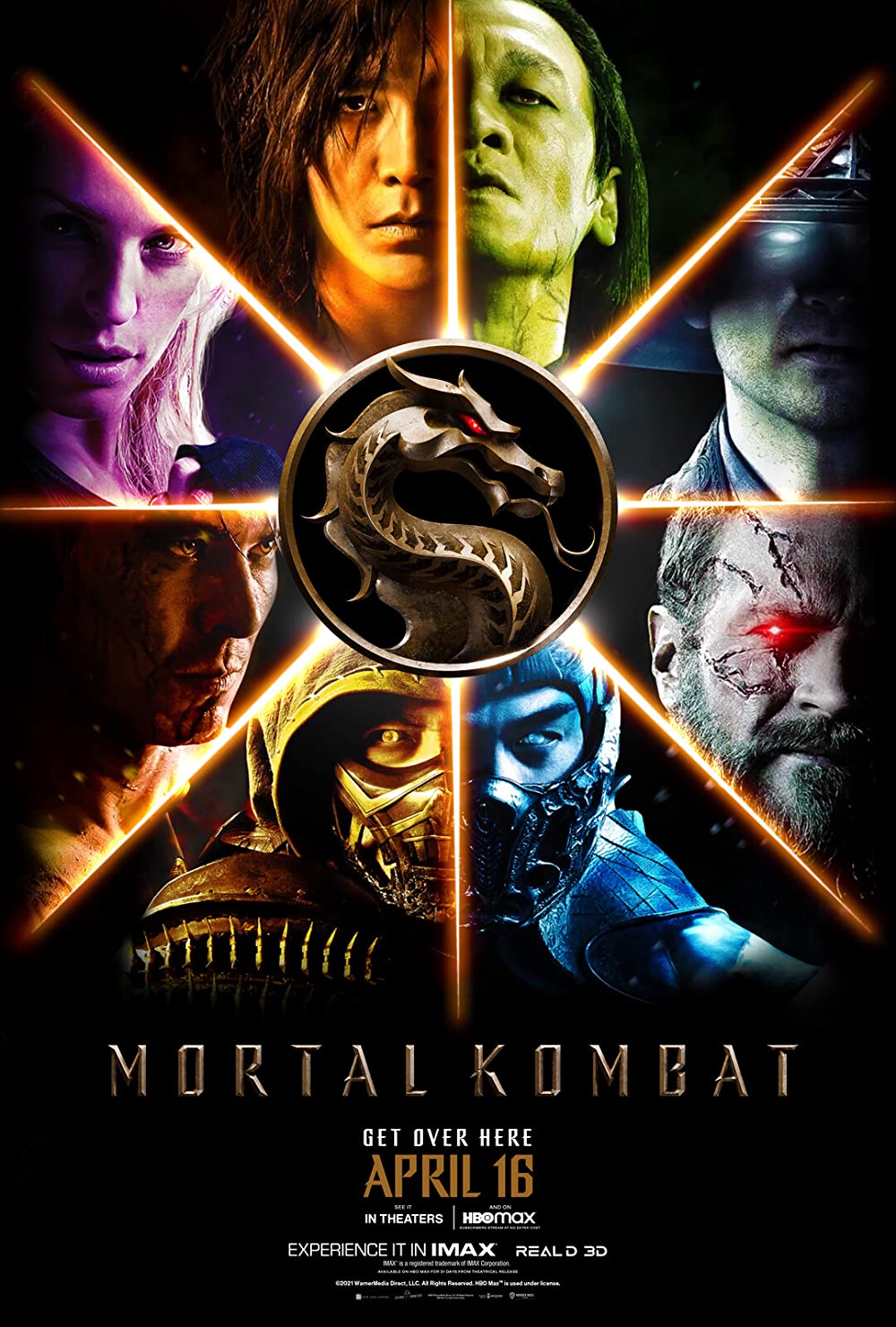} & \cellcolor[gray]{1} & \cellcolor[gray]{1} & \cellcolor[gray]{0} & \cellcolor[gray]{0} & \cellcolor[gray]{0} & \cellcolor[gray]{0} & \cellcolor[gray]{0} & \cellcolor[gray]{1} & \cellcolor[gray]{0} & \cellcolor[gray]{0}& \cellcolor[gray]{0}& \cellcolor[gray]{0}& \cellcolor[gray]{0}
& \cellcolor[gray]{0.89683} & \cellcolor[gray]{0.77096} & \cellcolor[gray]{0.00183} & \cellcolor[gray]{0.00076} & \cellcolor[gray]{0.02969} & \cellcolor[gray]{0.08091} & \cellcolor[gray]{0.18001} & \cellcolor[gray]{0.62568} & \cellcolor[gray]{0.12718} & \cellcolor[gray]{0.16251} & \cellcolor[gray]{0.0013} & \cellcolor[gray]{0.2935 } & \cellcolor[gray]{0.18951} \\ \hline

\includegraphics[width=0.015\linewidth, height=0.018\linewidth]{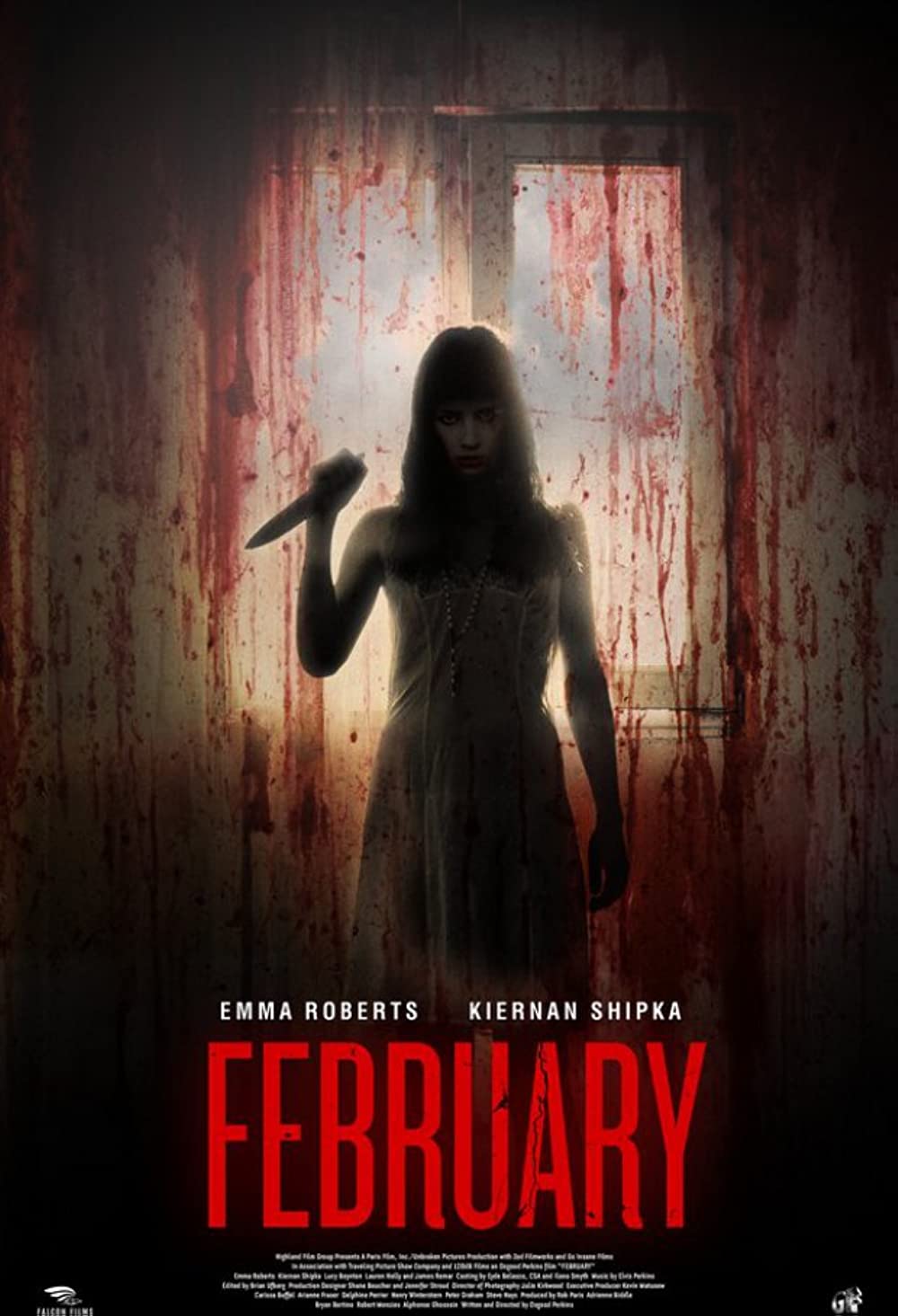} & \cellcolor[gray]{0} & \cellcolor[gray]{0} & \cellcolor[gray]{0} & \cellcolor[gray]{0} & \cellcolor[gray]{0} & \cellcolor[gray]{0} & \cellcolor[gray]{0} & \cellcolor[gray]{0} & \cellcolor[gray]{1} & \cellcolor[gray]{1}& \cellcolor[gray]{0}& \cellcolor[gray]{0}& \cellcolor[gray]{1}
& \cellcolor[gray]{0.08748} & \cellcolor[gray]{0.01577} & \cellcolor[gray]{0.00029} & \cellcolor[gray]{0.0016} & \cellcolor[gray]{0.1533} & \cellcolor[gray]{0.05036} & \cellcolor[gray]{0.21339} & \cellcolor[gray]{0.12512} & \cellcolor[gray]{0.99578} & \cellcolor[gray]{0.55865} & \cellcolor[gray]{0.0054 } & \cellcolor[gray]{0.01956} & \cellcolor[gray]{0.68148} \\ \hline

\includegraphics[width=0.015\linewidth, height=0.018\linewidth]{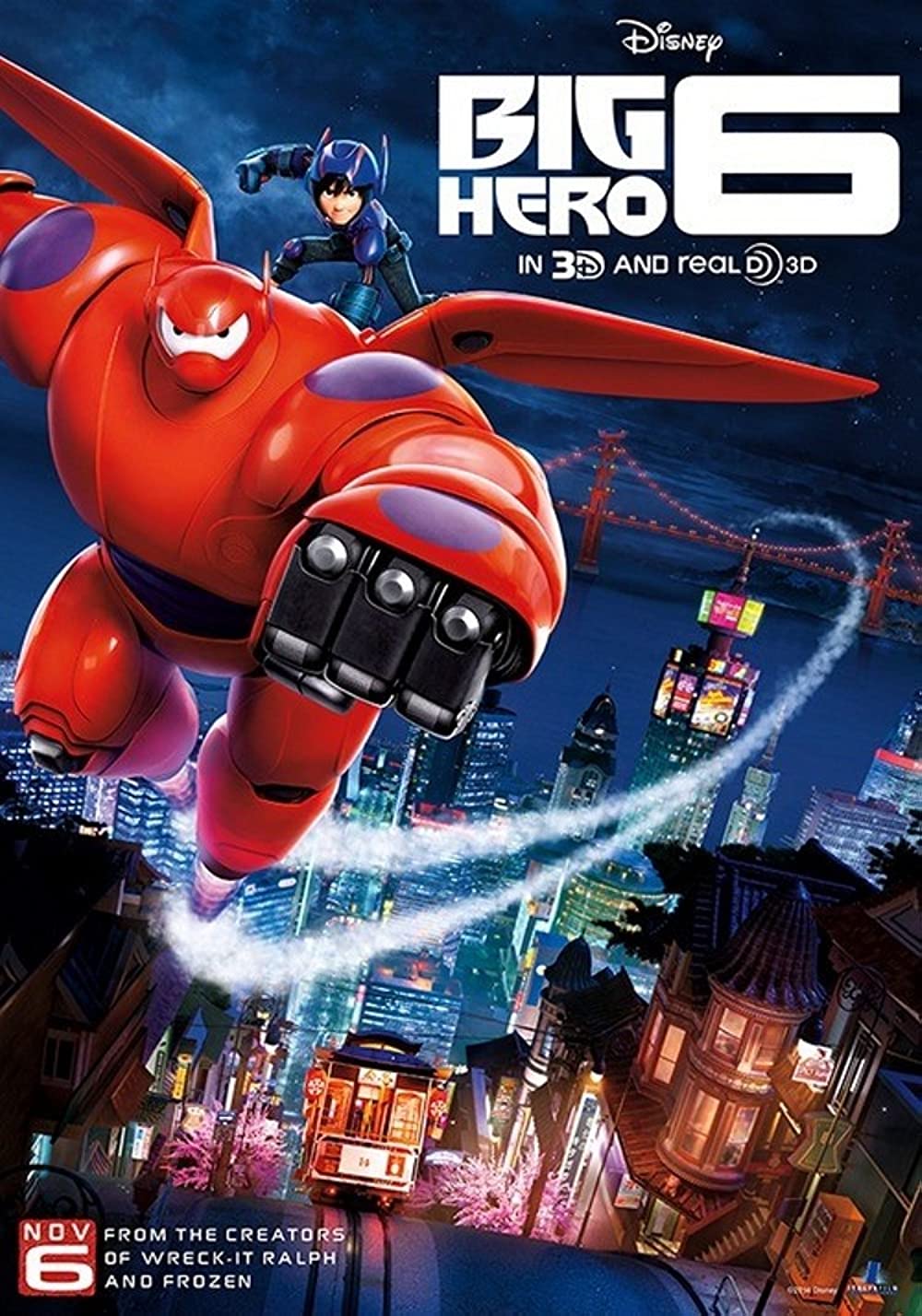} & \cellcolor[gray]{1} & \cellcolor[gray]{1} & \cellcolor[gray]{1} & \cellcolor[gray]{0} & \cellcolor[gray]{0} & \cellcolor[gray]{0} & \cellcolor[gray]{0} & \cellcolor[gray]{0} & \cellcolor[gray]{0} & \cellcolor[gray]{0}& \cellcolor[gray]{0}& \cellcolor[gray]{0}& \cellcolor[gray]{0}
& \cellcolor[gray]{0.95542} & \cellcolor[gray]{0.99458} & \cellcolor[gray]{0.99083} & \cellcolor[gray]{0.00047} & \cellcolor[gray]{0.78639} & \cellcolor[gray]{0.04461} & \cellcolor[gray]{0.00683} & \cellcolor[gray]{0.01735} & \cellcolor[gray]{0.00669} & \cellcolor[gray]{0.00178} & \cellcolor[gray]{0.00131} & \cellcolor[gray]{0.07634} & \cellcolor[gray]{0.02182} \\ \hline

\includegraphics[width=0.015\linewidth, height=0.018\linewidth]{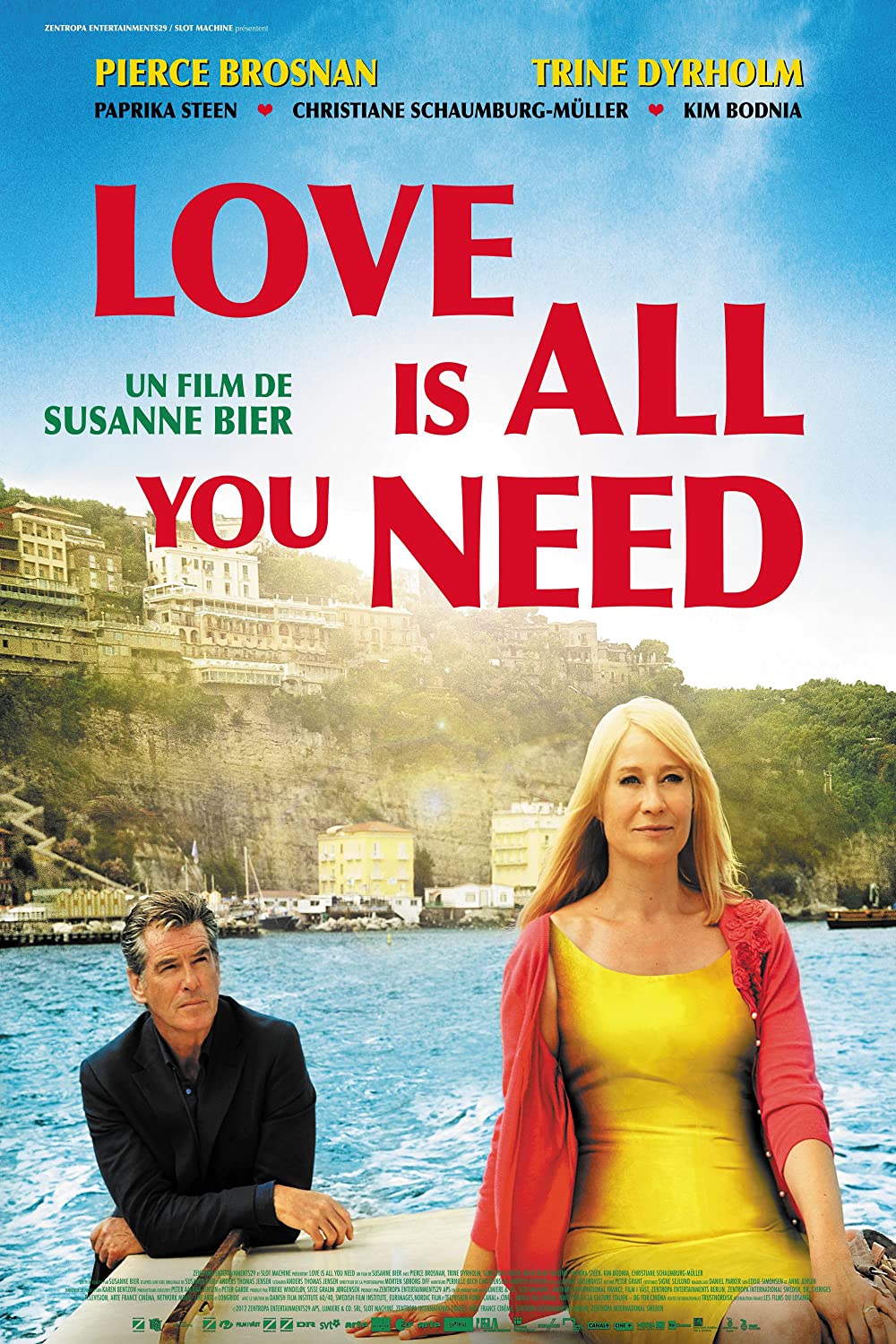} & \cellcolor[gray]{0} & \cellcolor[gray]{0} & \cellcolor[gray]{0} & \cellcolor[gray]{0} & \cellcolor[gray]{1} & \cellcolor[gray]{0} & \cellcolor[gray]{1} & \cellcolor[gray]{0} & \cellcolor[gray]{0} & \cellcolor[gray]{0}& \cellcolor[gray]{1}& \cellcolor[gray]{0}& \cellcolor[gray]{0}
& \cellcolor[gray]{0.02054} & \cellcolor[gray]{0.03893} & \cellcolor[gray]{0.00148} & \cellcolor[gray]{0.06049} & \cellcolor[gray]{0.92872} & \cellcolor[gray]{0.17461} & \cellcolor[gray]{0.89273} & \cellcolor[gray]{0.10082} & \cellcolor[gray]{0.0067} & \cellcolor[gray]{0.01327} & \cellcolor[gray]{0.96914} & \cellcolor[gray]{0.00428} & \cellcolor[gray]{0.02486} \\ \hline

\includegraphics[width=0.015\linewidth, height=0.018\linewidth]{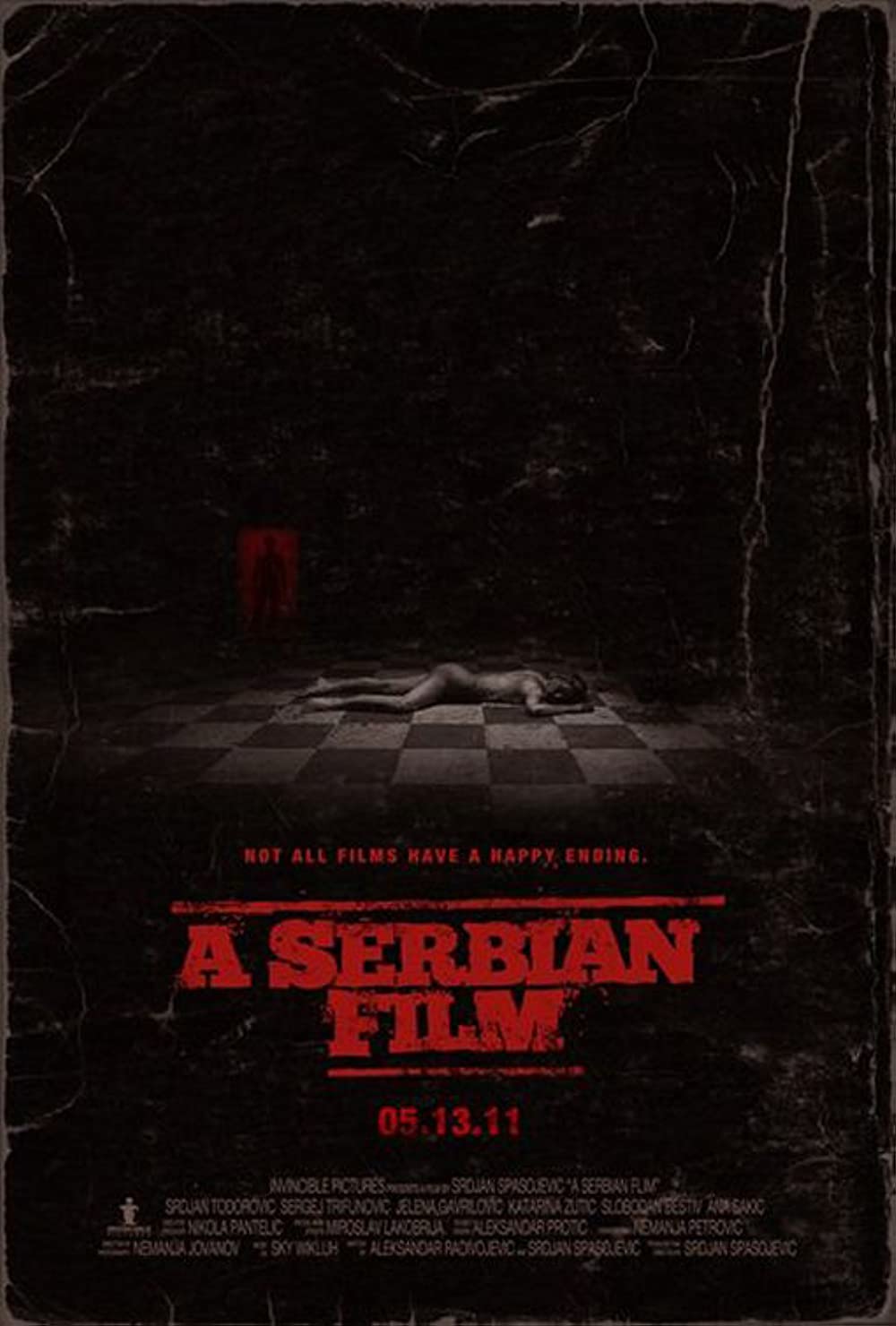} & \cellcolor[gray]{0} & \cellcolor[gray]{0} & \cellcolor[gray]{0} & \cellcolor[gray]{0} & \cellcolor[gray]{0} & \cellcolor[gray]{0} & \cellcolor[gray]{0} & \cellcolor[gray]{0} & \cellcolor[gray]{1} & \cellcolor[gray]{1}& \cellcolor[gray]{0}& \cellcolor[gray]{0}& \cellcolor[gray]{1}
& \cellcolor[gray]{0.04475} & \cellcolor[gray]{0.01484} & \cellcolor[gray]{0.00001} & \cellcolor[gray]{0.00244} & \cellcolor[gray]{0.00751} & \cellcolor[gray]{0.1962} & \cellcolor[gray]{0.442 } & \cellcolor[gray]{0.10097} & \cellcolor[gray]{0.99239} & \cellcolor[gray]{0.77578} & \cellcolor[gray]{0.00384} & \cellcolor[gray]{0.01122} & \cellcolor[gray]{0.62661} \\ \hline

\includegraphics[width=0.015\linewidth, height=0.018\linewidth]{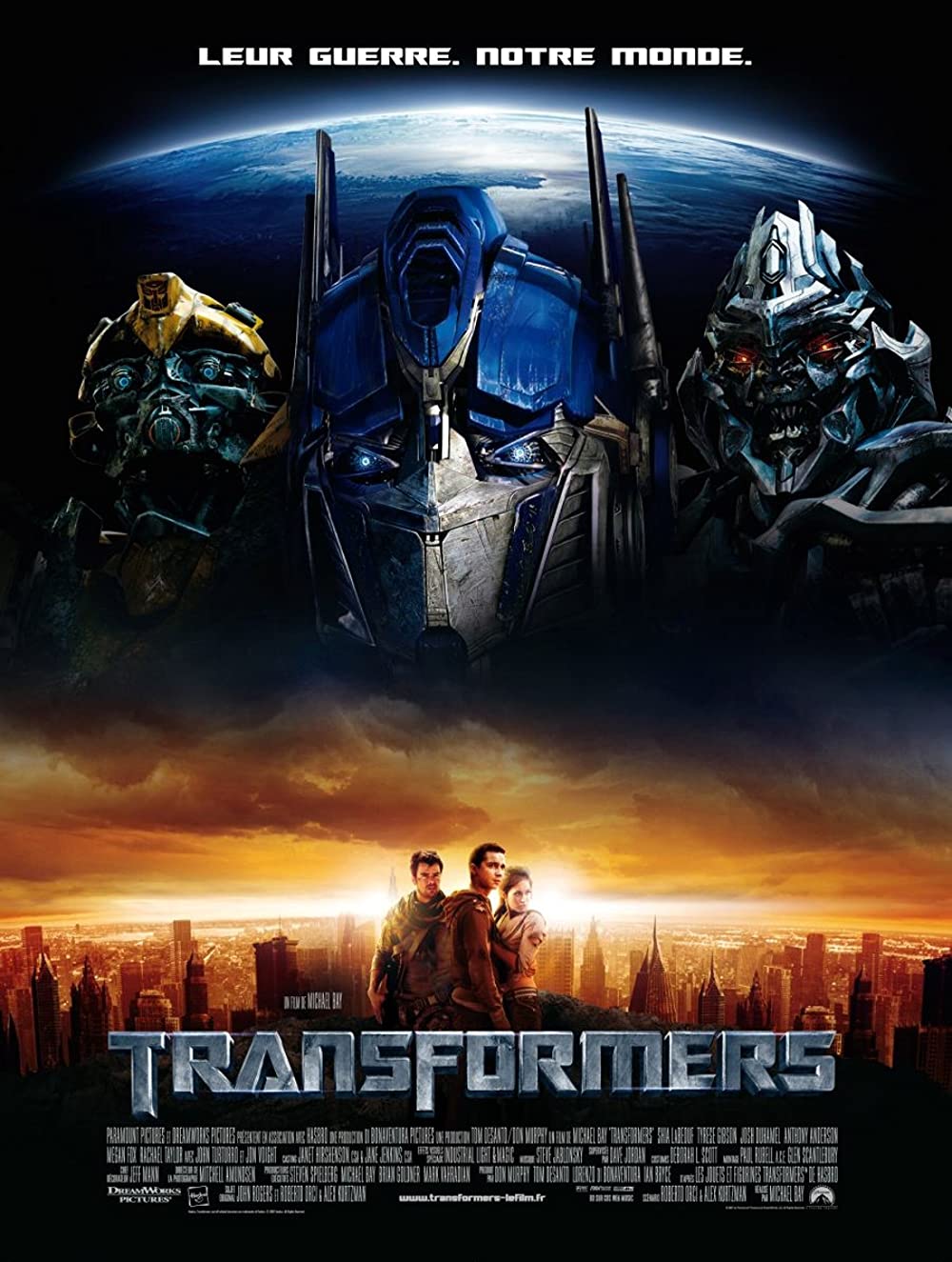} & \cellcolor[gray]{1} & \cellcolor[gray]{1} & \cellcolor[gray]{0} & \cellcolor[gray]{0} & \cellcolor[gray]{0} & \cellcolor[gray]{0} & \cellcolor[gray]{0} & \cellcolor[gray]{0} & \cellcolor[gray]{0} & \cellcolor[gray]{0}& \cellcolor[gray]{0}& \cellcolor[gray]{1}& \cellcolor[gray]{0}
& \cellcolor[gray]{0.99055} & \cellcolor[gray]{0.96854} & \cellcolor[gray]{0.02405} & \cellcolor[gray]{0.00394} & \cellcolor[gray]{0.04629} & \cellcolor[gray]{0.03564} & \cellcolor[gray]{0.12813} & \cellcolor[gray]{0.44616} & \cellcolor[gray]{0.04804} & \cellcolor[gray]{0.03185} & \cellcolor[gray]{0.00085} & \cellcolor[gray]{0.96595} & \cellcolor[gray]{0.04653} \\ \hline

\includegraphics[width=0.015\linewidth, height=0.018\linewidth]{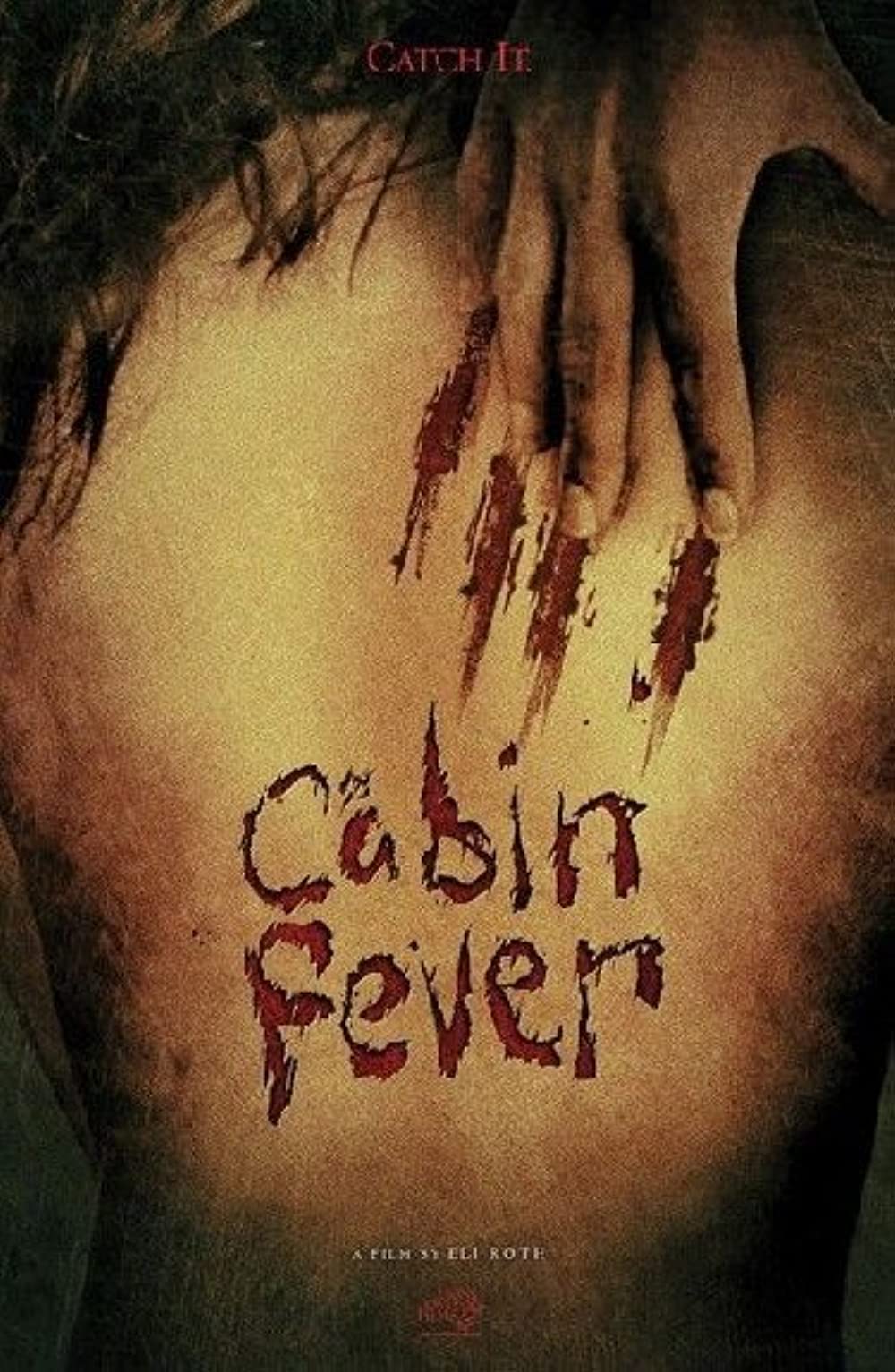} & \cellcolor[gray]{0} & \cellcolor[gray]{0} & \cellcolor[gray]{0} & \cellcolor[gray]{0} & \cellcolor[gray]{0} & \cellcolor[gray]{0} & \cellcolor[gray]{0} & \cellcolor[gray]{0} & \cellcolor[gray]{1} & \cellcolor[gray]{0}& \cellcolor[gray]{0}& \cellcolor[gray]{0}& \cellcolor[gray]{0}
& \cellcolor[gray]{0.20056} & \cellcolor[gray]{0.18501} & \cellcolor[gray]{0.00044} & \cellcolor[gray]{0.03351} & \cellcolor[gray]{0.28121} & \cellcolor[gray]{0.09654} & \cellcolor[gray]{0.3966} & \cellcolor[gray]{0.1372} & \cellcolor[gray]{0.95664} & \cellcolor[gray]{0.6543} & \cellcolor[gray]{0.13261} & \cellcolor[gray]{0.03736} & \cellcolor[gray]{0.21522} \\ \hline

\includegraphics[width=0.015\linewidth, height=0.018\linewidth]{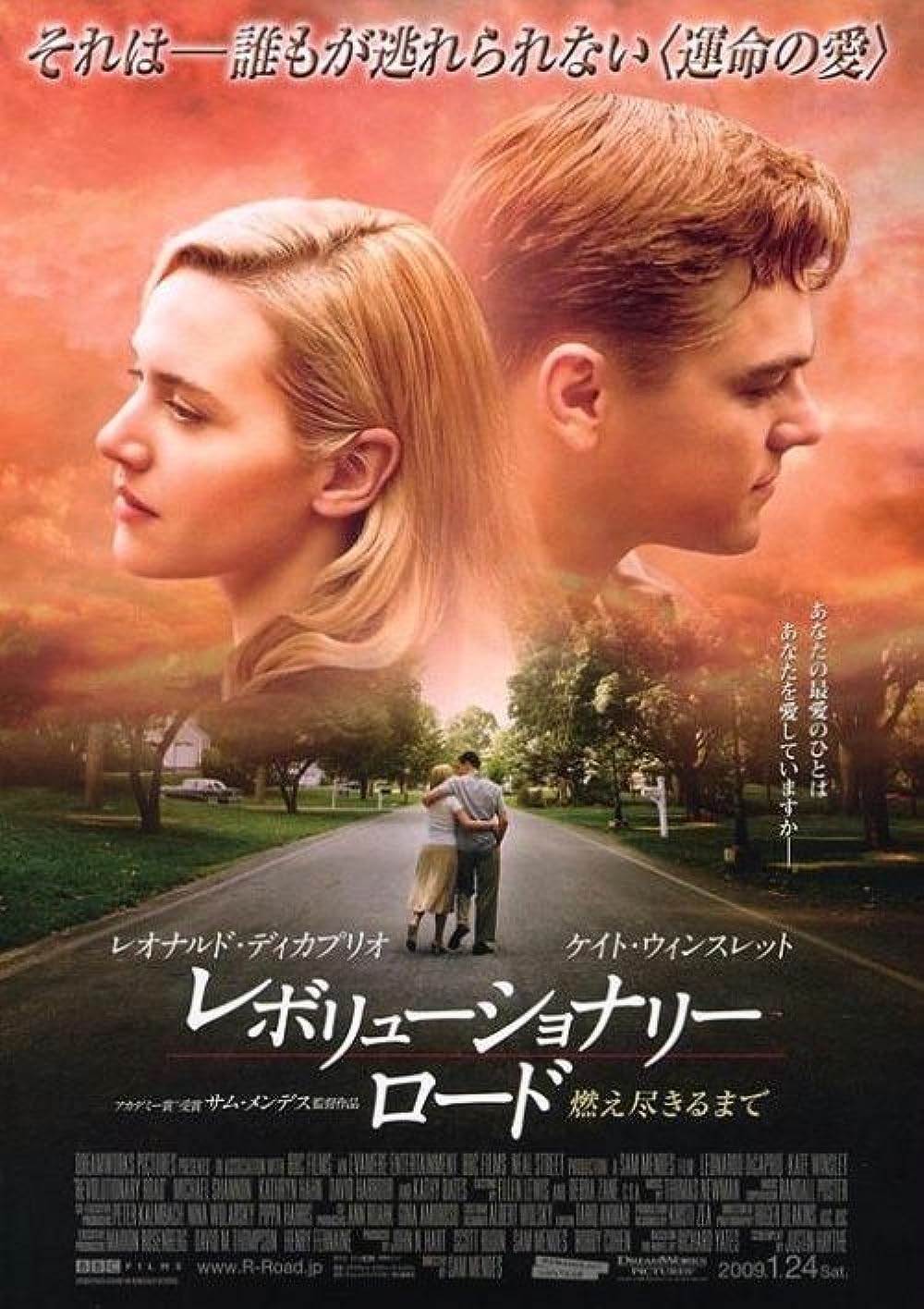} 
& \cellcolor[gray]{0} & \cellcolor[gray]{0} & \cellcolor[gray]{0} & \cellcolor[gray]{0} & \cellcolor[gray]{0} & \cellcolor[gray]{0} & \cellcolor[gray]{1} & \cellcolor[gray]{0} & \cellcolor[gray]{0} & \cellcolor[gray]{0} & \cellcolor[gray]{1} & \cellcolor[gray]{0} & \cellcolor[gray]{0} & \cellcolor[gray]{0.05458} & \cellcolor[gray]{0.07873} & \cellcolor[gray]{0.0004} & \cellcolor[gray]{0.21137} & \cellcolor[gray]{0.34106} & \cellcolor[gray]{0.11174} & \cellcolor[gray]{0.91329} & \cellcolor[gray]{0.15049} & \cellcolor[gray]{0.03933} & \cellcolor[gray]{0.21549} & \cellcolor[gray]{0.82826} & \cellcolor[gray]{0.00652} & \cellcolor[gray]{0.12775}
\\ \hline

\end{tabular}
\end{adjustbox}
\caption{Qualitative results for some sample posters: Ground-truth and ERDT-predicted confidence score heat-map encodings in gray color code}
    \label{fig:heatmap}
\end{figure*}

\noindent
In Fig. \ref{fig:heatmap}, we showcase the qualitative outcomes of ERDT through heat map encoding, as mentioned in Section 
IV-D of the main manuscript. 
We have selected 20 sample posters, and present the corresponding ground-truth and ERDT-predicted heat map encodings in grayscale, with white representing positive genres and black indicating negative ones.


\section{Performance Stagnation Analysis of PrERDT}
\label{app:model_perf_limit}

\begin{figure*}
\centering
\footnotesize
\begin{adjustbox}{width=1\linewidth}
\begin{tabular}{c|c|c|c | c|c|c|c}
\hline 
Comedy, Drama, & Action, Drama,   & Drama, Mystery, & Action, Adventure, & Action, Adventure, & Drama, Romance,  & Biography, Drama,  & Adventure, Comedy,  \\
Fantasy & Mystery   & Romance     & Comedy   & Fantasy  & \--- & \--- & Fantasy  \\
&&& &&&&  \\[\dimexpr-\normalbaselineskip+1.5pt]
\includegraphics[width=0.1\linewidth, height=0.145\linewidth]{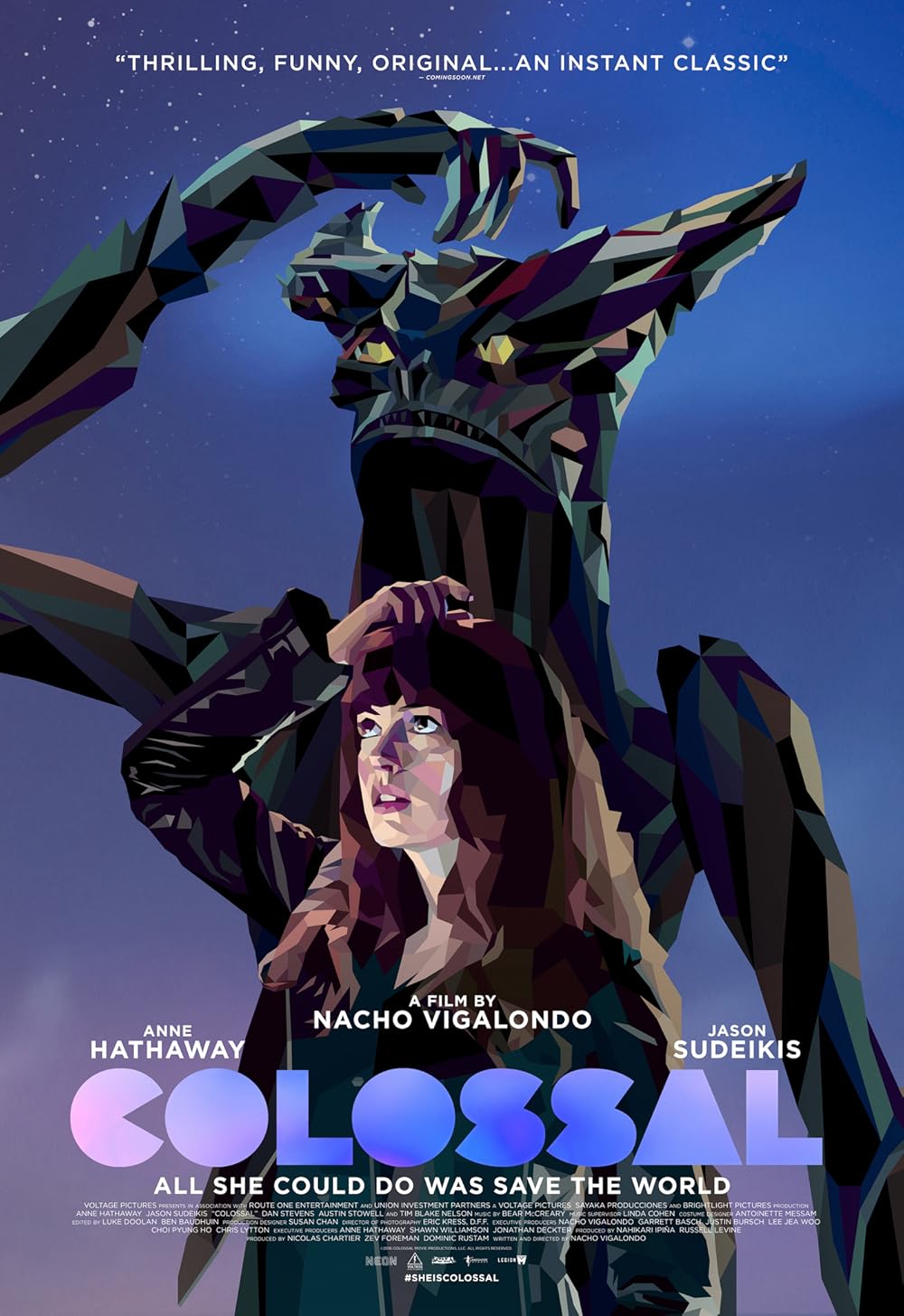} & 
\includegraphics[width=0.1\linewidth, height=0.145\linewidth]{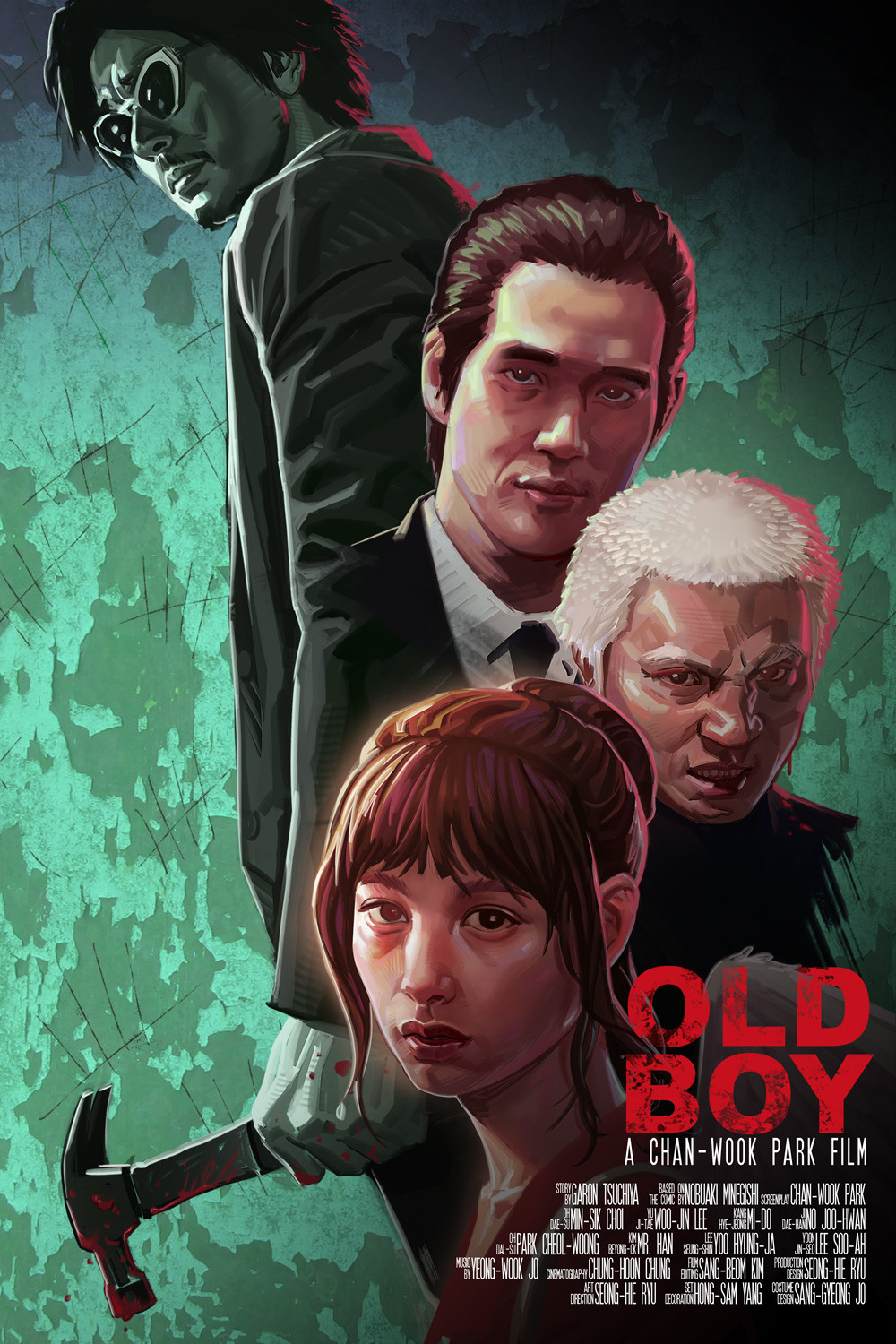} & 
\includegraphics[width=0.1\linewidth, height=0.145\linewidth]{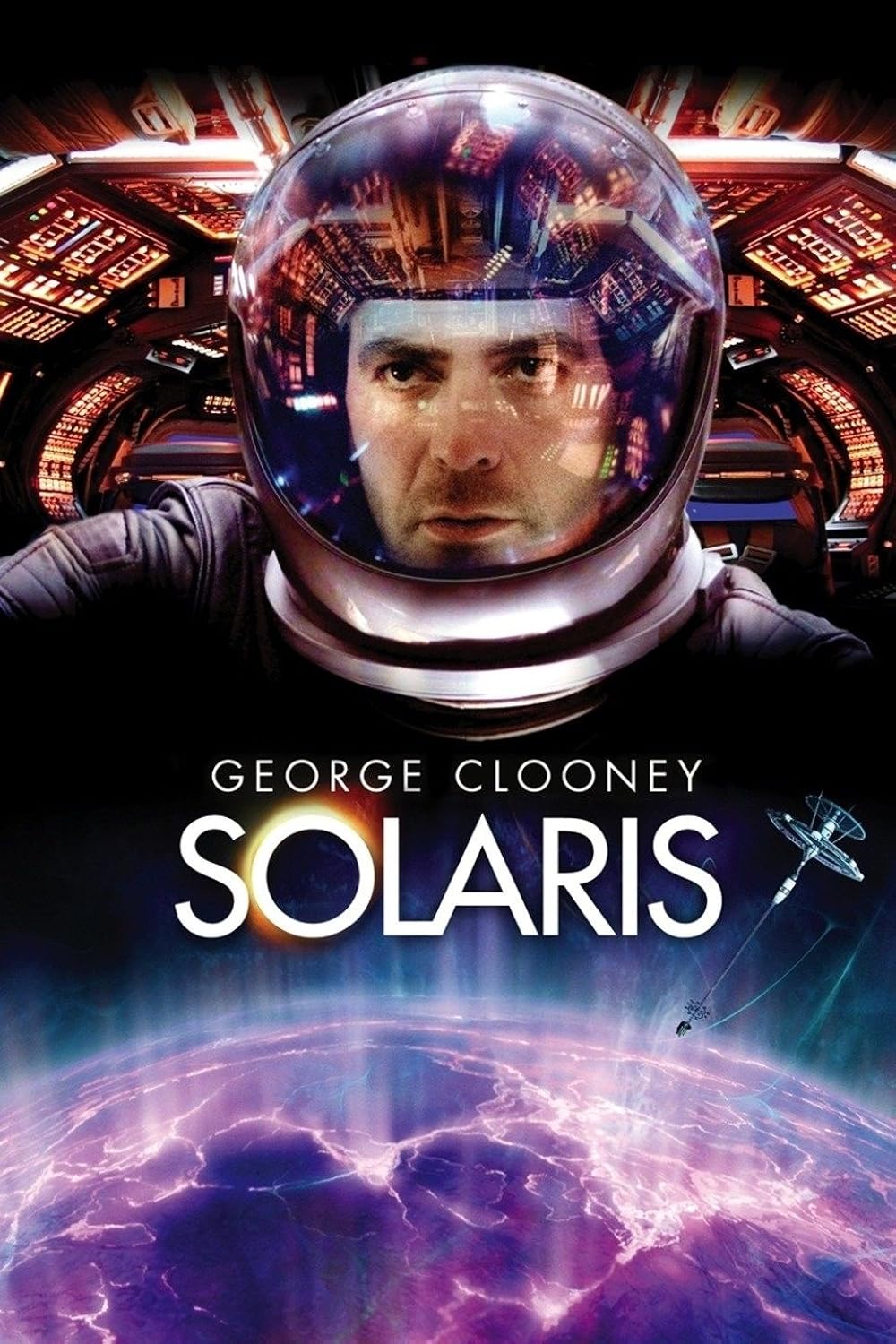} &
\includegraphics[width=0.1\linewidth, height=0.145\linewidth]{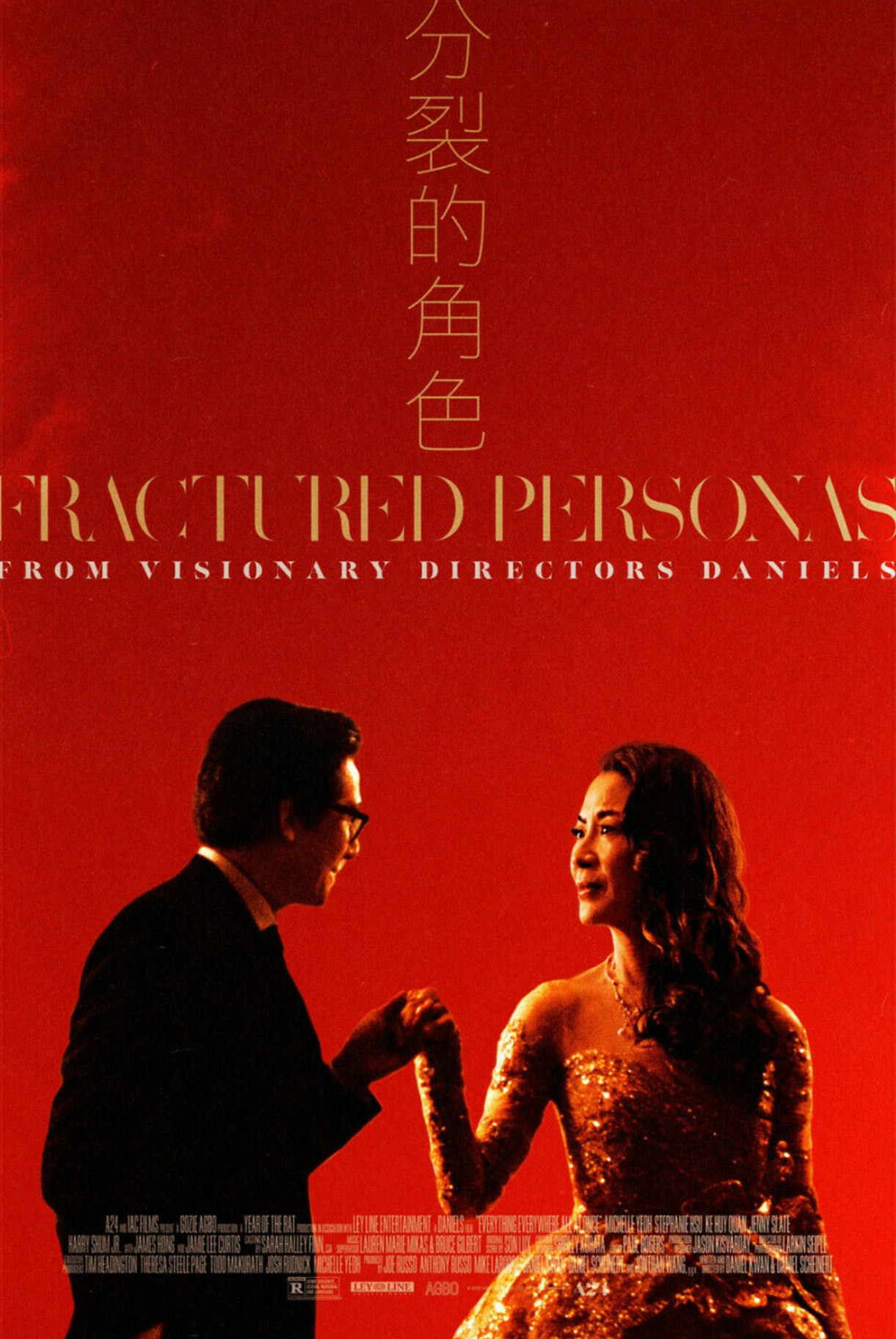} & 
\includegraphics[width=0.1\linewidth, height=0.145\linewidth]{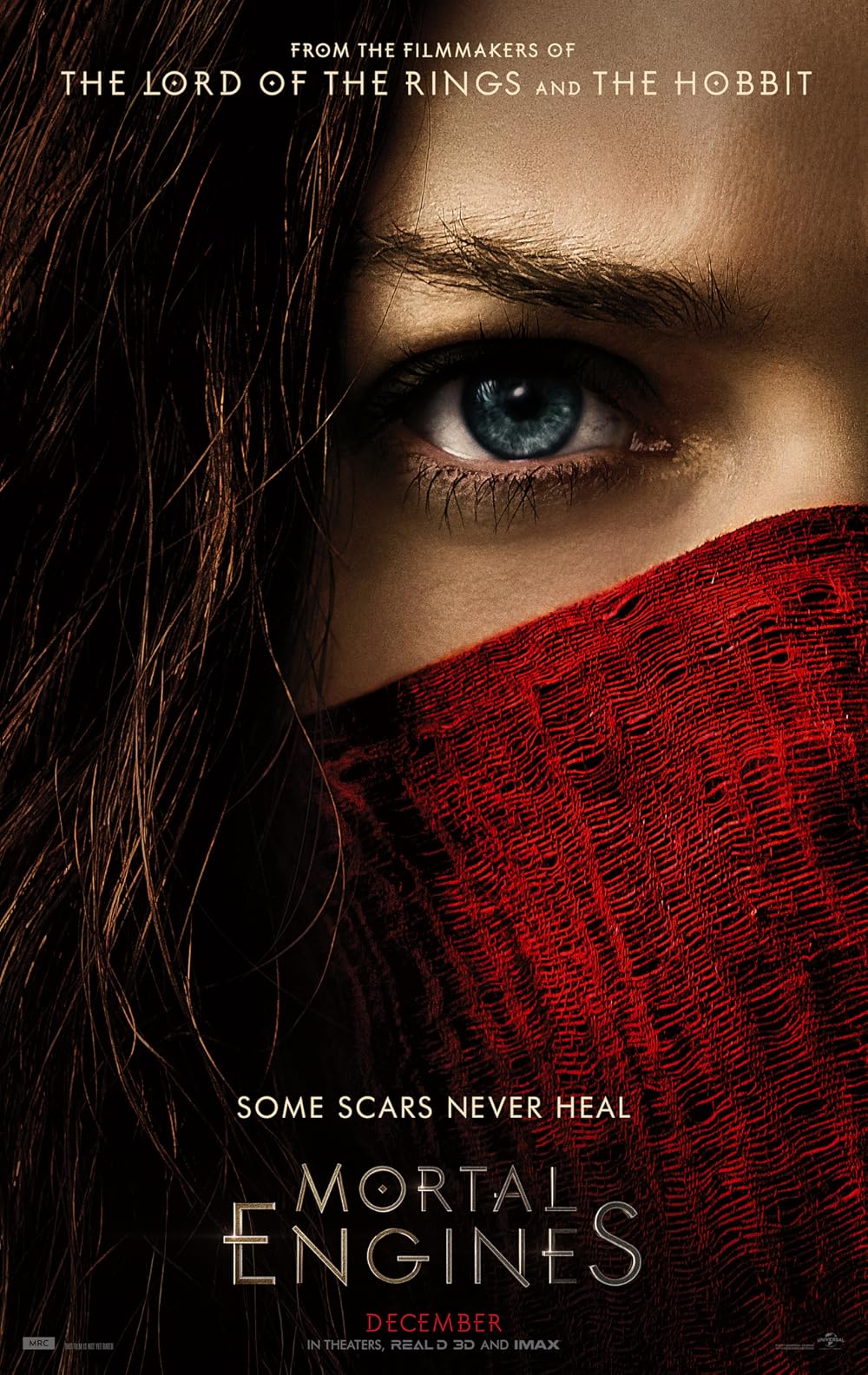} & 
\includegraphics[width=0.1\linewidth, height=0.145\linewidth]{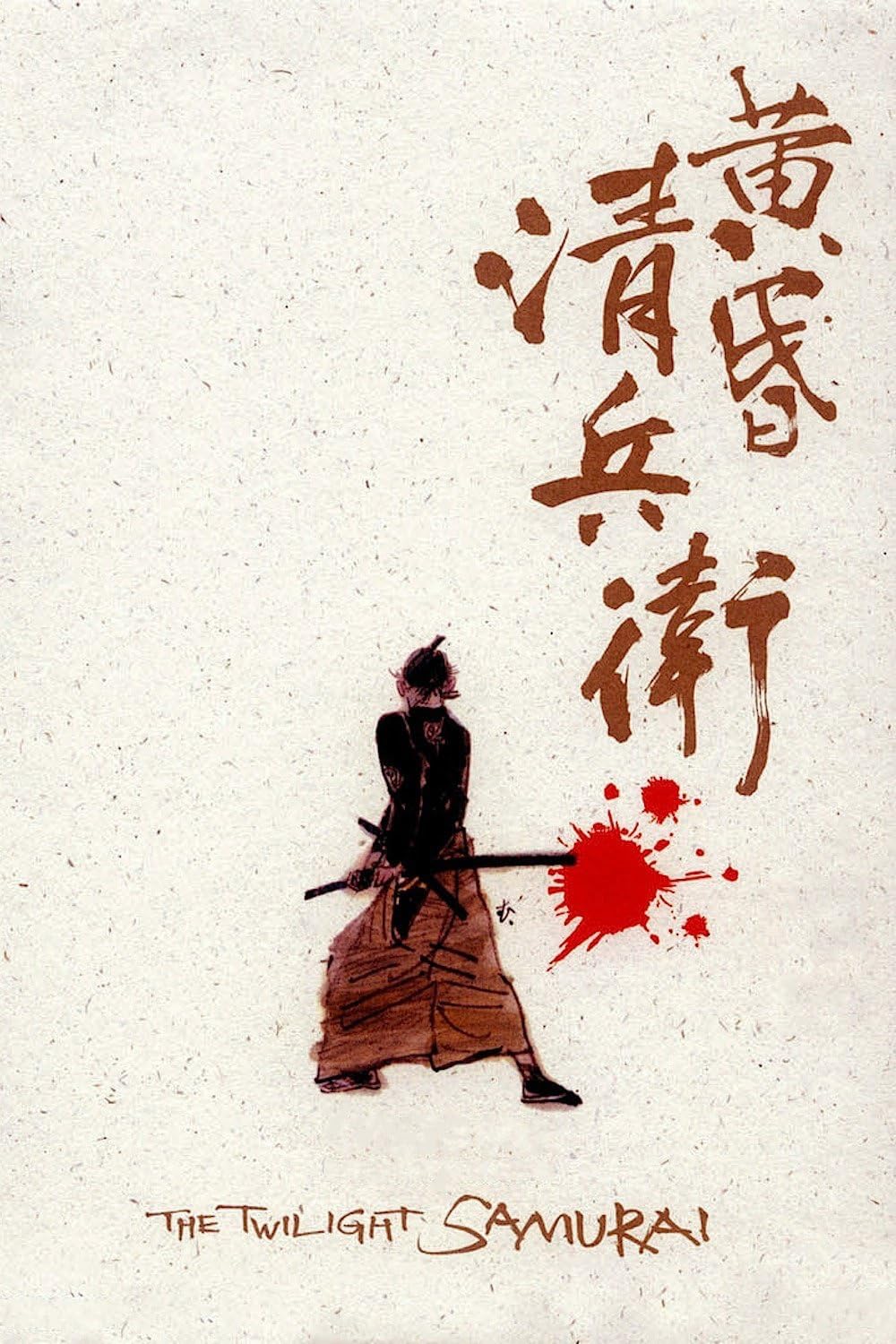} & 
\includegraphics[width=0.1\linewidth, height=0.145\linewidth]{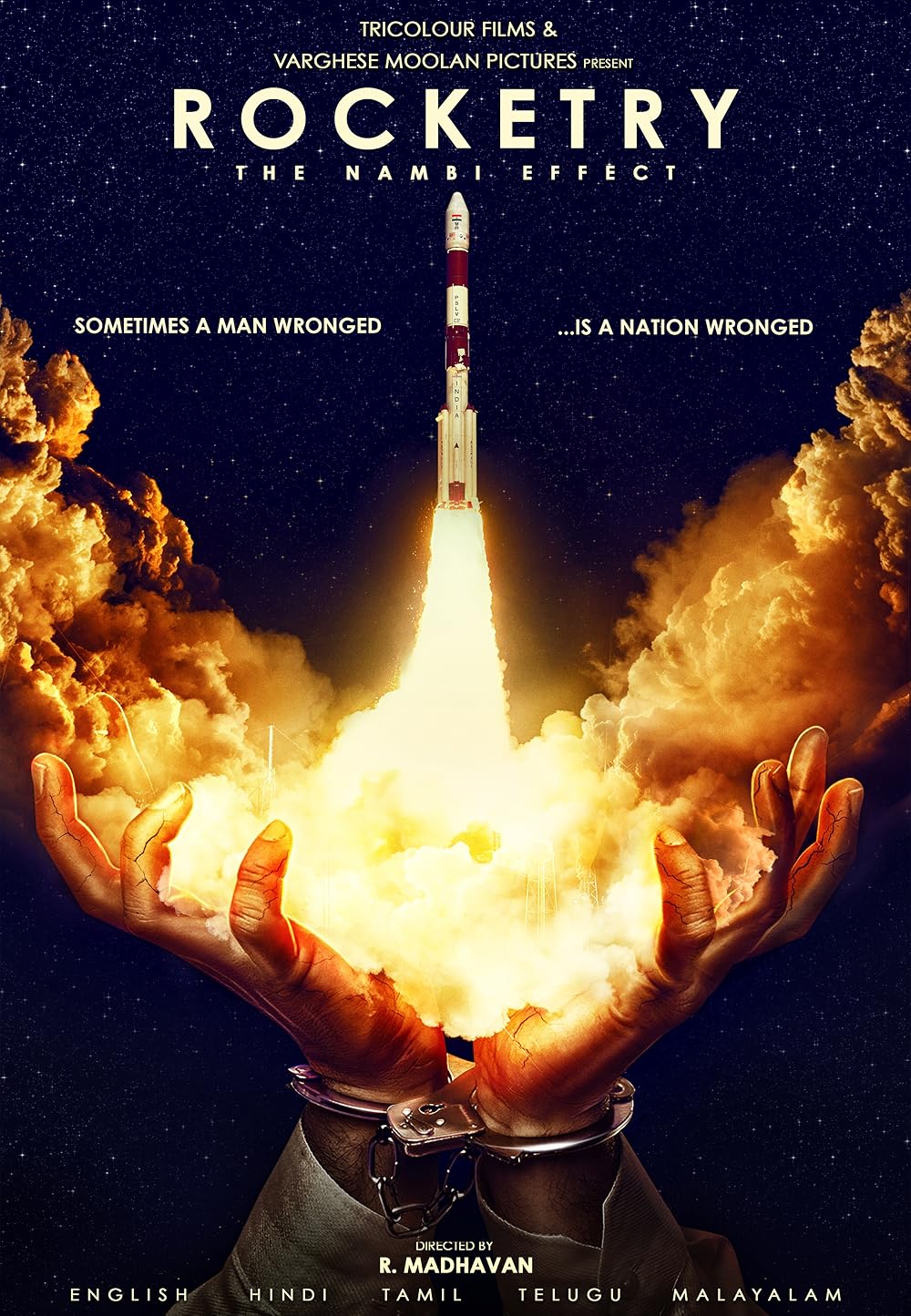} & 
\includegraphics[width=0.1\linewidth, height=0.145\linewidth]{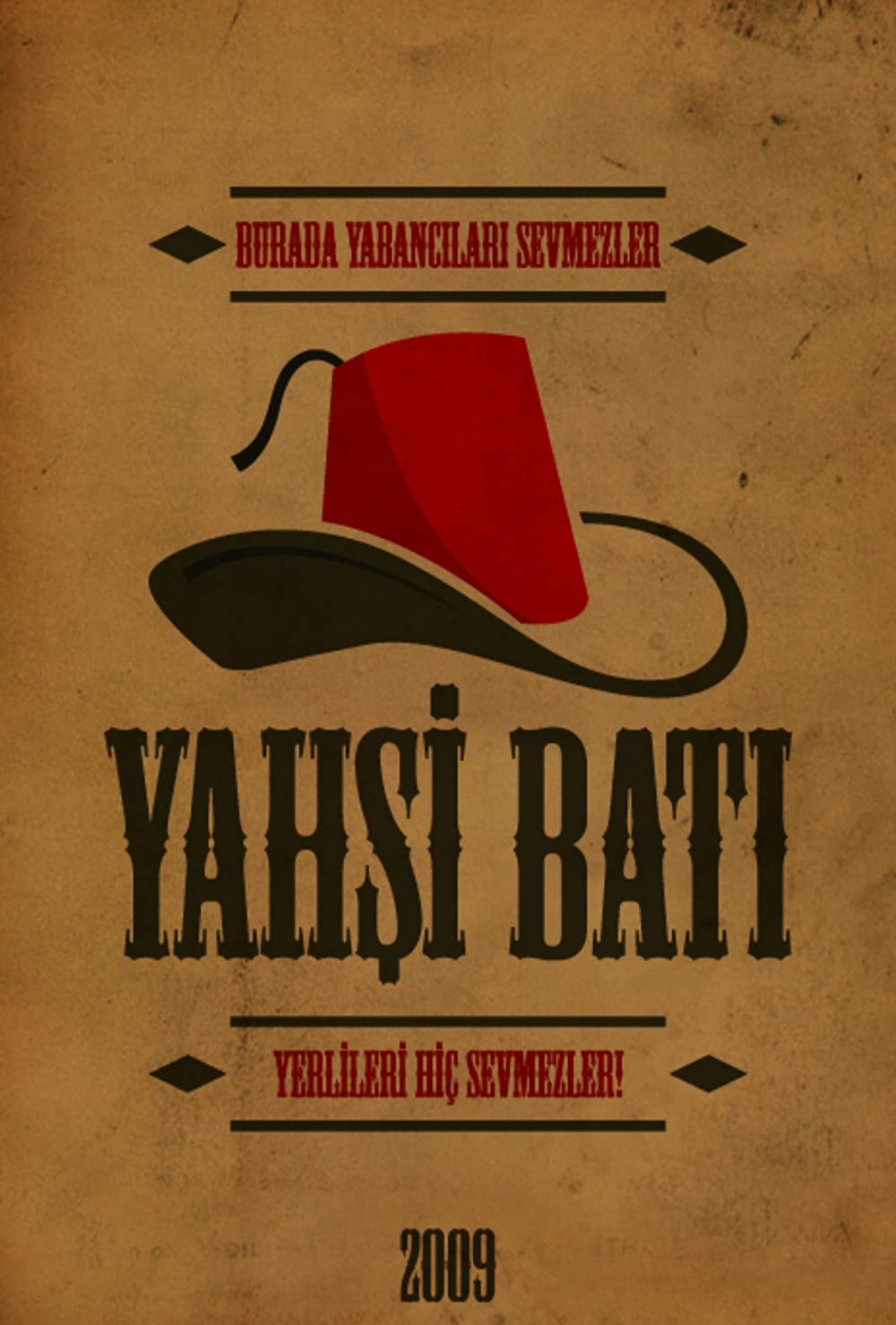} \\ 
 Action, Adventure, & Horror, Sci-Fi, & Action, Adventure, & Drama, Horror, & Drama, Horror, & Action, Horror,  & Action, Adventure,	& Drama, Horror, \\
 Animation & Thriller & Sci-Fi & Romance  & Mystery  & \--- &  Sci-Fi     & Mystery \\
\textcolor{blue}{\small\emph{i}}  & \textcolor{blue}{\small\emph{ii}}  & \textcolor{blue}{\small\emph{iii}}  & \textcolor{blue}{\small\emph{iv}}  & 
\textcolor{blue}{\small\emph{v}}  & \textcolor{blue}{\small\emph{vi}}  & \textcolor{blue}{\small\emph{vii}}  & \textcolor{blue}{\small\emph{viii}}   \\ 
\hline \hline 
&&& &&&&  \\[\dimexpr-\normalbaselineskip+1.5pt]
 Comedy, Fantasy  & Mystery, Thriller & Biography, Drama  & Action, Drama, & Comedy, Drama,  & Drama, Romance, & Drama, Fantasy, & Drama, Fantasy,\\
 \--- & \--- & \--- & Sci-Fi   & Fantasy    & Sci-Fi & Mystery & Mystery   \\
&&& &&&&  \\[\dimexpr-\normalbaselineskip+1.5pt]
\includegraphics[width=0.1\linewidth, height=0.145\linewidth]{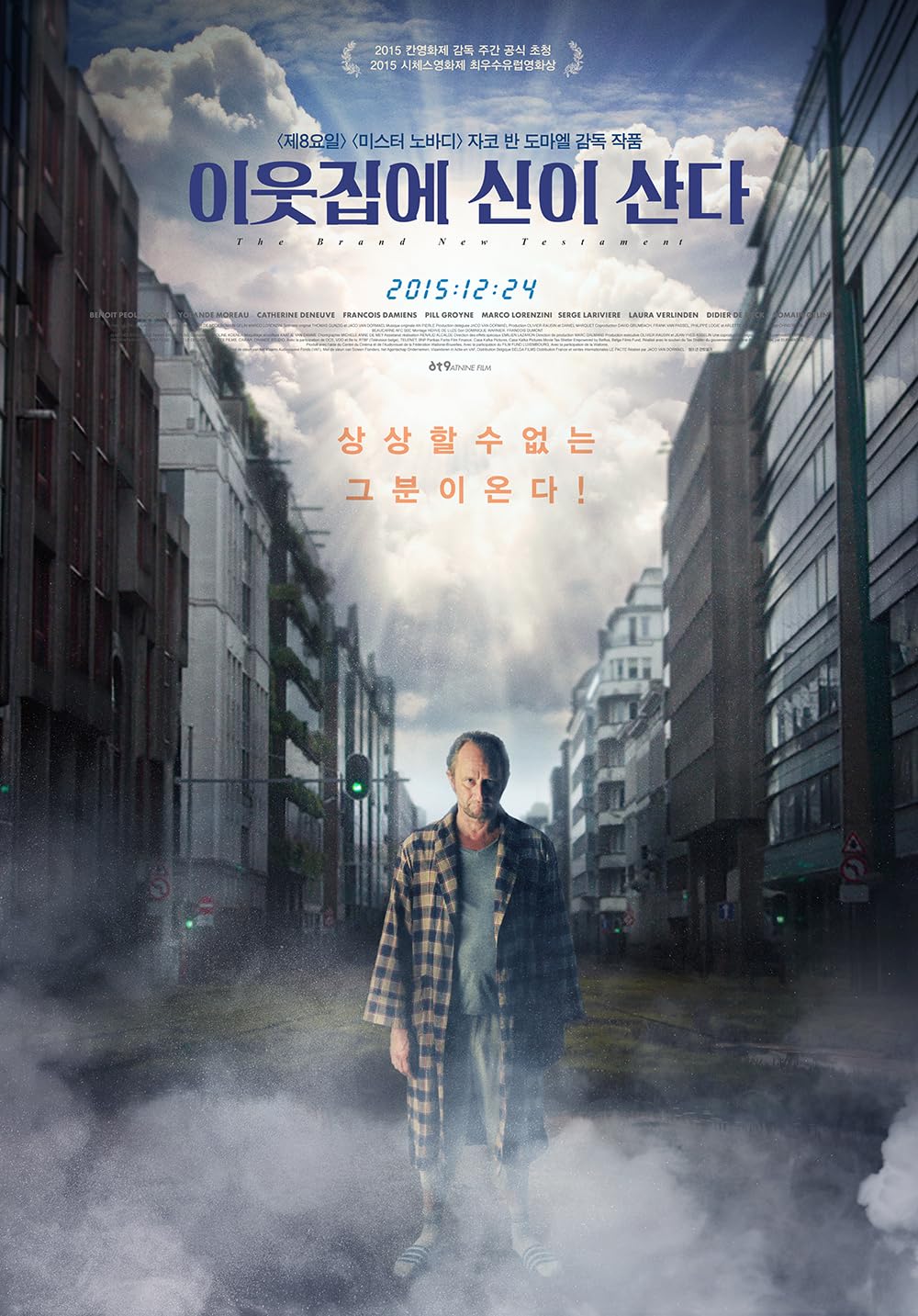} & 
\includegraphics[width=0.1\linewidth, height=0.145\linewidth]{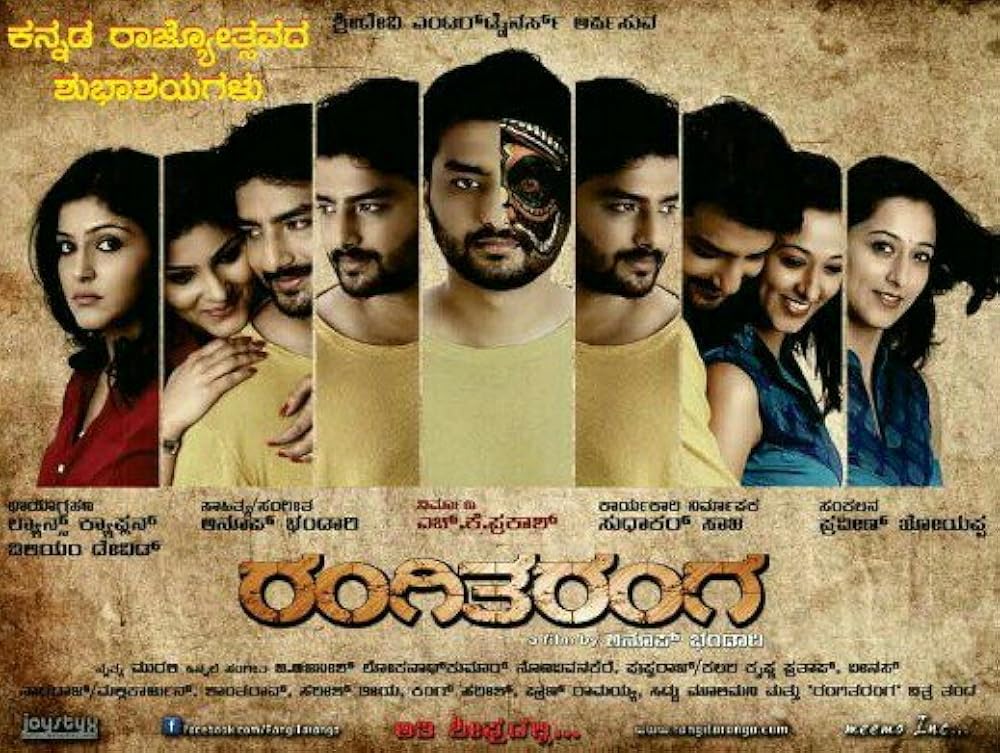} & 
\includegraphics[width=0.1\linewidth, height=0.145\linewidth]{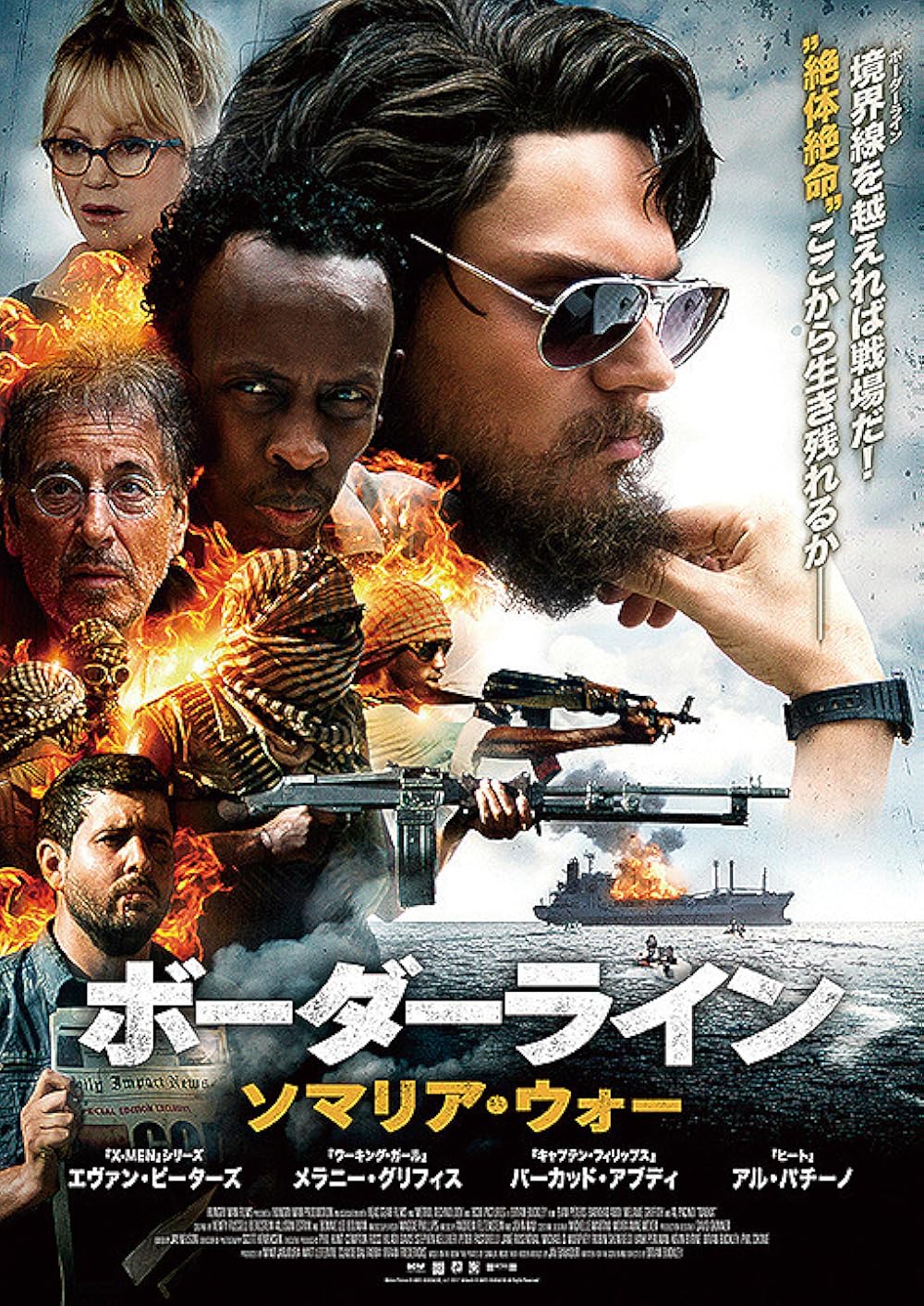} & 
\includegraphics[width=0.1\linewidth, height=0.145\linewidth]{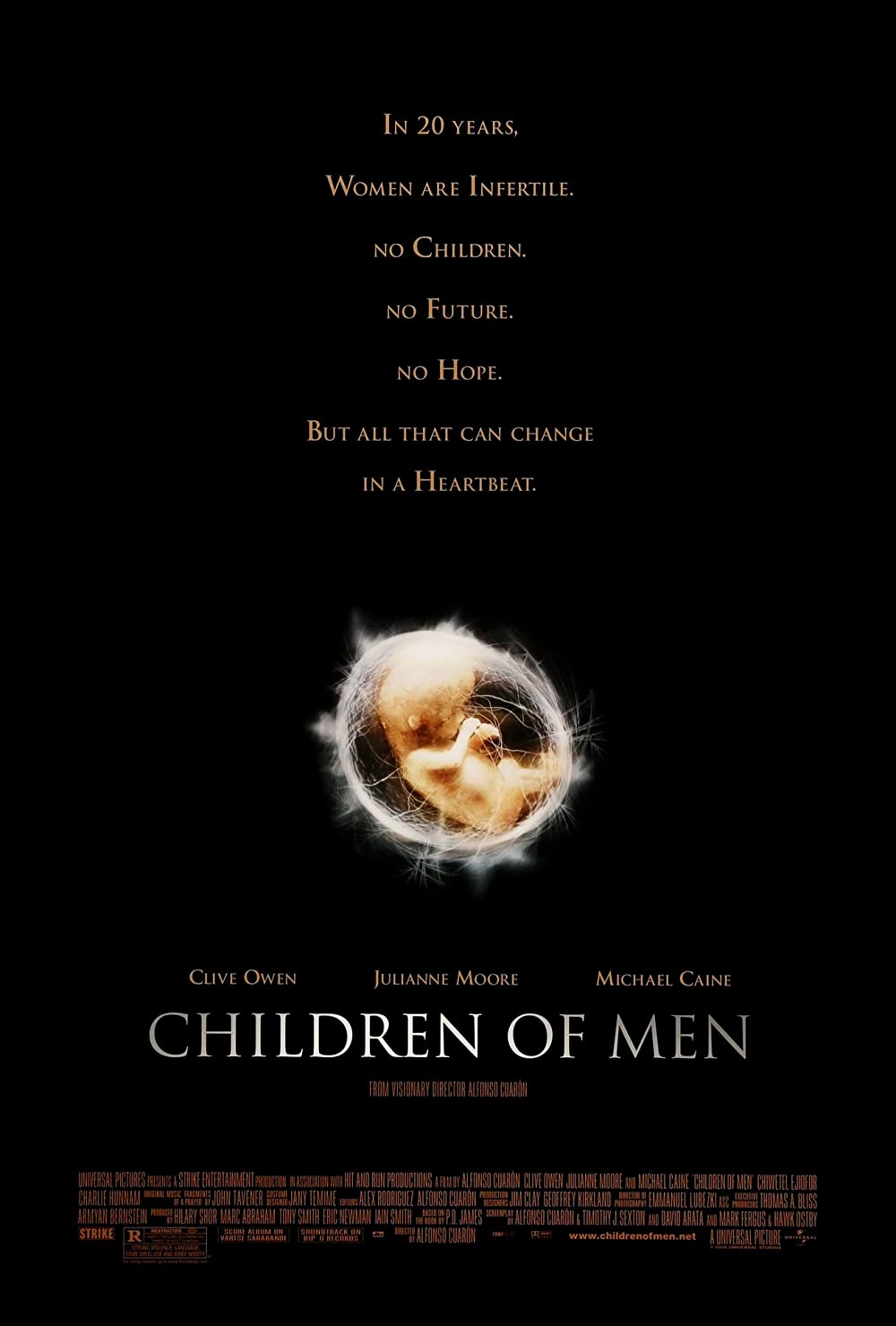} &
\includegraphics[width=0.1\linewidth, height=0.145\linewidth]{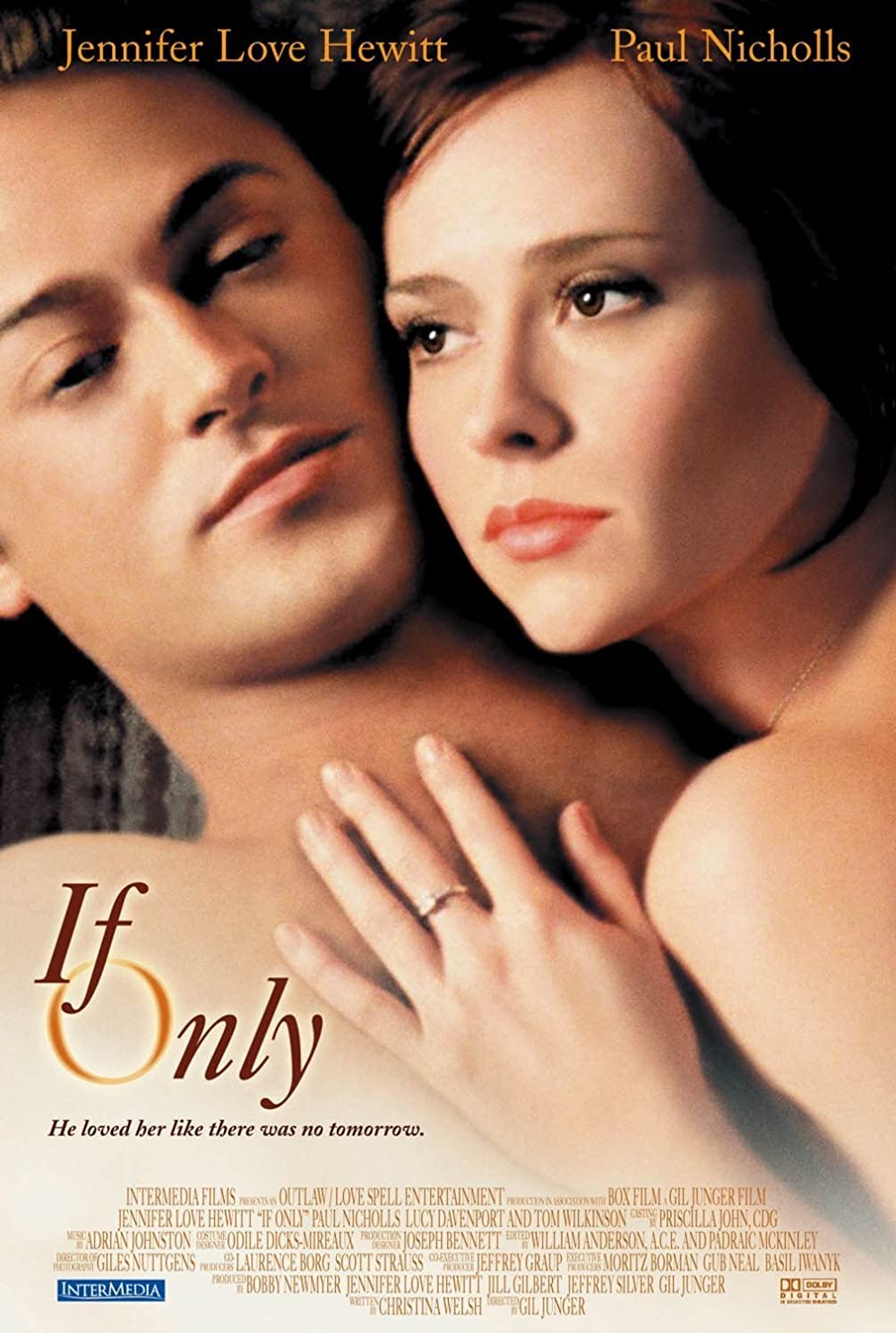} & 
\includegraphics[width=0.1\linewidth, height=0.145\linewidth]{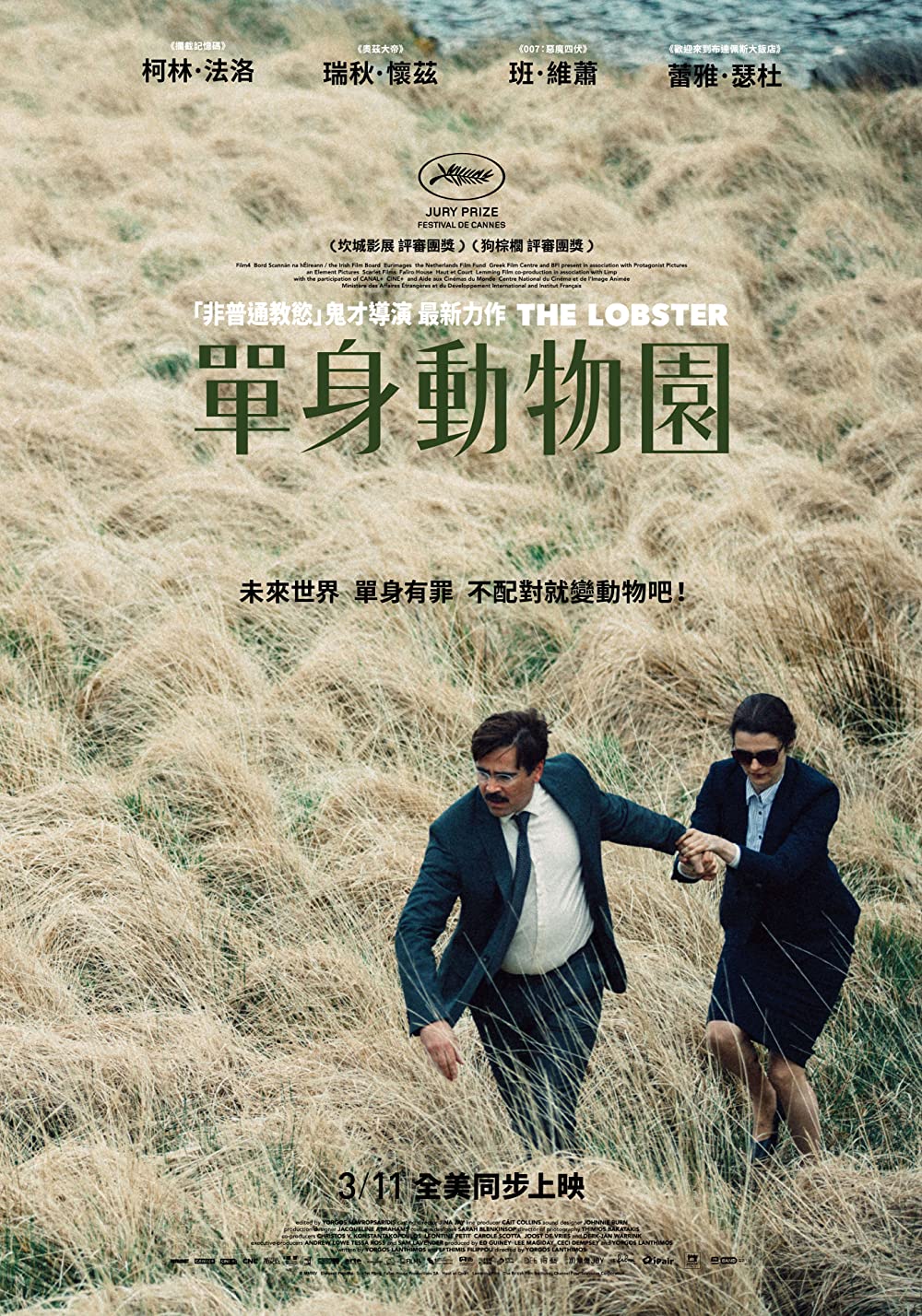} & 
\includegraphics[width=0.1\linewidth, height=0.145\linewidth]{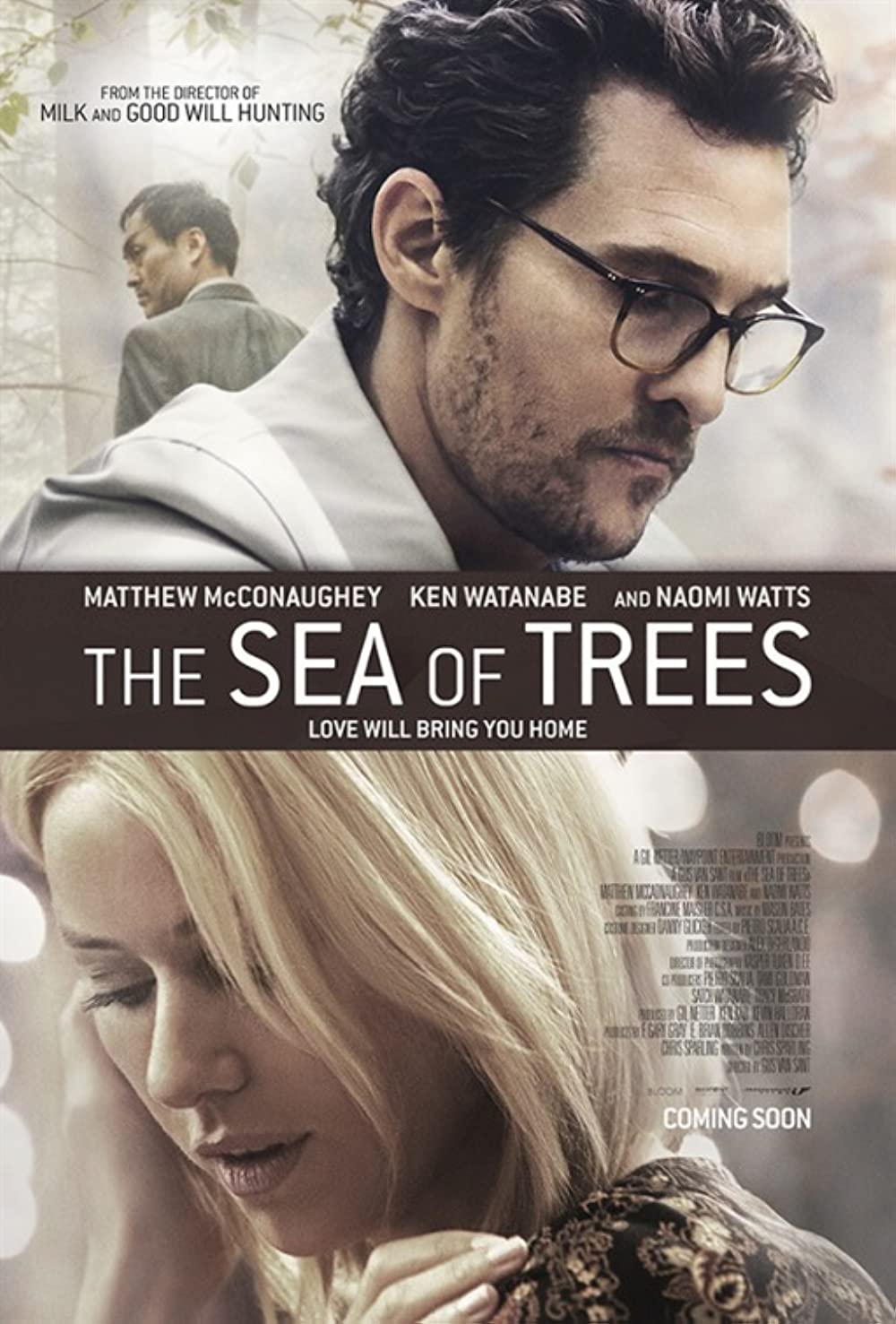} & 
\includegraphics[width=0.1\linewidth, height=0.145\linewidth]{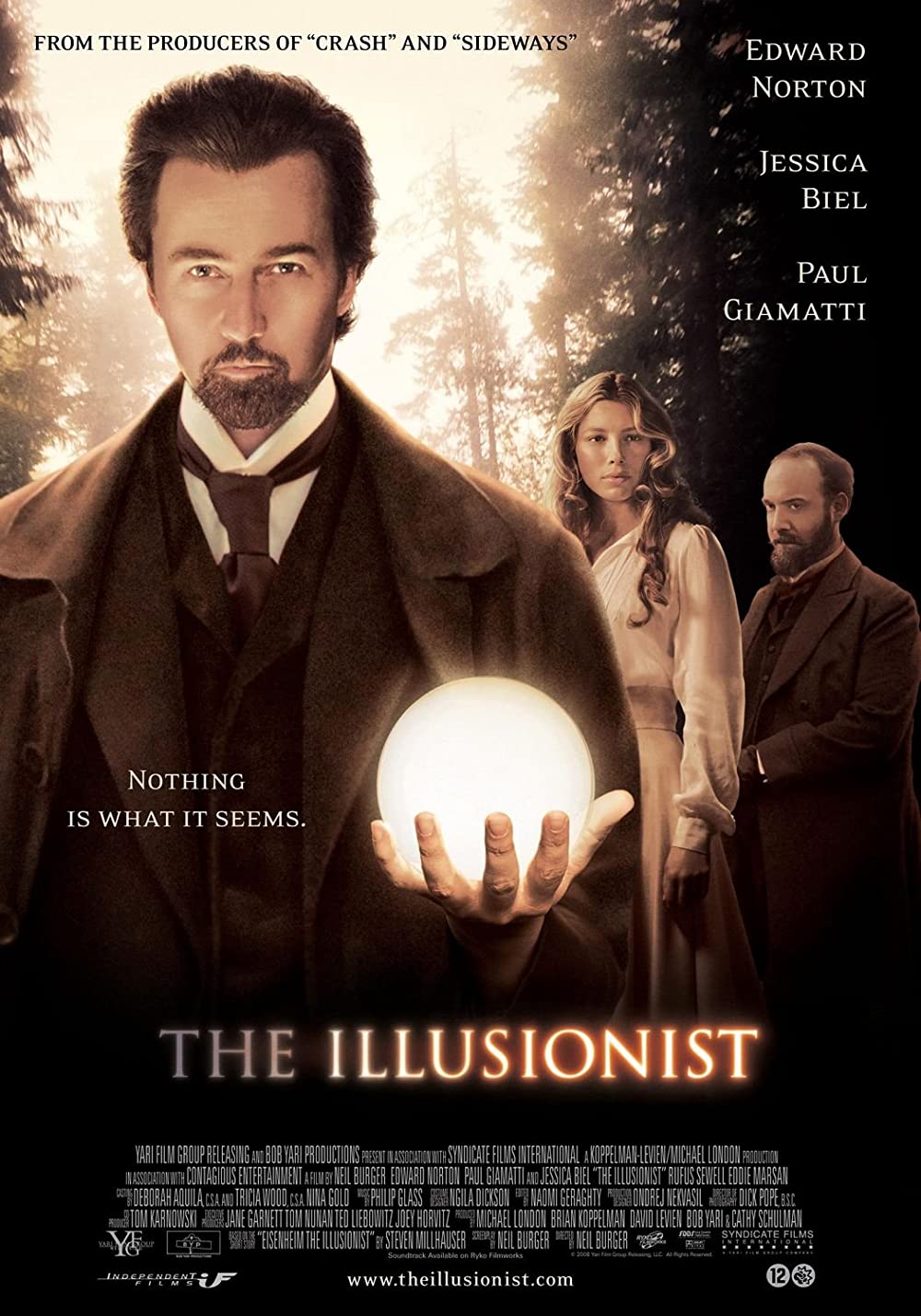} \\ 
Drama, Mystery,  & Comedy, Drama,   & Action, Adventure, & Biography, Drama,   & Biography, Drama,    & Biography, Crime,     & Biography, Drama,  & Biography, Drama, \\
\--- & Romance     & Thriller   & Mystery  & Romance & Drama  & Romance   & Mystery  \\ 
\textcolor{blue}{\small\emph{ix}}  & \textcolor{blue}{\small\emph{x}} & \textcolor{blue}{\small\emph{xi}}   & \textcolor{blue}{\small\emph{xii}}  & 
\textcolor{blue}{\small\emph{xiii}}  & \textcolor{blue}{\small\emph{xiv}}  & \textcolor{blue}{\small\emph{xv}}  & \textcolor{blue}{\small\emph{xvi}}  \\ 
\hline 
\end{tabular}
\end{adjustbox}
\caption{{Misprediction by PrERDT on incomprehensible posters (actual and predicted genres are mentioned above and below the posters, respectively)}
\label{fig:incomprehensible}}
\end{figure*}

\begin{table*}[]
\centering
\caption{{Genre indistinguishability analysis}}
\label{tab:cooccur}
\begin{adjustbox}{width=0.95\linewidth}
\begin{tabular}{l |c |c |c |c |c |c |c |c |c |c |c |c |c }
\cline{2-14}
 & {Action} & {Adventure} & {Animation} & {Biography} & {Comedy} & {Crime} & {Drama} & {Fantasy} & {Horror} & {Mystery} & {Romance} & {Sci-Fi} & {Thriller} \\ 
\hline
\multicolumn{14}{c}{\cellcolor[HTML]{D9D9D9}(a) \textbf{Cooccurrence}} \\
\hline
Action & 1 & 0.45 & 0.07 & 0.04 & 0.18 & 0.21 & 0.26 & 0.10 & 0.09 & 0.04 & 0.02 & 0.18 & 0.26 \\
Adventure & 0.61 & 1 & 0.28 & 0.03 & 0.39 & 0.05 & 0.23 & 0.11 & 0.04 & 0.02 & 0.02 & 0.15 & 0.05 \\
Animation & 0.29 & 0.88 & 1 & 0.01 & 0.63 & 0.02 & 0.11 & 0.03 & 0.01 & 0 & 0 & 0.02 & 0 \\
Biography & 0.17 & 0.08 & 0.01 & 1 & 0.17 & 0.31 & \textbf{0.87} & 0 & 0 & 0.01 & 0.13 & 0 & 0.03 \\
Comedy & 0.18 & 0.3 & 0.16 & 0.04 & 1 & 0.13 & 0.38 & 0.09 & 0.08 & 0.03 & 0.35 & 0.02 & 0.03 \\
Crime & 0.32 & 0.06 & 0.01 & 0.11 & 0.19 & 1 & 0.57 & 0.02 & 0.06 & 0.19 & 0.04 & 0.01 & 0.37 \\
Drama & 0.18 & 0.12 & 0.02 & 0.14 & 0.26 & 0.26 & 1 & 0.09 & 0.12 & 0.19 & 0.28 & 0.06 & 0.18 \\
Fantasy & 0.36 & 0.28 & 0.03 & 0 & 0.29 & 0.03 & 0.43 & 1 & 0.26 & 0.10 & 0.08 & 0.03 & 0.05 \\
Horror & 0.16 & 0.05 & 0 & 0 & 0.14 & 0.07 & 0.3 & 0.13 & 1 & 0.35 & 0.01 & 0.12 & 0.35 \\
Mystery & 0.09 & 0.03 & 0 & 0 & 0.06 & 0.26 & 0.55 & 0.06 & 0.4 & 1 & 0.05 & 0.09 & 0.32 \\
Romance & 0.04 & 0.02 & 0 & 0.06 & 0.61 & 0.04 & 0.72 & 0.04 & 0.01 & 0.04 & 1 & 0.03 & 0.04 \\
Sci-Fi & 0.57 & 0.34 & 0.01 & 0 & 0.07 & 0.02 & 0.26 & 0.02 & 0.22 & 0.14 & 0.05 & 1 & 0.19 \\
Thriller & 0.36 & 0.05 & 0 & 0.01 & 0.04 & 0.33 & 0.35 & 0.02 & 0.27 & 0.22 & 0.03 & 0.08 & 1 \\
\hline
\multicolumn{14}{c}{\cellcolor[HTML]{D9D9D9}(b) \textbf{Misprediction}} \\
\hline
Action & 0 & 0.02 & 0.01 & 0.02 & 0.06 & 0.06 & 0.2 & 0.02 & 0.06 & 0.06 & 0.05 & 0.01 & 0.07 \\
Adventure & 0.04 & 0 & 0 & 0.01 & 0.02 & 0.07 & 0.15 & 0.02 & 0.06 & 0.03 & 0.04 & 0.03 & 0.06 \\
Animation & 0.02 & 0.29 & 0 & 0 & 0.04 & 0.01 & 0.06 & 0.02 & 0.02 & 0.01 & 0.01 & 0.01 & 0.02 \\
Biography & 0.17 & 0.09 & 0 & 0 & 0.21 & 0.18 & \textbf{0.67} & 0.04 & 0.10 & 0.11 & 0.23 & 0.03 & 0.19 \\
Comedy & 0.06 & 0.07 & 0.01 & 0.03 & 0 & 0.05 & 0.11 & 0.02 & 0.03 & 0.04 & 0.02 & 0.01 & 0.06 \\
Crime & 0.08 & 0.06 & 0.01 & 0.05 & 0.10 & 0 & 0.34 & 0.01 & 0.09 & 0.11 & 0.12 & 0.05 & 0.14 \\
Drama & 0.06 & 0.05 & 0 & 0 & 0.02 & 0.02 & 0 & 0.03 & 0.02 & 0.02 & 0.01 & 0.03 & 0.04 \\
Fantasy & 0.18 & 0.12 & 0.02 & 0.05 & 0.07 & 0.09 & 0.36 & 0 & 0.15 & 0.17 & 0.07 & 0.14 & 0.14 \\
Horror & 0.12 & 0.10 & 0 & 0.01 & 0.03 & 0.12 & 0.2 & 0.03 & 0 & 0.06 & 0.03 & 0.03 & 0.10 \\
Mystery & 0.16 & 0.07 & 0 & 0.08 & 0.09 & 0.21 & 0.4 & 0.02 & 0.08 & 0 & 0.10 & 0.06 & 0.21 \\
Romance & 0.10 & 0.10 & 0.01 & 0.06 & 0.08 & 0.11 & 0.15 & 0.03 & 0.08 & 0.07 & 0 & 0.02 & 0.07 \\
Sci-Fi & 0.23 & 0.11 & 0.02 & 0.01 & 0.07 & 0.11 & 0.24 & 0.03 & 0.12 & 0.09 & 0.05 & 0 & 0.15 \\
Thriller & 0.10 & 0.10 & 0 & 0.07 & 0.09 & 0.10 & 0.28 & 0.02 & 0.06 & 0.11 & 0.07 & 0.04 & 0 \\
\hline
\multicolumn{14}{c}{\cellcolor[HTML]{D9D9D9}(c) \textbf{Poster Count}} \\
\hline
Action & 3724 & 1665 & 253 & 151 & 656 & 785 & 972 & 390 & 331 & 161 & 84 & 666 & 966 \\
Adventure & 1665 & 2725 & 768 & 69 & 1073 & 138 & 626 & 305 & 115 & 52 & 50 & 397 & 130 \\
Animation & 253 & 768 & 873 & 8 & 553 & 14 & 94 & 28 & 6 & 3 & 1 & 15 & 0 \\
Biography & 151 & 69 & 8 & \textbf{864} & 150 & 269 & \textbf{748} & 0 & 4 & 5 & 112 & 0 & 29 \\
Comedy & 656 & 1073 & 553 & 150 & 3561 & 477 & 1358 & 307 & 291 & 103 & 1232 & 85 & 112 \\
Crime & 785 & 138 & 14 & 269 & 477 & 2459 & 1392 & 37 & 155 & 472 & 87 & 20 & 900 \\
Drama & 972 & 626 & 94 & \textbf{748} & 1358 & 1392 & \textbf{5314} & 459 & 637 & 1002 & 1471 & 307 & 959 \\
Fantasy & 390 & 305 & 28 & 0 & 307 & 37 & 459 & 1074 & 279 & 105 & 91 & 27 & 49 \\
Horror & 331 & 115 & 6 & 4 & 291 & 155 & 637 & 279 & 2098 & 739 & 11 & 260 & 727 \\
Mystery & 161 & 52 & 3 & 5 & 103 & 472 & 1002 & 105 & 739 & 1828 & 89 & 168 & 591 \\
Romance & 84 & 50 & 1 & 112 & 1232 & 87 & 1471 & 91 & 11 & 89 & 2034 & 58 & 81 \\
Sci-Fi & 666 & 397 & 15 & 0 & 85 & 20 & 307 & 27 & 260 & 168 & 58 & 1162 & 220 \\
Thriller & 966 & 130 & 0 & 29 & 112 & 900 & 959 & 49 & 727 & 591 & 81 & 220 & 2707 \\
\hline 
\end{tabular}
\end{adjustbox}
\end{table*}

\subsection{Qualitative Analysis for Misprediction by PrERDT due to Incomprehensible Posters}

\noindent
{Our investigation into the misclassifications by PrERDT indicates that one possible reason for these errors is the incomprehensibility of certain posters. Even human evaluators might struggle to accurately discern the genres from the posters shown in Fig. \ref{fig:incomprehensible}.}

{We present some examples of misprediction by PrERDT in Fig. \ref{fig:incomprehensible}. 
It is evident that some posters in this set can lead to misunderstandings about their actual genres (e.g., Fig. \ref{fig:incomprehensible}: \emph{i}, \emph{iii}, \emph{iv}). 
Additionally, certain posters lack sufficient information to accurately identify the genres, resulting in potentially random genre mapping (e.g., Fig. \ref{fig:incomprehensible}: \emph{viii}, \emph{xii}).}

{Based on the above analysis, it can be recommended that these posters be redesigned to better attract users who have an affinity for the movies' actual genres. This will help prevent users from being misled by the poster design and ensure that they are more likely to engage with content that matches their interests.}

\subsection{Quantitative Analysis for Genre Indistinguishability}
\noindent
{Our further investigation into misclassifications by PrERDT reveals that the co-occurrence of genres contributes to potential errors. 
Table \ref{tab:cooccur}: (a) reports the co-occurrence of genre ${\cal{G}}_i$ in association with genre ${\cal{G}}_j$. 
In other words, the cell corresponding to row ${\cal{G}}_j$ and column ${\cal{G}}_i$ represents the ratio between the total number of training samples associated with both ${\cal{G}}_i$ and ${\cal{G}}_j$, and the total number of training samples associated with ${\cal{G}}_j$. 
Table \ref{tab:cooccur}: (b) demonstrates the misclassification rate by PrERDT, where each cell corresponding to row ${\cal{G}}_j$ and column ${\cal{G}}_i$ represents the ratio between the number of testing samples associated with ${\cal{G}}_j$ but wrongly identified as ${\cal{G}}_i$ (when ${\cal{G}}_i$ is not the actual genre) and the total number of testing samples belong to ${\cal{G}}_j$ that have been misclassified. 
Finally, Table \ref{tab:cooccur}: (c) presents the poster count of co-occurrences for each pair of genres in the training dataset.} 

{From Table \ref{tab:cooccur}, we observe that the high co-occurrence of two genres, combined with a significant imbalance in the number of training samples associated with one genre without the other, potentially causes a high misprediction rate. 
For instance, out of 348 movies in the \emph{biography} genre, 303 also fall into the \emph{drama} category. 
Now, if we consider the poster count instead of the movie count, in our training set, 748 out of 864 \emph{biography} posters are associated with \emph{drama} (refer to Table \ref{tab:cooccur}: (c)). This leaves only 116 \emph{biography} posters that are not categorized as \emph{drama}. In contrast, there are 4566 \emph{drama} posters that do not belong to the \emph{biography} genre, creating an imbalance between {non-drama} \emph{biography} samples and {non-biography} \emph{drama} samples.
Standard data augmentation techniques do not address this imbalance effectively, as increasing the number of samples for the \emph{biography} genre also increases the number of \emph{drama} samples. Consequently, we observed a misclassification rate of 0.67 for {non-drama} \emph{biography} samples, which our model incorrectly classified as \emph{drama} (refer to Table \ref{tab:cooccur}: (b)). A similar issue was noted with other genre pairs, such as \emph{mystery-drama} and \emph{fantasy-drama}. As a result, low balanced accuracy was observed for the \emph{biography}, \emph{fantasy}, and \emph{mystery} genres, as shown in Table VIII: (b) of Section IV-E in the main manuscript.}

{The above analysis highlights the need for a more sophisticated data augmentation technique that tackles pairwise genre co-occurrence. This issue can be further mitigated by incorporating additional movies that do not exhibit the usual co-occurrence patterns observed in the dataset.}

\subsection{Qualitative Analysis of Model Limitation}
\begin{figure}[!h]
\centering
\scriptsize
\begin{adjustbox}{width=0.49\textwidth}
\begin{tabular}{c|c|c}
Action, Animation, Drama      & Adventure, Biography, Drama     & Biography, Drama, Romance  \\ 
&&  \\[\dimexpr-\normalbaselineskip+1.5pt]
\includegraphics[width=0.25\linewidth, height=0.375\linewidth]{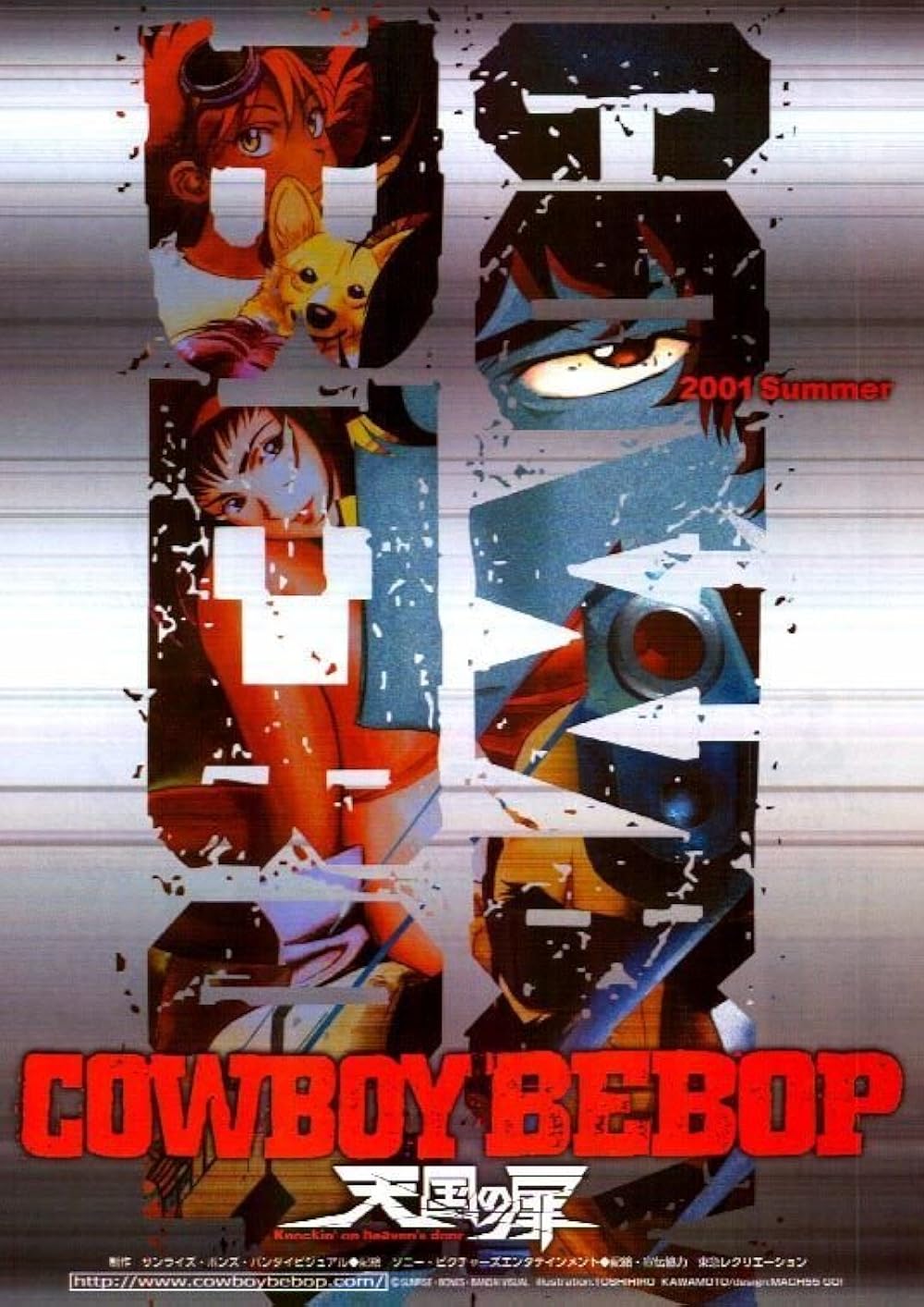} & 
\includegraphics[width=0.25\linewidth, height=0.375\linewidth]{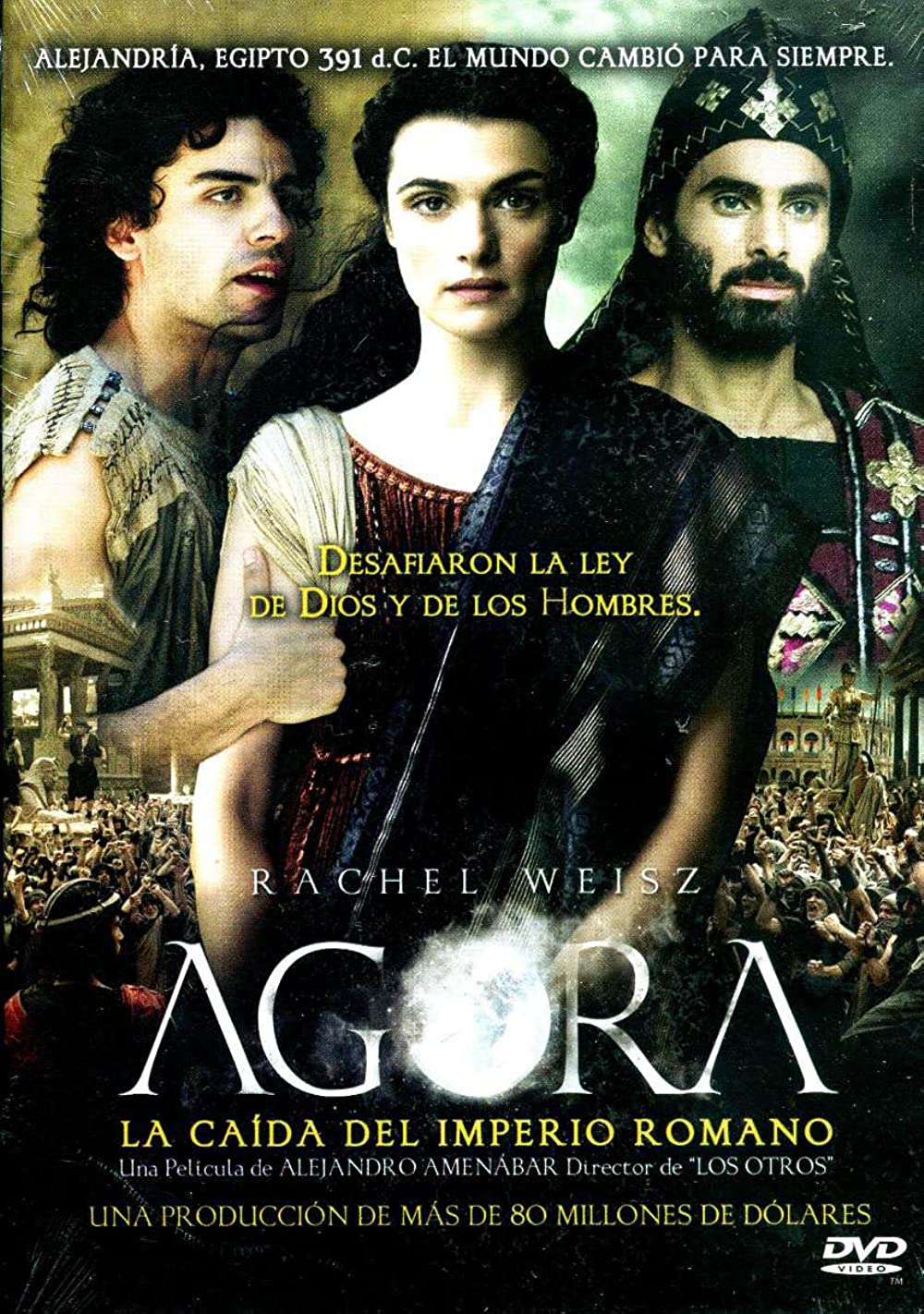} & 
\includegraphics[width=0.25\linewidth, height=0.375\linewidth]{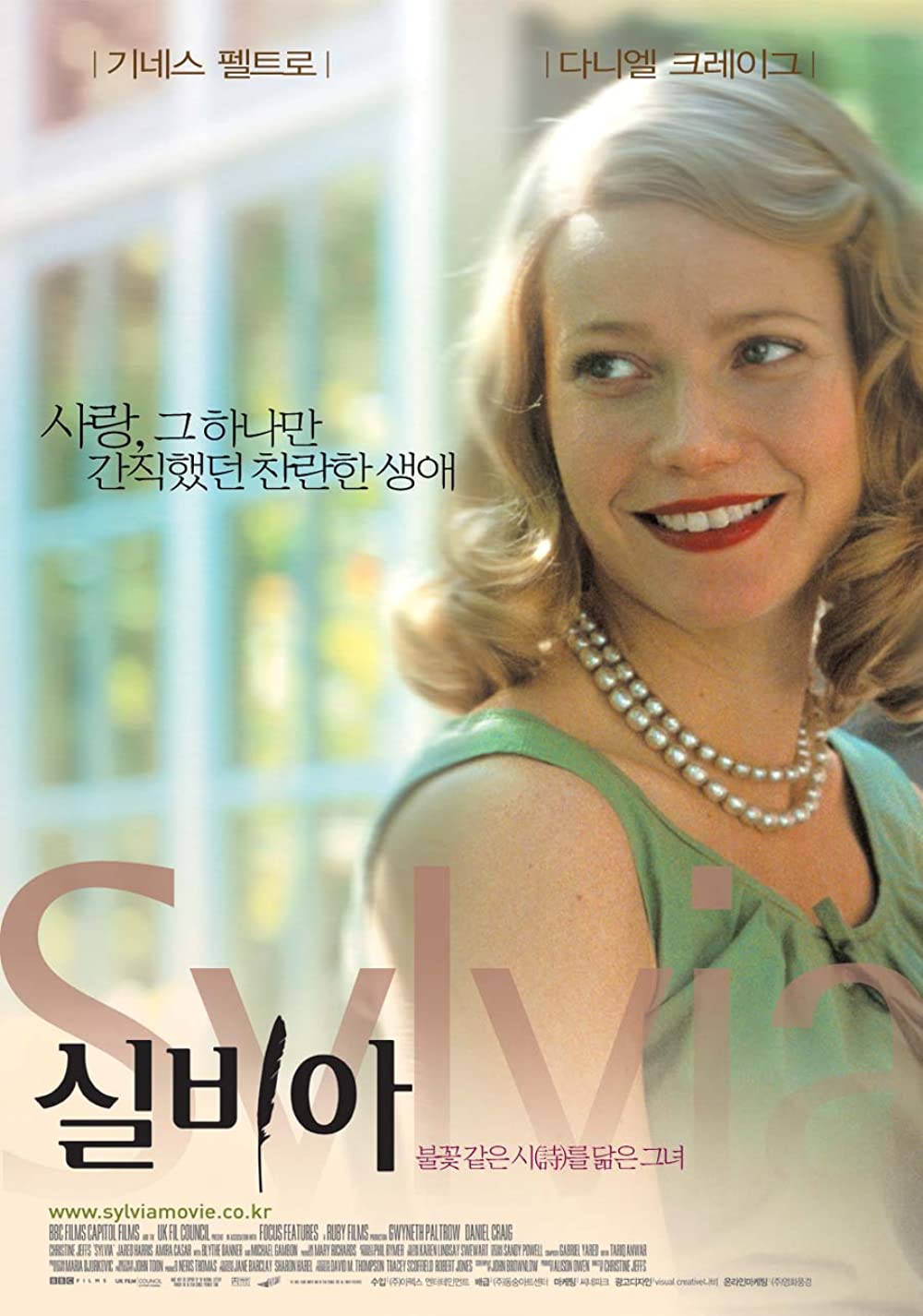} \\ 
Action, Adventure, Thriller   & Drama, Fantasy, Mystery         & Drama, Fantasy, Romance  \\
\textcolor{blue}{\small{(a)}} & \textcolor{blue}{\small{(b)}} & \textcolor{blue}{\small{(c)}} \\
\end{tabular}
\end{adjustbox}
\caption{Example of inaccurate predictions by PrERDT (actual and predicted genres are mentioned above and below the posters, respectively)}
\label{fig:limit2}
\end{figure}

\noindent
{Another major cause of misprediction by PrERDT is the model's limitation in comprehending genres from the moderate information revealed by posters. Fig. \ref{fig:limit2} illustrates several cases where our model failed to identify the correct genres from posters. For instance, in Fig. \ref{fig:limit2}: (a), the \emph{animation} characteristics are discernible only upon careful examination of the image inside the larger textual part. This analysis highlights areas for improvement in our model, which we intend to address in the future.}


\end{document}